\documentclass{article} % For LaTeX2e
\usepackage{iclr2024_conference,times}

\usepackage{amsmath,amsfonts,bm}

\def\eqref#1{equation~\ref{#1}}
\def\ceil#1{\lceil #1 \rceil}
\def\floor#1{\lfloor #1 \rfloor}
\def\1{\bm{1}}

\def\vzero{{\bm{0}}}

\def\vp{{\bm{p}}}

\def\vs{{\bm{s}}}

\def\vv{{\bm{v}}}
\def\vw{{\bm{w}}}
\def\vx{{\bm{x}}}

\def\mH{{\bm{H}}}

\def\mL{{\bm{L}}}

\def\mS{{\bm{S}}}

\def\mU{{\bm{U}}}
\def\mV{{\bm{V}}}
\def\mW{{\bm{W}}}

\DeclareMathAlphabet{\mathsfit}{\encodingdefault}{\sfdefault}{m}{sl}
\SetMathAlphabet{\mathsfit}{bold}{\encodingdefault}{\sfdefault}{bx}{n}

\def\gC{{\mathcal{C}}}

\def\gP{{\mathcal{P}}}

\def\gS{{\mathcal{S}}}

\newcommand{\E}{\mathbb{E}}

\newcommand{\R}{\mathbb{R}}

\DeclareMathOperator{\Tr}{Tr}

\usepackage{hyperref}
\usepackage{url}

\usepackage{graphicx}
\usepackage{wrapfig}

\usepackage{algorithm}
\usepackage{algpseudocode}

\usepackage{amsmath,amssymb}
\usepackage{enumitem}

\usepackage[capitalise,noabbrev]{cleveref}

\newtheorem{theorem}{Theorem}[section]

\DeclareTextFontCommand{\emph}{\em}

\newcommand{\quarot}{QuaRot}

\title{Pyramid Vector Quantization for LLMs}

\author{Tycho F. A. van der Ouderaa \thanks{work completed during internship at MSR} \\
Imperial College London\\
\texttt{tycho.vanderouderaa@imperial.ac.uk} \\
\And
Maximilian L. Croci \\
Microsoft AI\\
\texttt{mcroci@microsoft.com} \\
\And
Agrin Hilmkil\\
Microsoft Research\\
\texttt{agrinhilmkil@microsoft.com}
\And
James Hensman\\
Microsoft Research\\
\texttt{jameshensman@microsoft.com}
}

\iclrfinalcopy % Uncomment for camera-ready version, but NOT for submission.

\begin{document}

\maketitle
%\vspace{2em}

\begin{abstract}
Recent works on compression of large language models (LLM) using quantization considered reparameterizing the architecture such that weights are distributed on the sphere. This demonstratively improves the ability to quantize by increasing the mathematical notion of coherence, resulting in fewer weight outliers without affecting the network output. In this work, we aim to further exploit this spherical geometry of the weights when performing quantization by considering \textit{Pyramid Vector Quantization} (PVQ) for large language models. Arranging points evenly on the sphere is notoriously difficult, especially in high dimensions, and in case approximate solutions exists, representing points explicitly in a codebook is typically not feasible due to its additional memory cost. Instead, PVQ uses a fixed integer lattice on the sphere by projecting points onto the 1-sphere, which allows for efficient encoding and decoding without requiring an explicit codebook in memory. To obtain a practical algorithm, we propose to combine PVQ with scale quantization for which we derive theoretically optimal quantizations, under empirically verified assumptions. Further, we extend pyramid vector quantization to use Hessian information to minimize quantization error under expected feature activations, instead of only relying on weight magnitudes. Experimentally, we achieve state-of-the-art quantization performance with pareto-optimal trade-off between performance and bits per weight and bits per activation, compared to competitive methods. On weight-only, we find that we can quantize a Llama-3 70B model to 3.25 bits per weight and retain 98\% accuracy on downstream tasks.
\end{abstract}

\vspace{-0.0em}
\section{Introduction}

Quantization enables compression of large language models (LLMs) by reducing the number of bits per weight required to represent weights. Weight outliers can make quantization difficult, as they cause weight distributions to not match the implicit evenly distributed grids used in many quantization methods. To overcome this, recent works have proposed to reparameterize architectures by rotating weights in a way that leaves the network as a function unchanged \citep{ashkboos2024quarot,chee2024quip}.

In this work, we aim to exploit the spherical geometry in weights when performing quantization by considering Pyramid Vector Quantization (PVQ) \citep{fischer1986pyramid} for LLMs. In PVQ, weights are quantized on a hyper-pyramidal lattice that allows efficient encoding and decoding without having to explicitly represent a codebook in memory. By projecting the lattice onto the hyper-sphere, a quantization grid is obtained that accurately approximates a uniform grid on the spherical domain.

PVQ has been very successful in well-known audio \citep{valin2012definition} and video codecs \citep{daede2016daala}. We demonstrate that the same algorithm allows practical quantization of large language models, by proposing a group-wise quantization scheme and further extending PVQ to use Hessian information accounting for curvature in the loss. We also propose a scheme to quantize the normalized scales (amplitudes) of each group according to theoretically derived quantiles, which we verified to closely match the empirical weight distributions of pretrained LLMs in practice. Our contribution extends beyond \citet{liguori2017pyramid}, which considered PVQ on weights of small neural networks.

Experimentally, we find that our proposed PVQ quantization scheme outperforms the state-of-the-art in terms of bits per weight and bits per activation. We do not only perform simulated quantization, but also provide kernels that allow hardware accelerated encoding and decoding of PVQ. We achieve state-of-the-art quantization on the most prominent Llama-3, Phi-3 and Mistral architectures in terms of performance against bits per weight (BPW). In particular, we demonstrate 3.25 bit weight quantization at a negligible 1-3\% drop in performance, as measured in accuracy on downstream tasks.

\section{Background}

Before discussing our approach on using pyramid vector quantization to quantize LLMs, we provide an overview of vector quantization, spherical geometry of LLM weights, and describe classic PVQ.

\subsection{Quantization}
Quantization is a compression technique for machine learning models by storing weights ($\mW$) or activations ($\vx$) in a some chosen lower bit representation, such as lower precision floats or scaled integers ($\widehat\mW$). A common conversion is to minimize a second-order layer-wise proxy loss \citep{nagel2020up,frantar2022gptq},
\begin{align}
L(\widehat{\mW}) = \E_{\vx} \left[ || \mW \vx - \widehat{\mW} \vx ||_2 \right] = \Tr\left( (\mW - \widehat{\mW}) \mH (\mW - \widehat{\mW}) \right)
\label{eq:proxy-loss}
\end{align}
where the layer-wise Hessian $\mH = \E_{\vx} \left[ \vx_n \vx_n^T \right] \approx \frac{1}{N} \sum_{n=1}^N \vx_n \vx_n^T$ is empirically estimated using a calibration dataset $\{ \vx_n \}_{n=1}^N$. The objective is optimal in the sense that it minimizes the layer-wise output at each layer, but a crude approximation with respect to the actual training loss. Some recent works have proposed to also use gradients to improve the approximation to the true training objective \citep{van2023llm}. Although it would be interesting to combine ideas presented in this paper with gradients, we stick to a layer-wise loss \cref{eq:proxy-loss} for simplicity and because this yields a faster quantization method that does not require backpropagation.

\subsection{Vector quantization}

Instead of individually quantizing weights, as done in scalar quantization, vector quantization aims to simultaneously quantize multiple weights. It can be shown that, even for completely independent Gaussian sources, this typically results in much higher theoretical signal-to-noise ratios, leading recent works to consider vector quantization for LLMs \citep{van2024gptvq, liu2024vptq,egiazarian2024extreme,tseng2024quip,tseng2024qtip}. Yet, vector quantization is not widely adopted in practice because of two practical problems. Firstly, naive vector quantization requires constructing an explicit codebook using clustering (such as K-means), quickly becoming infeasibly large for higher number of dimensions. Secondly, quantization requires an expensive search over this explicit codebook, which can not practically be used on-the-fly on activations. Although application to LLMs is limited, it is common in vector quantization \citep{gray1998quantization} to use an \textbf{implicit codebook} which does not have to be explicitly instantiated in memory. PVQ is such a vector quantization and comes with the additional benefit of being \textbf{search-free}, allowing encoding and decoding of vectors without having to perform an explicit and exhaustive lookup that at least linearly scales in algorithmic complexity with the size of the codebook. As a result, it can be applied on-the-fly and not only on the neural network weights, but also to the activations during inference. 

\subsection{Weights on the sphere}
\label{sec:weights-on-sphere}

Instead of viewing weight vectors in Euclidean $D$-space $\vw \in \mathbb{R}^D$, they can be interpreted as scaled points $\vw = s \vv$ on the unit sphere $\vv \in \Omega_D$, with $\Omega_D = \{ \vv = (v_1, v_2, \ldots, v_D) \in \mathbb{R}^D : \|\vv\|_2 = 1\}$. We refer to this spherical decomposition as the \textit{direction} $\vv = \frac{\vw}{\|\vw\|_2}$ and the \textit{amplitude} $s = \|\vw\|_2$ of a vector. Recent works that use such a spherical perspective on LLM weights have offered new insights in properties of the training dynamics and guide algorithmic improvements. For example, \citep{kosson2023rotational} noted that LLM weights under weight decay or popular deep learning optimizers converge to an equilibrium on the sphere, theoretically predicting the magnitude of the amplitude after training. Recent works have shown that only updating direction components can be beneficial in low-rank adaptation \cite{liu2024dora} and training itself \citep{loshchilov2024ngpt}. Yet, weights are not always uniformly distributed making quantization hard and giving rise to outliers. We use \textit{coherence processing} \citep{chee2024quip} to reparameterize weights on the sphere in a way that reduces outliers without functionally changing the output of the network. Further, in high dimensions the information in the amplitude is negligible compared to the information in the direction \citep{kipnis2021gaussian}. The observation that LLM weights are in practice uniformly distributed across the sphere is the primary motivation behind exploring pyramid vector quantization, which allows us to construct an efficient spherical quantization code.

\paragraph{Hadamard coherence processing}
Recent works have shown that rotating weights \citep{chee2024quip,ashkboos2024quarot} can improve weight coherence and reduce the number of outliers without altering the network's output,
$\widetilde{\mW} = \mU \mW \mV$
where $\mU \in \R^{R \times R}$ and $\mV \in \R^{C \times C}$ are both orthogonal matrices. Since the transpose of orthogonal matrices equals their inverse, the reparameterisation can simply be undone in the forward pass by left- and right multiplying with matrix transposes, $\mW = \mU^T \widetilde{\mW} \mV^T$. We follow \cite{chee2024quip,ashkboos2024quarot} and use random Hadamard matrices for $\mU$ and $\mV$, which can be implemented very efficiently.

Representing weights on the sphere through incoherence processing has shown to improve existing quantization methods by preventing outliers in weight distributions. Yet, these methods still quantize weights to Euclidean grids and do not exploit the spherical geometry of underlying weight distributions. We explore pyramid vector quantization to construct a quantization grid that is tailored to the spherical geometry by being approximately uniform on the sphere.

\paragraph{Absorbing rotation matrices}
In most cases, the additional rotation matrices do not result in a memory or compute overhead during inference since rotation matrices can be absorbed into weight matrices: depending on the placement of the rotation matrix in the architecture, it can be left- or right- multiplied with an adjacent weight matrix. This principle was used in \quarot~\citep{ashkboos2024quarot}, which describes how rotation matrices can be efficiently absorbed into attention and fully-connected layers in commonly used LLM architectures.

\subsection{Classic Pyramid Vector Quantization}
\label{sec:pvq-background}

\begin{figure}[t]
\hspace{-0.1\linewidth}
\includegraphics[width=1.15\linewidth]{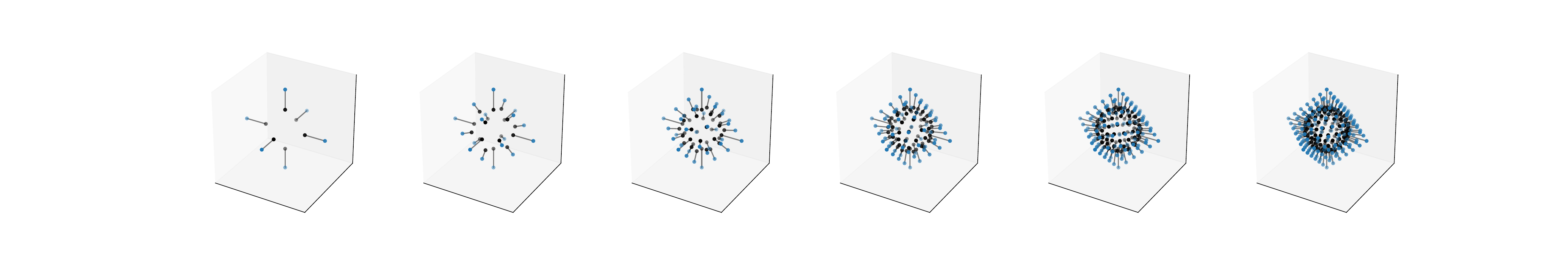}
\vspace{-2.5em}
\caption{Illustration of the PVQ integer lattice in $d=3$ dimensions with increasing pulses $k$ from 1 to 6. Points on the pyramid $\gP_{3, k}$ are projected onto the sphere $\gS_{3, k}$.}
\vspace{-0.5em}
\label{fig:3d-pvq}
\end{figure}

To quantize points on the sphere, we would like to construct \textit{spherical code}, a finite subset on the unit sphere $\gS \subset \Omega_D$. Packing a set of points on the surface of a sphere such that their distance is maximized is a notoriously hard problem in mathematics, famously dating back to the Dutch botanist \citet{tammes1930origin}. Just like sphere packing, optimal spherical codes are not generally known, with the exception of some specific dimensions, similar to the E8 packing in 8 dimensions exploited in \citep{tseng2024quip}. Even though good but sub-optimal spherical codes exist \citep{conway2013sphere}, there is not always an efficient method to enumerate the packing without requiring an explicit codebook, resulting in costly quantization. PVQ \citep{fischer1986pyramid} provides a solution to both of these issues simultaneously, allowing good spherical codes to be constructed in arbitrary dimension that can be efficiently encoded and decoded without having to maintain an explicit codebook in memory. This is achieved by projecting an integer lattice on the $l_1$ ball $\gP_{D, K}$ onto the hypersphere $\gS_{D, K}$. We now formalize these concepts by providing an overview of classic PVQ algorithm.

\paragraph{The integer pyramid lattice} Formally, the integer lattice of PVQ on the $D$-dimensional hyper-sphere is obtained by starting from a set of points on the $l_1$ ball of radius $K$, and projecting the set of points onto the sphere. We denote the set of integer points on the $l_1$ ball as $\gP_{D, K}$,
\begin{align}
\gP_{D, K} = \left\{ \vp \in \mathbb{Z}^D : ||\vp||_1 = \sum_{d=1}^D |\vp_d| = K \right\} 
\end{align}
%For historical reasons, we also refer to $P_{D, K}$ as the integer points that lie on \textit{the pyramid}, instead of $l_1$ ball, even though in mathematics this term is often described to a strictly more general class of geometric objects\james{the $l_1$ ball is a "slice" of a pyramid. Let's discuss how to present this here}, and refer to $k$ as the \textit{number of pulses}.
To obtain our spherical code $\gS_{D, K}$, we can project the points $\gP_{D, K}$ onto the sphere, by normalizing,
\begin{align}
\gS_{D, K} = \left\{ \frac{\vp}{||\vp||_2} : \vp \in \gP_{D, K} \right\}\,.
\end{align}
The number of codes $N(D, K) = |\gS_{D, K}| = |\gP_{D, K}|$ can be written as
\begin{align}
N(D, K) = 
2 D \cdot {}_2 F_1(1-D, 1-K, 2 ; 2)
\end{align}
where
${}_2F_1(a, b, c ; z)$ is the hypergeometric function \citep{terriberry2007pulse}, and for specific $D, K \in \mathbb{Z}$ can be computed through the following recurrence relation,
\begin{align}
\label{eq:size_table}
N(D, K) = N(D - 1, K) + N(D, K - 1) + N(D, K)
\end{align}
where we define $N(D, 0)=1$ for all $D \geq 0$ and $N(0, K) = 0$ for all $K\geq 1$.

%An illustration of the integer lattice can be found in for $d=2$ dimensions in \cref{fig:pvq-codebook} and $d=3$ dimensions in \cref{fig:3d-pvq}. It can be observed that distributions are not uniform on the sphere, with slightly denser grids close to the poles. Even though this distortion implies that PVQ obtains lower error rates for Laplacian sources, it can also be effectively applied to data from a Gaussian source \citep{fischer1986pyramid}, as also evidenced by the empirical validation carried out in this work.

%Lastly, we discuss power projections in \cref{sec:power-projections} which can further improve PVQ to lower the error on Gaussian source data.

\subsection{Subroutines of classic PVQ}

\paragraph{Quantizing the direction}
To quantizing a vector $\vw \in \R^D$, we map it to the closest point on the pyramid $\gP_{D, K}$ using an iterative procedure that projects the vector $\vw_0 = \vw$ onto the $l_1$ ball of radius $K$, and round it to the closest integer,
\begin{align}
\vw_{t+1} = \text{quantize\_step}_K(\vw_{t}) = \text{round}\left(\frac{K}{||\vw_t||_1} \vw_t \right)
\end{align}
After this step, we check whether the norm satisfies $||\vw||_1 = K$. If this is the case, we're done. If not, we either decrease by 1 or increase by 1, the element in $\vw_t$ with the biggest deviation $|\vw_i| - K$. We then call $\text{quantize\_step}_K(\cdot)$ again, and repeat this process until convergence, which should happen within at most $T < D$ steps. The resulting vector lies on the pyramid $\vw_T \in \gP_{D, K} \subset \mathbb{Z}^D$, meaning it is integer-valued and the absolute values sum to 1. We provide pseudocode of the quantization algorithm in \cref{alg:encode} in \cref{sec:encode-alg} to be fully self-contained.

% \paragraph{Closed-form amplitudes}
% 
% After quantizing a vector onto the pyramid, we multiply it by a single scale parameter $s \in \R$ that minimizes the L2 distance to the original points. This is akin to the scale used in virtually any other quantization methods, like round-to-nearest in practice. In PVQ, we interpret a set of weights as a vector $\vw$ of which $s$ changes the radius, while keeping the angles fixed. The optimal scale that minimizes our proxy objective \cref{eq:proxy-loss} can be found in closed-form,
% \begin{align}
% s = \frac{\widehat{\vw}^T\widehat{\vw}}{\widehat{\vw}^T \vw}
% \end{align}
% Thus, we can write a set of weights as $\vw = s \widehat{\vw}$, referring to (quantized) vector $\widehat{\vw}$ as the `\textit{direction}' and scalar $s$ as the `\textit{amplitude}'. In \cref{sec:amplitude-quantization}, we propose a scheme to also quantize the amplitudes.

\begin{wrapfigure}{r}{0.37\textwidth}
    \vspace{-1.2em}
    \centering
    \includegraphics[width=13em]{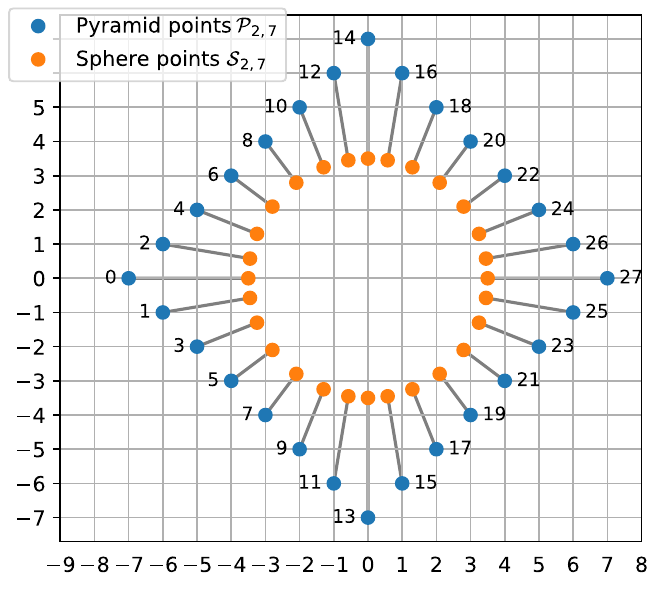}
    \vspace{-0.40em}
    \caption{Illustration of points on pyramid $\gP_{2, 7}$, their projections onto the sphere $\gS_{2, 7}$ and codes in $\gC_{2,7}$.}
    \label{fig:pvq-codebook}
    \vspace{-2.0em}
\end{wrapfigure}

\paragraph{Pyramid encoding} We can encode points on the pyramid $\vp \in \gP_{D, K}$ to integer codes $c \in [1, \ldots, N(D, K)-1]$ using an efficient algorithm, which avoids having to build an explicit table in memory to represent the codebook. The original PVQ paper \citep{fischer1986pyramid} describes an encoding scheme, of which we provide pseudocode \cref{alg:encode} in \cref{sec:encode-alg}, for the purpose of being self-contained. The algorithm provides a bijective mapping from the points of the pyramid $\gP_{D,K}$ to the set of integer indices $\gC_{D,K}$, providing a compact representation that allow vectors to be efficiently stored in as few bits as possible. In \cref{fig:pvq-codebook}, we provide an example illustration of points on $\gP_{2,7}$ and indices $\gC_{2,7}=[0, 27]$. The table of $N(d, k)$ can be precomputed for $0 \leq d \leq D$ and $0 \leq k \leq K$ and reused to avoid recomputing the same quantity.

\paragraph{Pyramid decoding}

We decode the integer codes $c \in \gC$ of PVQ to their associated vectors $\vp \in \gP_{D, K}$ through a decoding algorithm which performs the inverse operation of the encoding algorithm above. In \cref{sec:decode-alg}, we provide a corrected version of the decoding algorithm described in the original PVQ paper \citep{fischer1986pyramid}. The original paper contains an error and misses a line after setting $x_i \gets -j$, resulting in wrong decodings when $\vp$ contains negative values, except for when the last value is negative (which happens to be the case for the example given in the original paper).

% \paragraph{Power projections (optional improvement)}
% \label{sec:power-projections}
% 
% Power projections offers an optional extention to PVQ proposed by \citet{duda2017improving}, relying on a non-linear change-of-variables transforming the PVQ quantization grid to one that is more uniformly distributed across the sphere. This can be helpful, as the original PVQ uses an $l_1$-norm which has especially good error rates for Laplacian source data, whereas we in LLMs we can expect weights that are closer to a Gaussian source, based on empirical observations (see ...). Power projections are simple to implement, by using an adapted quantization step:
% \begin{align}
% \vw' = |\vw|^{\frac{1}{p}} * |\vw|
% \end{align}
% where $p \in \R$ denotes the power. In ..., we empirically find that the optimal power for Gaussian source data is roughly 1.2.  

\newpage
\section{PVQ for LLM compression}

Before we present our overall algorithm to perform PVQ to LLMs in \cref{sec:pvq-llm-overview}, we discuss the motivations and practical benefits of PVQ for LLM quantization in \cref{sec:why-pvq-for-llm} and analyse the theoretical signal-to-noise ratio of PVQ \cref{sec:theoretical-analysis}.

\subsection{Practical advantages of PVQ}
\label{sec:why-pvq-for-llm}

PVQ offers several practical advantages over competing methods. Firstly, PVQ is a vector quantization method which means it can achieve higher signal-to-noise ratios than scalar quantization methods that quantize weights independently. Secondly, \textbf{PVQ uses an implicit codebook}, which means that it does not require an explicit codebook to be constructed in memory. This makes the approach more memory efficient, but more importantly, the implicit codebook size can reach far beyond the memory that would have been required with an explicit codebook. To illustrate, explicitly storing a codebook that quantizes 16 bit precision vectors of a groupsize of 128 to 4 bit per weight would require approximately $2.7\cdot 10^{154}$ bytes, exceeding the estimated information capacity of the observable universe. With PVQ, we can use implicit codebooks of this size because encoding and decoding are done by an efficient algorithm, not a table lookup. Thirdly, \textbf{PVQ is search-free}, which means that vectors can be encoded without having to search over a codebook. This is significant apart from the computational benefits, because it allows for on-the-fly quantization of activations and opens the door to quantization at train time. Lastly, the desired bits per weight after quantization can also be fractions (e.g. $b_{\text{direction}} = 3.5$) and are not limited to integers. The ability to choose the groupsize, and bits for the direction make PVQ highly flexible and can be chosen such that the most optimal trade-off between compression and performance is achieved. We find that PVQ outperforms competitive quantization methods in terms of weight-only and activation quantization.

In this work, our focus is on post-training quantization of weights and activations. Some current works quantize LLMs during training \citep[e.g.][]{ma2024era}. We anticipate that PVQ could be used during training because of the advantages outlined above, though we leave a thorough investigation and comparisons to future work. 
% There have been some recent works reporting very low quantization bit rates (e.g. 1.58 in \citet{ma2024era}) \textit{weight-only} quantization. Although this is impressive, these results are limited to the weights and such rates have not been shown for weights and activations. Further, these methods are not always as amenable to fast implementations in low-level code or even in hardware.
% \tycho{can we cite something here?} \james{} \tycho{Do we want to mention issues with `matmul' free \citep{zhu2024scalable}? Why would these methods fail?? and would PVQ be better?). (all in on activations?)}

%Further, the ability to quantize activations together with fast decoding without codebook opens the door to quantizing at training time. We deem this method a very significant contribution in the light of existing efforts towards quantization at training time, beyond the achieved state-of-the-art performance for quantization on both weights and activations.

% \subsection{Implementation}
% \label{sec:implementation}

\begin{wrapfigure}{r}{0.61\textwidth}  % 'r' for right, '0.5\textwidth' for the width
\vspace{-1.5em}
\resizebox{\linewidth}{!}{
\includegraphics[width=0.94\linewidth]{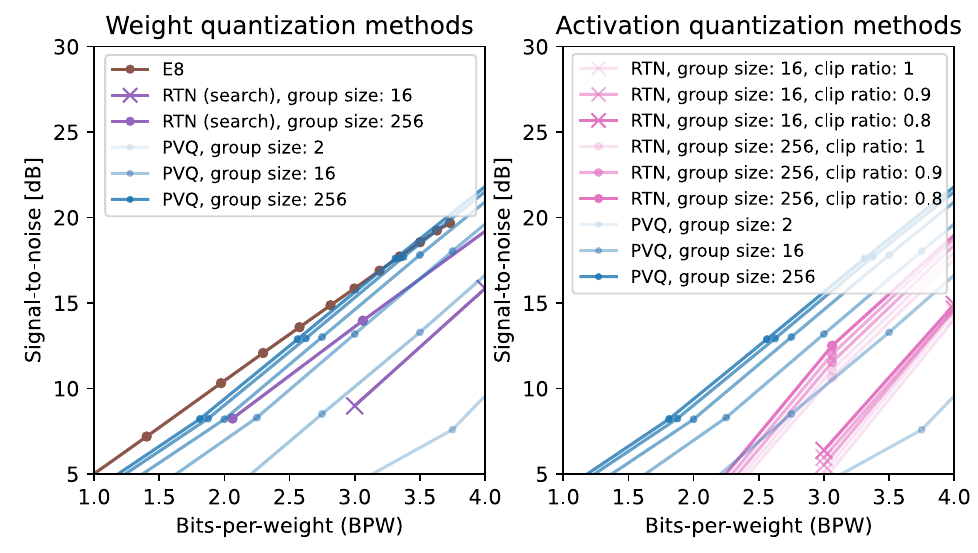}
}
\vspace{-1.5em}
\caption{Signal-to-quantization-noise-ratio (QSNR) of quantization methods on standard Gaussian source. PVQ achieves high QSNR close to E8, which uses an optimal packing on a uniform source. PVQ uses an implicit codebook and is search-free, thereby amenable to quantization of weight and activations.}
\vspace{-1em}
\label{fig:snr-plot}
\end{wrapfigure}

\subsection{Signal-to-quantization-noise on ideal Gaussian source}
\label{sec:theoretical-analysis}

To assess the theoretical effectivity between different quantization algorithms are, we start by comparing empirical estimates of their performance on an idealized standard Gaussian source, zero mean unit variance. We measure the signal-to-quantization-noise ratio QSNR by averaging the mean squared error between true and quantized signal over 1000 samples in \cref{fig:snr-plot}. We compare PVQ with E8, a method that has optimal packing of uniformly distributed weights in $D=8$ dimensions only \citep{tseng2024quip} and naive rounding-to-nearest (RTN) scalar quantization. We find that PVQ achieves QSNR ratios close to the optimal E8 method. Further, PVQ can also be applied to activation quantization, in which case we compare to RTN without search as a baseline that is suitable to quantization of both weights and activations.

% Further advantages of PVQ are discussed in \cref{sec:pvq-method}.

%\begin{figure}[H]
%\vspace{-1em}
%\resizebox{\linewidth}{!}{
%\includegraphics[width=0.44\linewidth]{figures/snr_plot-fixed.pdf}
%}
%\vspace{-2em}
%\caption{Signal-to-noise-ratio of weight and activation quantization methods on a Gaussian source. PVQ achieves similar high signal-to-noise as state-of-the-art E8 weight-only vector quantization. As PVQ is search-free and uses an implicit codebook, it can practically quantize both weight and activations.}
%\vspace{-1em}
%\label{fig:snr-plot}
%\end{figure}

\newpage
%\section{Pyramid vector quantization for LLMs}
\subsection{Pyramid vector quantization for LLMs}
\label{sec:pvq-llm-overview}

This section describes the overall method for quantizing an LLM using PVQ. 

\paragraph{Step 0. Choose the desired BPW.}
\label{sec:bpw}
In PVQ, the trade-off between performance and effective bits per weight (BPW) is controllable through the groupsize $D$, which must divide the number of columns so that $\mW \in \R^{N \times DG}$, direction bits $b_{\text{direction}} \in \mathbb{N}$ and amplitude bits $b_{\text{amplitude}} \in \mathbb{N}$:
\begin{align}
\text{BPW} = b_{\text{direction}}/D + b_{\text{amplitude}}/G \end{align}
As the number of direction bits per weight $b_{\text{direction}}/D$, and amplitude bits per weight $b_{\text{amplitude}}/G$ are fractions, it is easy to choose a non-integer number of bits per weight in PVQ.

\paragraph{Step 1. Coherence processing.} Before we begin quantizing the weights, we perform coherence processing using efficient Hadamard rotation matrices proposed in \citep{chee2024quip}, which can be fused into the architecture as described in \quarot \citep{ashkboos2024quarot}.

\paragraph{Step 2. Determine the number of pulses $K$.}
To determine the number of pulses $K$ such that the number of bits required to encode PVQ vectors remains within the desired maximum, we find the largest $K$ by increasing it such that it still satisfies $\left\lceil \log_2(N(D, K))/D \right\rceil < b_{\text{direction}}$. Here, $N(D, K)$ is computed using the recursive algorithm outlined in Section \ref{sec:pvq-background}. In practice, the number $N(D, K)$ can exceed regular integer types and may require arbitrary precision integers.

\paragraph{Step 3. Quantizing the direction.}

We quantize all LLM weight matrices making up key, query, value and fully-connected components. We write the normalized weight matrix $\mW = \begin{bmatrix} \mW_1 &\mW_2 &\ldots &\mW_G \end{bmatrix} \in \R^{N \times GD}$, with $N$ features, $G$ groups and a groupsize of $D$. Given our choice of $K$, we quantize groups $\mW_g \in \mathbb{R}^{N \times D}$ using the quantization procedure of \cref{sec:pvq-background} yielding a quantized direction matrix $\widehat{\mW}_g$ on the pyramid $\gP_{D, K}$, ie. elements are rounded integers and absolute values of rows sum to 1. This operation can be parallelized over features and groups.
%The quantization is highly parallelisable over both features and groups, and can be implemented in a dedicated CUDA/C++ kernel.

\paragraph{Step 4. Computing the amplitude.}

For each quantized row vector $\widehat{\vw} \in \R^D$ in a group, we can find an optimal rescaling by $s \in \R$ that minimizes the Euclidean distance to the original weight $\vw$ in closed-form $ s = \widehat{\vw}^T\widehat{\vw} / \widehat{\vw}^T \vw$. Repeating this for all features and groups yields an amplitude matrix $\mS \in \R^{N \times G}$.
%Thus, we can compute the optimal scale for all row vectors of each groups, to find a vector of scales $\vs_g \in \R^N$, such that $\mW_g \approx \vs_g^T \widehat{\mW}_g$, and repeated for all groups results in a scale matrix $\mS \in \R^{N \times G}$.
%The operation can be parallelized over features and groups. %We have that the full weight matrix $\mW \approx \widehat{\mW} \odot \mS'$, by broadcasting the scales $\mS' = \begin{bmatrix} \mS & \mS \odots & \mS \end{bmatrix} \times \R^{N \times GD}$.
We refer to quantized $\widehat{\mW}$ as the `\textit{direction}' and the scales $\mS$ as the `\textit{amplitude}'.

\paragraph{Step 5. Quantizing the amplitude (optional)}

For small groupsizes, we can optionally quantize amplitudes as described in \cref{sec:amplitude-quantization}. For each row $\vs \in \R^G$ in $\mS$, this entails quantizing normalized elements $\floor{\text{CDF}(s_i^2 / || \vs ||_2^2) \cdot 2^b}$ using the CDF of the $\text{Beta}(D/2, D(G-1)/2)$ distribution. The normalizing constant $||\vs||_2^2$ needs to be stored for dequantization, but because it is shared across groups its contribution to the total $<0.01$ bits per weight is negligibly small.

\paragraph{Step 6. Correcting the quantization error.}

For each quantized group of weights indexed by the quantized columns $\vs_g \widehat{\mW}_g \in \R^{N \times D}$, we can update the other remaining columns $\mW_{\neg g} \in \R^{N \times (G-1)D}$ to compensate the quantization error that minimizes the proxy loss $L$ of \cref{eq:proxy-loss}:
\begin{align}
\mW_{\neg g} \leftarrow \mW_{\neg g} - \mH^{-1}_{\neg g, \neg g} \mH_{\neg g, g} (\mW_g - \vs_g \widehat{\mW}_g)
\end{align}
where $\mH_{\neg g, \neg g}$ is a square submatrix of the inverse Hessian with rows and columns corresponding to remaining weights and $\mH^{-1}$ is a rectangular submatrix of the inverse Hessian with rows and columns that correspond to quantized and remaining weights respectively. We can avoid inverting the full Hessian for every update by working with its Cholesky decomposition $\mH = \mL \mL^T$, as detailed in GPTQ \citep{frantar2022gptq} which proposed the update in the context of scalar quantization.

\paragraph{Step 7. Encoding the direction.}

To encode a quantized weight $\widehat{\mW} \in \R^{N \times GD}$ (e.g. 32 bit float for $\R$) into the actual low bit integer representation $\overline{\mW} \in \mathbb{N}^{N \times G}$, sometimes referred to as the `code', we can use the PVQ encoding algorithm described in \cref{alg:encode}. This is problematic, since the number of bits required to store a vector may exceed the bit width of integer types in many languages (e.g. when $\ceil{\log_2(N(D, K))} > 128 $ bits, it can not be represented in a 128 bit integer). To overcome this, we implemented arbitrary precision arithmetic in CUDA/C++ to support arbitrary bit width integer types to allow encoding and decoding kernels.

\newpage
\subsection{amplitude quantization}
\label{sec:amplitude-quantization}

\begin{figure}[t]
    \vspace{-2em}
    \includegraphics[width=1.0\linewidth]{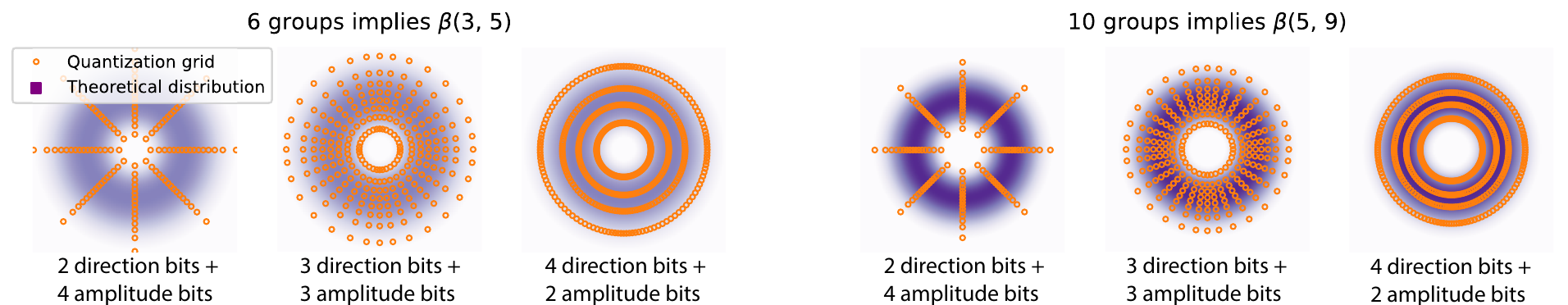}
    \vspace{-1.5em}
    \caption{Effect of different direction and amplitude bits on effective PVQ quantization grid. The distribution automatically matches the theoretical $\text{Beta}$ weight distribution.}
\label{fig:direction-amplitude-plot}
    \vspace{-1em}
\end{figure}

\paragraph{Normalized amplitudes follow the Beta distribution.}
Smaller groupsizes lead to a lower quantization error, but lead to an increase in the number of bits required to store amplitude parameters. This is because the amplitude $\mS \in \R^{N \times G}$ grows linearly with the number of groups $G$, and therefore inversely proportional to the chosen groupsize $D$. To overcome this issue, we propose a theoretically and empirically motivated scheme to quantize amplitude parameters. In \cref{thm:beta-dist}, we note that row vectors of normalized amplitudes follow a Beta distribution $\text{Beta}(\frac{D}{2}, \frac{D(G-1)}{2})$ of which the coefficients depend on the groupsize $D$ and the number of groups $G$. We empirically confirm that our theory meets practice, as we find that the Beta distribution matches normalized weight distributions of pretrained LLM models, after performing the rotation described in \cref{sec:weights-on-sphere}, as shown for selection of layers in a pretrained LLM in \cref{fig:direction-amplitude-plot}. To exploit the observation that amplitudes are Beta distributed, we propose to use the quantiles of this distribution to quantize amplitudes.

\begin{theorem}
\label{thm:beta-dist}
Let $\vw \in \R^{GD}$ be a normally distributed vector that can be grouped in $G$ equally sized vectors $\vw = \begin{bmatrix} \vv_1 & \vv_2 & \cdots & \vv_G \end{bmatrix}$ where each of the vectors $\vv_g$ has the same dimensionality equal to the groupsize $\vv_g \in \R^D$. Then the normalized radius (the `amplitude') of each group $s_g = \vv_g^T \vv_g/||\vw||_2^2$ follows the $s_g \sim \text{Beta}(\frac{D}{2}, \frac{G(D-1)}{2})$ distribution. (Proof in \cref{sec:beta-proof})
\end{theorem}

% Even though \cref{thm:beta-dist} relies on the assumption that weights are normally distributed, which may not necessarily be the case, we find that spherical weights closely match our theoretical distribution. We verified this by inspecting the empirical weight distributions in LLMs after applying the rotation described in \cref{sec:reparameterising-on-sphere}. We find empirically that normalized weights indeed very closely match the Beta distribution across layers, as can be seen in \cref{fig:hist-example} (the black line closely matches the blue distribution). We find that this observation holds robustly across layers, and we provide histograms for all LLM layers in \cref{sec:additional-amplitude-histograms}.
% \begin{figure}[h]
%     \begin{minipage}{0.6\textwidth}
%         \centering
%         \includegraphics[width=\linewidth]{figures/hist-example.png}
%         \caption{The theoretical Beta distribution of \cref{sec:amplitude-quantization} closely matches the empirical distribution over amplitudes after weight matrices have been rotated.}
%         \label{fig:hist-example}
%     \end{minipage}\hfill
%     \begin{minipage}{0.4\textwidth}
%         \centering
%         \includegraphics[width=0.75\linewidth]{figures/quantile_fig.pdf}
%         \hspace{-2em}
%         \caption{Quantiles of Beta(2, 6).}
%         \label{fig:beta-cdf}
%     \end{minipage}
% \end{figure}

\begin{figure}[h]
 \centering
 %\vspace{-1.5em}
 \includegraphics[width=\linewidth]{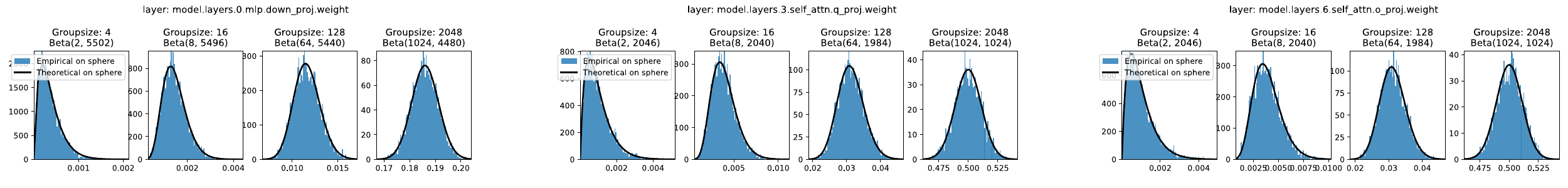}
 \vspace{-1.0em}
 \caption{Theoretical Beta distribution of \cref{thm:beta-dist} closely match amplitudes of rotated weights in trained LLMs, here demonstrated for empirical weight distributions of a pretrained Llama-v3 8B.}
 \vspace{-0.5em}
 \label{fig:hist-example}
\end{figure}

% \begin{figure}[h]
%  \centering
%  \vspace{-1em}
%  \includegraphics[width=\linewidth]{figures/histograms.pdf}
%  \vspace{-1.5em}
%  \caption{Theoretical Beta distribution of \cref{thm:beta-dist} closely match amplitudes of rotated weights in trained LLMs. Here demonstrated for empirical weight distributions of a pretrained Llama-v3 8B.}
%  \vspace{-0.5em}
%  \label{fig:hist-example}
% \end{figure}

\begin{wrapfigure}{r}{0.37\textwidth}
\vspace{-1.5em}
\hspace{0.5em}
  \includegraphics[width=0.9\linewidth]{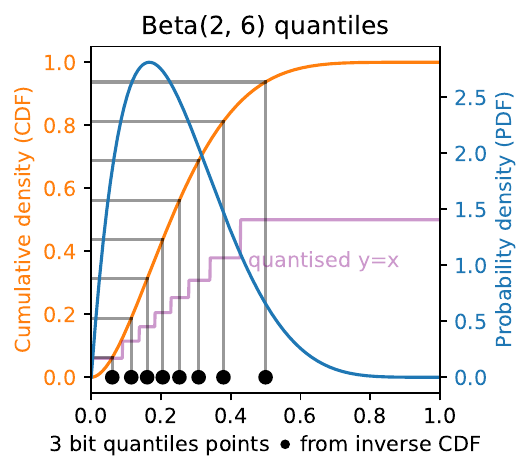}
\vspace{-0.5em}
  \caption{Quantiles of Beta(2, 6).}
\vspace{-1.0em}
  \label{fig:beta-cdf}
\end{wrapfigure}

\paragraph{Quantizing amplitudes using Beta quantiles.}
The observation that normalized amplitudes are Beta distributed is important, as it suggests that we can quantize amplitudes efficiently by mapping centers of linearly spaced regions through the quantile function of a Beta distribution, without introducing additional hyper-parameters. To obtain a $b$ bit quantizer, we take a regular uniform grid of $2^b$ points after first transforming elements through a change-of-variables given by the CDF of the Beta distribution and dequantizing using the inverse CDF (see \cref{fig:beta-cdf}):
\begin{align}
\text{quantize}(x) &= \floor{\text{CDF}(x) \cdot 2^b} \\
\text{dequantize}(x) &= \text{PPF}((x + 0.5) / 2^b)
\end{align}
where $\floor{\cdot}$ denotes the floor function, $\text{CDF}(\cdot)$ the cumulative density function of the Beta distribution and its inverse $\text{PPF}(\cdot)$, known as the percentile point function of the Beta distribution.

%Consider weight matrix $\mW = \begin{bmatrix}\mB_1 & \mB_2 & \cdots & \mB_B \end{bmatrix} \in \R^{N \times X}$ equally split up into groups, such that each group $b \in [1, B]$ has a size of $\mB_b \in \R^{N \times G}$ with group size $G = X / B$. We normalise the row vector in each group 

% We can write vectors $\mB = \vs^T \mX = \vs^T \text{decode}(\bar{\mX})$, where $\bar{\mX} \in [0, \text{bits}]^{N \times X}$ where $\text{decode}(\cdot)$ maps from compressed $[0, \text{bits}]^{N \times X}$ to the hyper-pyramoidal lattice in $\R^{0, 1}$.

% We may denote normalise the weights by a scalar $\mV = \mW / ||\mW||^2 \in R^{N \times X}$

% \begin{algorithm}
% \caption{Pyramid vector quantization for LLMs}\label{alg:cap}
% \begin{algorithmic}
% \Require $n \geq 0$
% \Ensure $y = x^n$
% \State $y \gets 1$
% \State $X \gets x$
% \State $N \gets n$
% \While{$N \neq 0$}
% \If{$N$ is even}
%     \State $X \gets X \times X$
%     \State $N \gets \frac{N}{2}$  \Comment{This is a comment}
% \ElsIf{$N$ is odd}
%     \State $y \gets y \times X$
%     \State $N \gets N - 1$
% \EndIf
% \EndWhile
% \end{algorithmic}
% \end{algorithm}

% \subsection{Hardware acceleration}
% 
% The PVQ algorithm allows for efficient encoding and decoding of weights without having to store the associated codebook explicitly in memory. To fully leverage the savings in bits per width in practice, we need this encoding and decoding to be implemented in hardware. TODO: explore kernel acceleration?

\newpage
\section{Results}
\label{sec:results}

\subsection{Weight-only quantization (without amplitude quantization)}
\label{sec:weight-only}

Since most prior work focuses on weight-only quantization, we begin by comparing the performance of weight-only PVQ quantization with common weight-only quantization baselines. Following prior work \citep{frantar2022gptq,ashkboos2024quarot}, we measure test perplexity (PPL) and average accuracy on a range of zero-shot downstream tasks after quantizing common open-source LLM models of Phi, Mixtral and Llama families. We compare groupsizes in [16, 32, 64, 128, 256], keep the amplitude in 16 bit, and use direction bits in [3, 3.5, 4, 4.5, 5, 5.5, 6, 7, 8] for PVQ and [3, 4, 5, 6, 7, 8] for other methods. Unlike most scalar quantization methods, PVQ more easily supports non-integer number of direction bits as we can choose an integer number of bits per codeword that is not divisible by the number of groups $G$ -- for instance, with groupsize $D=16$ and 40 bits per group results in $40/16=2.5$ direction bits).

\begin{figure}[h]
\vspace{-0.5em}
\resizebox{\linewidth}{!}{
\includegraphics[width=1.15\linewidth]{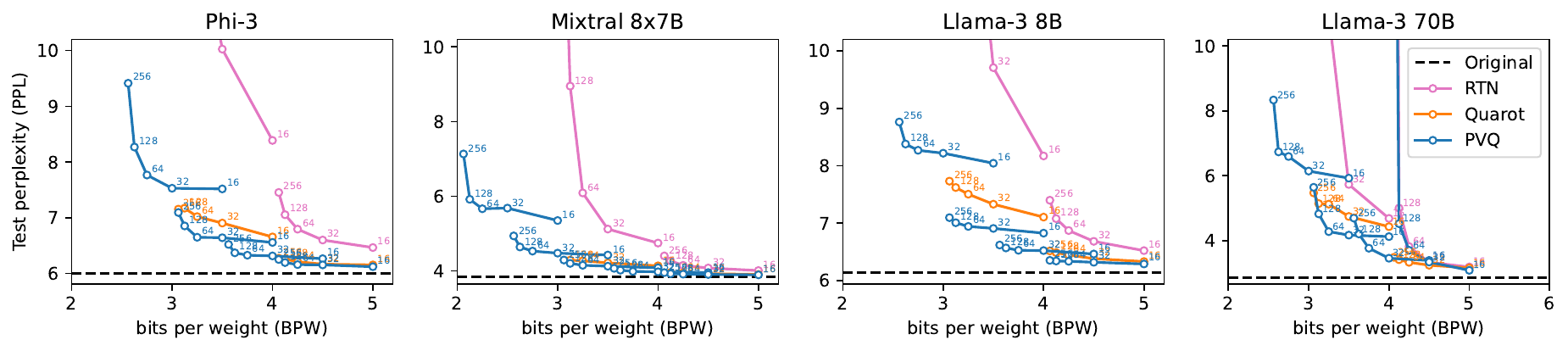}
}
\vspace{-1.5em}
\caption{Weight only quantization. Test perplexities (PPL) after quantizing with different methods at group size settings (within connected set) and various direction bits (between connected sets).}
\vspace{-0.5em}
\label{fig:weight-only}
\end{figure}

\begin{table}[h!]
\resizebox{\linewidth}{!}{
%\begin{tabular}{l r c c c c | cc | cc | cc cc cc}
\begin{tabular}{l r c c c c | cc | cc | cc cc }
& & & & & 
& \multicolumn{2}{c}{Phi-3-mini-4k}
& \multicolumn{2}{c}{Mixtral-8x7B} 
& \multicolumn{2}{c}{Llama-3-8B} 
& \multicolumn{2}{c}{Llama-3-70B}
%& \multicolumn{2}{c}{Llama-3.1-405B} 
\\
Method & & Hessian & Spherical & Groupsize & BPW &
PPL ($\downarrow$) & Avg. Acc ($\uparrow$) &
PPL ($\downarrow$) & Avg. Acc ($\uparrow$) &
PPL ($\downarrow$) & Avg. Acc ($\uparrow$) &
PPL ($\downarrow$) & Avg. Acc ($\uparrow$)
% & PPL ($\downarrow$) & Avg. Acc ($\uparrow$) 
\\
\hline
Original & & & & & 16 & 6.01 & 0.72 & 3.84 & 0.78 & 6.13 & 0.73 & 2.85 & 0.80
%& -1.00 & -1.00
\\
RTN & & & & 128 & 3.125 & 19.03 & 0.53 & 8.95 & 0.66 & 29.41 & 0.41 & 487.94 & 0.45
%& -1.00 & -1.00
\\
GPTQ & & \checkmark & & 128 & 3.125 & 7.36 & 0.65 & 8.40 & 0.52 & 17.77 & 0.40 & nan$^{*}$ & nan$^{*}$
%& -1.00 & -1.00
\\
\quarot & & \checkmark & \checkmark & 128 & 3.125 & 7.17 & 0.67 & 4.29 & 0.76 & 7.62 & 0.69 & 5.14  & 0.76
% & -1.00 & -1.00
\\
\hline PVQ & & \checkmark & \checkmark & 128 & 3.125 & \textbf{6.85} & \textbf{0.68} & \textbf{4.20} & \textbf{0.77} & \textbf{7.01} & \textbf{0.72} & \textbf{4.82} & \textbf{0.77}
% & \textbf{-1.00} & \textbf{-1.00}
\\
\hline
\end{tabular}
}
\vspace{-0.0em}
%\caption{Weight-only quantization performance in 4 bits. We compare perplexity (PPL) and average zero-shot (Avg. Acc) performance after quantizing different open source LLM models using various post-training quantization methods. PVQ yields the highest performance after quantization.}
\caption{Weight-only quantization in sub-4 bits. Post-quantization test perplexity (PPL) and average zero-shot (Avg. Acc) performance. PVQ yields the highest performance after quantization. Details and additional results in \cref{sec:additional-weight-only}. nan$^{*}$ indicates GPTQ fails due to non-psd Hessian.}
\vspace{-0em}
\label{tab:weight-only-acc}
\end{table}

\subsection{Weight-only quantization (direction and amplitude quantization)}
\label{sec:results-direction-and-amplitude}

A benefit of PVQ is that the number of bits for the direction and bits for the amplitudes can be chosen flexibly, even to not-integer ratios. To evaluate the effect of different amplitude bits, we repeat the same experiment as before, but rather than varying the groupsize, we fix the groupsize to 16 and vary the amount of bits used for the amplitude. 

\begin{figure}[h]
% \vspace{-2em}
%\resizebox{\linewidth}{!}{
%\includegraphics[width=1.0\linewidth]{figures/gains-plot-ppl.pdf}
%}
%\vspace{-2.5em}
%\resizebox{\linewidth}{!}{
%\includegraphics[width=1.0\linewidth]{figures/gains-plot-acc.pdf}
%}
\resizebox{\linewidth}{!}{
\includegraphics[width=1.0\linewidth]{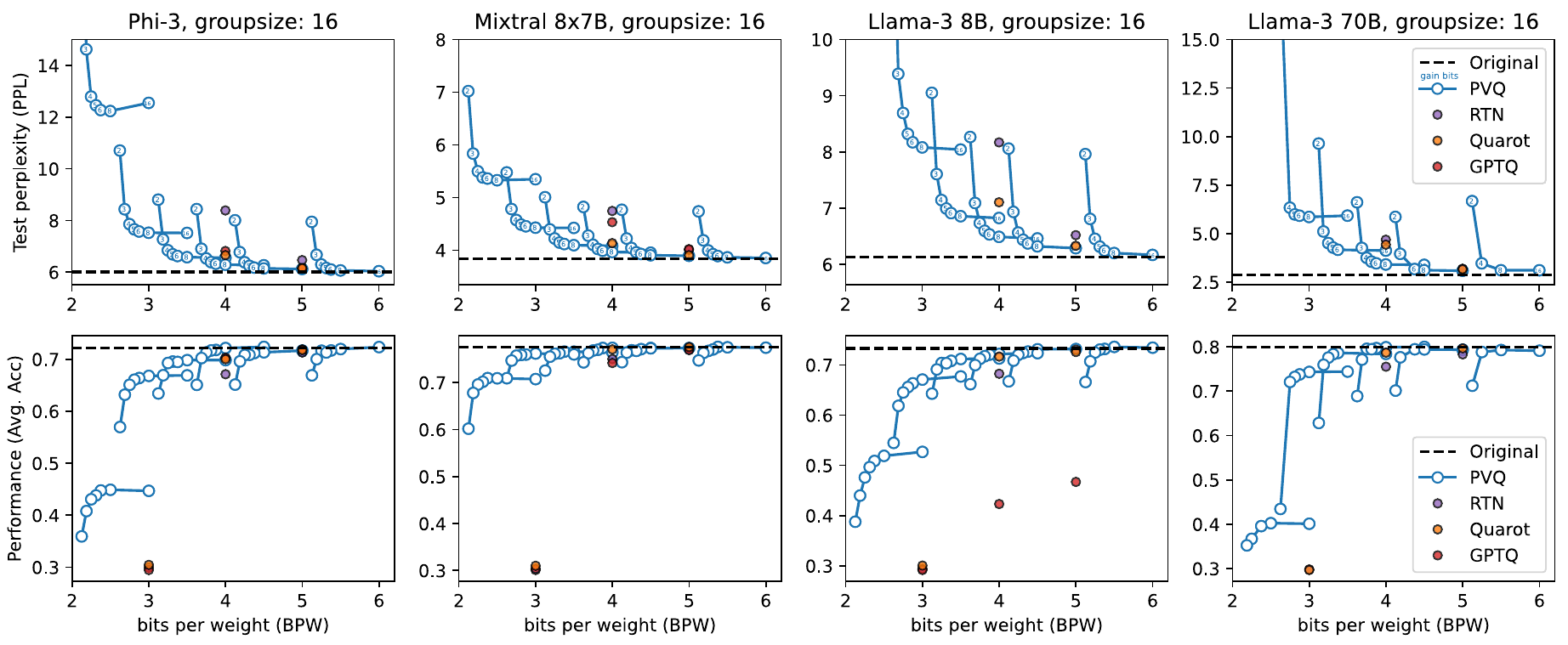}
}
\vspace{-1.5em}
%\caption{direction and amplitude quantization. Test perplexities (PPL) and test accuracies (Avg.Acc.) at different bits for amplitudes (within connected sets) and various direction bits (between sets).}
\caption{direction and amplitude quantization. Test perplexities (PPL) and test accuracies (Avg. Acc
) at different bits for amplitudes (within connected sets) and various direction bits (between sets).}
\label{fig:gains-plot-ppl-and-acc}
\end{figure}
% 
% \begin{figure}[h]
% \resizebox{\linewidth}{!}{
% \includegraphics[width=1.0\linewidth]{figures/gains-plot-acc.pdf}
% }
% \vspace{-2em}
% \caption{direction and amplitude quantization. Average accuracies (Avg. Acc.) at different bits for amplitudes (within connected set) and various direction bits (between connected sets).}
% \vspace{-1em}
% \label{fig:gains-plot-acc}
% \end{figure}

\begin{table}[h!]
\vspace{-1em}
\resizebox{\linewidth}{!}{
%\begin{tabular}{l r c c c c | cc | cc | cc cc cc}
\begin{tabular}{l r c c c c | cc | cc | cc cc }
& & & & &  
& \multicolumn{2}{c}{Phi-3-mini-4k}
& \multicolumn{2}{c}{Mixtral-8x7B} 
& \multicolumn{2}{c}{Llama-3-8B} 
& \multicolumn{2}{c}{Llama-3-70B}
%& \multicolumn{2}{c}{Llama-3.1-405B} 
\\
Method & & Groupsize & Hessian & Spherical & BPW &
PPL ($\downarrow$) & Avg. Acc ($\uparrow$) &
PPL ($\downarrow$) & Avg. Acc ($\uparrow$) &
PPL ($\downarrow$) & Avg. Acc ($\uparrow$) &
PPL ($\downarrow$) & Avg. Acc ($\uparrow$)
% & PPL ($\downarrow$) & Avg. Acc ($\uparrow$) 
\\
\hline
Original & & & & & 16.00 & 6.01 & 0.72 & 3.84 & 0.78 & 6.13 & 0.73 & 2.85 & 0.80 % & -1.00 & -1.00
\\
\hline PVQ [2.5 bit directions, 16 bit amplitudes] & & 16 & \checkmark & \checkmark & 3.50 & \textbf{7.52} & \textbf{0.67} & \textbf{4.42} & \textbf{0.76} & \textbf{8.04} & \textbf{0.68} & \textbf{5.92} & \textbf{0.74}
%& \textbf{-1.00} & \textbf{-1.00}
\\
PVQ [3 bit directions, 4 bit amplitudes] & & 16 & \checkmark & \checkmark & 3.25 & \textbf{6.85} & \textbf{0.69} & \textbf{4.22} & \textbf{0.76} & \textbf{7.14} & \textbf{0.70} & \textbf{4.51} & \textbf{0.78}
% & \textbf{-1.00} & \textbf{-1.00}
\\
\hline
\end{tabular}
}
\vspace{-0em}
%\caption{Quantizing direction and amplitude. We compare perplexity (PPL) and average zero-shot performance (Avg. Acc) after quantizing different open source LLM models using various post-training quantization methods. PVQ yields the highest performance after quantization.}
\caption{Quantizing direction and amplitude. We compare post-training perplexity (PPL) and average zero-shot performance (Avg. Acc). PVQ yields the highest performance after quantization. Details and additional results in \cref{sec:additional-direction-and-amplitude}.}
\vspace{-0em}
\label{tab:direction-and-amplitude}
\end{table}

\newpage
\subsection{Weights and Activation quantization}
\label{sec:weight-and-acts}

As PVQ use an implicit codebook and is search-free, it can
be applied to not only the weights but also the activations during inference and reduce computational requirements of the forward-pass. In \cref{fig:activations-plot} we evaluate the final test perplexity for different settings of effective bits per weight and bits per activations. In \cref{tab:weight-and-acts}, we compare the resulting test perplexities to other methods for quantizing weight and activation. We also compare to naive round-to-nearest (RTN) without search, which is also amenable to quantization of weights and activations but has much lower signal to noise. We find that PVQ obtains state-of-the-art performance across the considered LLM architectures, and that this holds generally across different settings of bit rates.

\begin{figure}[H]
\vspace{-1.2em}
\resizebox{\linewidth}{!}{
\includegraphics[width=1.15\linewidth]{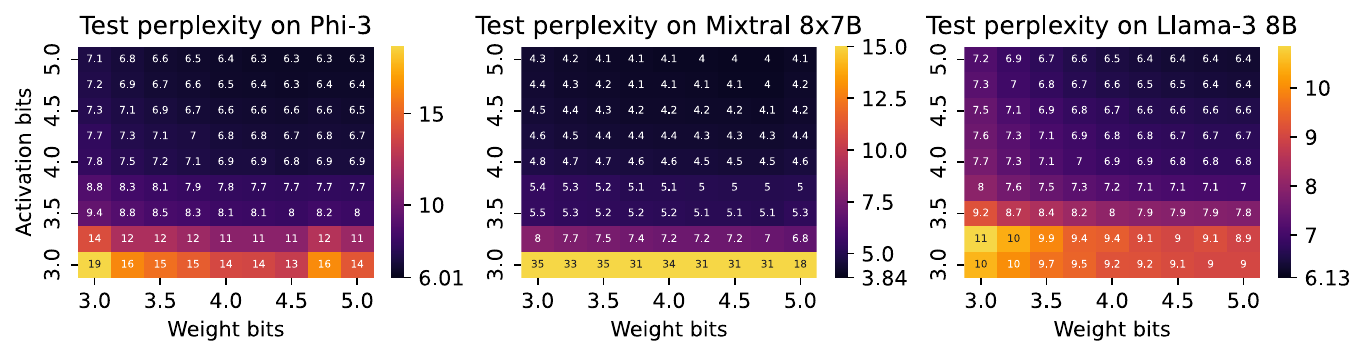}
}
\vspace{-2.0em}
\caption{Weights and activations. Comparing test perplexity at different bits per weight and bits per activations. From minimal compression (top right) to high levels of compression (bottom left).}
\vspace{-1em}
\label{fig:activations-plot}
\end{figure}

\begin{table}[h!]
\resizebox{\linewidth}{!}{
%\begin{tabular}{l l r c c c c c | cc |cc | cc cc cc}
%\begin{tabular}{l l r c c c c c | cc |cc | cc cc }
\begin{tabular}{l l r c c c c c | cc |cc | cc  }
\multicolumn{2}{c}{Method} & & & & & &
& \multicolumn{2}{c}{Phi-3-mini-4k}
& \multicolumn{2}{c}{Mixtral-8x7B} 
& \multicolumn{2}{c}{Llama-3-8B} 
%& \multicolumn{2}{c}{Llama-3-70B}
%& \multicolumn{2}{c}{Llama-3.1-405B} 
\\
Weights & Activations & & Groupsize & Hessian & Spherical & BPW & BPA &
PPL ($\downarrow$) & Avg. Acc ($\uparrow$) &
PPL ($\downarrow$) & Avg. Acc ($\uparrow$) &
PPL ($\downarrow$) & Avg. Acc ($\uparrow$)
%& PPL ($\downarrow$) & Avg. Acc ($\uparrow$)
%& PPL ($\downarrow$) & Avg. Acc ($\uparrow$) 
\\
\hline
Original & - & & & & & 16 & 16 & 6.01 & 0.72 & 3.84 & 0.78 & 6.13 & 0.73 
%& 2.85 & 0.80
%& -1.00 & -1.00
\\
%RTN & RTN & & 128 & & & -1 & -1 & -1.00 & -1.00 & -1.00 & -1.00 & -1.00 & -1.00 & -1.00 & -1.00 & -1.00 & -1.00 \\
GPTQ & RTN & & 128 & \checkmark & & 4.125 & 4.125 & 8.36 & 0.63 & 7.41 & 0.59 & 4747.68 & 0.36
%& &
%& -1.00 & -1.00
\\
\quarot & RTN & & 128 & \checkmark & \checkmark & 4.125 & 4.125 & 7.48 & 0.67 & 4.43 & 0.75 & 7.34 & 0.70
%& -1.00 & -1.00
%& -1.00 & -1.00
\\
\hline PVQ & RTN & & 128 & \checkmark & \checkmark & 4.125 & 4.125 & 7.37 & 0.66 & \textbf{4.40} & \textbf{0.76} & 7.16 & 0.70
%&  &  
% & \textbf{-1.00} & \textbf{-1.00}
\\
PVQ & PVQ & & 128 & \checkmark & \checkmark & 4.125 & 4.125 & \textbf{6.94} & \textbf{0.68} & 4.57 & 0.75 & \textbf{6.89} & \textbf{0.71}
%& \textbf{nan} & \textbf{0.29} 
% & \textbf{-1.00} & \textbf{-1.00}
\\
\hline
\end{tabular}
}
\vspace{-0em}
\caption{Quantizing weight and activation in 4 bits. We compare perplexity (PPL) and average zero-shot performance (Avg. Acc) after quantizing different open source LLM models using various post-training quantization methods. PVQ yields the highest performance after quantization. Additional results in \cref{sec:additional-weights-and-activations}.}
\vspace{-1em}
\label{tab:weight-and-acts}
\end{table}

% \subsection{Use bits for direction or amplitude?}
% \label{sec:performance-analysis}
% 
% The PVQ algorithm has configurable amount of bits that can be used for direction and amplitude, introducing a trade-off. We empirically explore this trade-off by ...

% \paragraph{Empirical amplitudes are beta distributed} The amplitude quantization scheme proposed in \cref{sec:amplitude-quantization} relies on the theoretical realisation that amplitudes after rotation follow a particular Beta distribution. This theoretical distribution assumes that rotated weights are normally distributed, which we find to hold very well in practice. To assess how well our theoretical Beta distribution matches amplitudes of different groupsizes, we compare the empirical weight distributions of pre-trained neural networks, both before and after applying the rotations of \cref{sec:reparameterising-on-sphere} For each weight matrix $\mW$ in the network, we compute amplitudes $\vv_1^T \vv_1/||\vw||_2^2$ of the first group $\vv_1 \in \R^G$ in $\vw = [\vv_1 \vv_2 \ldots \vv_K]$. We repeat this for each row vector $\vw$ in $\mW$, plot the resulting amplitudes in a histogram, and report full results in \cref{sec:additional-amplitude-histograms}. We find that the theoretical distribution always closely matches the empirical distribution closely. In cases where rotation caused a shift in amplitude histogram, the theoretical distribution remains a particularly good match, as can be observed in \cref{fig:hist-example}.

\subsection{Downstream zero-shot tasks} 

In \cref{tab:downstream-llama3}, we provide results on downstream zero-shot tasks split out per task. We report weight-only PVQ with both direction and amplitude quantization on a Llama-3-8B model. Additional results including other LLM models can be found in \cref{sec:additional-downstream-tasks}.

\begin{table}[h!]
\vspace{-0.0em}
\resizebox{\linewidth}{!}{
\begin{tabular}{l r c c c c | c | cccccc | c}
& & & & &  
& \multicolumn{7}{c}{Llama-3-8B} \\
Method & & Groupsize & Hessian & Spherical & BPW &
PPL $\downarrow$ &
PQ $\uparrow$ &
WG $\uparrow$ &
HS $\uparrow$ &
A-e $\uparrow$ &
A-c $\uparrow$ &
LA $\uparrow$ &
Avg. $\uparrow$ \\
\hline
Original & & & & & 16.000 & 6.13 & 0.81 & 0.73 & 0.79 & 0.78 & 0.53 & 0.76 & 0.73 \\
\hline RTN & & & & & 3.125 & 29.41 & 0.64 & 0.55 & 0.42 & 0.41 & 0.25 & 0.22 & 0.41 \\
GPTQ & & 128 & \checkmark & & 3.125 & 17.77 & 0.63 & 0.59 & 0.35 & 0.43 & 0.26 & 0.17 & 0.40 \\
\quarot & & 128 & \checkmark & \checkmark & 3.125 & 7.62 & 0.77 & 0.71 & 0.73 & 0.75 & 0.46 & 0.71 & 0.69 \\
\hline PVQ [3 bit directions, 16 bit amplitudes] & & 128 & \checkmark & \checkmark & 3.125 & 7.01 & 0.80 & 0.73 & 0.76 & 0.78 & 0.50 & 0.75 & 0.72 \\
\hline
\end{tabular}
}
\vspace{-0.0em}
\caption{Performance on downstream tasks. We compare performance on zero-shot downstream tasks after quantizing weights using different weight quantization methods.}
\label{tab:downstream-llama3}
\vspace{-0em}
\end{table}

\newpage
\subsection{Optimizing PVQ for CUDA}
\label{sec:results:optimizing_pvq_for_cuda}
For large $D$ and $K$, PVQ results in large precision integer codes $c$, far surpassing the native 32-bit integer operations on CUDA. As highlighted in \cref{sec:pvq-llm-overview}, even 128-bit operations introduced with CUDA 11.5 are not sufficient. We therefore implement the CUDA kernels for PVQ on custom subroutines for arbitrary precision integer arithmetic relying on PTX instructions using the \textit{carry-forward} registry (\texttt{CC.CF}) for multi-word integer addition, subtraction and bit-shifting. We use a word-minor memory layout to ensure that the memory access can be coalesced. \autoref{tab:cuda_complexity} presents the time and I/O complexity for our implementations.

For quantization, encoding and decoding we parallelize the work across the batch dimension. Further optimizations are possible, especially for small batch sizes $B$. In particular, the inner loops of each algorithm (\cref{sec:pseudocode}) can be further parallelized by first sorting or computing the cumulative sum of $|x_i|$ respectively using a CUDA optimized reduction \citep{harris2007gems}.

The recurrent formulation for the size table ($N\left[D, K\right]$) quickly becomes prohibitive as its naive implementation requires $O(D \cdot K \cdot G)$ serial operations, each requiring overlapping reads and writes. Instead, we reformulate the algorithm to perform $K$ operations in parallel using a CUDA optimized Hillis-Steele type scan reduction~\citep{hillis1986data} which accounts for the $\log(K)$ factor in \autoref{tab:cuda_complexity}, computing each row in parallel (\cref{appendix:implementation:cuda}). This has the further advantage of only having to read from global memory for synchronization following \citep{xiao2010inter}, otherwise the threads keep the values in registry and exclusively writes to global memory.

\begin{table}[h]
\begin{tabular}{lll}
\textbf{Function}     & \textbf{Time complexity}          & \textbf{I/O complexity}                                    \\ \hline
Quantization          & $O(B \cdot D)$                   & $O(B \cdot D)$                                                          \\
Encoding / Decoding   & $O(B \cdot D \cdot G)$           & $O(B \cdot D \cdot G)$ reads, $O(B \cdot D)$ ordered writes \\
Size table generation & $O(K \cdot \log(K) \cdot D \cdot G)$ & $O(K \cdot D \cdot G)$ ordered writes                     
\end{tabular}
\caption{\label{tab:cuda_complexity} The time and I/O complexity on the HBM for our CUDA optimized PVQ kernels, where $B$ is the batch size and $G$ the number of words comprising each integer. All functions have $O(G)$ space complexity. We highlight ordered I/O operations as this more efficient use of the memory bandwidth as they can be coalesced into fewer operations as opposed to random ones.}
\end{table}
% Outline: Agrin
% -------------
% . quantization
% . big-integer operations: PTX + CC.CF for addition, subtraction, bitshift and cumsum
% . encoding and decoding: readily parallelizable across the batch dimension, trick allows even faster encoding.
% . size table: reformulate reccurrent relationship, exploit warp shuffles, shared memory across warps, quote block synchronization through global memory from Inter-Block GPU Communication via Fast Barrier Synchronization
% . IO and runtime complexity table:
% .. quantization ??
% .. size table 0 writes, O(D * K) coalesced writes, O(K * log(K) * N) compute
% .. encoding / decoding: O(N * D * G) random reads, O(N * G) writes, O(N * D * G) compute. Fast version of encode does O(N * D * log(D) * G) compute.
% Appendix:
%
% Pseudocode for scan table based on:
% def build_scan_table(D, K):
%    table = torch.zeros(D, K, dtype=torch.int64)
%    if D <= 0:
%        return table
%
%    # Need to handle the first row separately
%    row = torch.full((K,), 2, dtype=torch.int64)
%    table[0, :] = row
%    if D <= 1:
%        return table
%
%    for i in range(1, D):
%        row[1:] += row[:-1].clone()  # Diagonal relationships
%        row[0] = 2 * (i + 1)
%        row = row.cumsum(0)
%        table[i, :] = row
%    return table
% Pseudocode for fast encoding
% - compute cumsum(|x_i|)
% - compute cumsum(N[., k])
% - compute each inner loop independently

\section{Conclusion}

This work explored pyramid vector quantization (PVQ) for quantization of weights and activations in large language models (LLMs). PVQ is a vector quantization method that allows high signal-to-noise ratios without having to build an explicit codebook or perform search. This results in state-of-the-art quantization performance in terms of the most favourable performance to bits-per-weight trade-off, and is amenable to quantization of activation. This has direct practical benefit for post-training model compression, but also opens the door towards quantization at train time. We propose to quantize LLMs using an implicit PVQ codebook on the unit sphere, which can be flexibly configured for codesize and dimensions. In addition, we propose a theoretically and empirically motivated way to also quantize amplitudes enabling small groupsizes in practice. Lastly, we incorporate Hessian information throughout the process to minimize feature error due to quantization. This yields a novel and highly parallelisable algorithm for LLM weights and activations. We demonstrate state-of-the-art quantization performance in terms of superior performance after quantizing pre-trained models, on both weight-only and weight and activation quantization. 

\newpage
\bibliography{iclr2024_conference}
\bibliographystyle{iclr2024_conference}

\appendix

\newpage
\section{Mathematical details}

%\subsection{Hadamard matrices}
%\label{sec:hadamard-background}

\subsection{Proof that amplitudes are Beta distributed}
\label{sec:beta-proof}

Assume we have weights ( $w_1, w_2, \ldots, w_N $) that are normally distributed: $ w_i \sim \mathcal{N}(0, \sigma^2) $. The sum of the squares of these weights for a subset of size ( D ) is chi-squared distributed $ u = w_1^2 + w_2^2 + \ldots + w_G^2 \sim \sigma^2 \chi^2_B $. Now, consider two independent chi-squared distributed variables: $ a \sim \sigma^2 \chi^2_A, b \sim \sigma^2 \chi^2_B $. It can be shown that the ratio of these two variables follows a Beta distribution: $ \frac{a}{a + b} \sim \text{Beta}\left(\frac{A}{2}, \frac{B}{2}\right) $ \citep{frankl1990some}. This result is independent of the scale parameter ( $\sigma^2$ ). To normalize the weights in a group, we consider the ratio: $ \frac{w_1^2 + w_2^2 + \ldots + w_D^2}{w_12 + w_2^2 + \ldots + w_N^2} $, where $D < N$. This can be rewritten as: $ \frac{w_1^2 + w_2^2 + \ldots + w_D^2}{(w_12 + w_2^2 + \ldots + w_D^2) + (w_{D+1}^2 + \ldots + w_N^2)} $. Given that $( w_1^2 + w_2^2 + \ldots + w_D^2) \sim \sigma^2 \chi^2_D$ and $(w_{D+1}^2 + \ldots + w_D^2) \sim \sigma^2 \chi^2_{N-D}$, the ratio follows a Beta distribution: $ \frac{w_1^2 + w_2^2 + \ldots + w_B^2}{w_12 + w_2^2 + \ldots + w_N^2} \sim \text{Beta}\left(\frac{D}{2}, \frac{N-D}{2}\right) $. Thus, the normalized weights in a group follow a Beta distribution.

\newpage
\section{Classic PVQ subroutines}
\label{sec:pseudocode}

The quantization, encoding and decoding algorithms of classic PVQ are provided below. They are equivalent to the algorithms originally proposed in \citep{fischer1986pyramid}, and included to be self-contained. We also fixed a small bug in the original description of the decoding algorithm.

\subsection{Algorithm 1: PVQ Quantization}
\label{sec:quantize-alg}

\begin{algorithm}[H]
\begin{algorithmic}[1]
\State $\widehat{\vv} \gets \vv$
\While{$||\widehat{\vv}||_1 \neq K$}
    \State $i \gets \text{argmax}_i(|\vv_i|)$
    \State $\vv_i \gets \vv_i - \text{sign}(\vv_i)$
    \State $\widehat{\vv} \gets \text{round}\left(\frac{K}{||\widehat{\vv}||_1} \widehat{\vv} \right)$
    %\State \text{CHECK THIS}
\EndWhile \\
\Return $\widehat{\vv}$
\end{algorithmic}
\caption{PVQ quantization: $\vv \mapsto \widehat{\vv}, \hspace{1em}  \R^D \to \mathbb{Z}^D$.}
\label{alg:quantize}
\end{algorithm}

\subsection{Algorithm 2: PVQ Encoding}
\label{sec:encode-alg}

\begin{algorithm}
\begin{algorithmic}[1]
\State $c \gets 0$, $i \gets 1$, $d \gets D$, $k \gets K$
\While{$k!=0$}
% OLD VERSION, DIDN"T WANT TO DELETE JUST IN CASE // AGRIN
%    \If{$|x_i|=0$}
%        \State $x_i \gets x_i + N(d-1,k) + \frac{1 - \text{sgn}(x_i)}{2} N(d-1,k-1)$
%    \Else
%        \State $x_i \gets x_i + N(d-1,k) + 2 \sum_{j=1}^{|x_i|-1}N(d-1,k-j) + \frac{1-\text{sgn}(x_i)}{2}N(d-1,k-|x_i|)$
%    \EndIf
%    \State $k \gets k - |x_i|$
%    \State $d \gets d - 1$
%    \State $i \gets i + 1$
    \If{$|x_i| = 1$}
        \State $c \gets c + N(d-1,k) + \frac{1 - \text{sgn}(x_i)}{2} N(d-1,k-1)$
    \EndIf
    \If{$|x_i| > 1$} % No ElseIf seems availble
        \State $c \gets c + N(d-1,k) + 2 \sum_{j=1}^{|x_i|-1}N(d-1,k-j) + \frac{1-\text{sgn}(x_i)}{2}N(d-1,k-|x_i|)$ \label{alg:encode:sum}
    \EndIf
    \State $k \gets k - |x_i|$
    \State $d \gets d - 1$
    \State $i \gets i + 1$
\EndWhile \\
\Return $c$
\end{algorithmic}
\caption{PVQ encoding: $\vp \mapsto c, \hspace{1em} \mathbb{Z}^D \to [1, N(D, K)]$.}
\label{alg:encode}
\end{algorithm}

\newpage
\subsection{Algorithm 3: PVQ Decoding}
\label{sec:decode-alg}

\begin{algorithm}[H]
\begin{algorithmic}[1]
\State $\vx = \vzero$
\State $z \gets 0$, $i \gets 1$, $d \gets D$, $k \gets K$
\While{$k > 0$}
    \If{$c=z$}
        \State $x_i \gets 0$ 
        \If{$k > 0$}
            \State $x_D \gets k - |x_i|$
            \State $k \gets 0$
        \EndIf
    \Else
        \If{$c < cb + N(d-1, k)$}
            \State $x_i \gets 0$
        \Else
            \State $z = z + N(d-1,k)$
            \State $j \gets 1$
            \While{$c \geq z + 2 N(d-1,k-j)$}
                \State $z \gets z + 2 N(d-1,k-j)$
                \State $j \gets j + 1$
            \EndWhile
            \If{$c < z + N(d-1,k-j)$}
                \State $x_i \gets j$
            \Else
                \State $x_i \gets -j$
                \State $z + N(d-1,k-j)$
            \EndIf
        \EndIf
    \EndIf
\EndWhile \\
\Return $\vx$
\end{algorithmic}
\caption{PVQ decoding: $c \mapsto \vp, \hspace{1em} [1, N(D, K)] \to \mathbb{Z}^D$.}
\label{alg:decode}
\end{algorithm}

The decode algorithm \cref{alg:decode} is akin to that described in the original PVQ paper \citep{fischer1986pyramid} and provided to be self-contained. It also fixes a missing line 23, setting $x_i \gets -j$. Not including this line results in wrongly decoded vectors when $\vp$ contains negative values, except for when the last value is negative (which is why the example provided in the original paper does not fail).

\section{Implementation details}

\subsection{Dataset}
We ran all methods on exactly the same data to ensure fair comparison. We follow the Quarot paper \citep{ashkboos2024quarot}, and use the same 128 samples of the WikiText-2 dataset and hold-out validation data in all experiments. 

\subsection{Parallelization using CUDA kernels}
\label{appendix:implementation:cuda}
Here, we provide additional details for the CUDA implementation described in \cref{sec:results:optimizing_pvq_for_cuda}.

The quantization, encoding and decoding operations described in \cref{sec:pseudocode} can all be parallelized across the batch dimension. To utilize this, we implementing these operations in custom CUDA kernels. To achieve the time complexities of \autoref{tab:cuda_complexity} we must re-write the summation in the encoding operation (Algorithm \ref{alg:encode}, line \ref{alg:encode:sum}). This is done by in addition to the size table $N$ computing an additional cumulative sum

\begin{equation}
V(d, k) = \sum_{i=1}^k N(d, i) \quad \forall d = 0, \dots, D\, \text{ and } k = 0, \dots,  K\,.
\end{equation}

Using V, the expression becomes

\begin{align*}
c \gets c &+ N(d-1,k) \\
&+ 2 \left( V(d - 1, k - 1) - V(d - 1, k - |x_i|) 
 \right) \\
 &+ \frac{1-\text{sgn}(x_i)}{2}N(d-1,k-|x_i|)\,,
\end{align*}

which has constant time and I/O complexity.

Furthermore, the size table function can be parallelized across the $K$ axis, by recognizing that the contributions $N(D - 1, K) + N(D, K - 1)$ to $N(D, K)$ from \autoref{eq:size_table} correspond to adding the cumulative sum $V(D - 1, k)$ to each $k$ at row $D$ which may be done efficiently in CUDA~\citep{harris2007gems}. The pseudocode for each thread is shown in \autoref{alg:fast_size_table}. Notice how the threads only write to global memory, except for any synchornization reads.

\begin{algorithm}[H]
\begin{algorithmic}[1]
\State $N(D, K) = \vzero$
\State $d \gets 0$
\State $k \gets \text{thread index}$
\State $v_k \gets 2$  \Comment{Value in thread registry}
\While{$d < D$}
    \If{$k=0$}
        \State $v \gets d << 1$  \Comment{Recurrent relationship does not hold for $k = 0$, set to $2d$}
    \Else
        \State $v_k \gets v_k + v_{k - 1}$  \Comment{Add diagonal relationships through warp shuffle}
    \EndIf
    \State $v_k \gets \sum_{i=1}^k v_{i}$  \Comment{Add cumulative sums through a parallelized scan}
    
    \State $N(d, k) \gets v_k$  \Comment{Write to size table in global memory}
    \State $d \gets d + 1$
\EndWhile \\
\Return $N$
\end{algorithmic}
\caption{Size table computation.}
\label{alg:fast_size_table}
\end{algorithm}

% Pseudocode for scan table based on:
% def build_scan_table(D, K):
%    table = torch.zeros(D, K, dtype=torch.int64)
%    if D <= 0:
%        return table
%
%    # Need to handle the first row separately
%    row = torch.full((K,), 2, dtype=torch.int64)
%    table[0, :] = row
%    if D <= 1:
%        return table
%
%    for i in range(1, D):
%        row[1:] += row[:-1].clone()  # Diagonal relationships
%        row[0] = 2 * (i + 1)
%        row = row.cumsum(0)
%        table[i, :] = row
%    return table
% Pseudocode for fast encoding
% - compute cumsum(|x_i|)
% - compute cumsum(N[., k])
% - compute each inner loop independently
\subsection{amplitude quantization}

For scale quantization, we pre-compute a list of 10000 points using \texttt{scipy.stats.beta.cdf} onto GPU. We then directly index neighbouring points on this list which we interpolate linearly to obtain quantized values. To dequantize, we use \texttt{scipy.stats.beta.ppf} to construct an explicit codebook on GPU, which we can index in parallel to perform dequantization.

\newpage
\section{Empirical amplitude histograms}
\label{sec:additional-amplitude-histograms}

To assess how well our theory of  theoretical \cref{sec:amplitude-quantization} matches the empircial weight distribuitons of pretrained LLM models, we compare the empirical weight histograms of all layers of a pretrained LLM model with the expected $\text{Beta}\left(\frac{D}{2}, \frac{D(G-1)}{2}\right)$ distribution. We consider the weights of a pretrained Llama-v2-7b after coherence processing and provide the histograms of all weight matrices in the model below. We find that the Beta distribution closely matches the empirical weight distributions in practice.
\begin{table}[H]
%\resizebox{!}{\linewidth}{
\begin{tabular}{llll}
\includegraphics[width=0.22\linewidth]{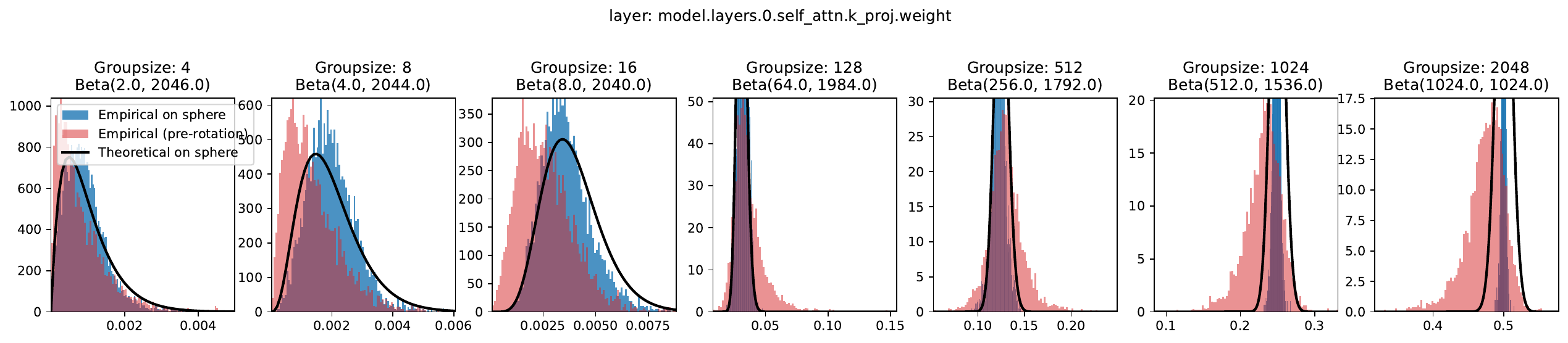} &
\includegraphics[width=0.22\linewidth]{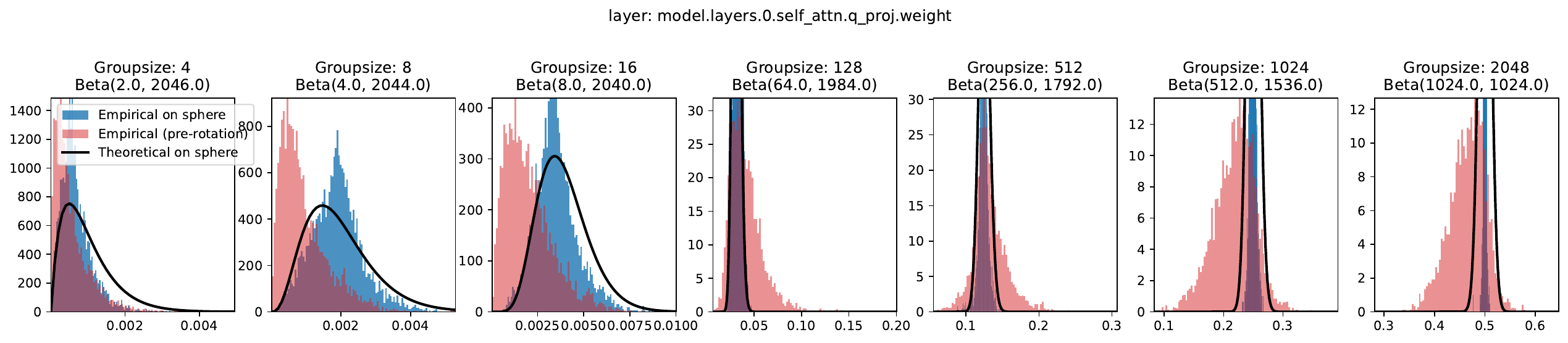} &
\includegraphics[width=0.22\linewidth]{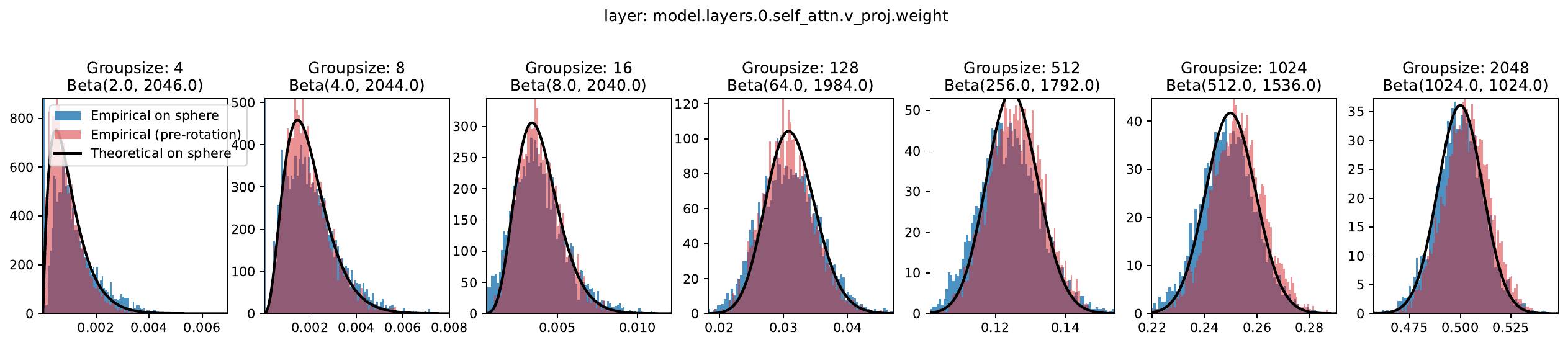} &
\includegraphics[width=0.22\linewidth]{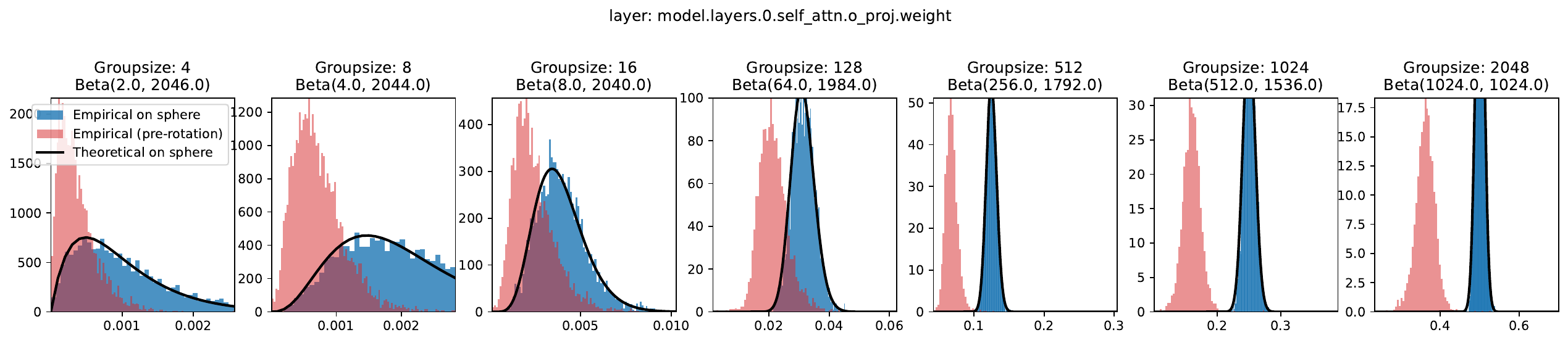} \\
\includegraphics[width=0.22\linewidth]{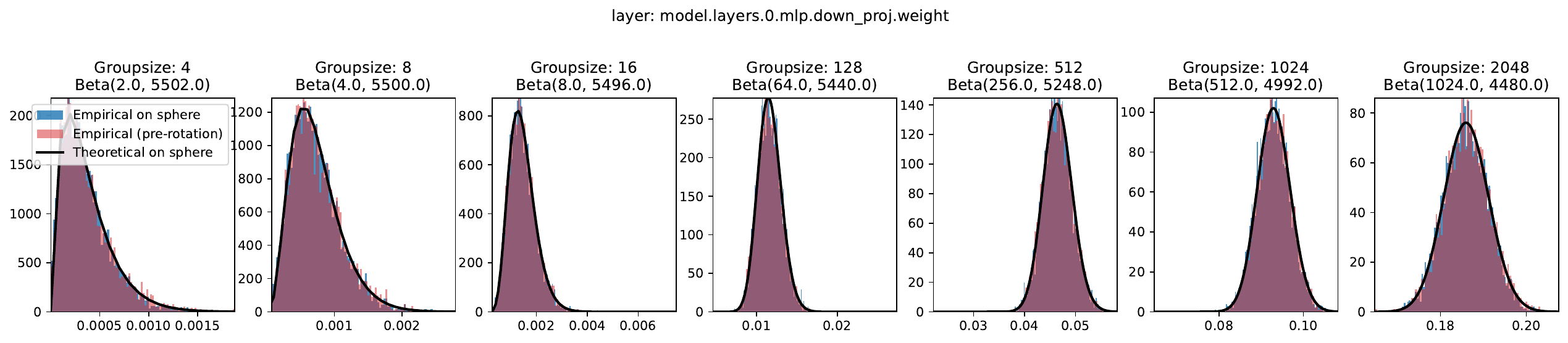} &
\includegraphics[width=0.22\linewidth]{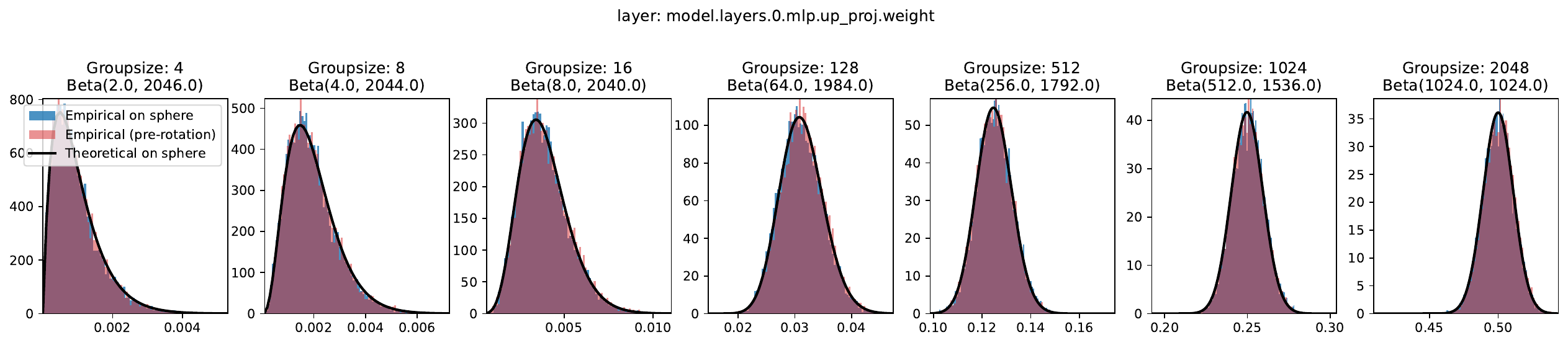} &
\includegraphics[width=0.22\linewidth]{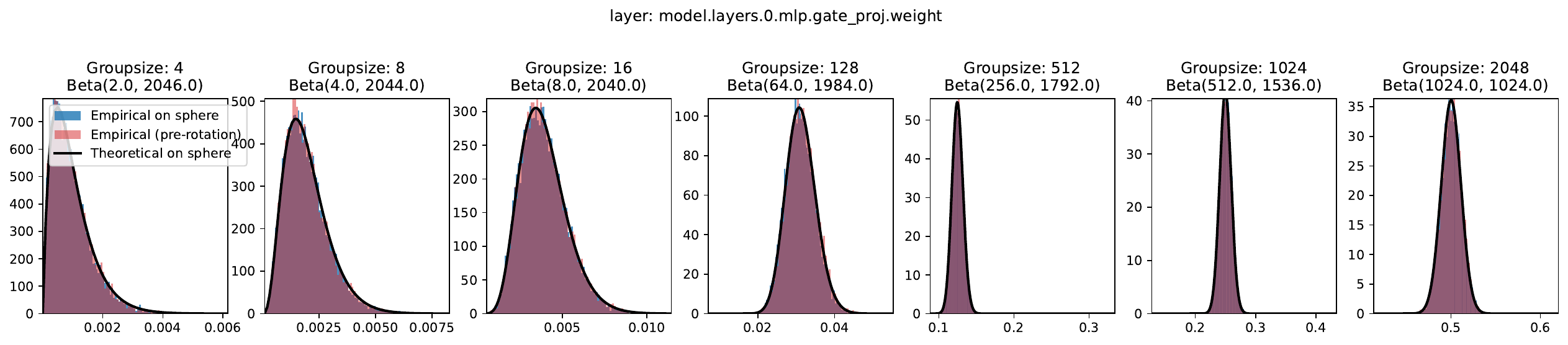} \\
\includegraphics[width=0.22\linewidth]{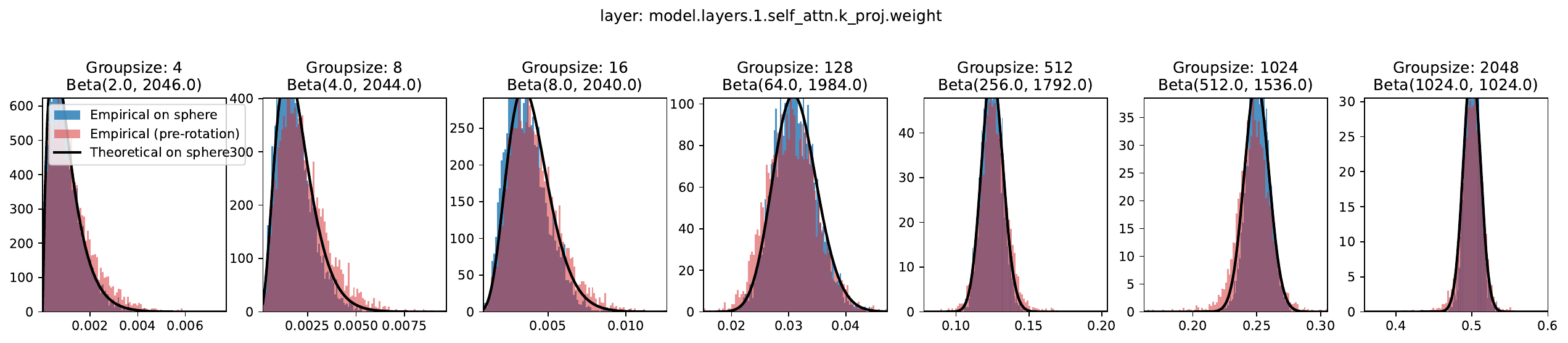} &
\includegraphics[width=0.22\linewidth]{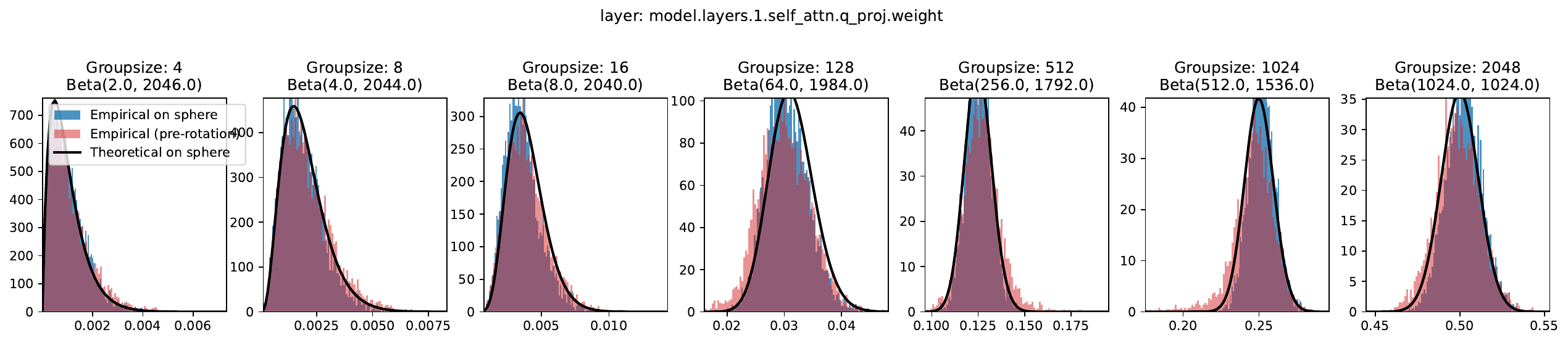} &
\includegraphics[width=0.22\linewidth]{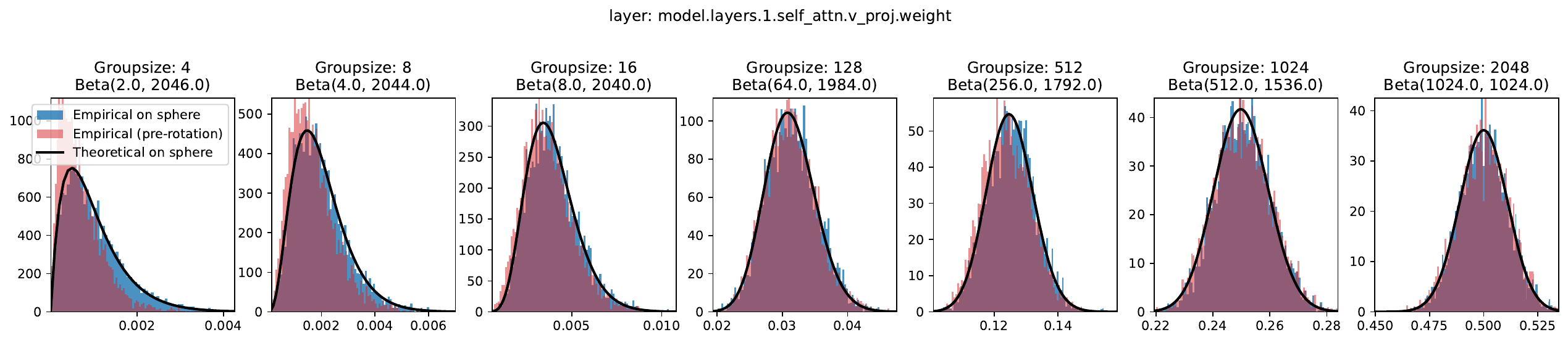} &
\includegraphics[width=0.22\linewidth]{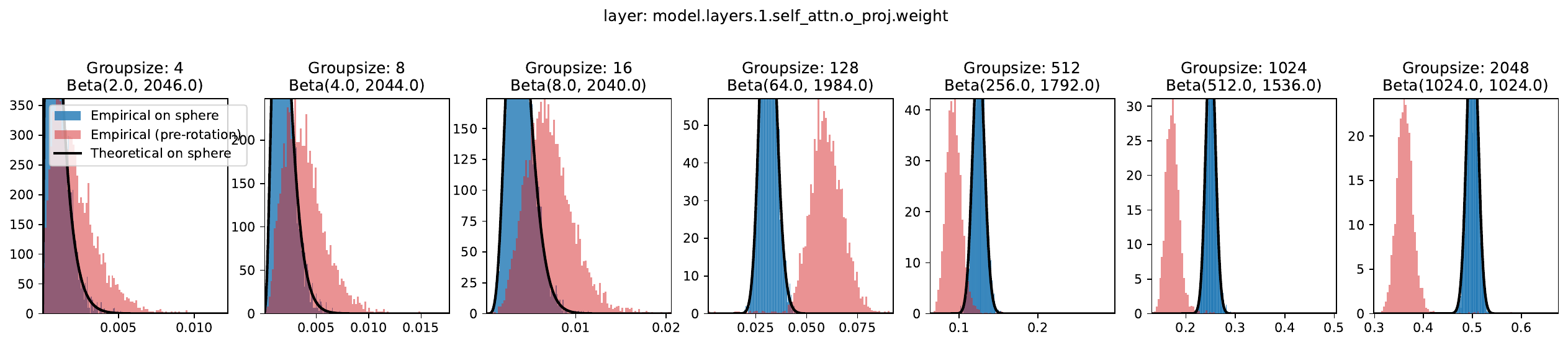} \\
\includegraphics[width=0.22\linewidth]{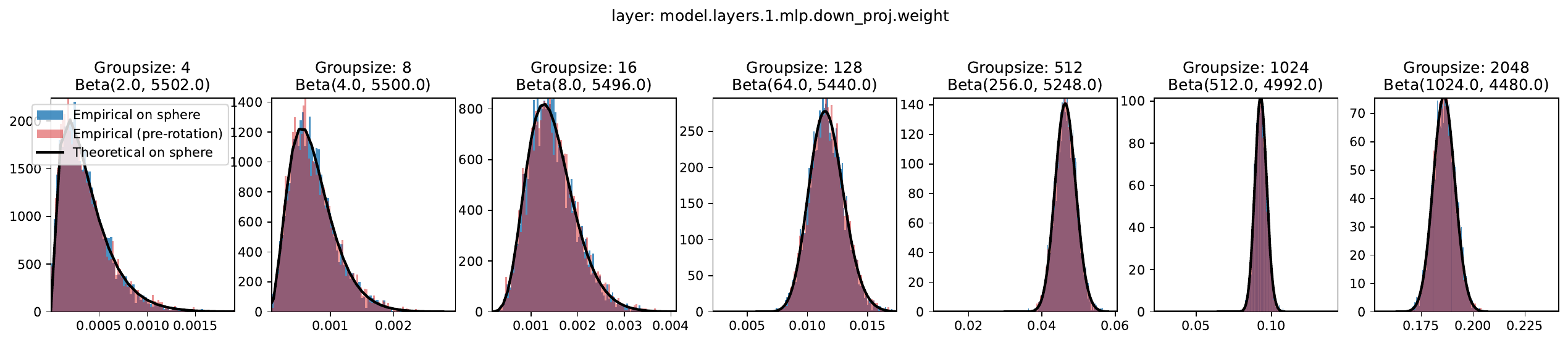} &
\includegraphics[width=0.22\linewidth]{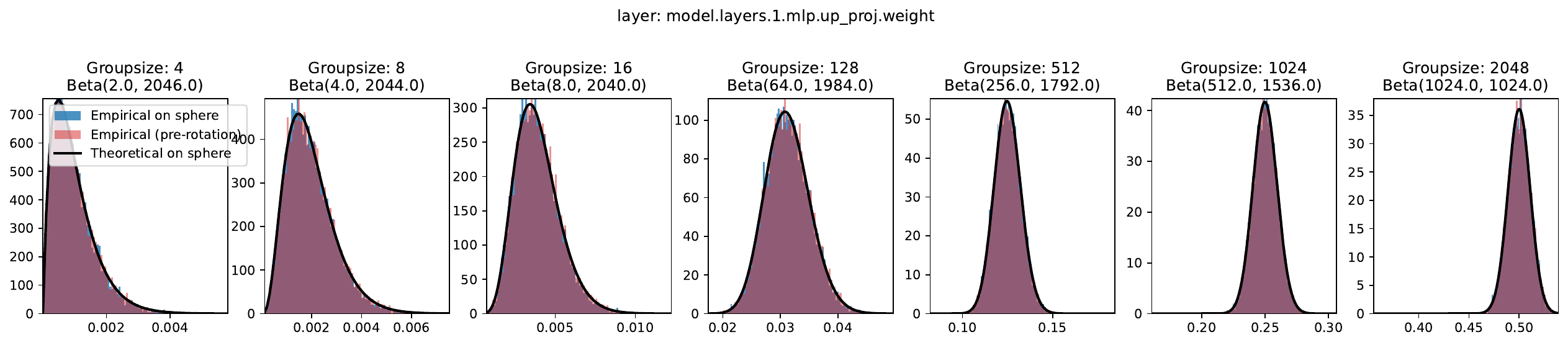} &
\includegraphics[width=0.22\linewidth]{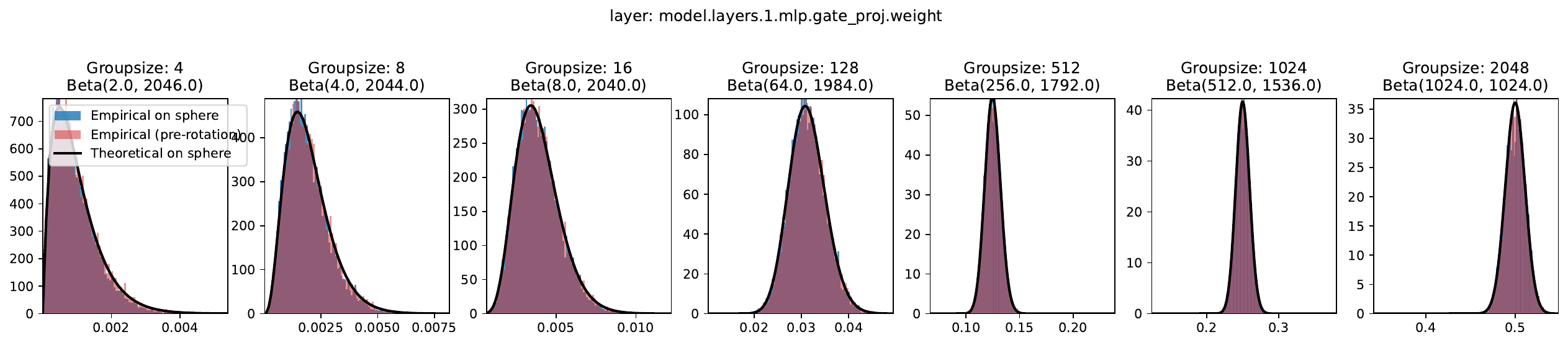} \\
\includegraphics[width=0.22\linewidth]{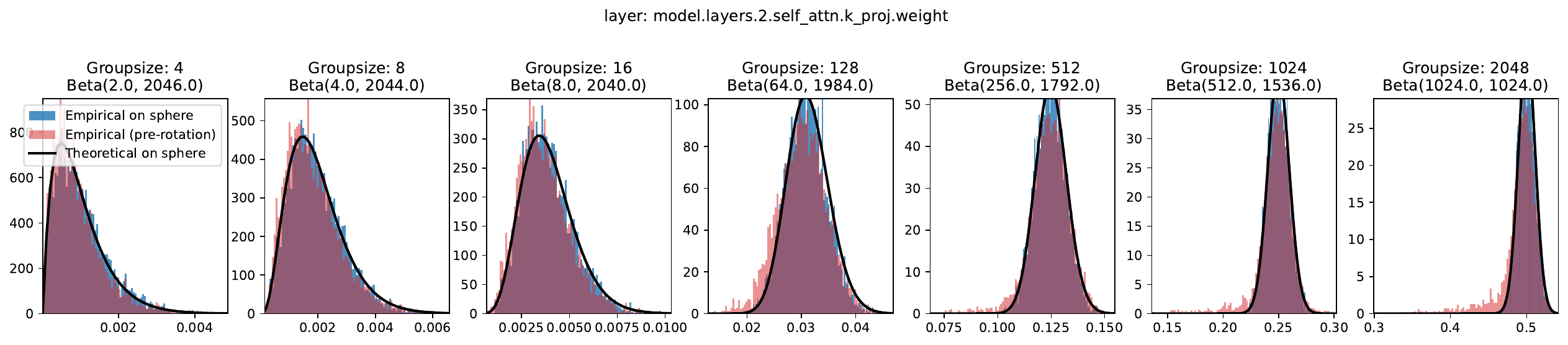} &
\includegraphics[width=0.22\linewidth]{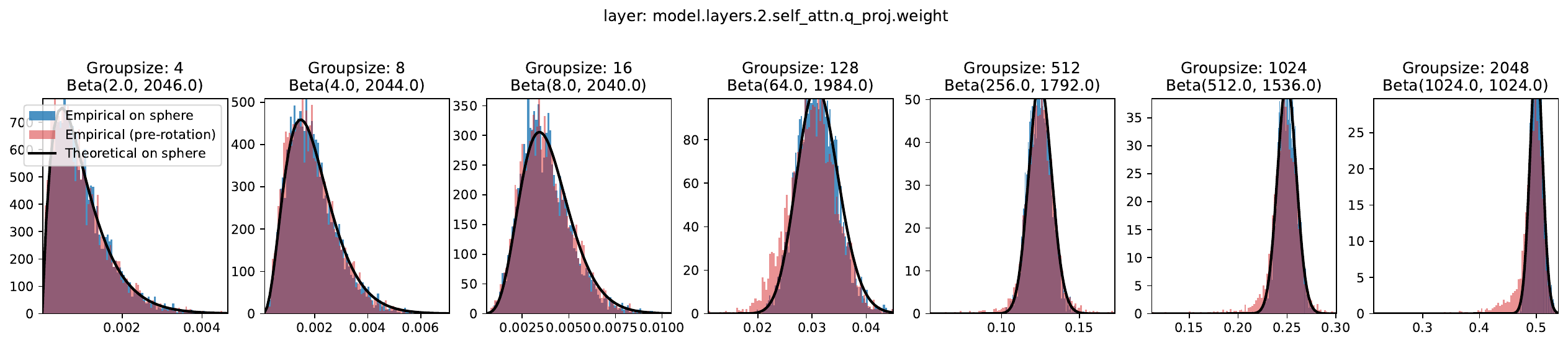} &
\includegraphics[width=0.22\linewidth]{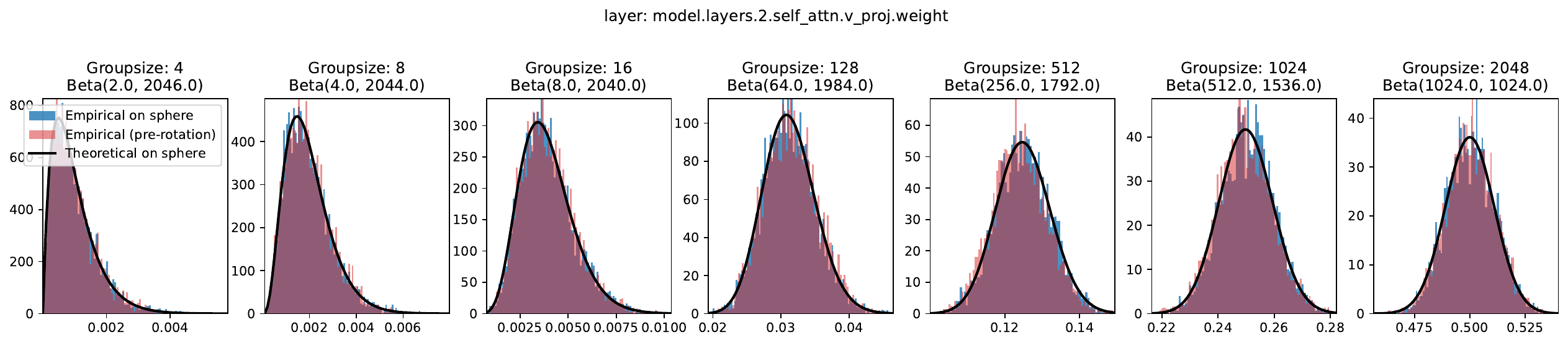} &
\includegraphics[width=0.22\linewidth]{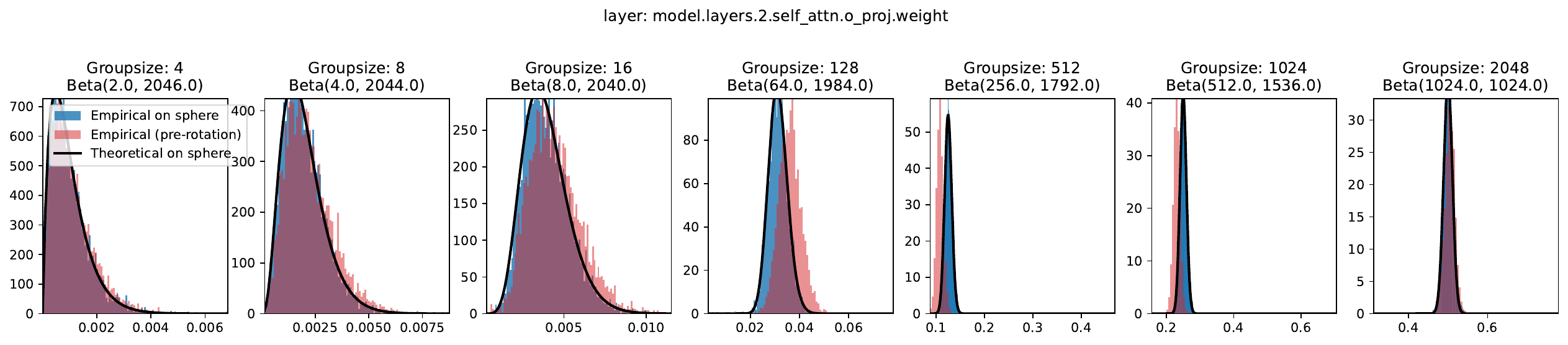} \\
\includegraphics[width=0.22\linewidth]{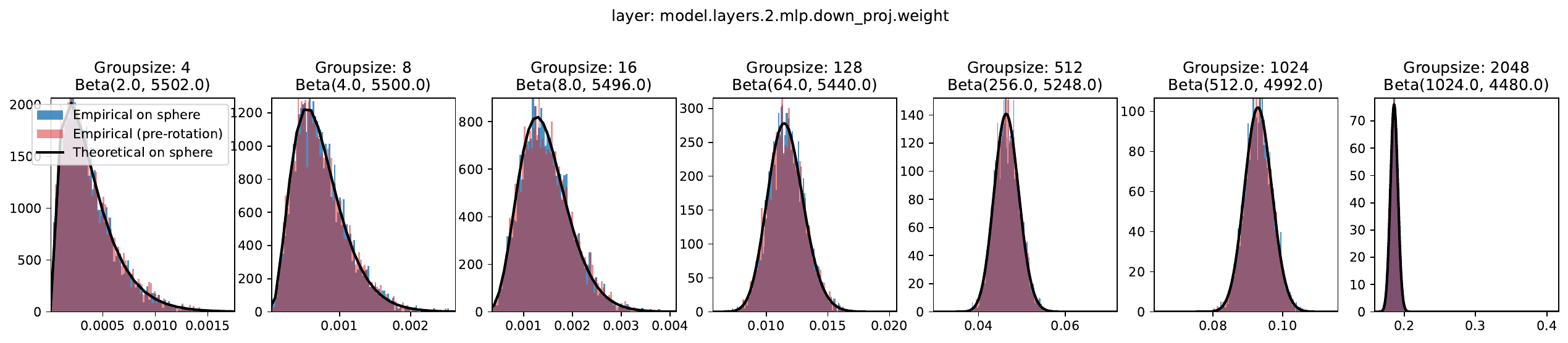} &
\includegraphics[width=0.22\linewidth]{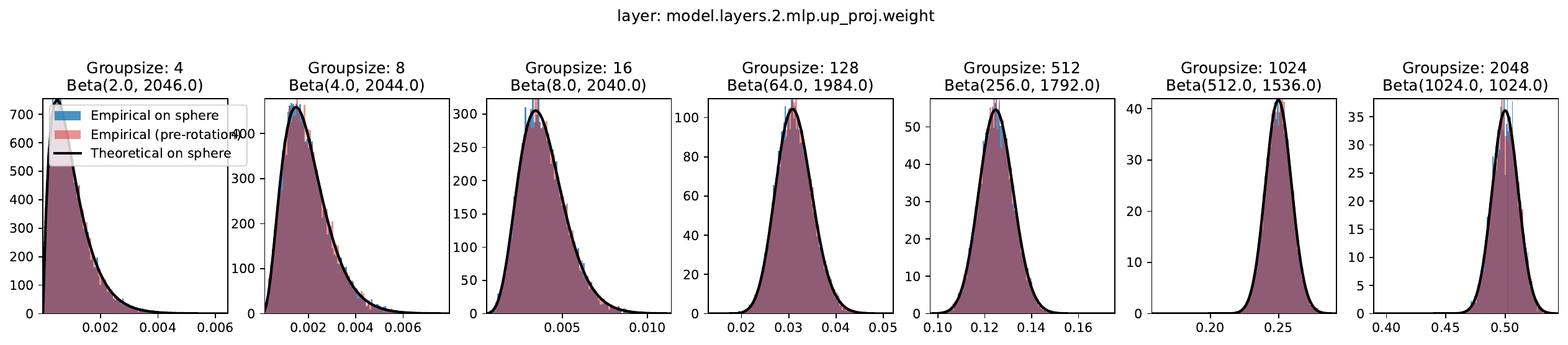} &
\includegraphics[width=0.22\linewidth]{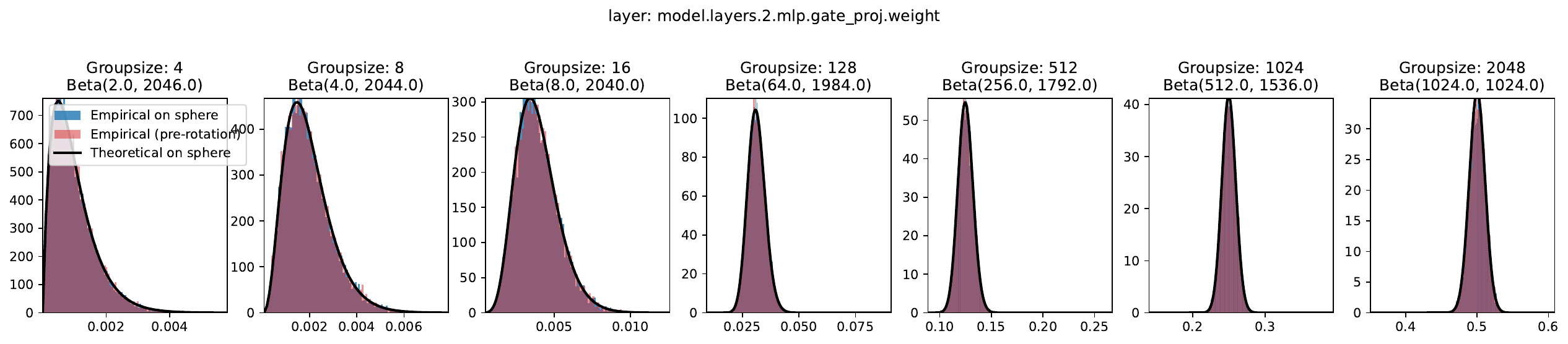} \\
\includegraphics[width=0.22\linewidth]{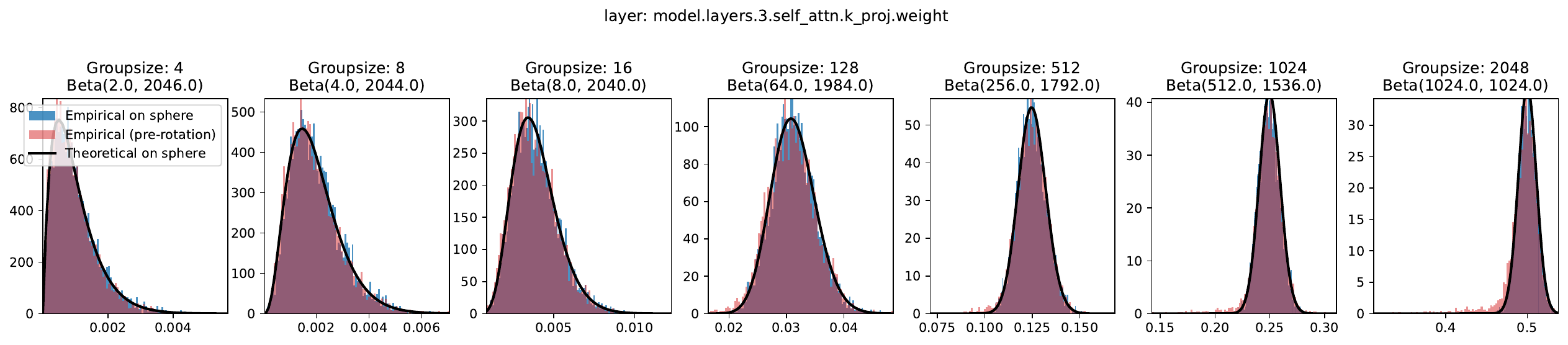} &
\includegraphics[width=0.22\linewidth]{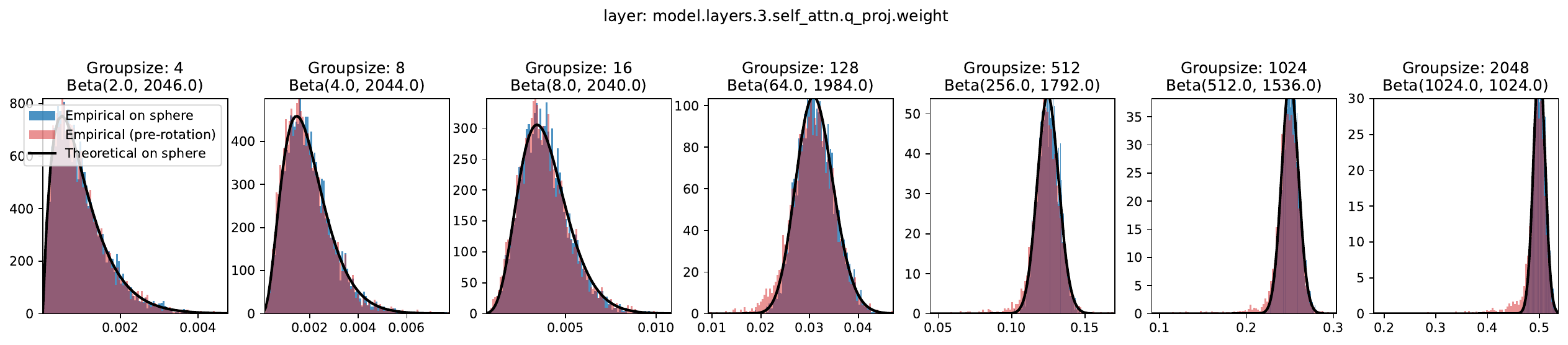} &
\includegraphics[width=0.22\linewidth]{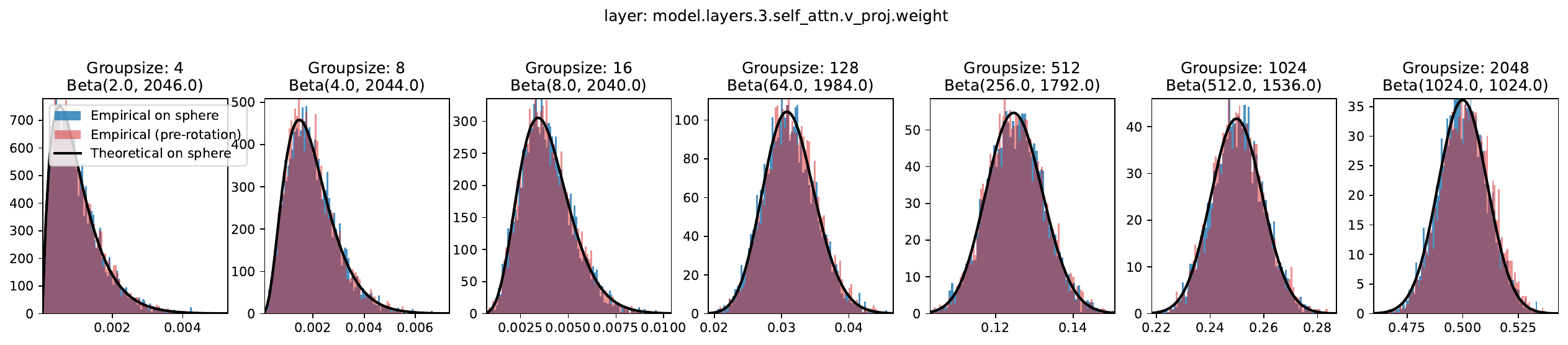} &
\includegraphics[width=0.22\linewidth]{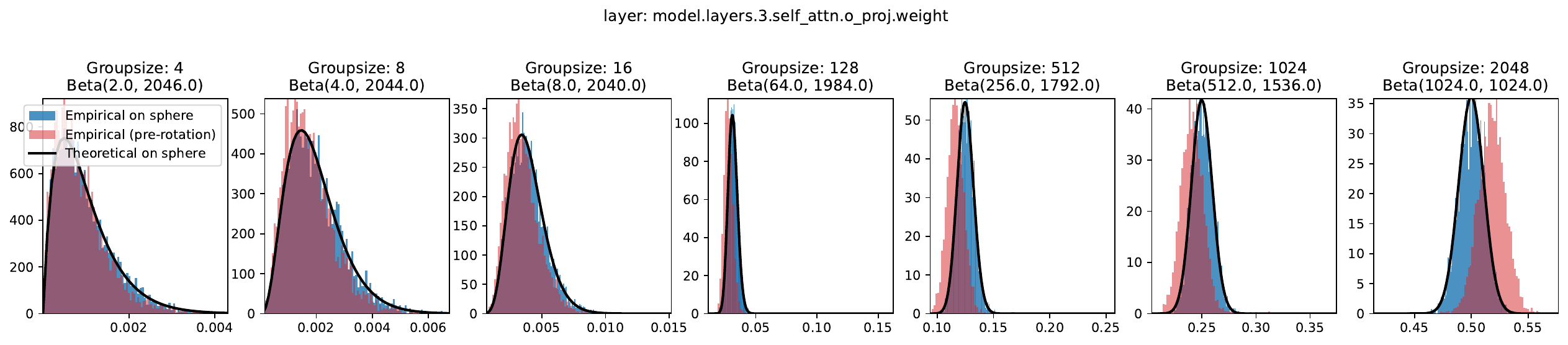} \\
\includegraphics[width=0.22\linewidth]{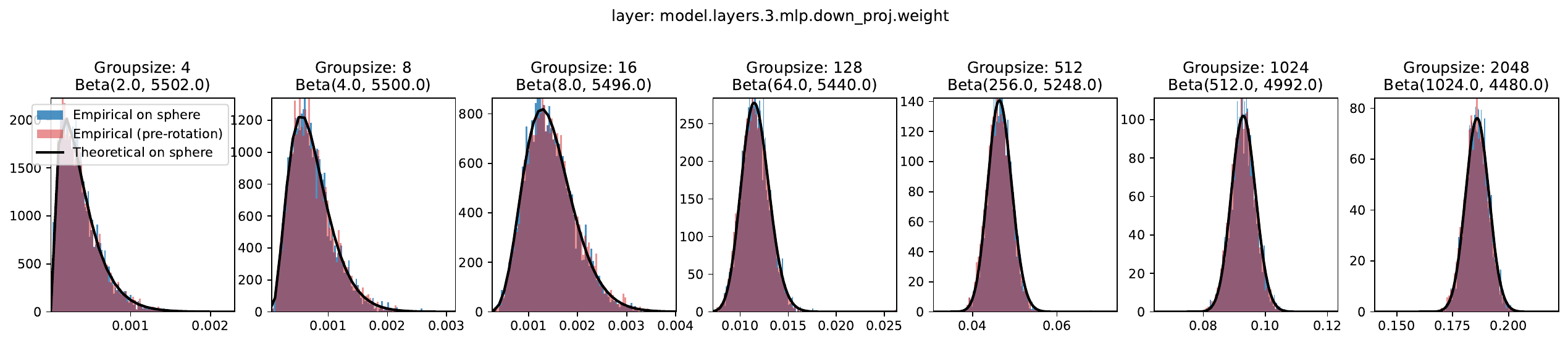} &
\includegraphics[width=0.22\linewidth]{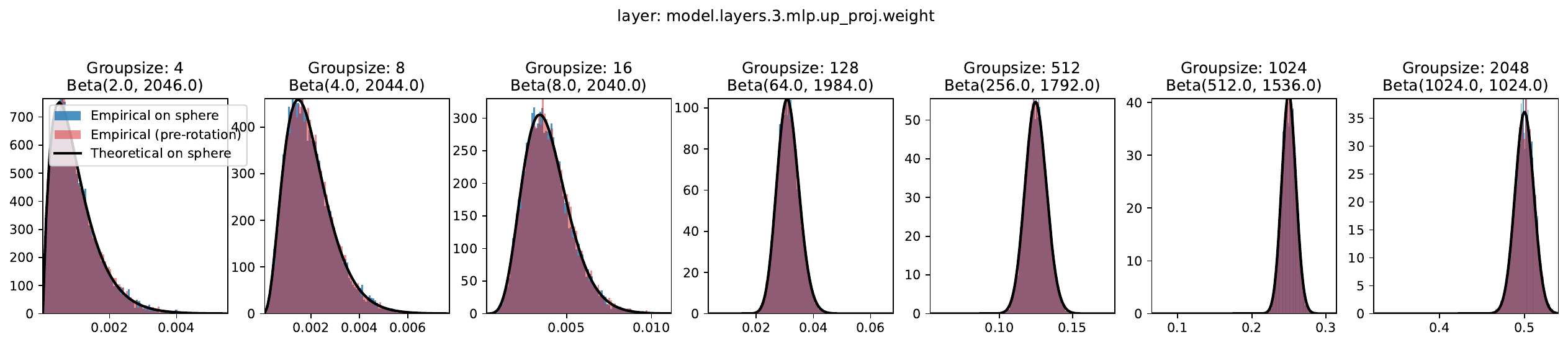} &
\includegraphics[width=0.22\linewidth]{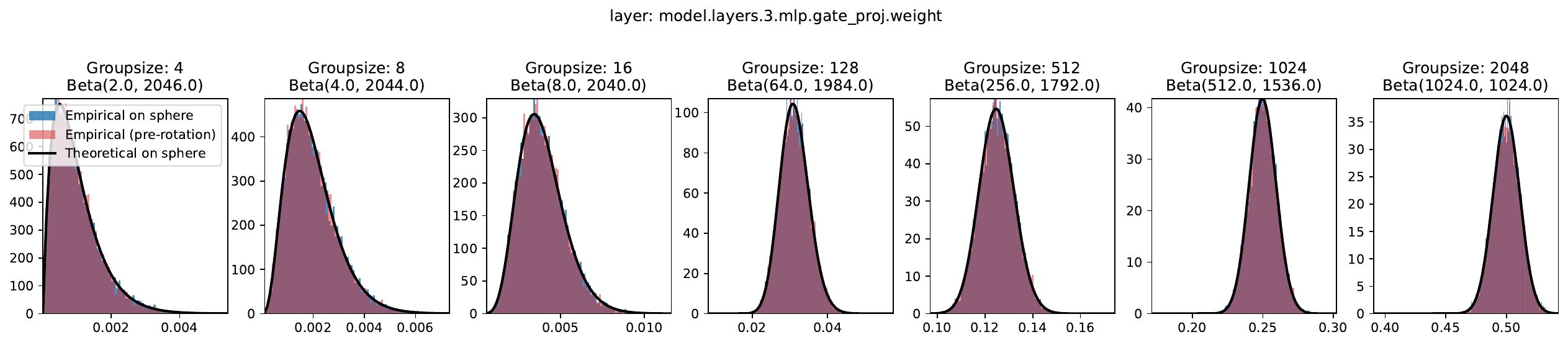} \\
\includegraphics[width=0.22\linewidth]{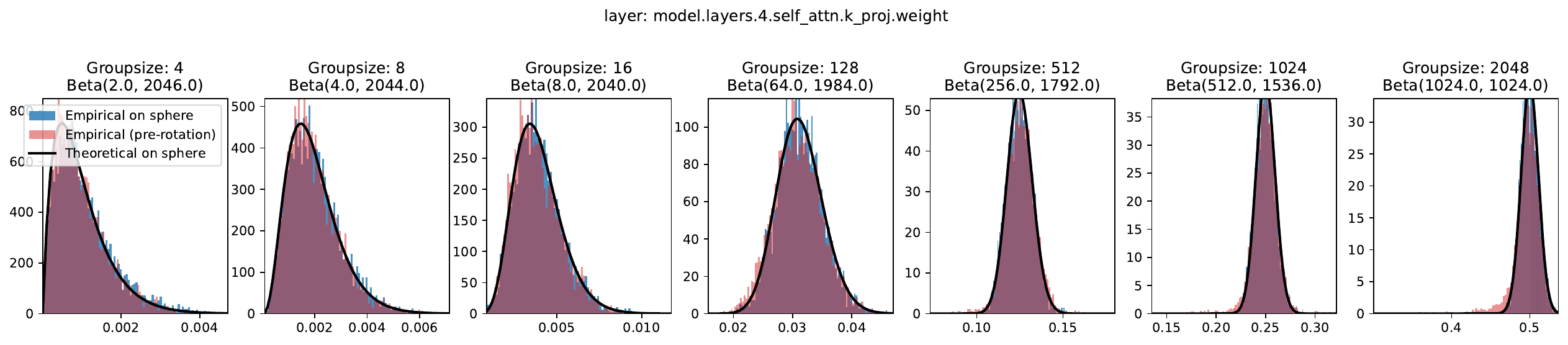} &
\includegraphics[width=0.22\linewidth]{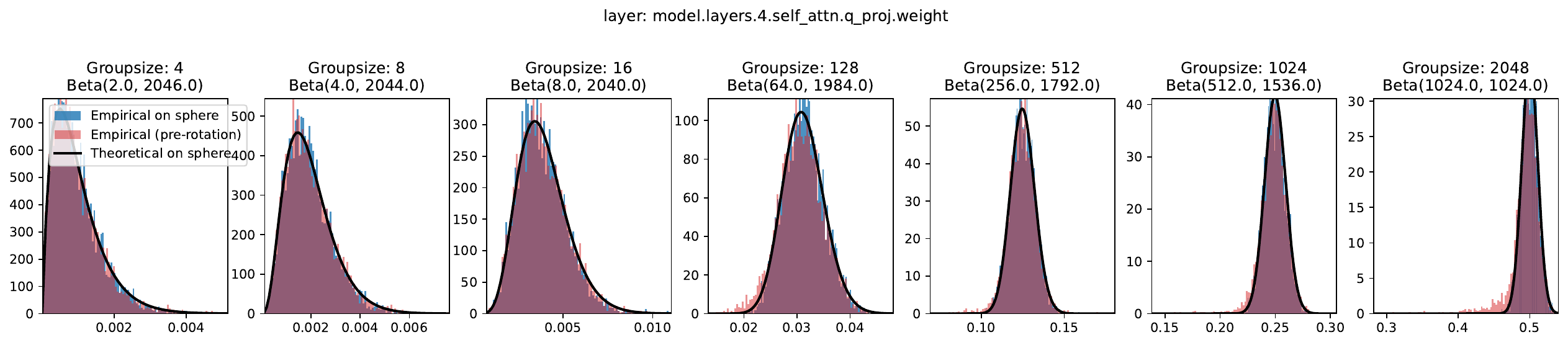} &
\includegraphics[width=0.22\linewidth]{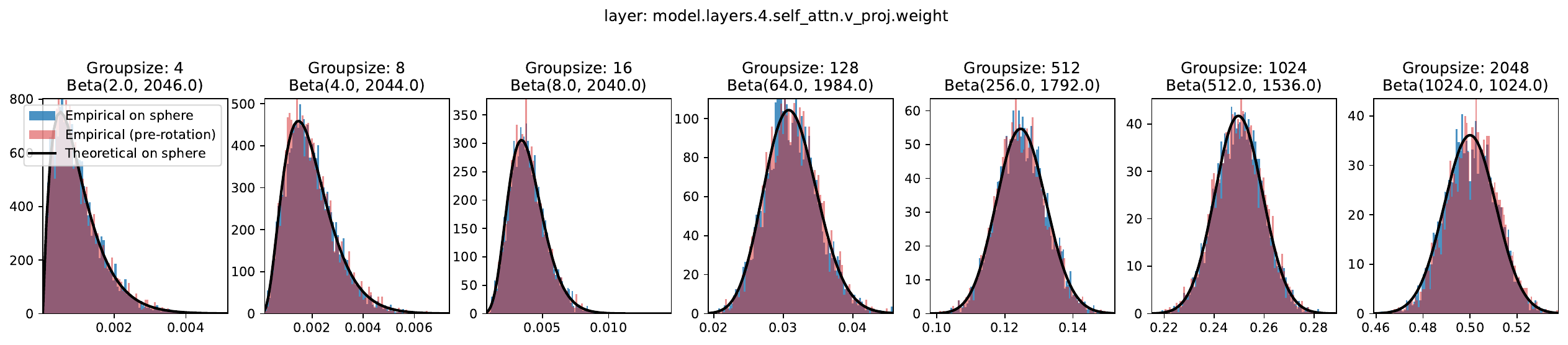} &
\includegraphics[width=0.22\linewidth]{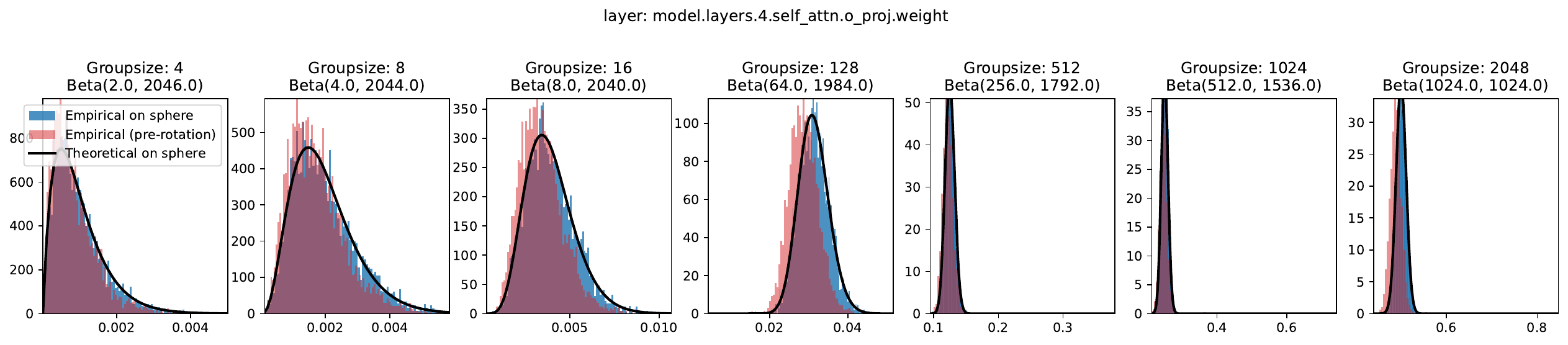} \\
\includegraphics[width=0.22\linewidth]{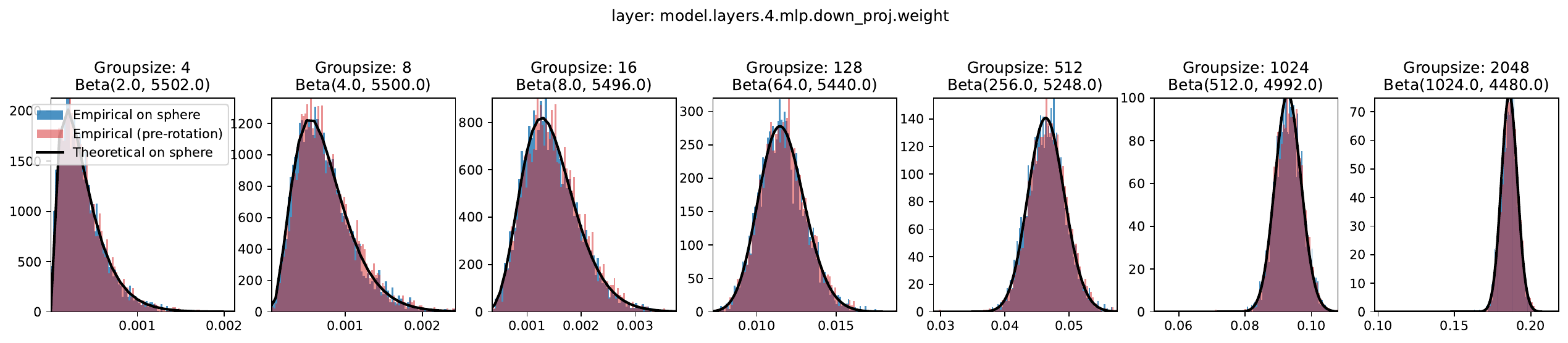} &
\includegraphics[width=0.22\linewidth]{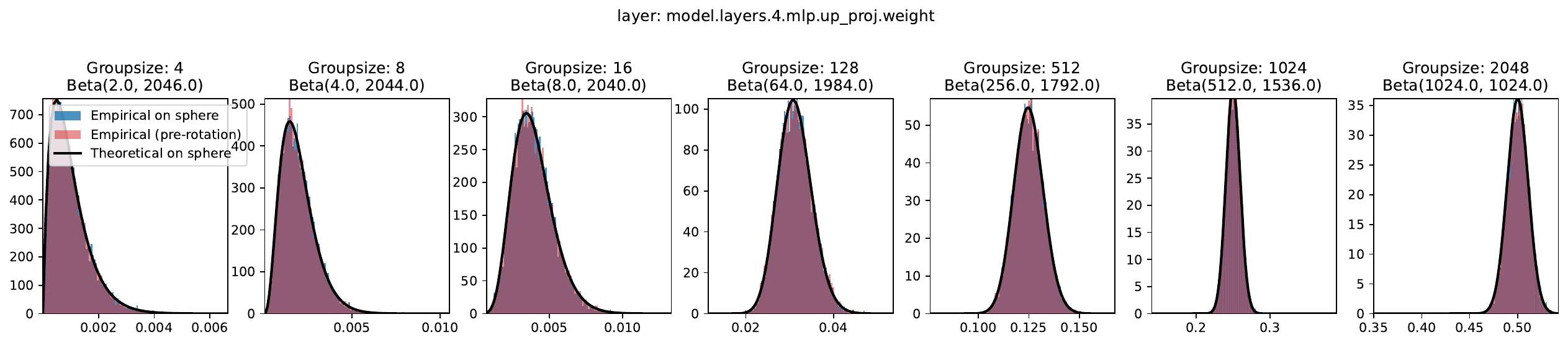} &
\includegraphics[width=0.22\linewidth]{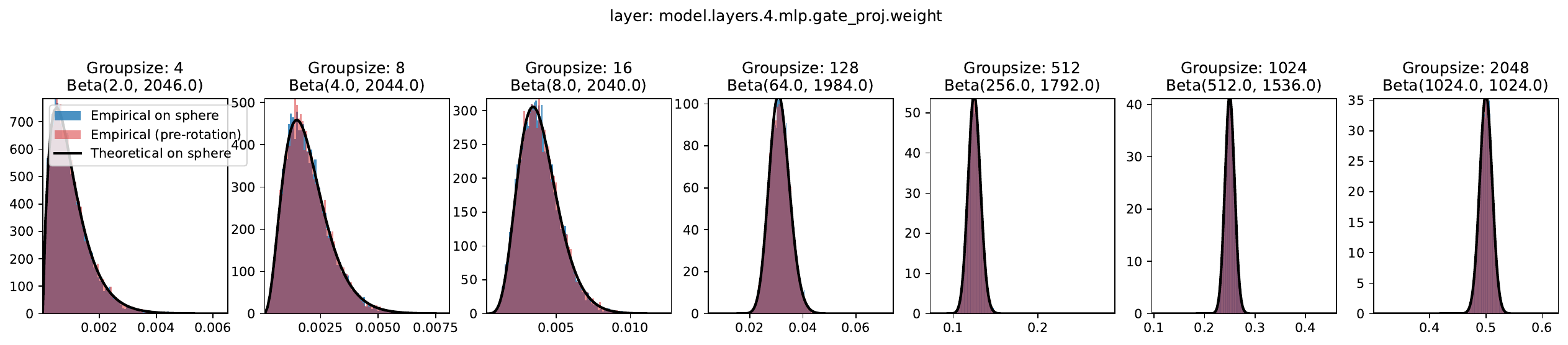} \\
\includegraphics[width=0.22\linewidth]{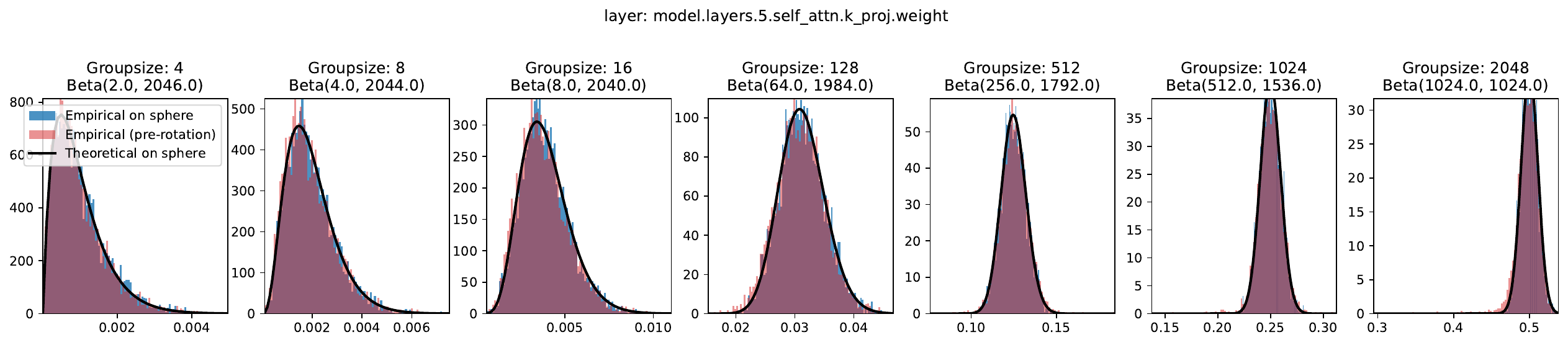} &
\includegraphics[width=0.22\linewidth]{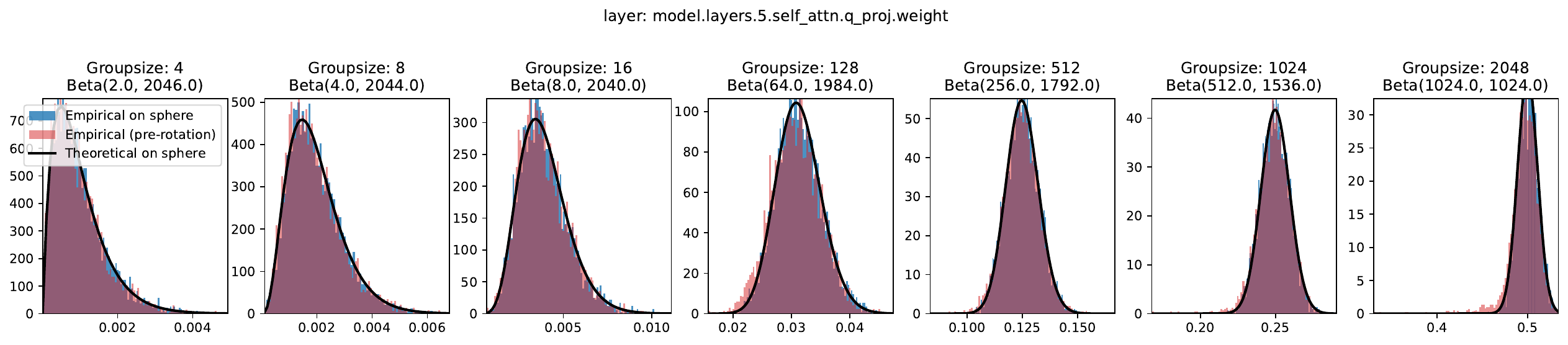} &
\includegraphics[width=0.22\linewidth]{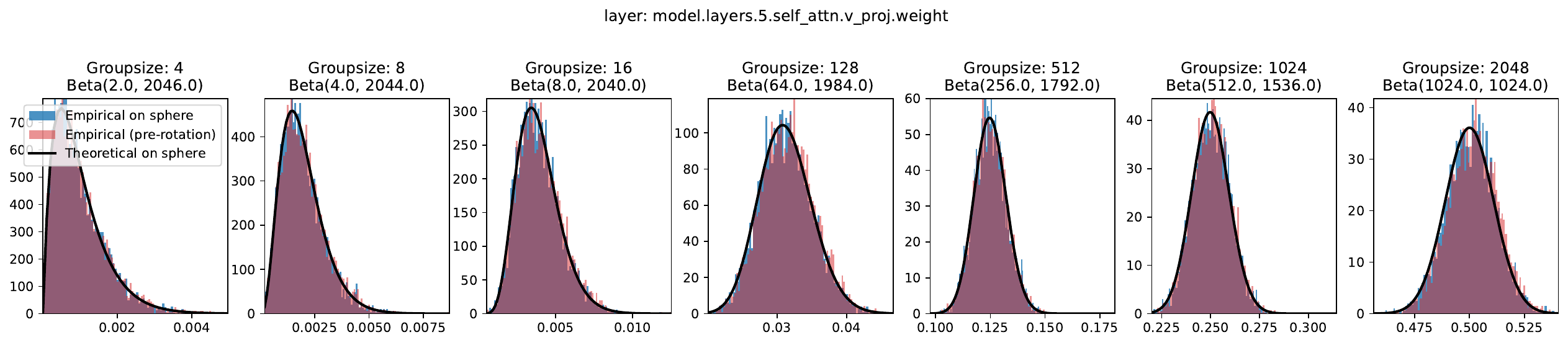} &
\includegraphics[width=0.22\linewidth]{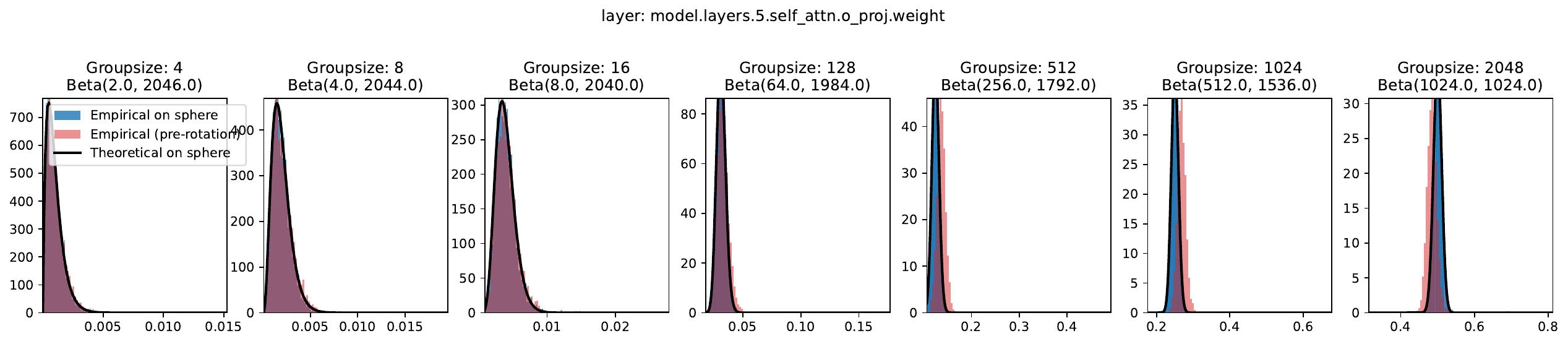} \\
\includegraphics[width=0.22\linewidth]{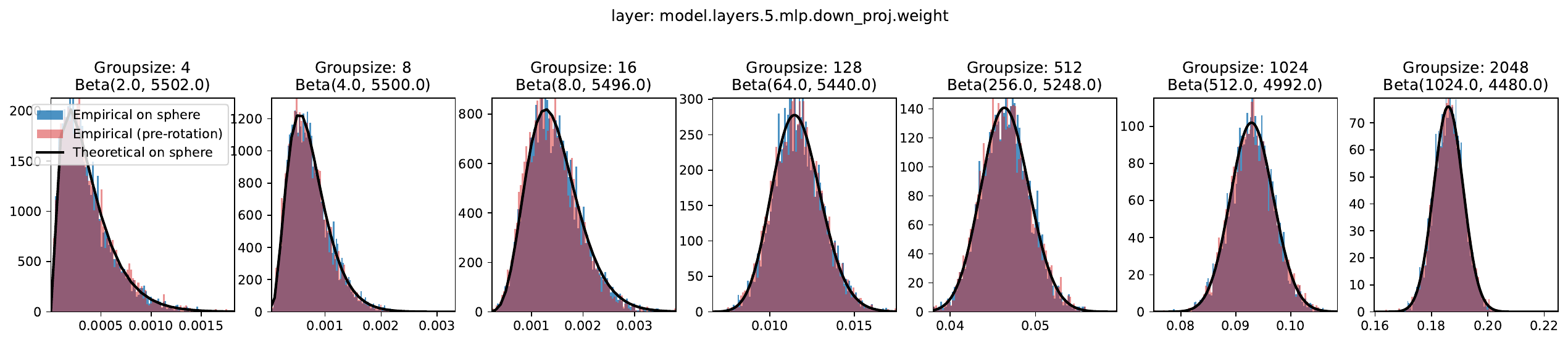} &
\includegraphics[width=0.22\linewidth]{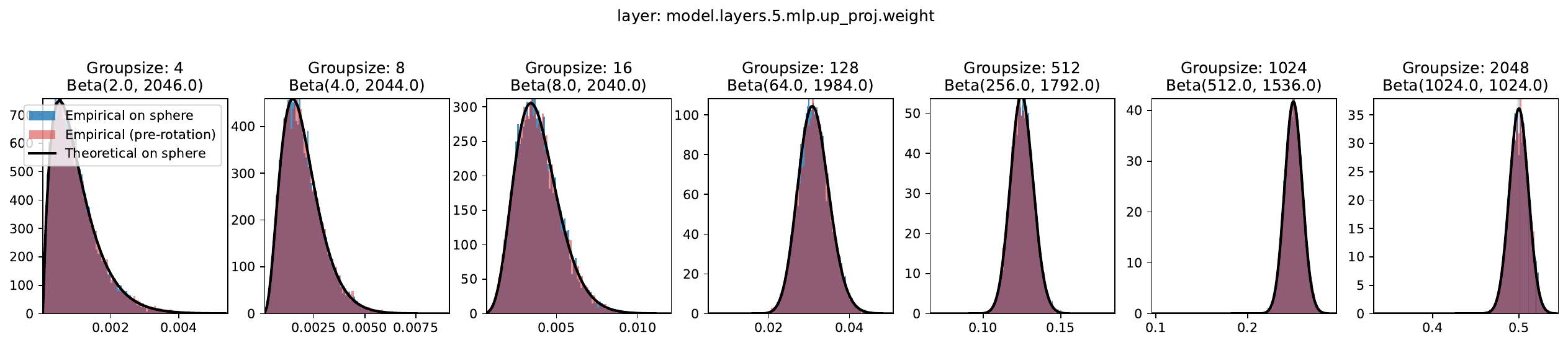} &
\includegraphics[width=0.22\linewidth]{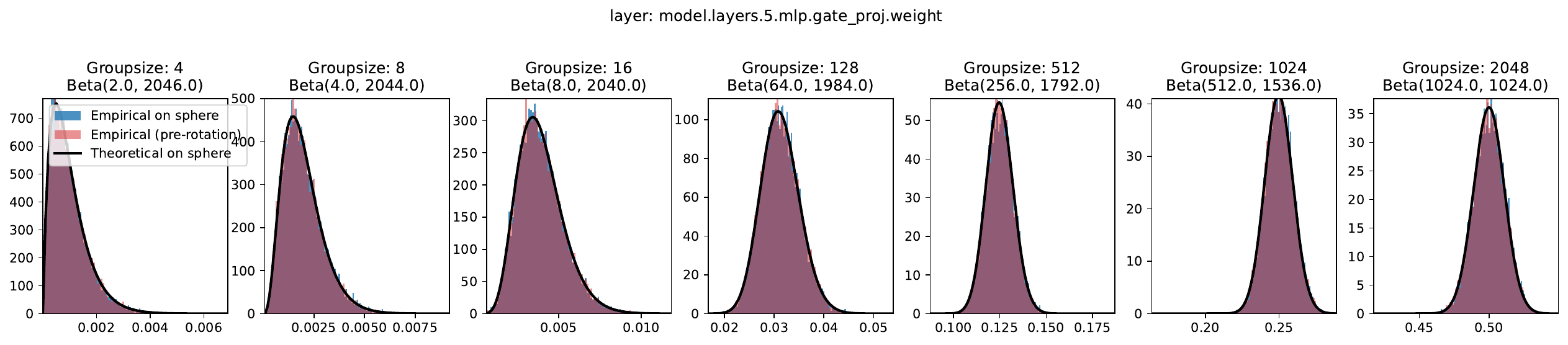} \\
\includegraphics[width=0.22\linewidth]{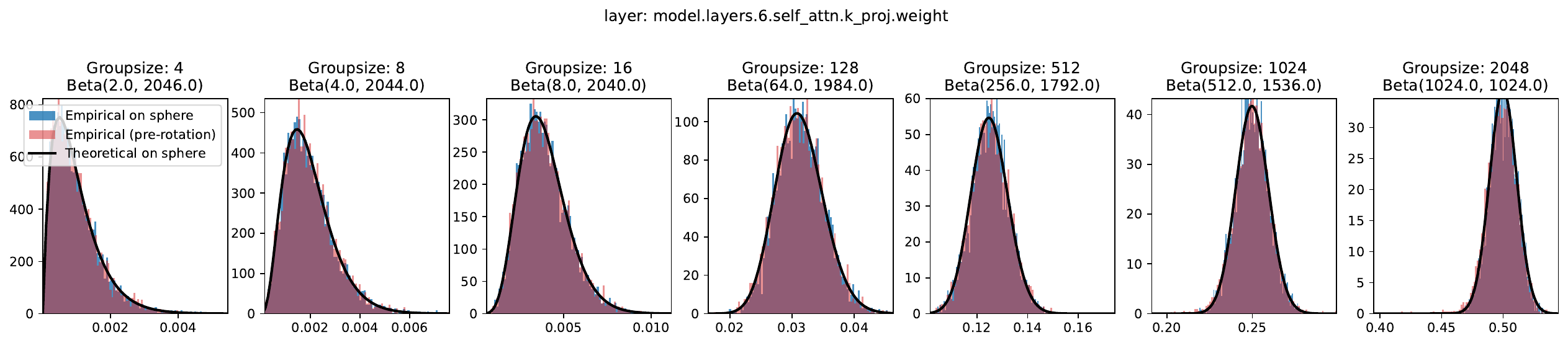} &
\includegraphics[width=0.22\linewidth]{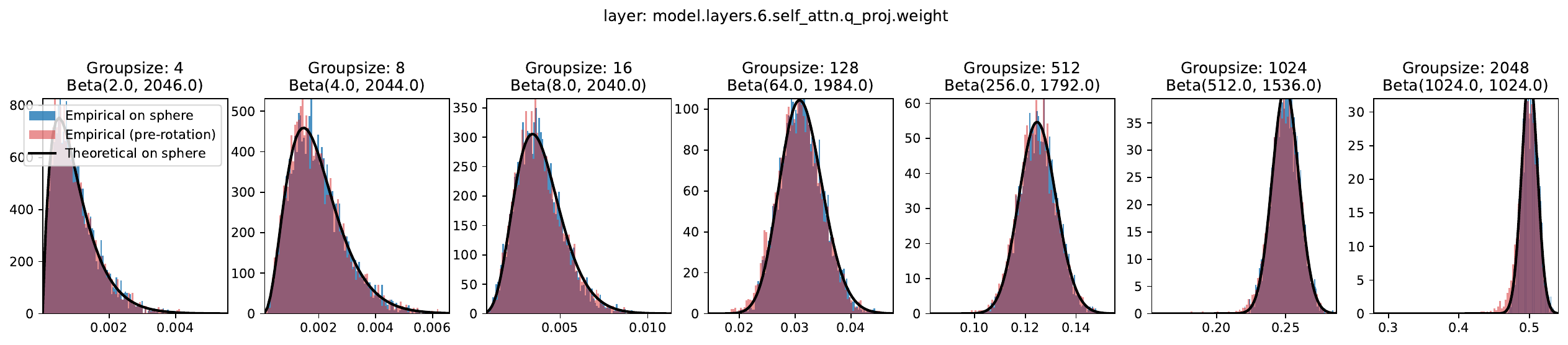} &
\includegraphics[width=0.22\linewidth]{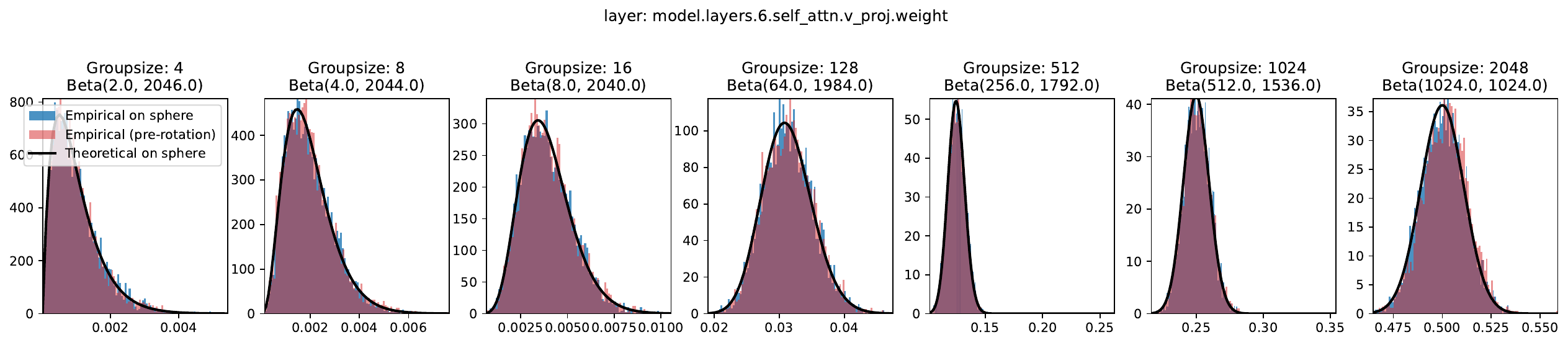} &
\includegraphics[width=0.22\linewidth]{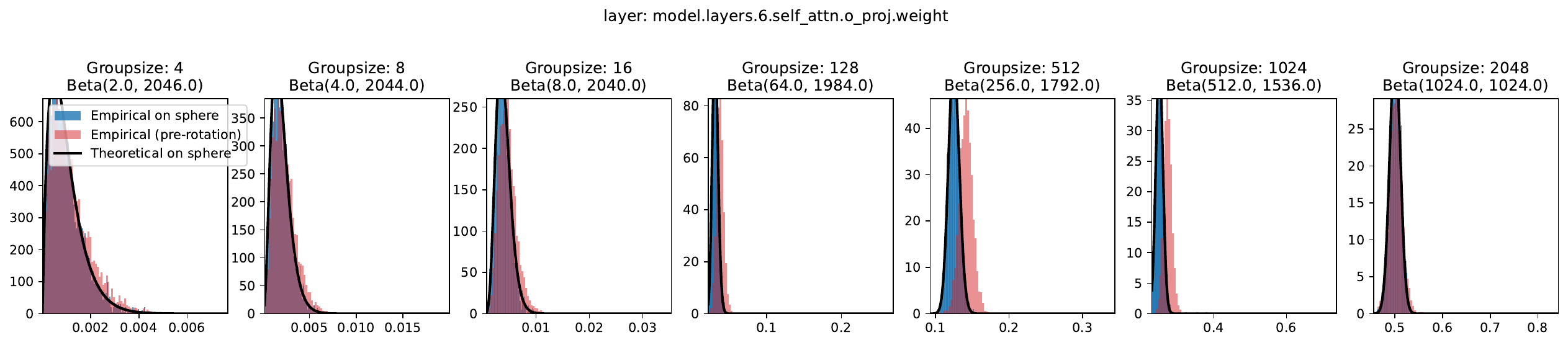} \\
\includegraphics[width=0.22\linewidth]{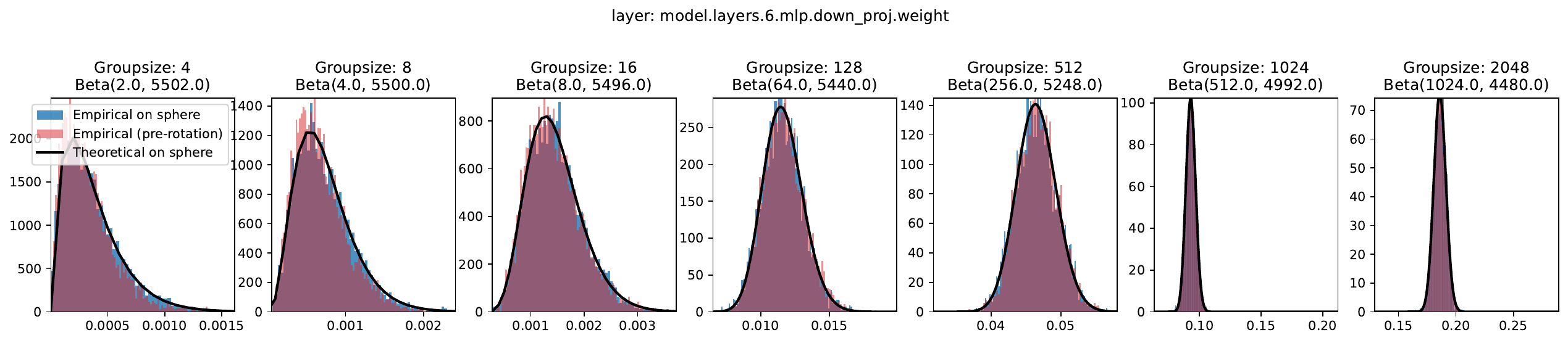} &
\includegraphics[width=0.22\linewidth]{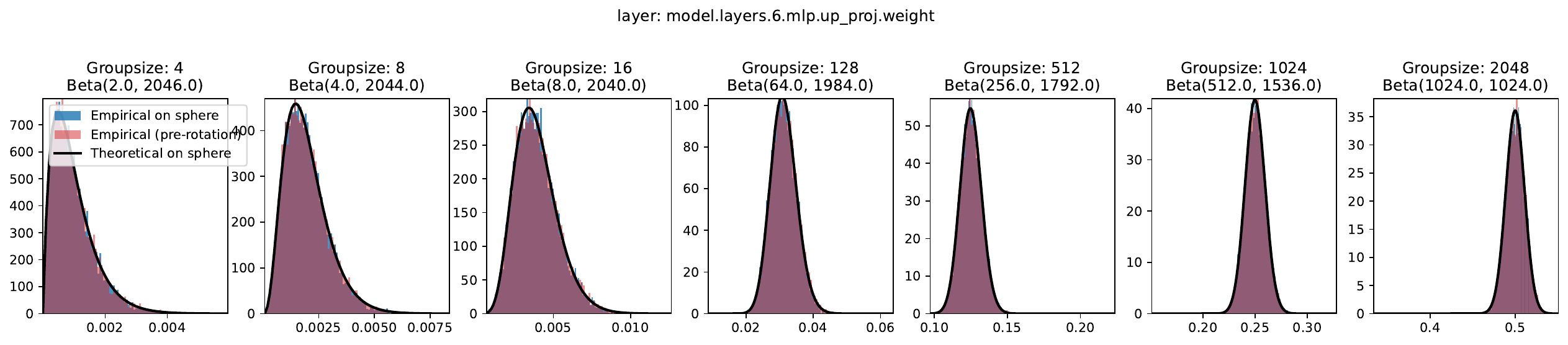} &
\includegraphics[width=0.22\linewidth]{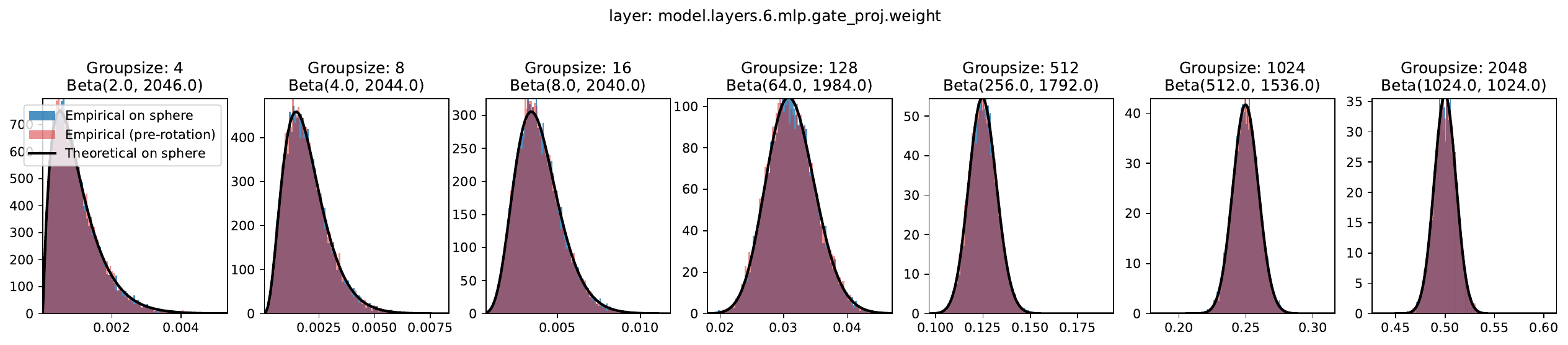} \\
\end{tabular}
\end{table}
\newpage
\begin{table}[H]
%\resizebox{!}{\linewidth}{
\begin{tabular}{llll}
\includegraphics[width=0.22\linewidth]{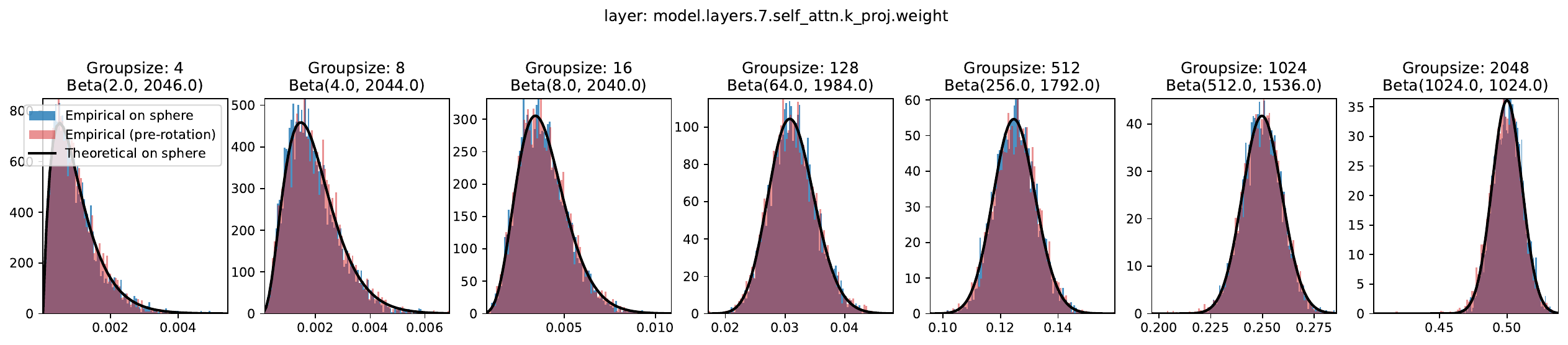} &
\includegraphics[width=0.22\linewidth]{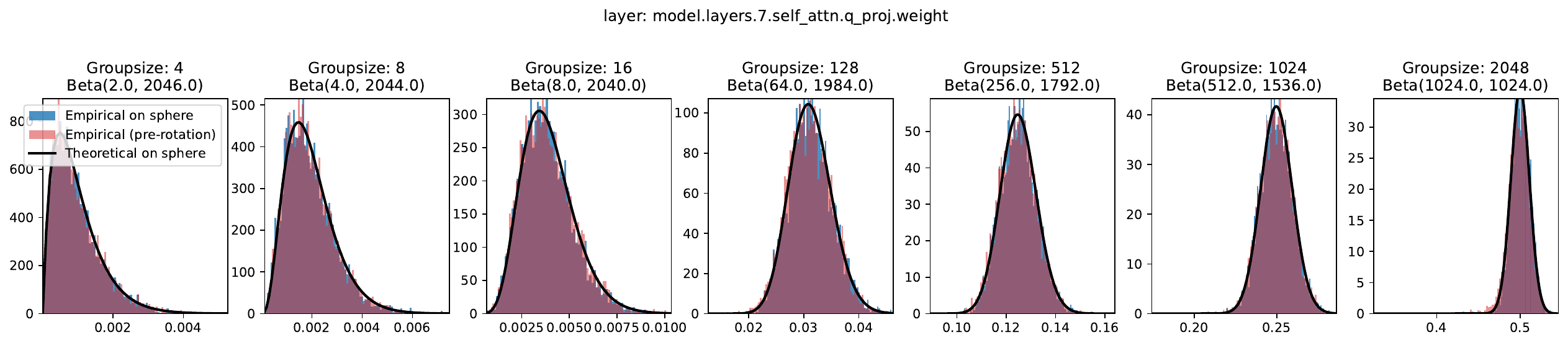} &
\includegraphics[width=0.22\linewidth]{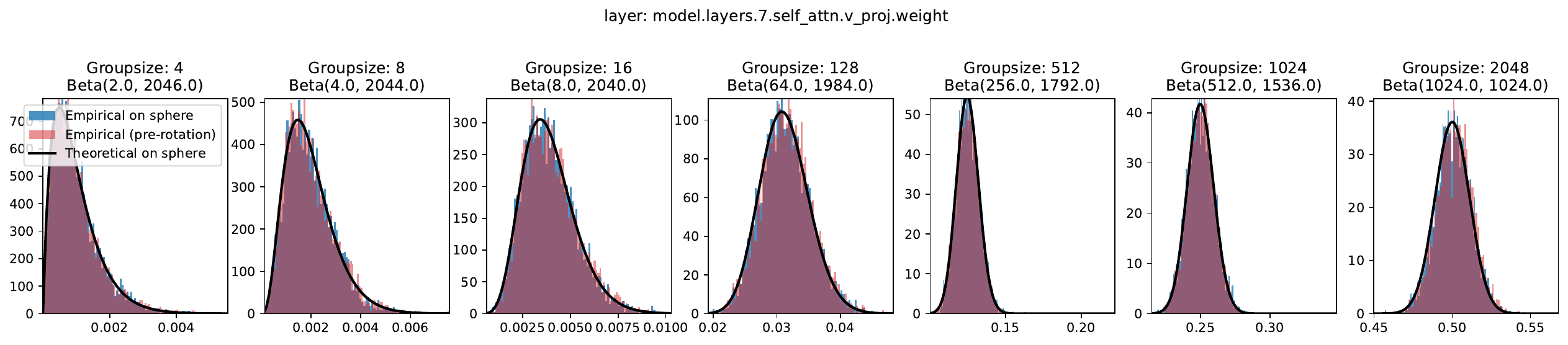} &
\includegraphics[width=0.22\linewidth]{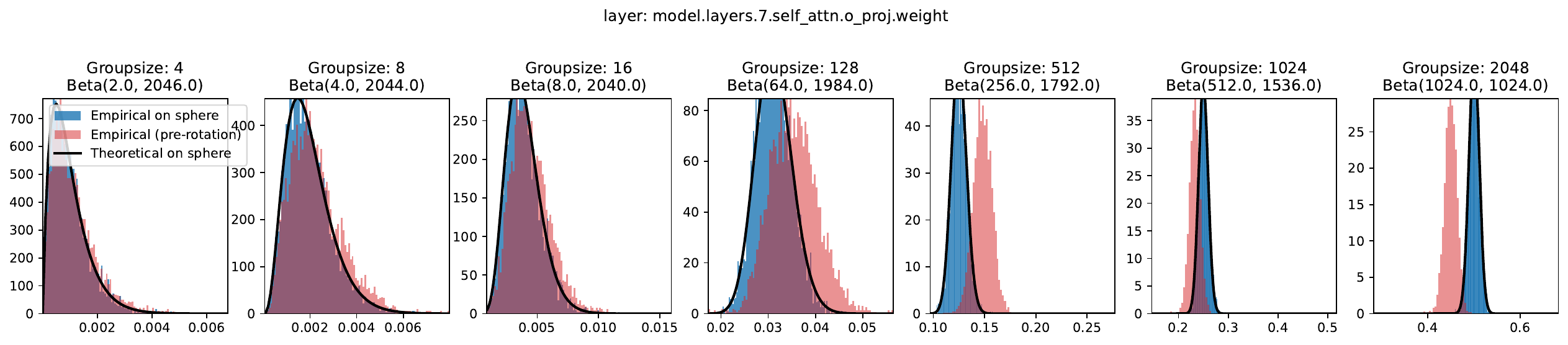} \\
\includegraphics[width=0.22\linewidth]{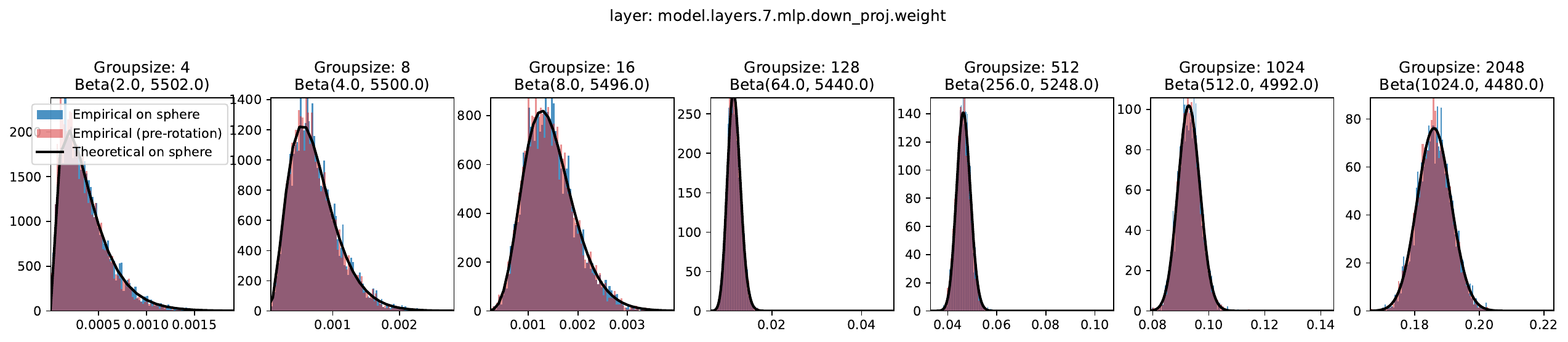} &
\includegraphics[width=0.22\linewidth]{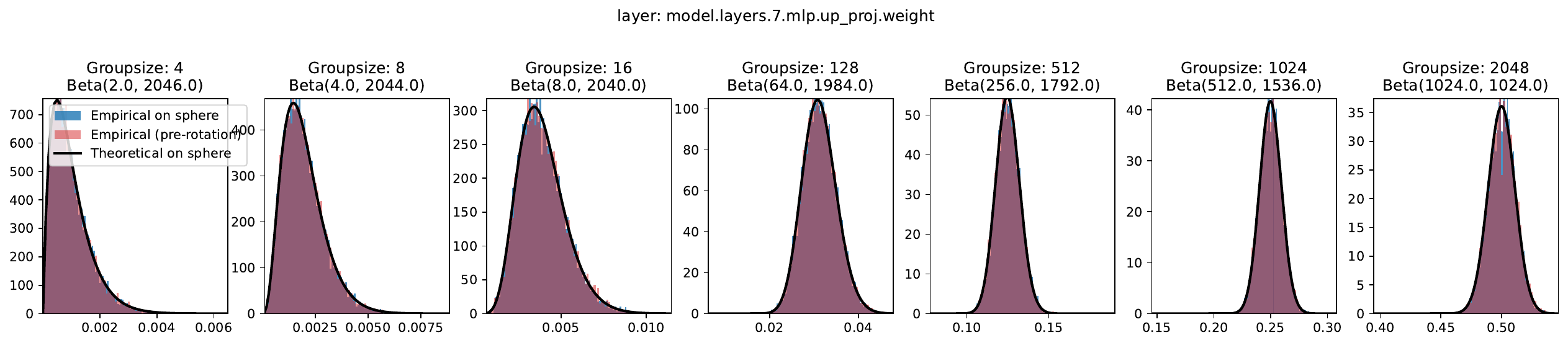} &
\includegraphics[width=0.22\linewidth]{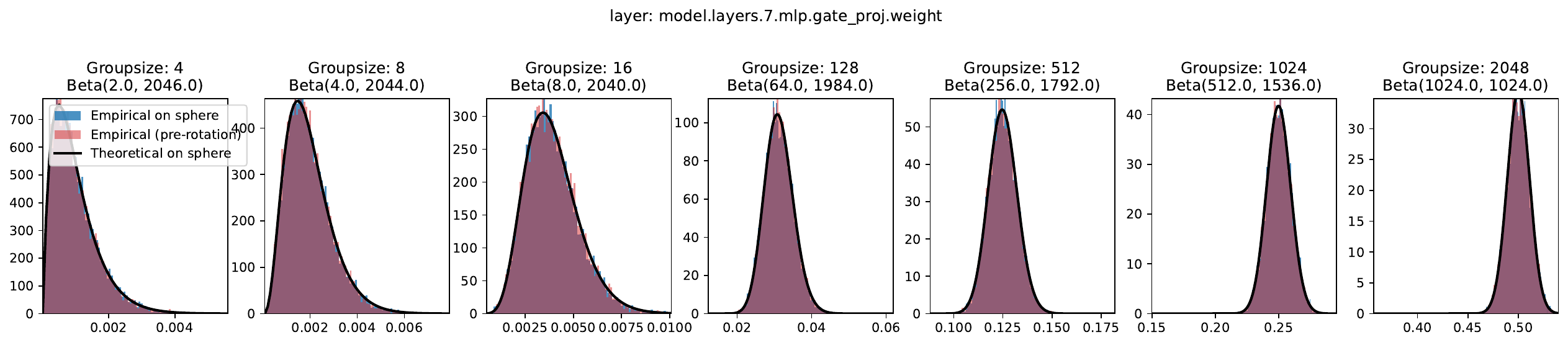} \\
\includegraphics[width=0.22\linewidth]{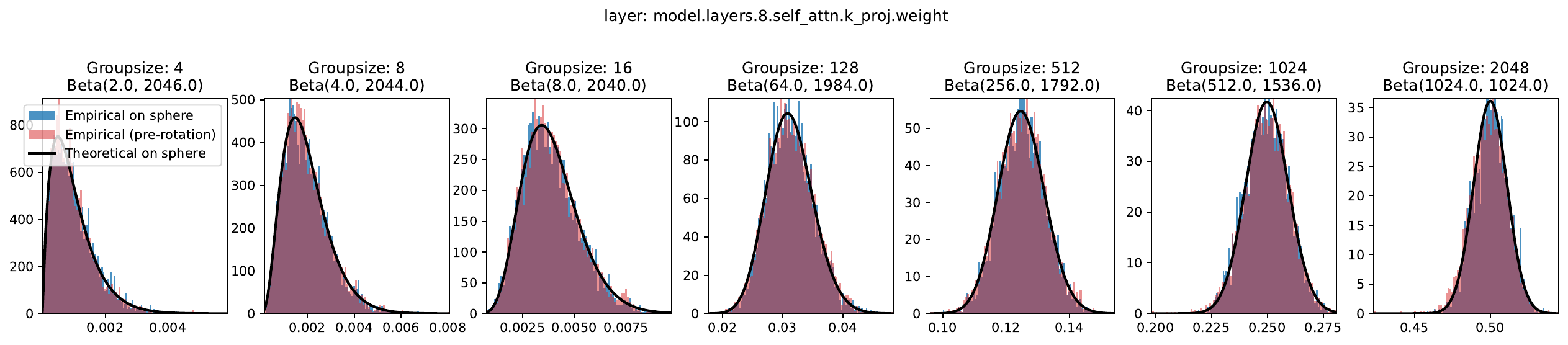} &
\includegraphics[width=0.22\linewidth]{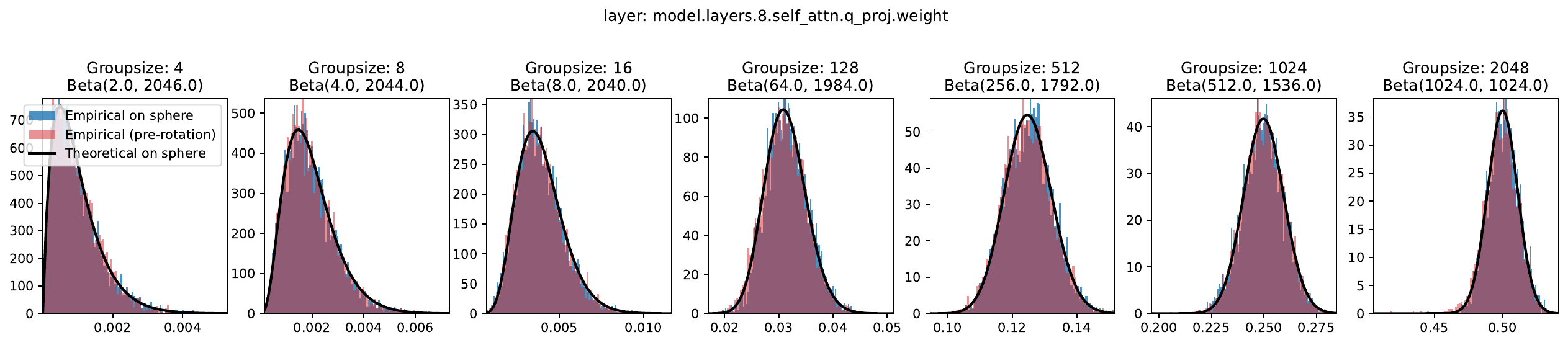} &
\includegraphics[width=0.22\linewidth]{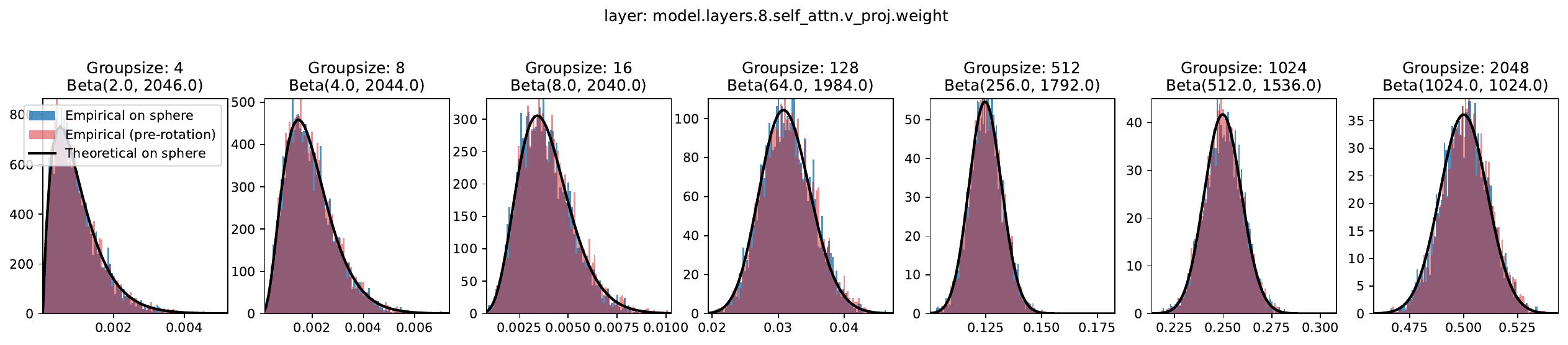} &
\includegraphics[width=0.22\linewidth]{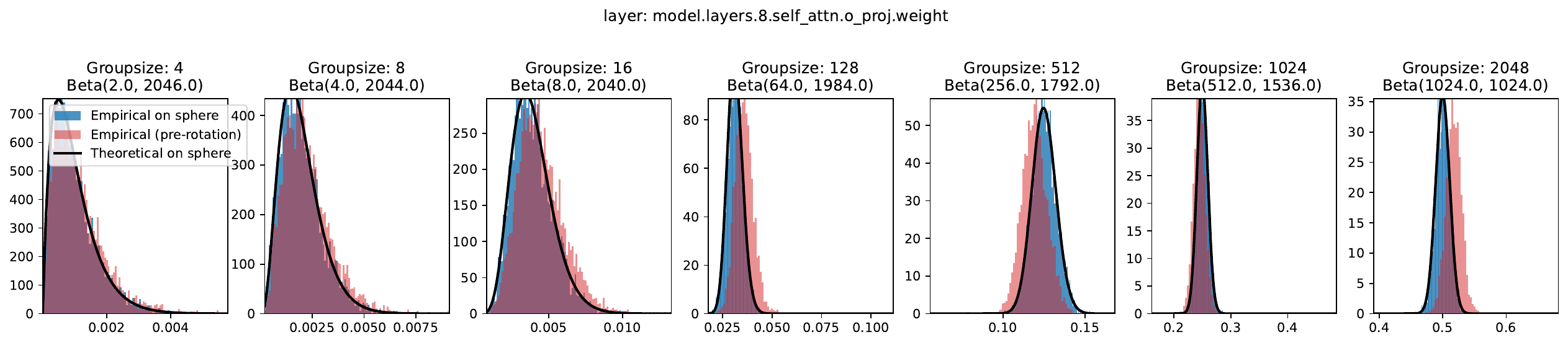} \\
\includegraphics[width=0.22\linewidth]{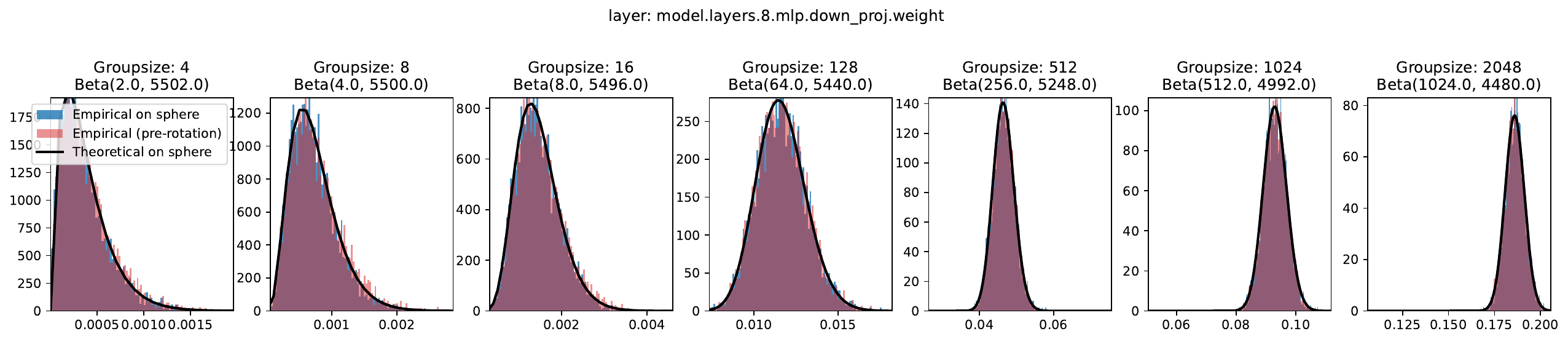} &
\includegraphics[width=0.22\linewidth]{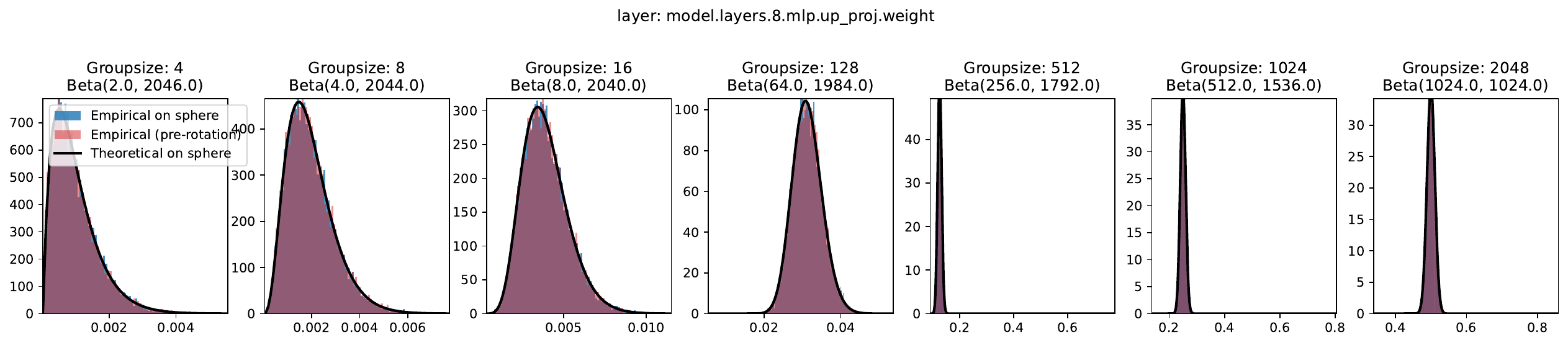} &
\includegraphics[width=0.22\linewidth]{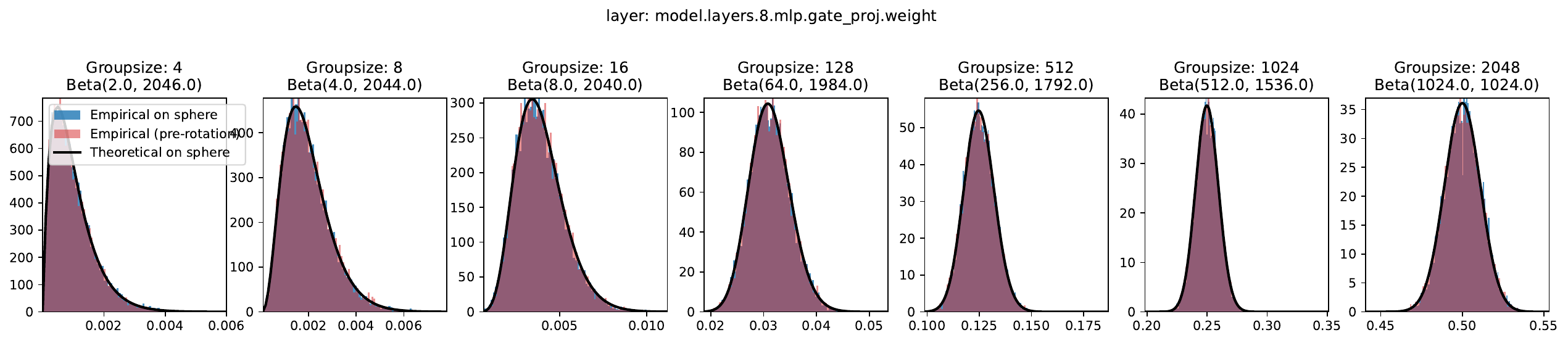} \\
\includegraphics[width=0.22\linewidth]{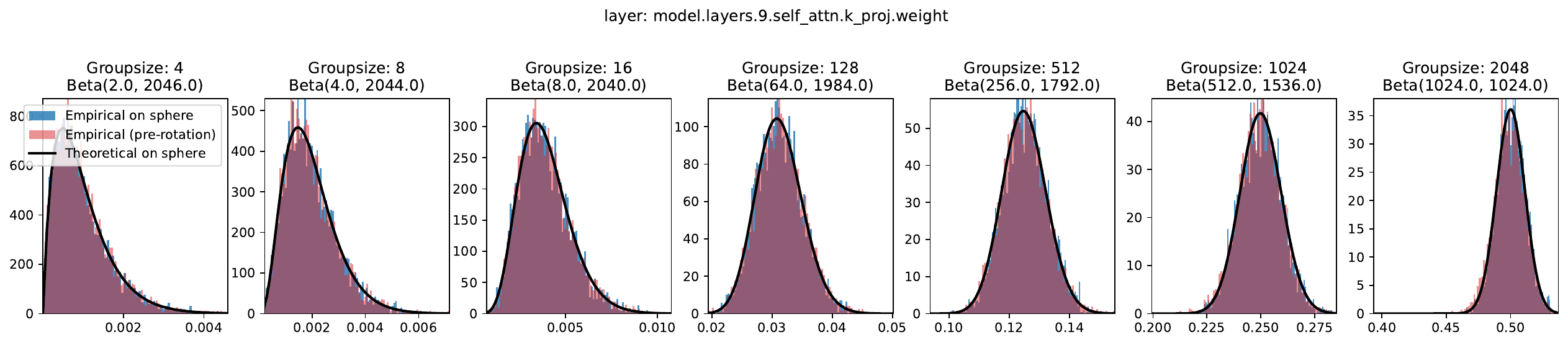} &
\includegraphics[width=0.22\linewidth]{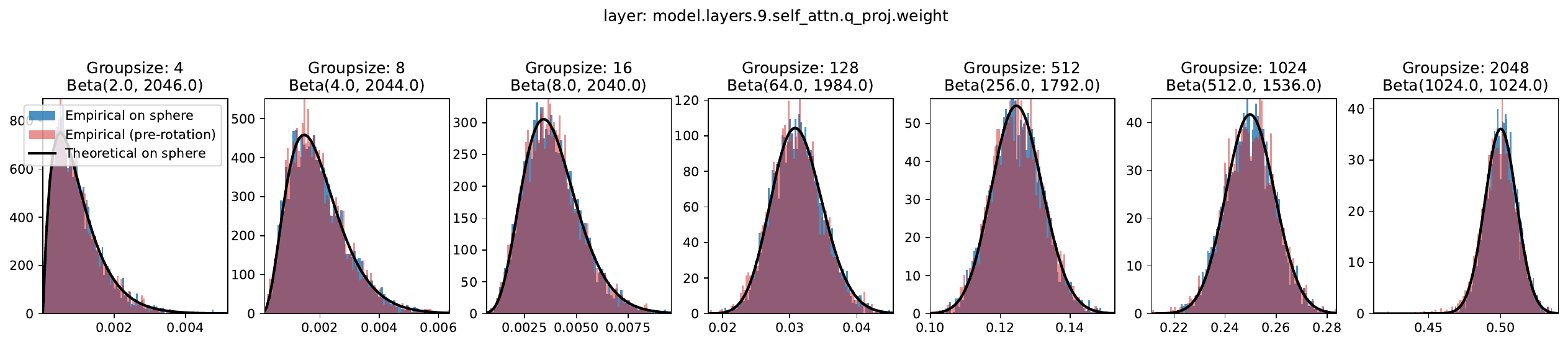} &
\includegraphics[width=0.22\linewidth]{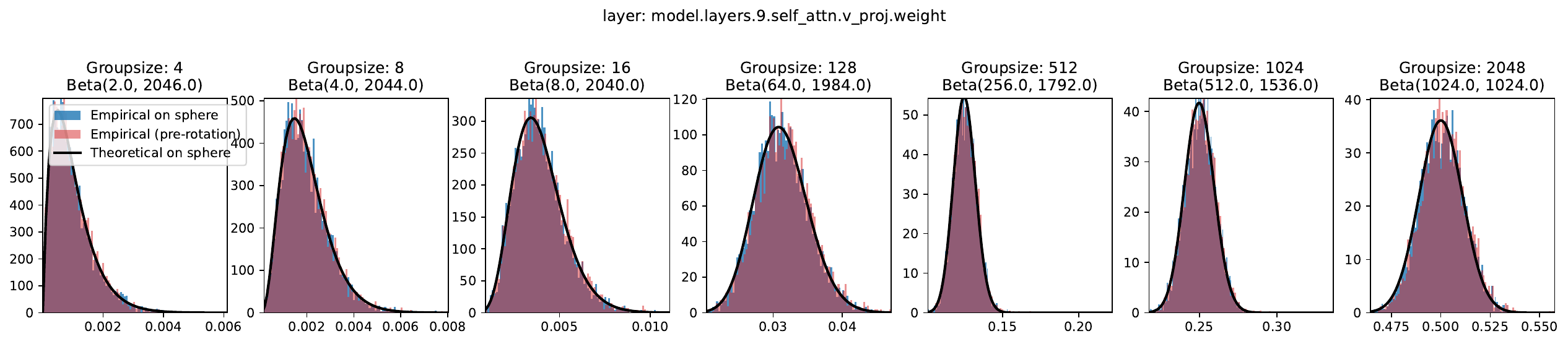} &
\includegraphics[width=0.22\linewidth]{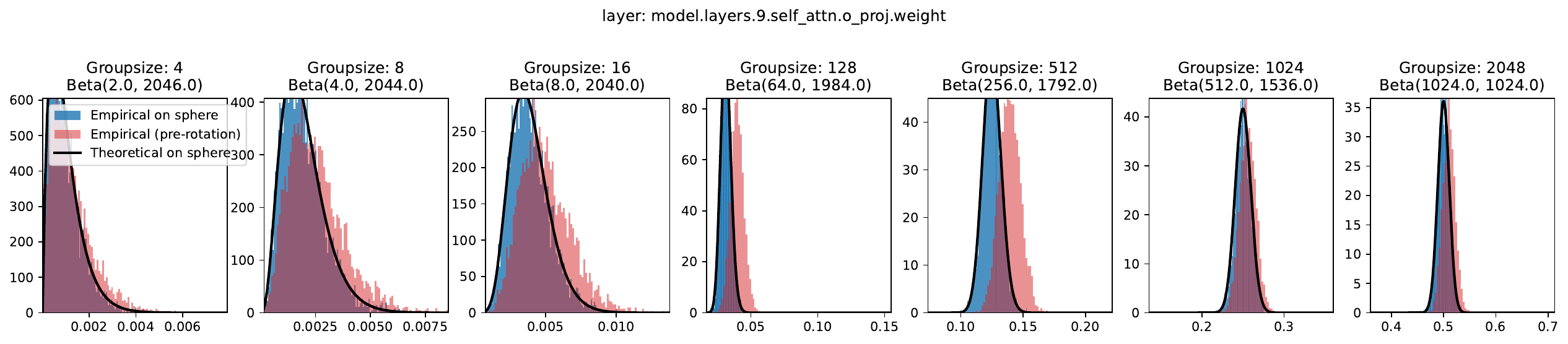} \\
\includegraphics[width=0.22\linewidth]{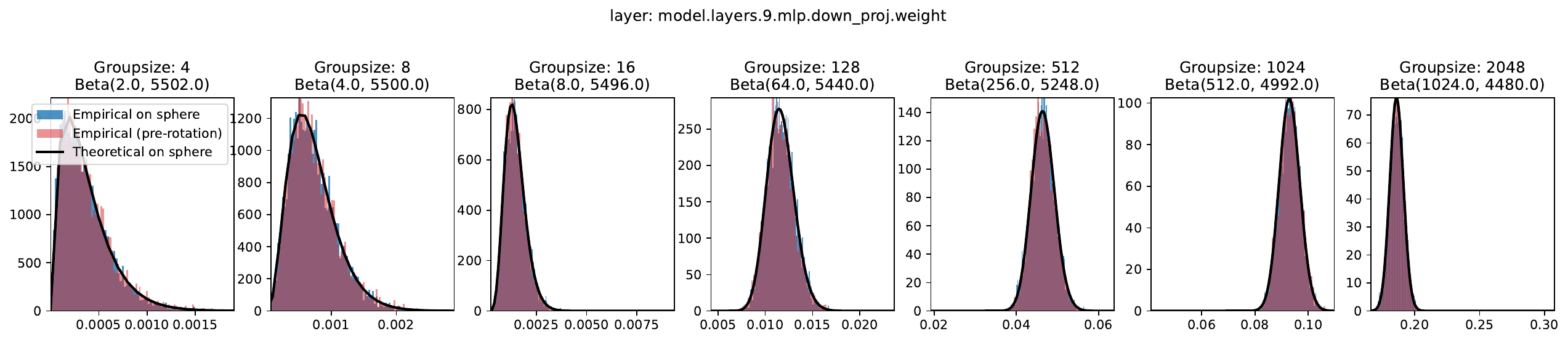} &
\includegraphics[width=0.22\linewidth]{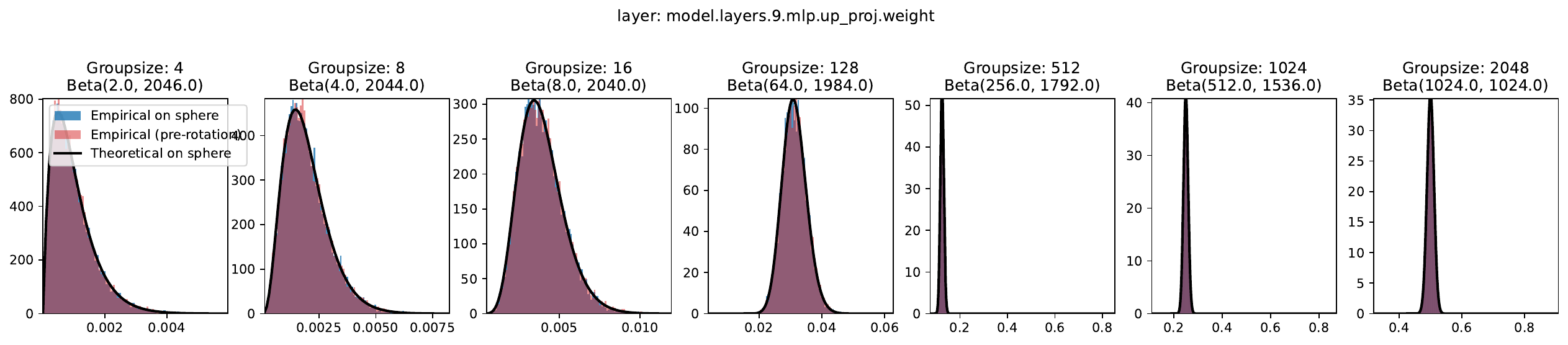} &
\includegraphics[width=0.22\linewidth]{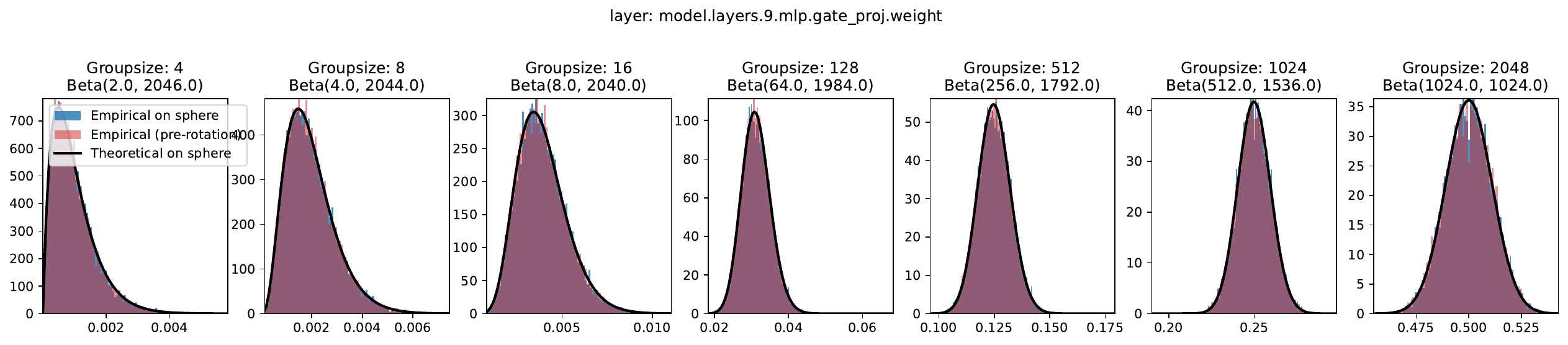} \\
\includegraphics[width=0.22\linewidth]{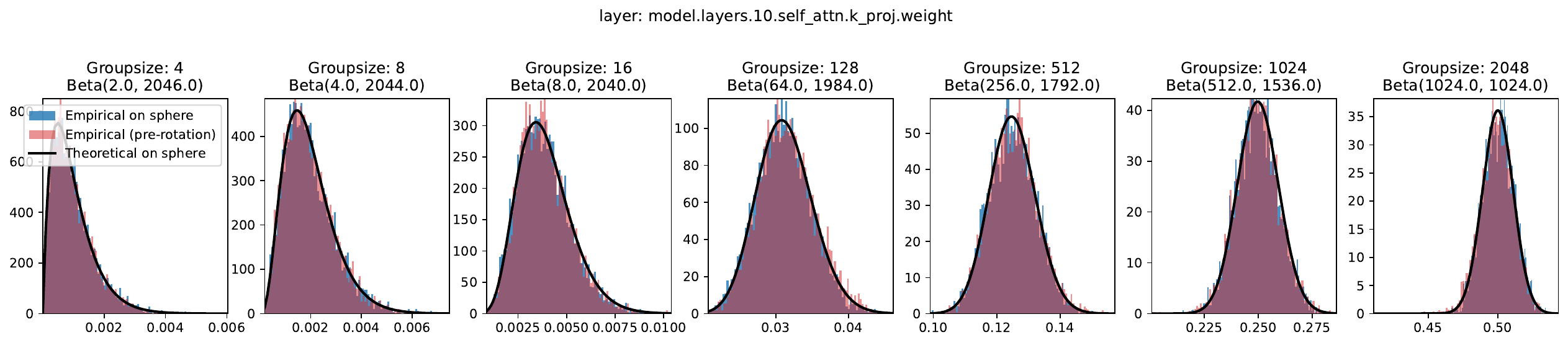} &
\includegraphics[width=0.22\linewidth]{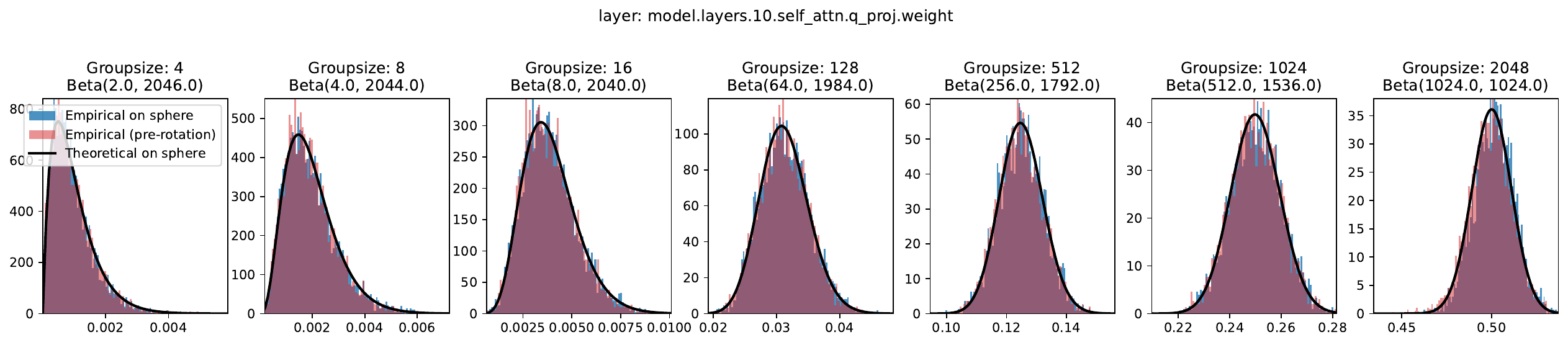} &
\includegraphics[width=0.22\linewidth]{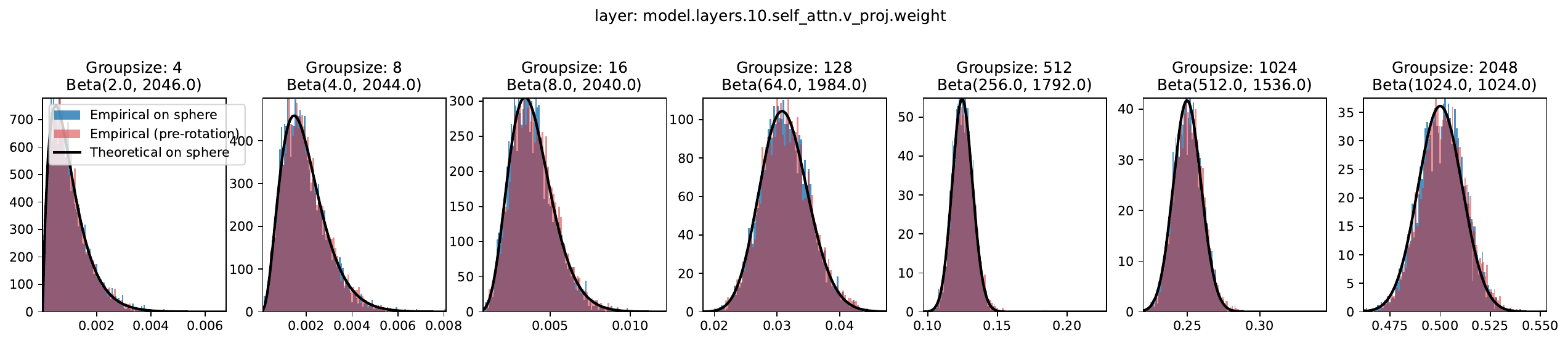} &
\includegraphics[width=0.22\linewidth]{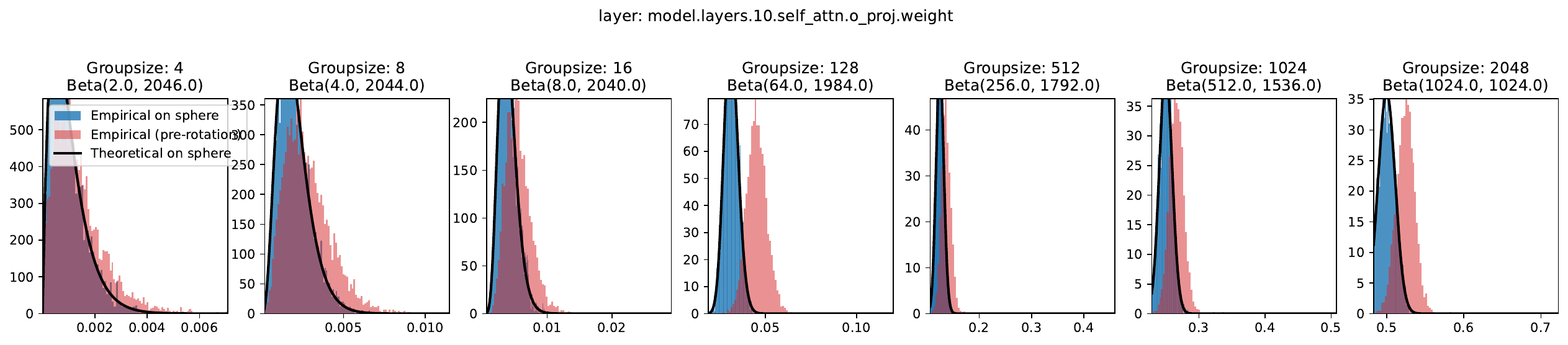} \\
\includegraphics[width=0.22\linewidth]{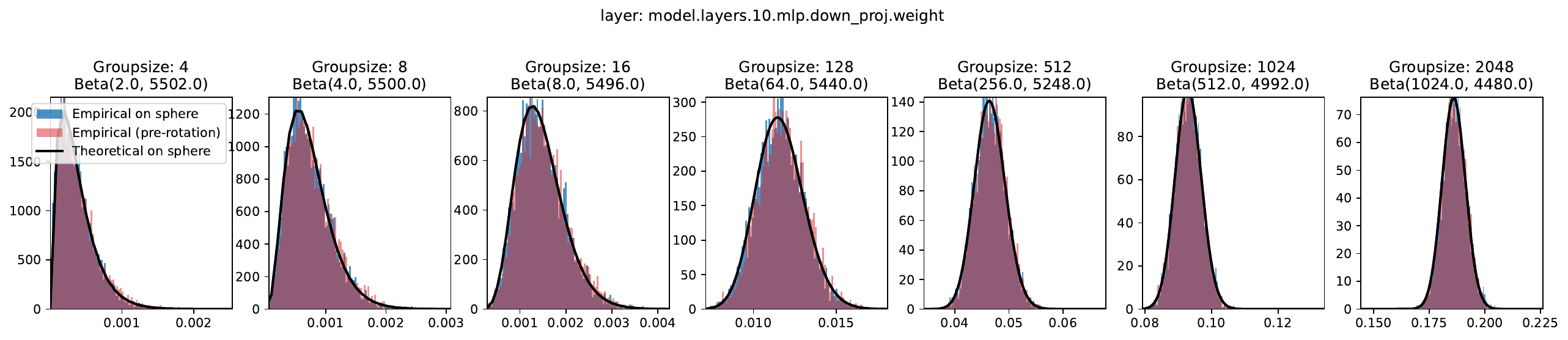} &
\includegraphics[width=0.22\linewidth]{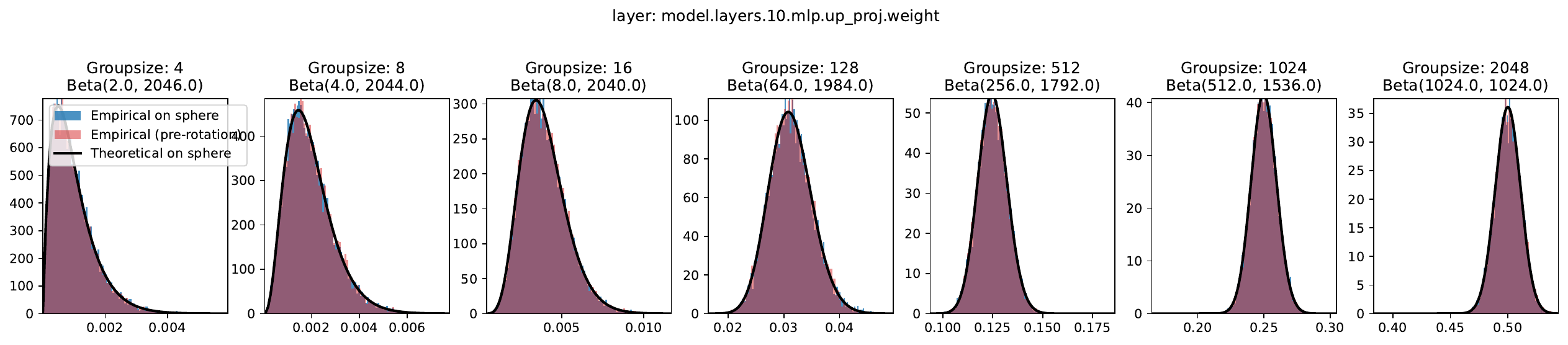} &
\includegraphics[width=0.22\linewidth]{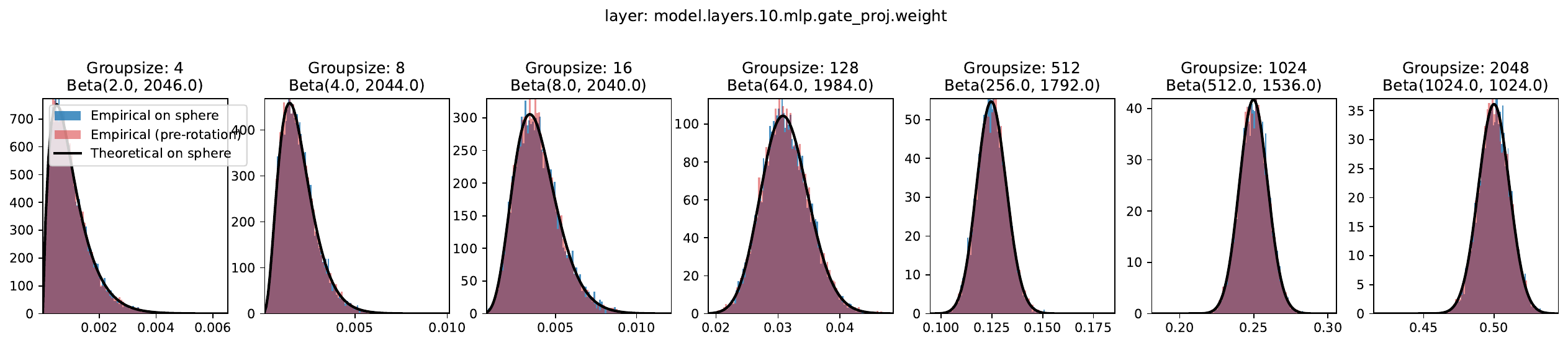} \\
\includegraphics[width=0.22\linewidth]{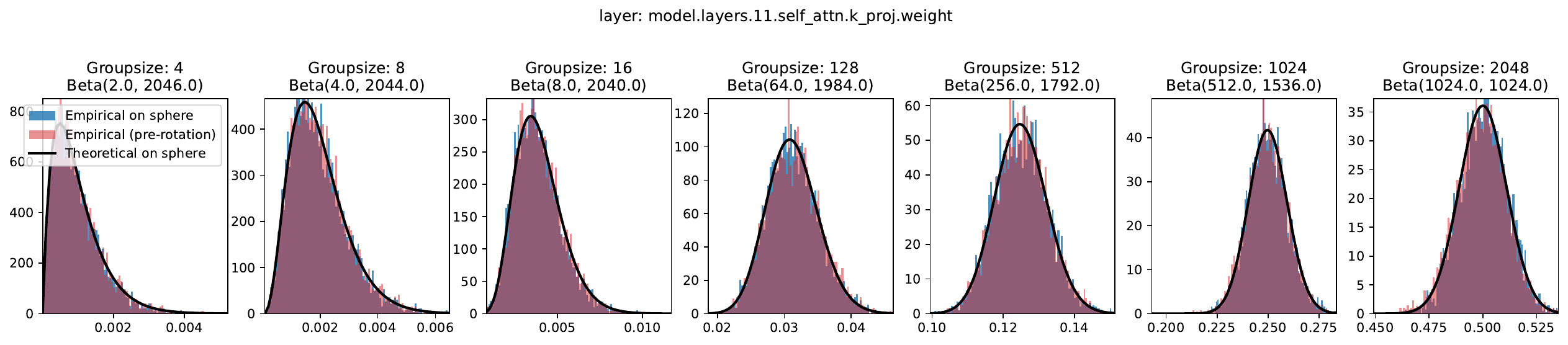} &
\includegraphics[width=0.22\linewidth]{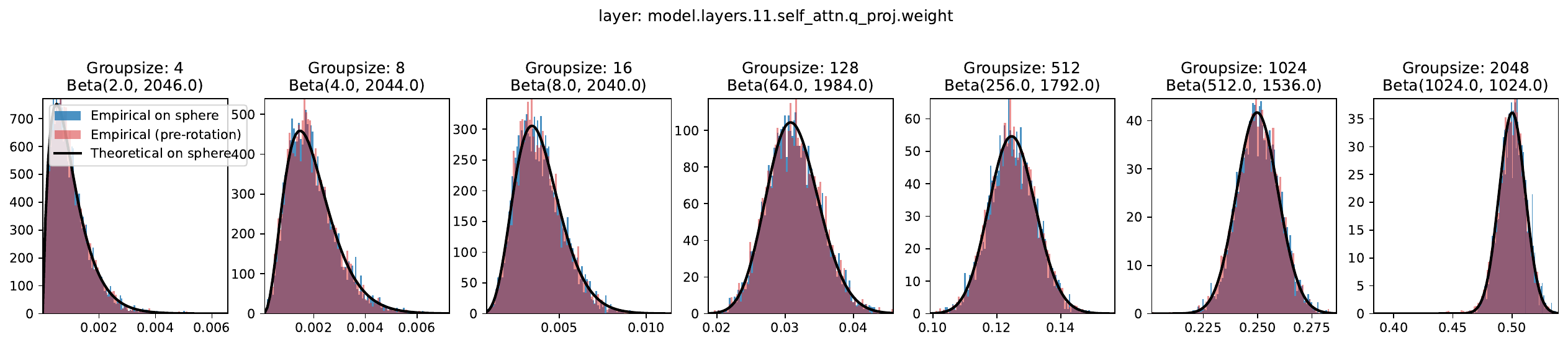} &
\includegraphics[width=0.22\linewidth]{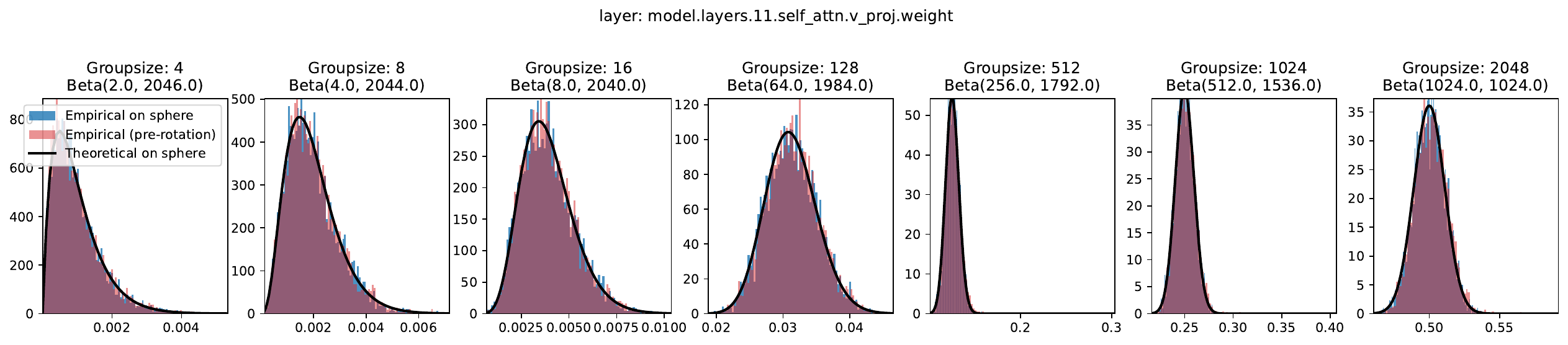} &
\includegraphics[width=0.22\linewidth]{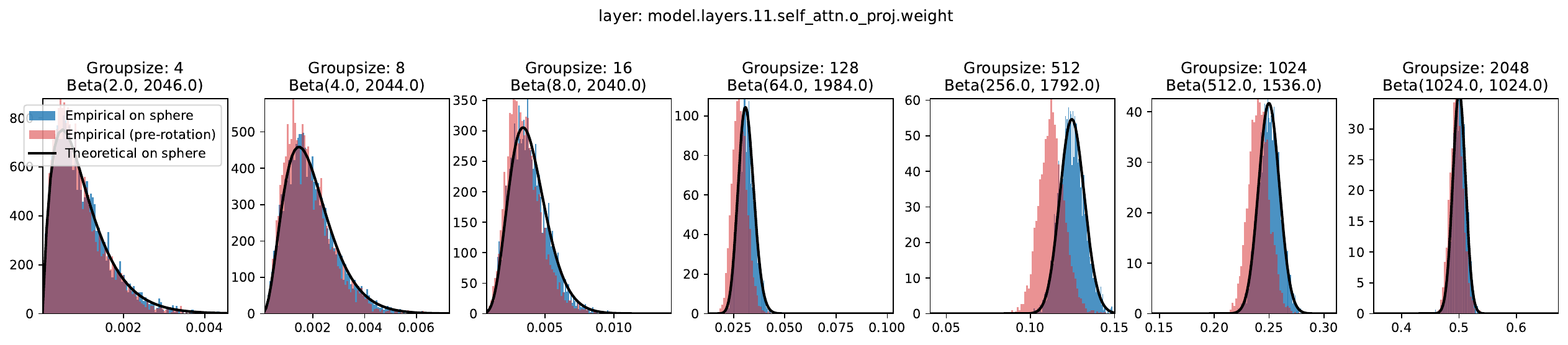} \\
\includegraphics[width=0.22\linewidth]{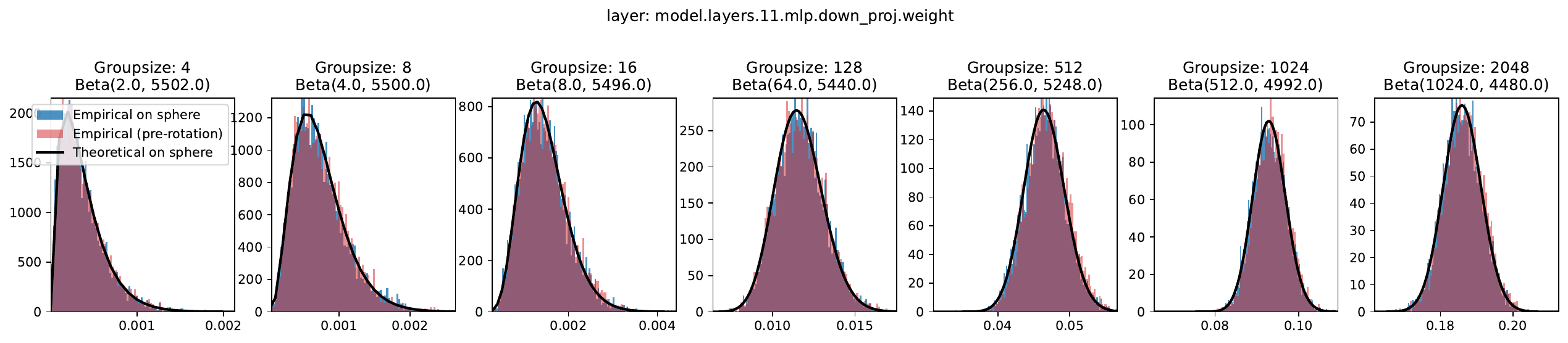} &
\includegraphics[width=0.22\linewidth]{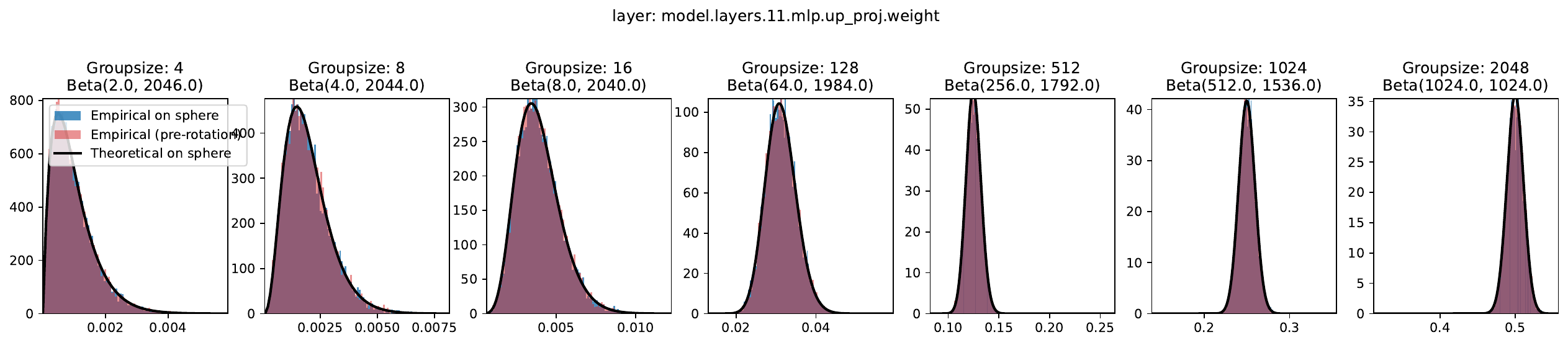} &
\includegraphics[width=0.22\linewidth]{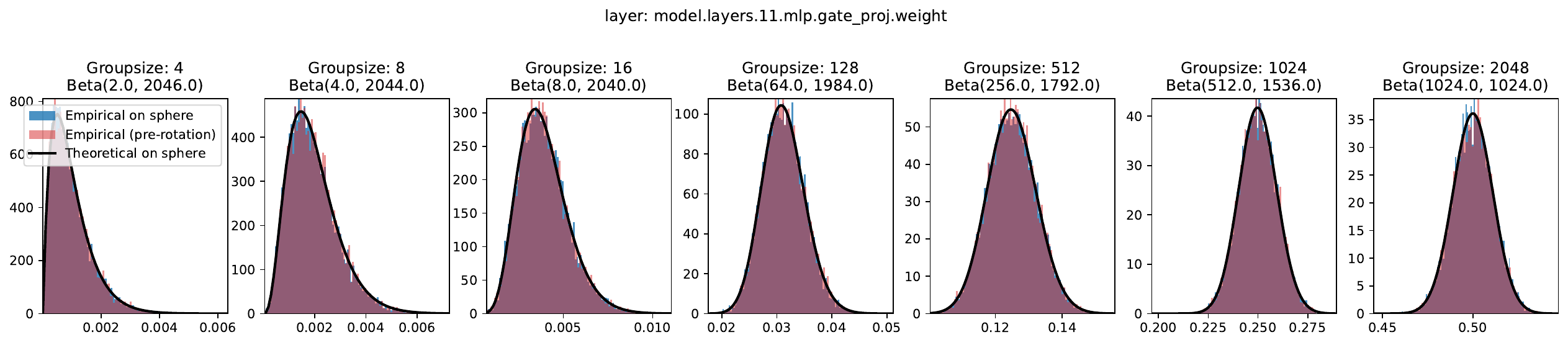} \\
\includegraphics[width=0.22\linewidth]{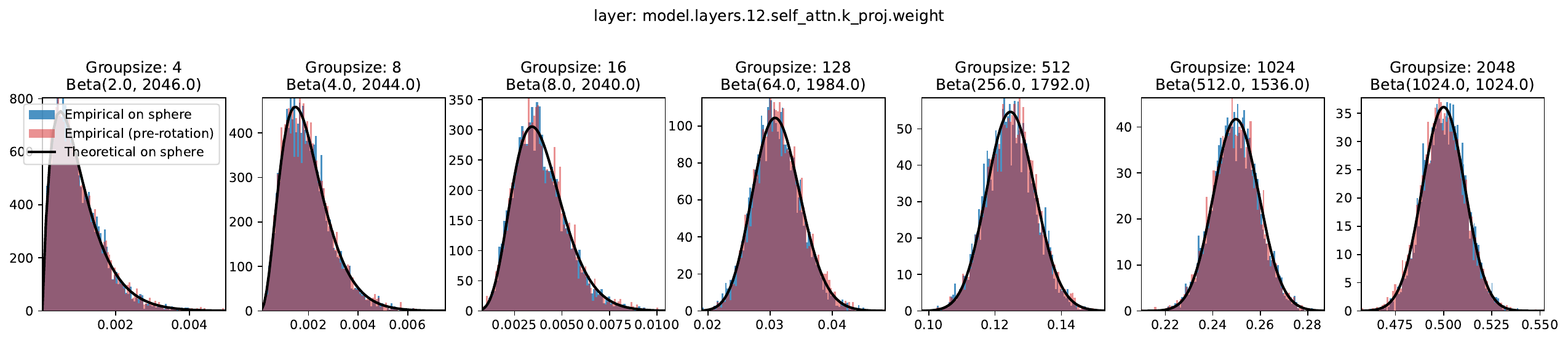} &
\includegraphics[width=0.22\linewidth]{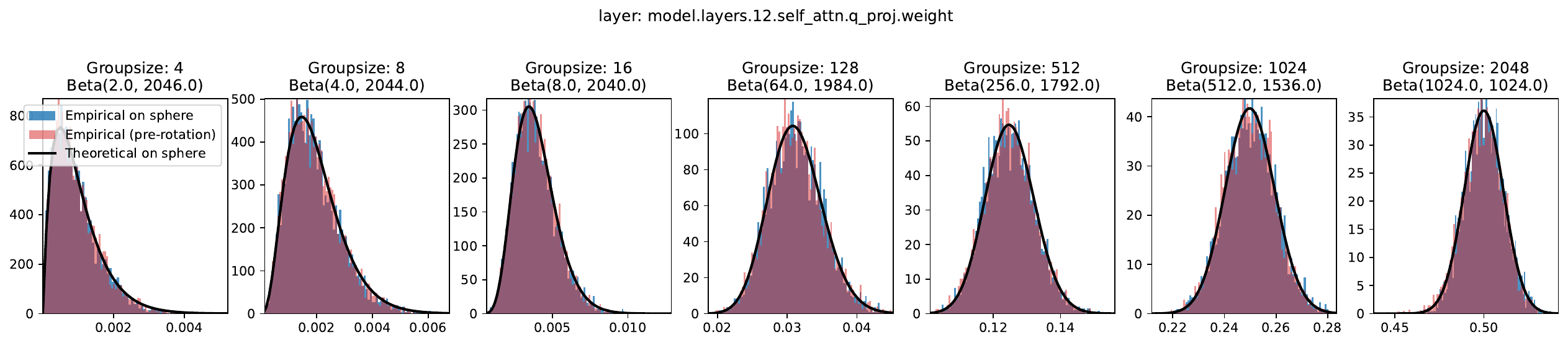} &
\includegraphics[width=0.22\linewidth]{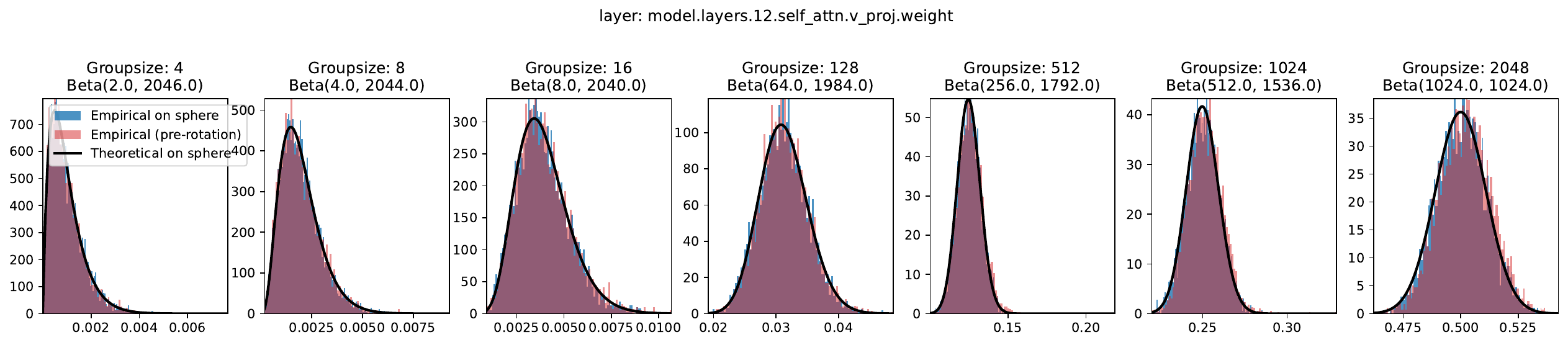} &
\includegraphics[width=0.22\linewidth]{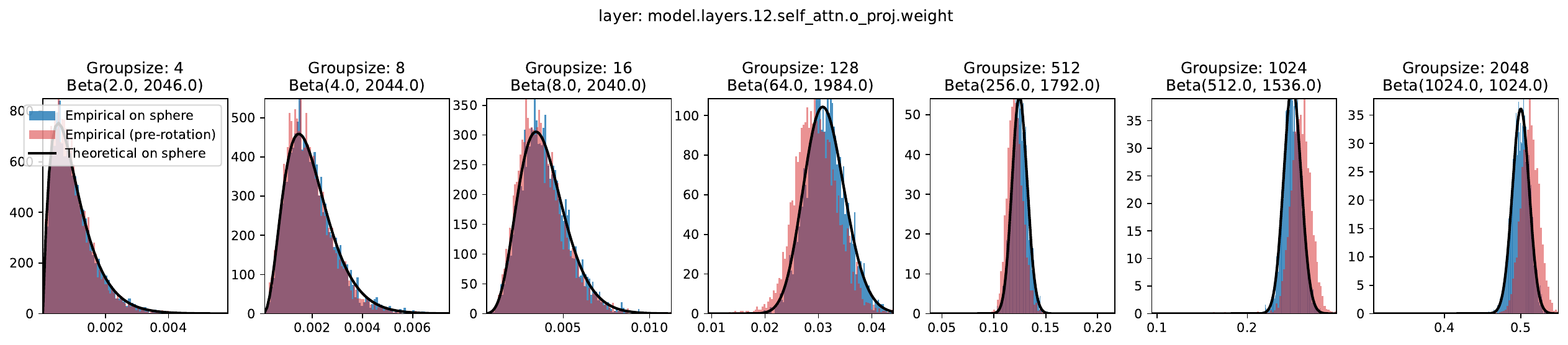} \\
\includegraphics[width=0.22\linewidth]{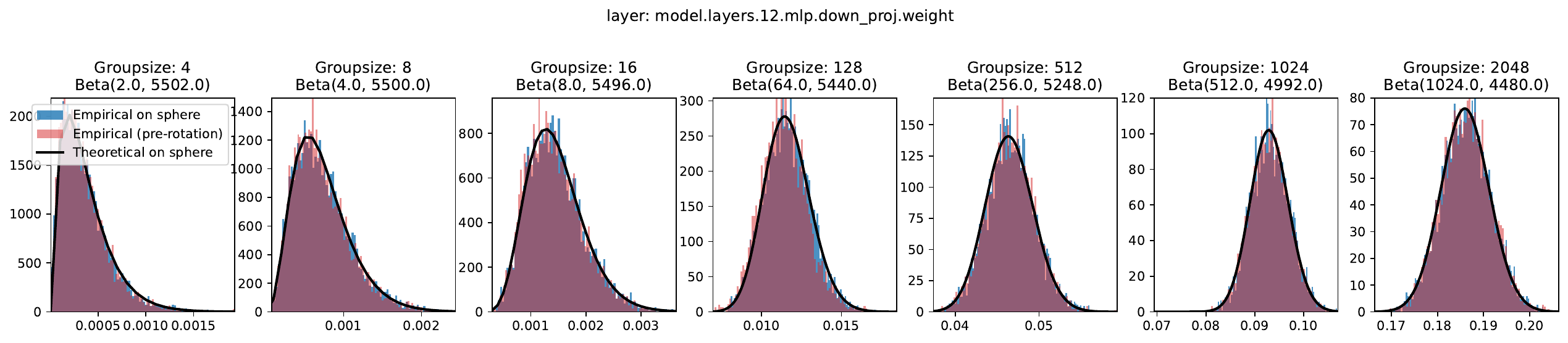} &
\includegraphics[width=0.22\linewidth]{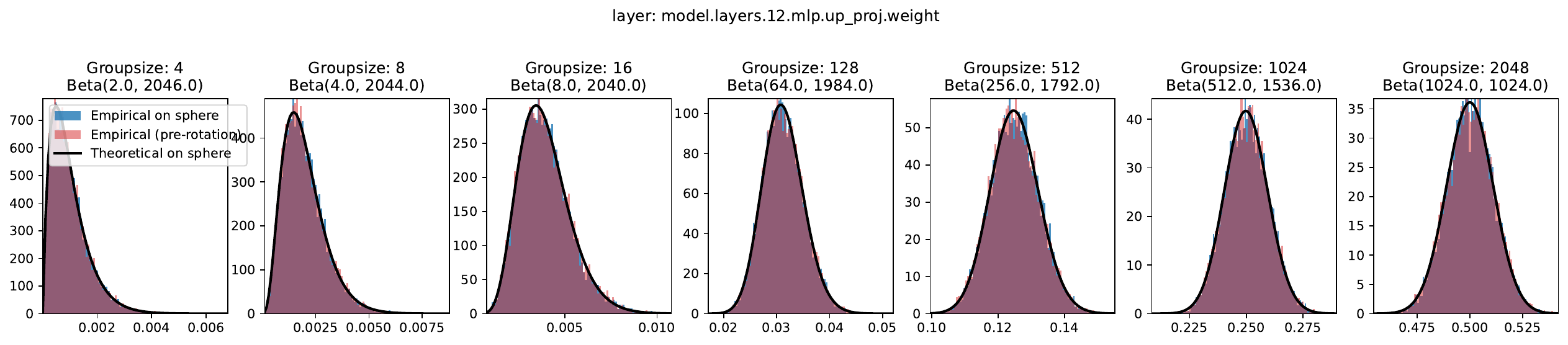} &
\includegraphics[width=0.22\linewidth]{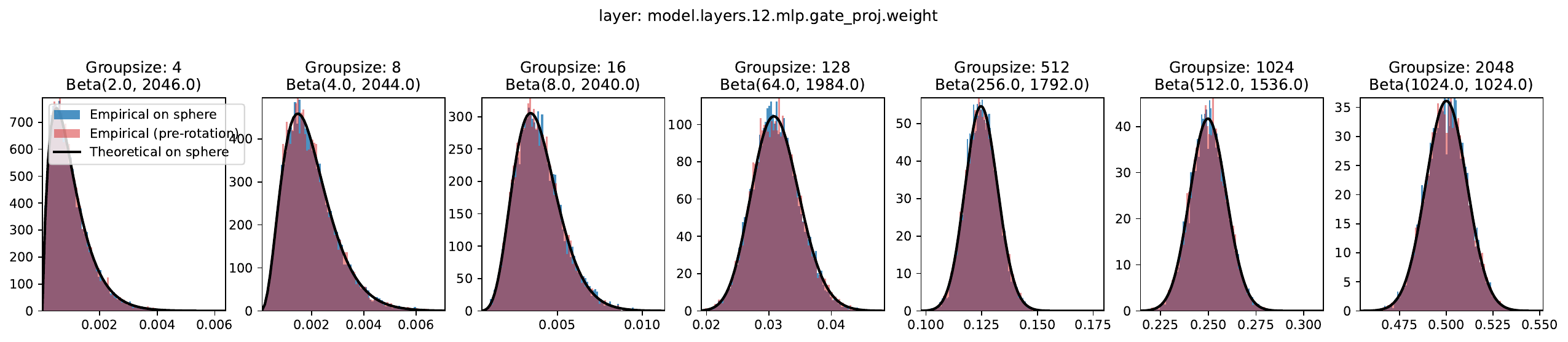} \\
\includegraphics[width=0.22\linewidth]{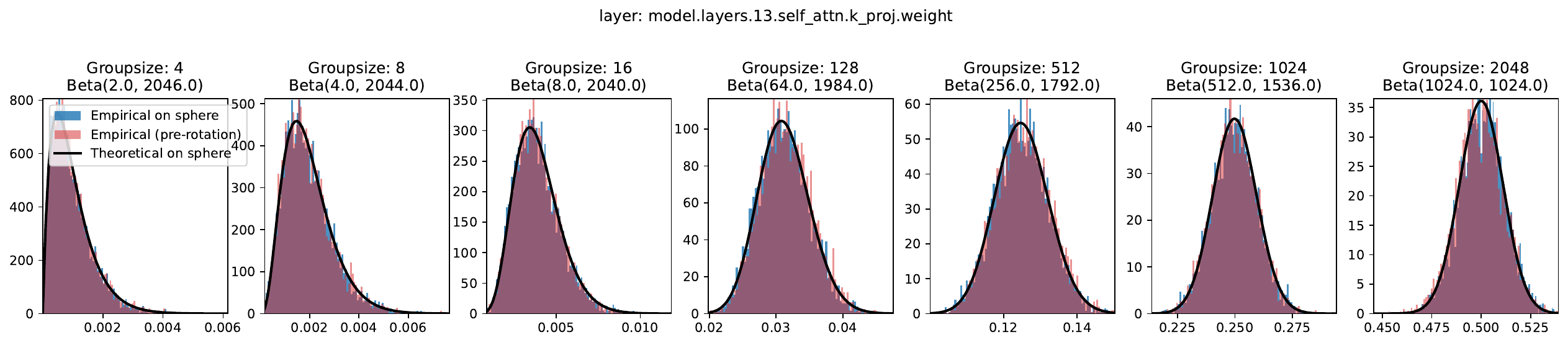} &
\includegraphics[width=0.22\linewidth]{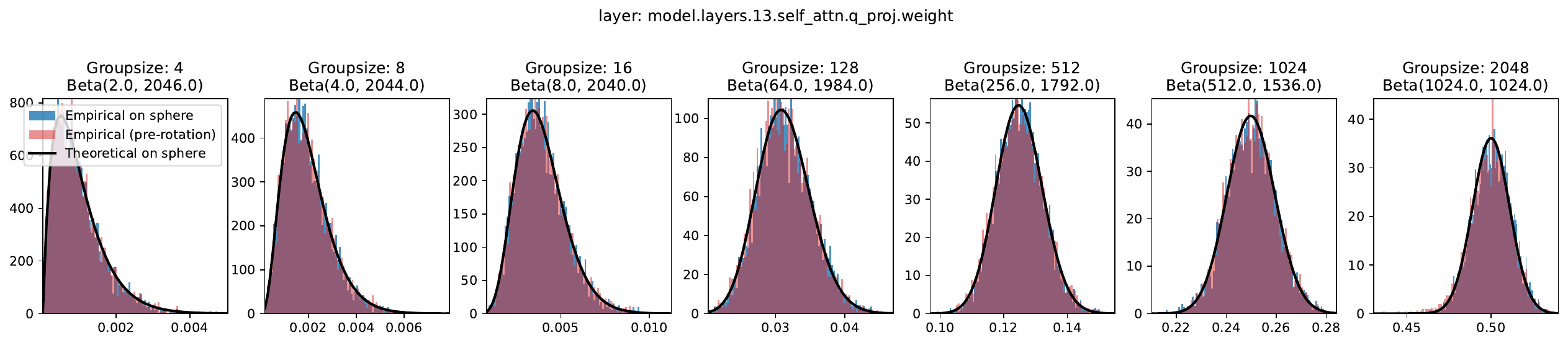} &
\includegraphics[width=0.22\linewidth]{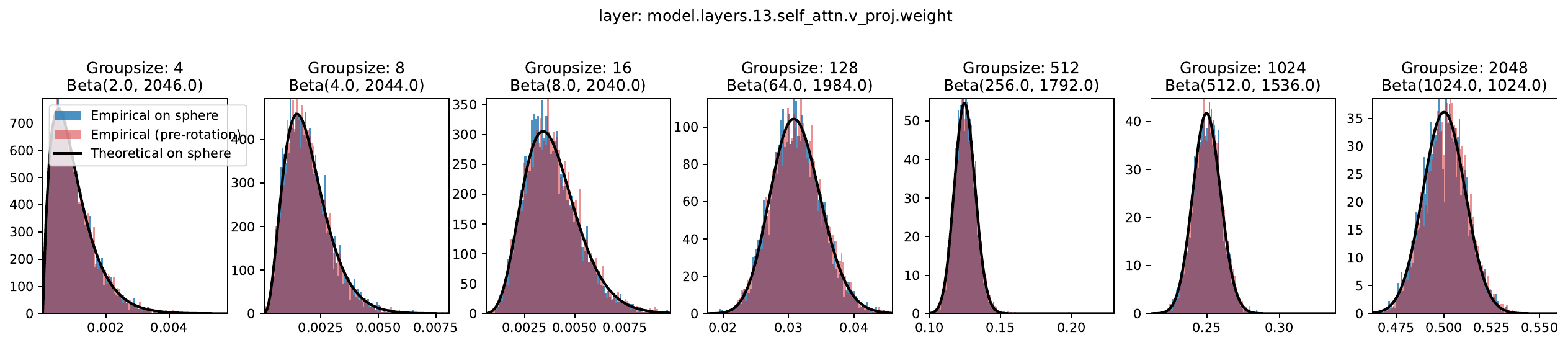} &
\includegraphics[width=0.22\linewidth]{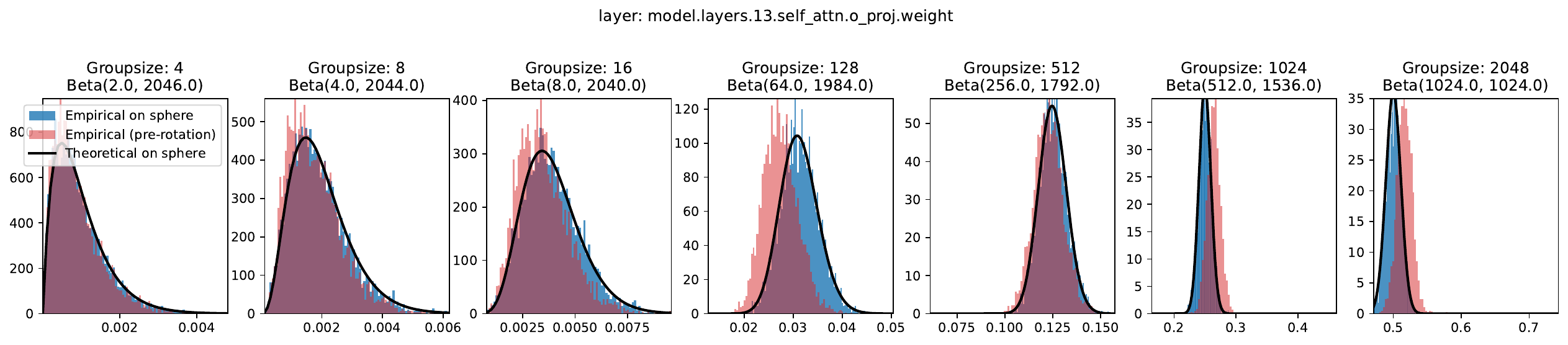} \\
\includegraphics[width=0.22\linewidth]{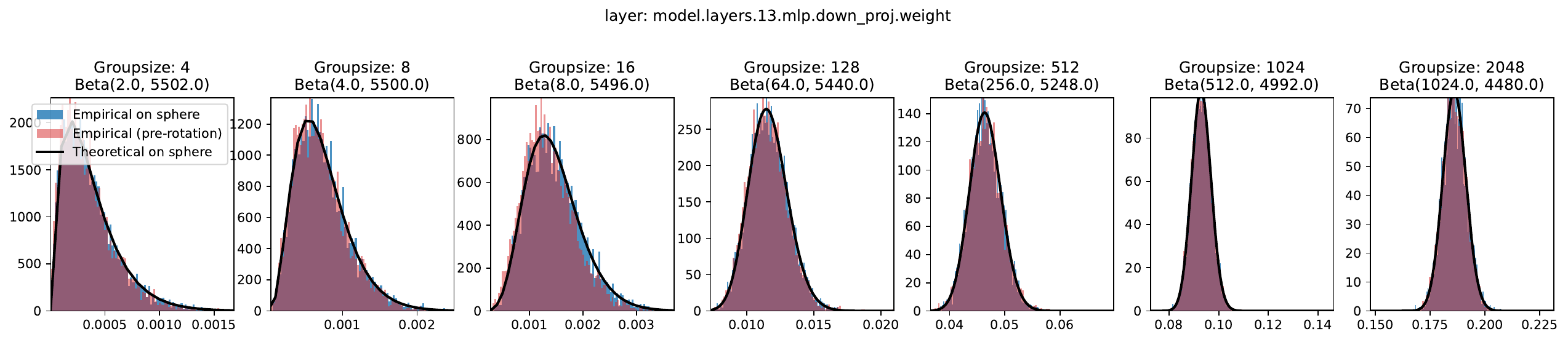} &
\includegraphics[width=0.22\linewidth]{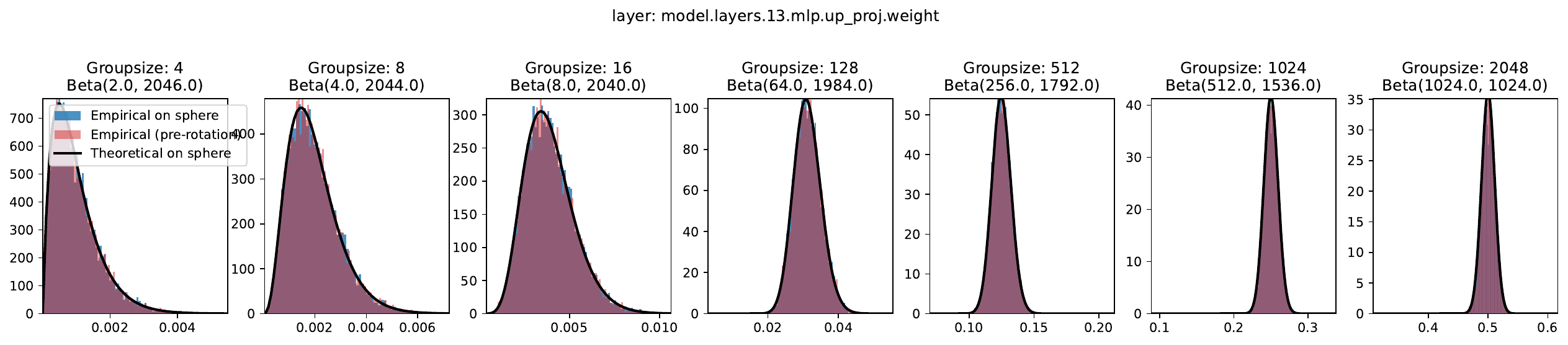} &
\includegraphics[width=0.22\linewidth]{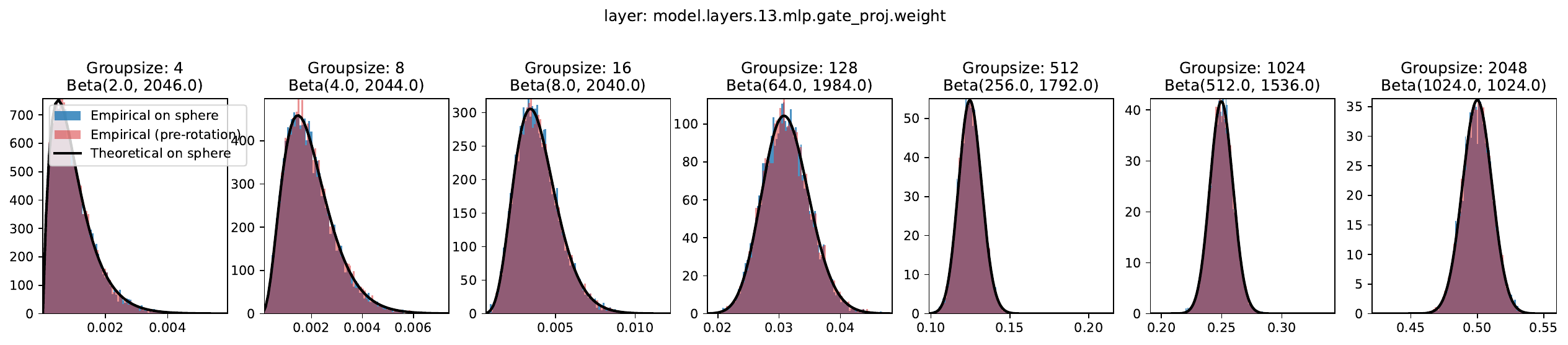} \\
\includegraphics[width=0.22\linewidth]{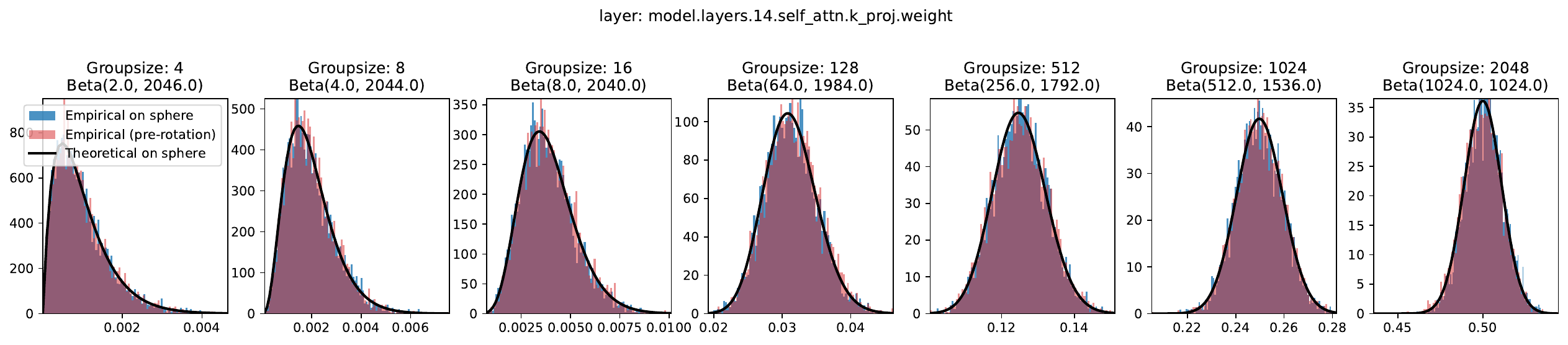} &
\includegraphics[width=0.22\linewidth]{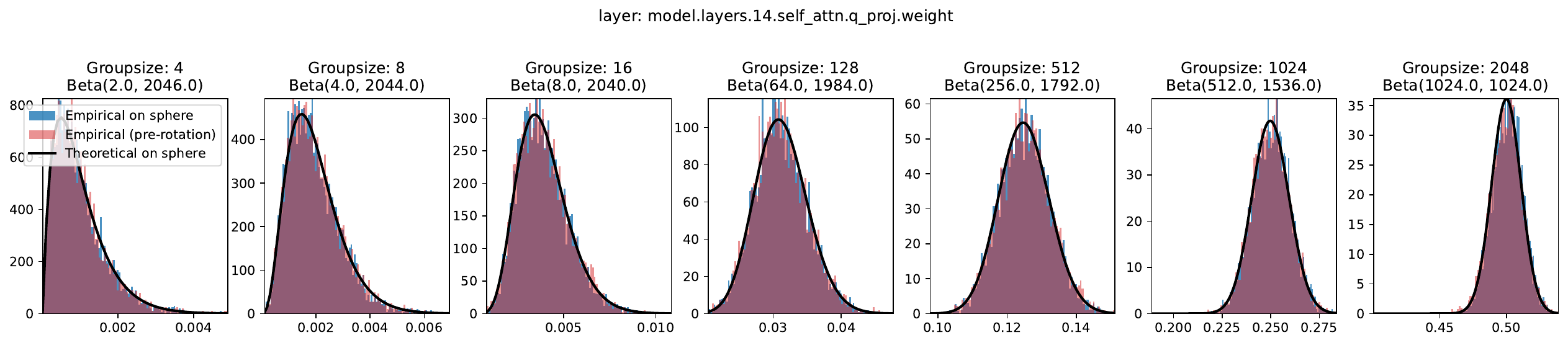} &
\includegraphics[width=0.22\linewidth]{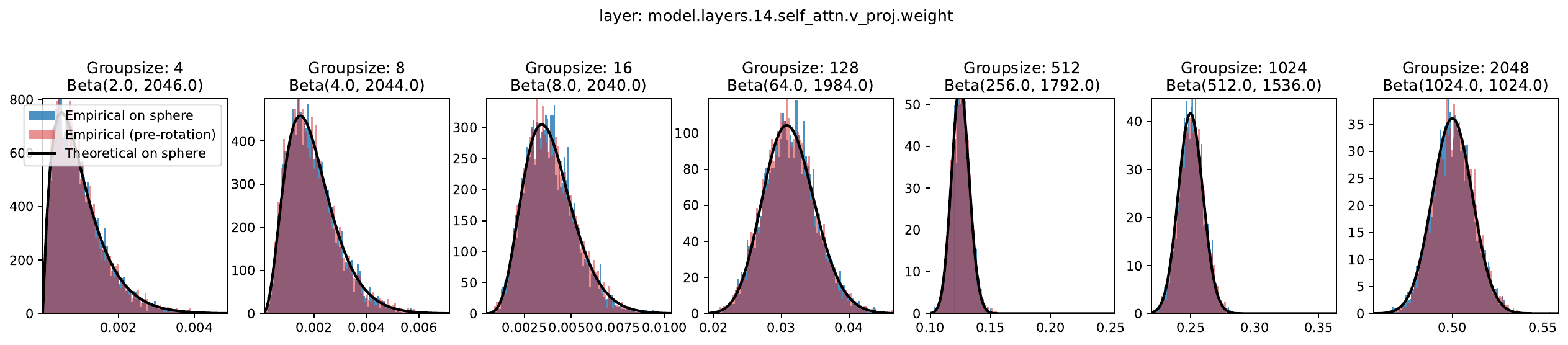} &
\includegraphics[width=0.22\linewidth]{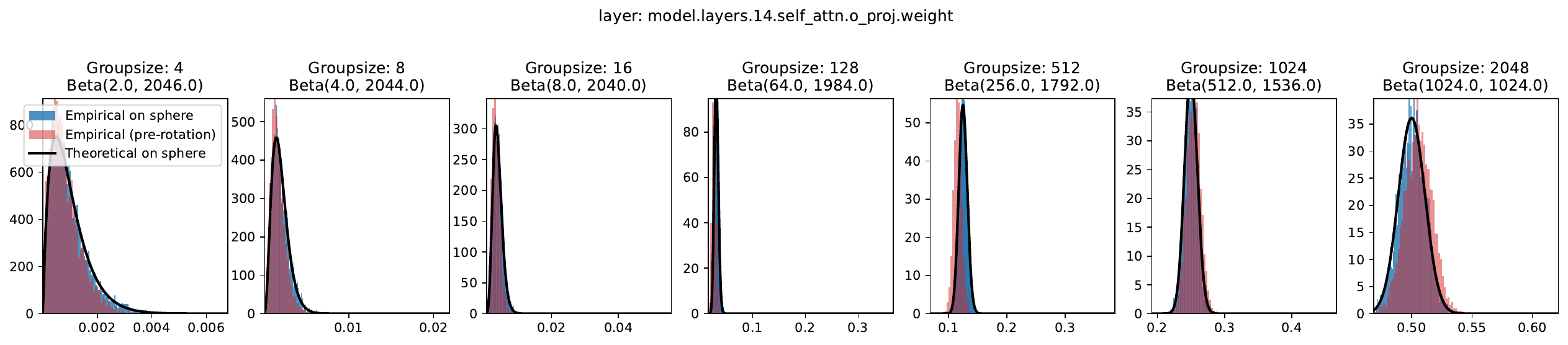} \\
\includegraphics[width=0.22\linewidth]{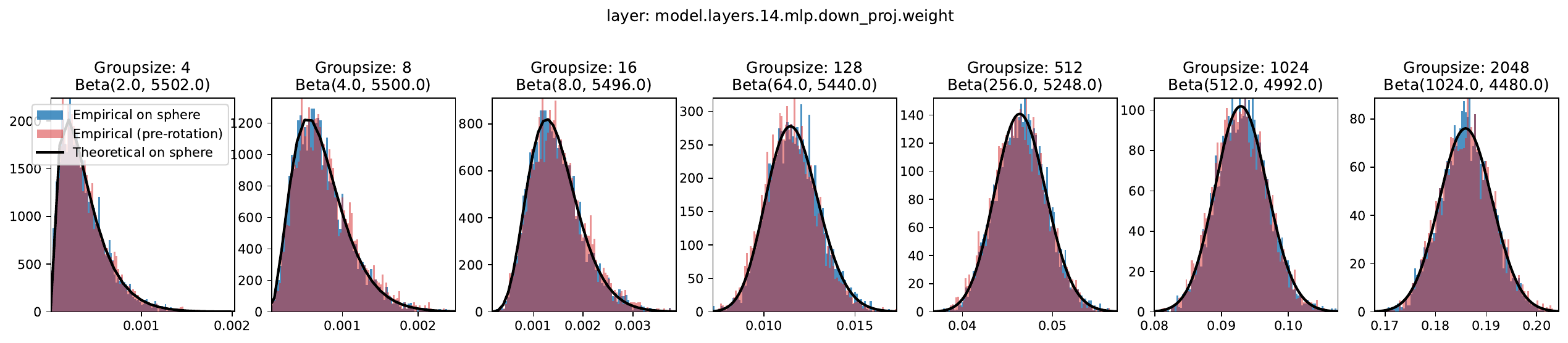} &
\includegraphics[width=0.22\linewidth]{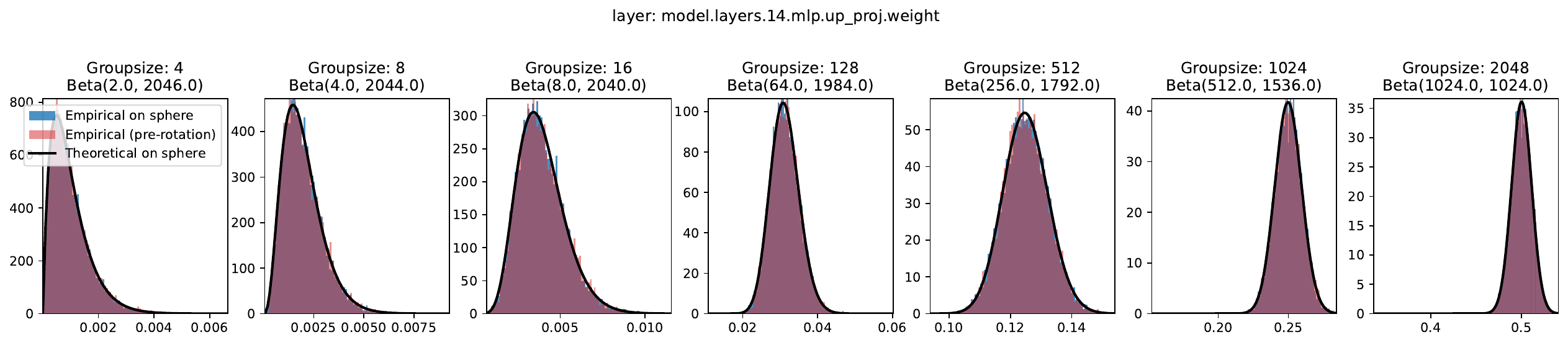} &
\includegraphics[width=0.22\linewidth]{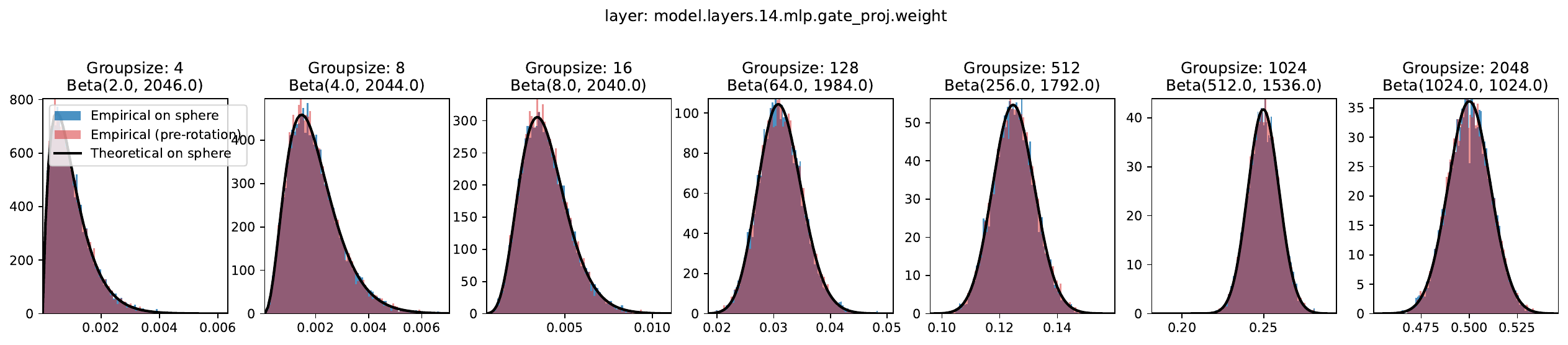} \\
\end{tabular}
\end{table}
\newpage
\begin{table}[H]
%\resizebox{!}{\linewidth}{
\begin{tabular}{llll}
\includegraphics[width=0.22\linewidth]{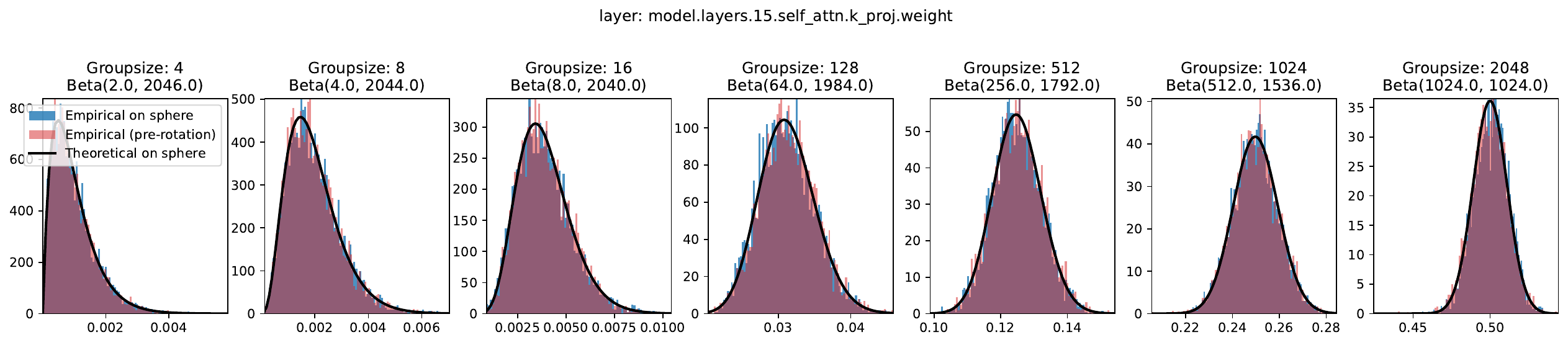} &
\includegraphics[width=0.22\linewidth]{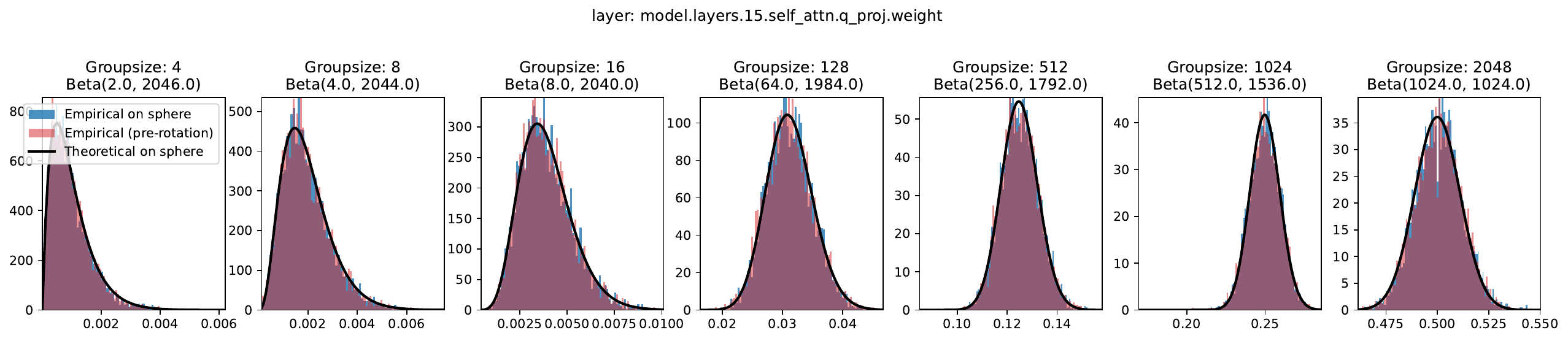} &
\includegraphics[width=0.22\linewidth]{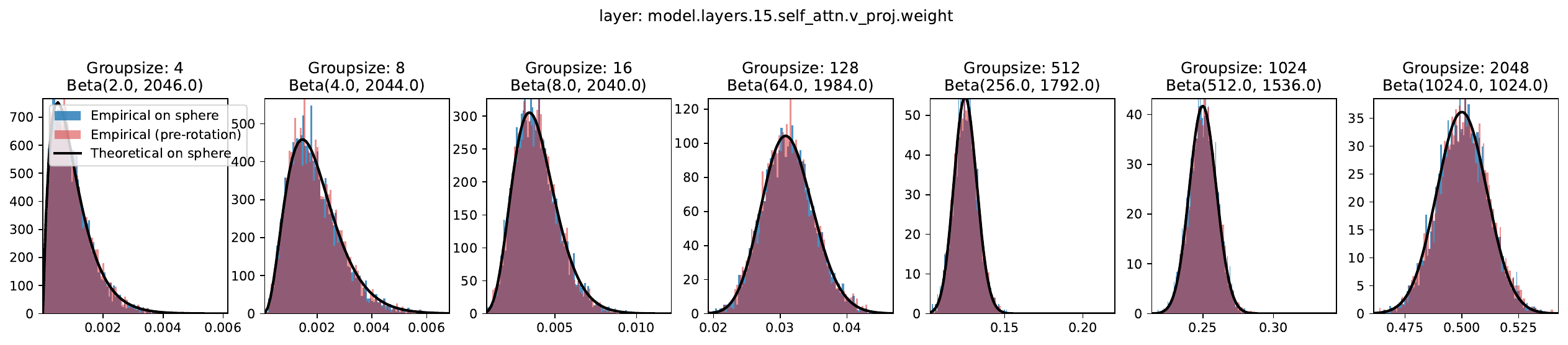} &
\includegraphics[width=0.22\linewidth]{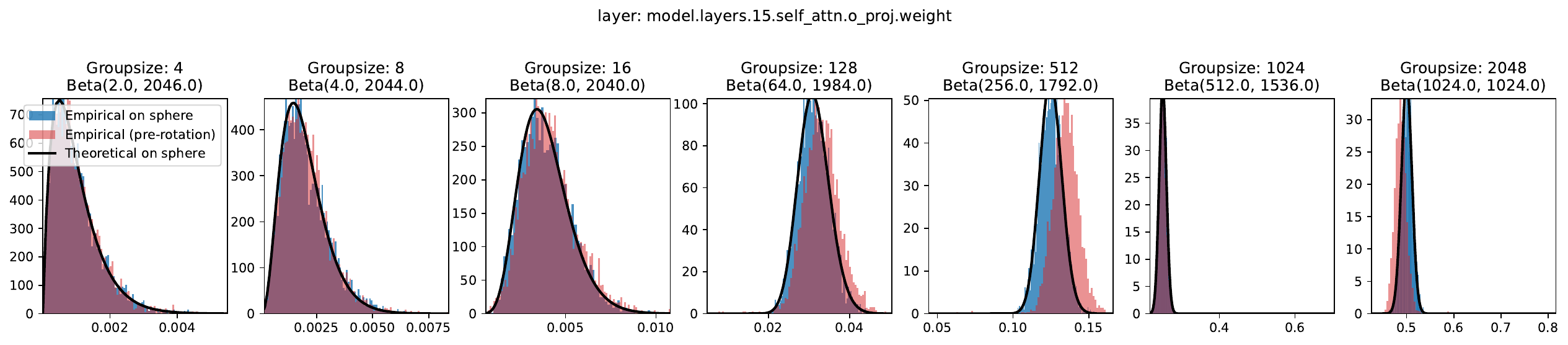} \\
\includegraphics[width=0.22\linewidth]{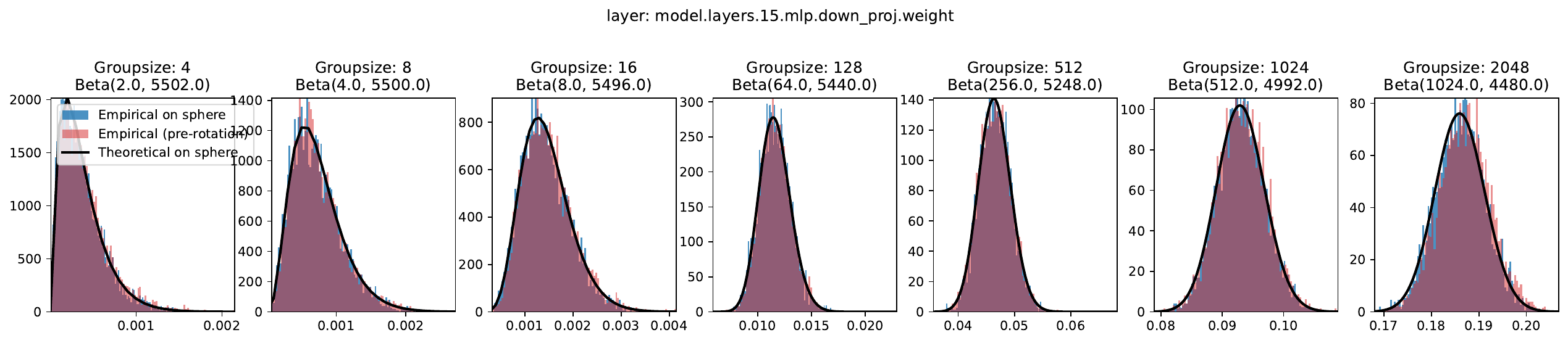} &
\includegraphics[width=0.22\linewidth]{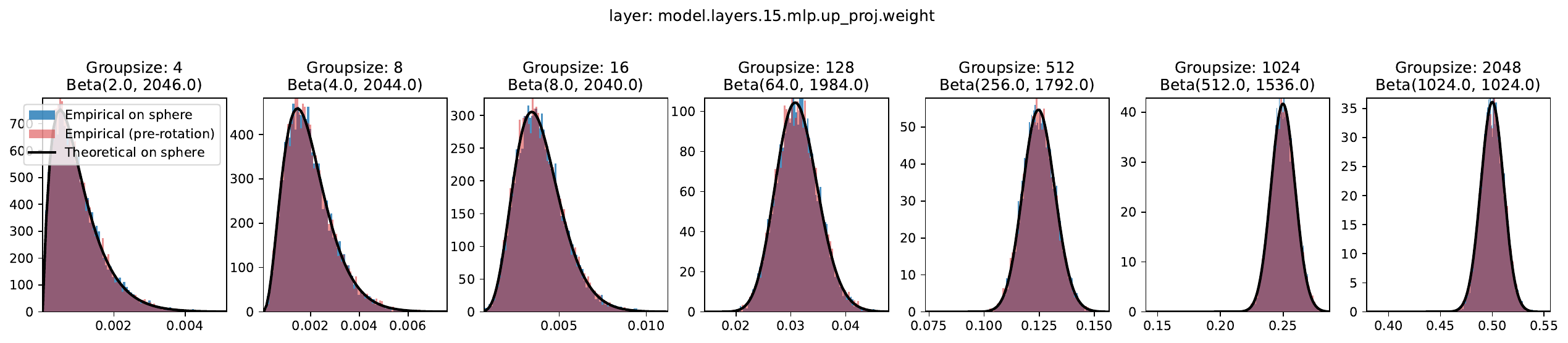} &
\includegraphics[width=0.22\linewidth]{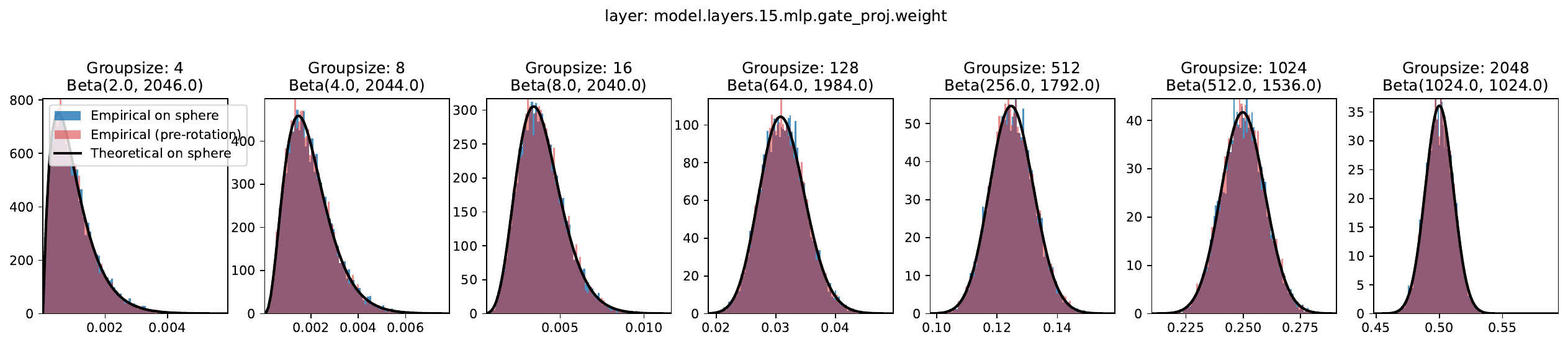} \\
\includegraphics[width=0.22\linewidth]{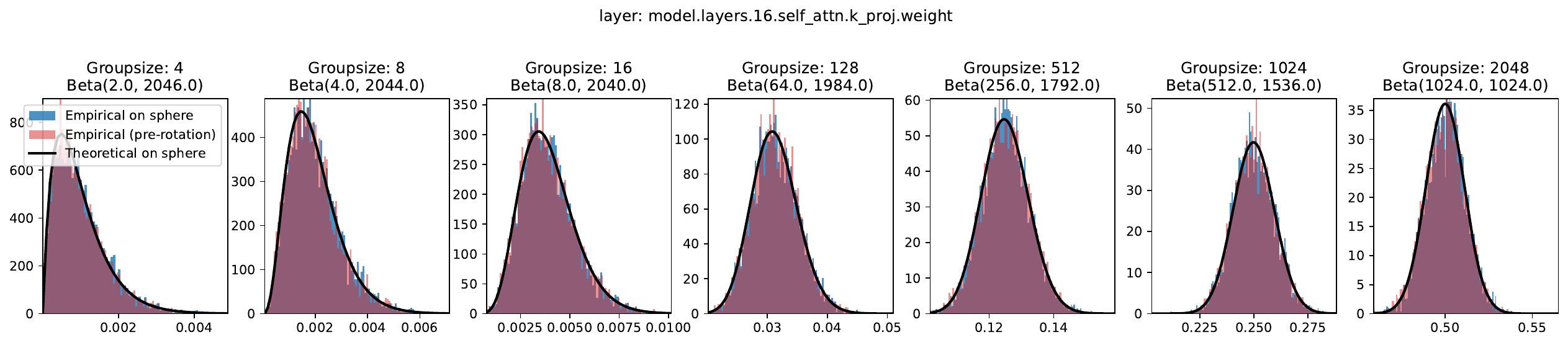} &
\includegraphics[width=0.22\linewidth]{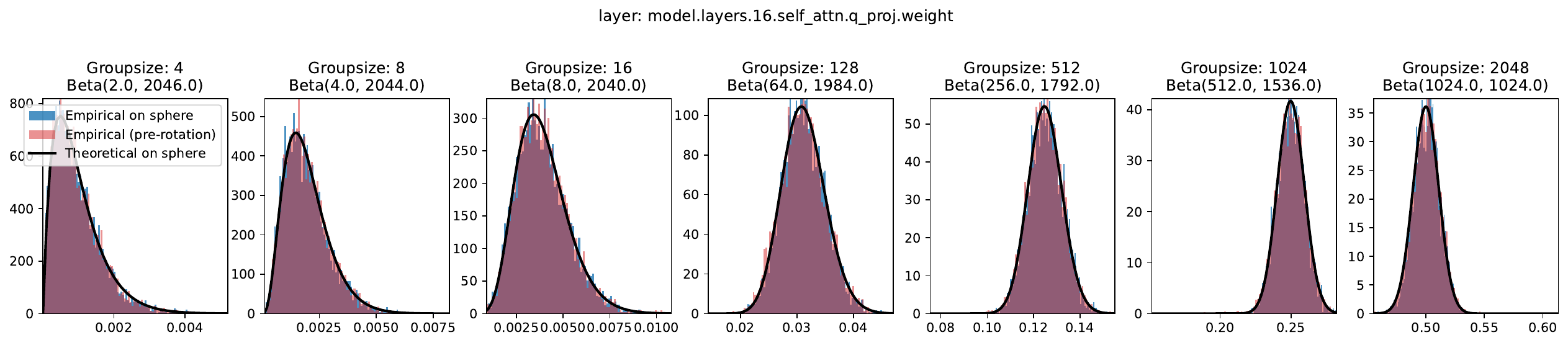} &
\includegraphics[width=0.22\linewidth]{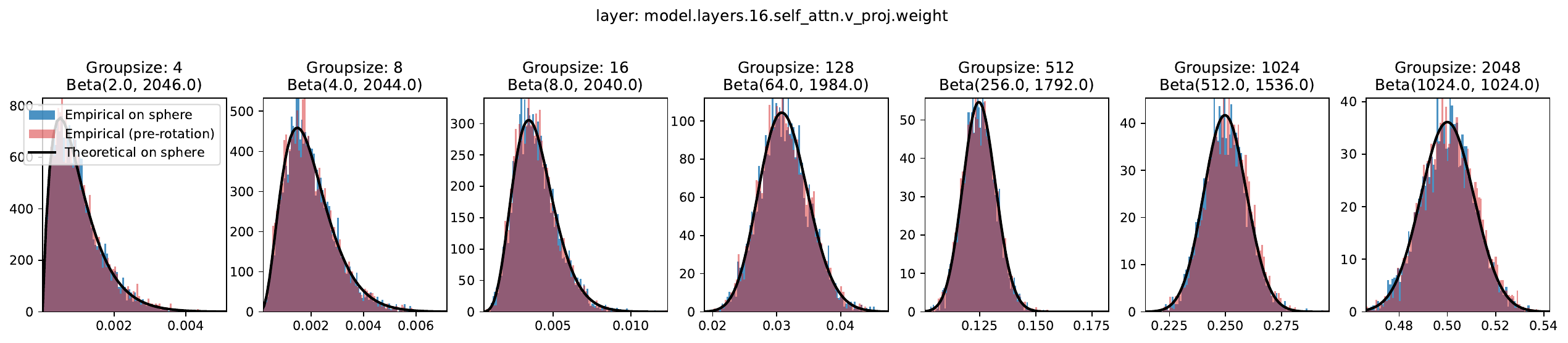} &
\includegraphics[width=0.22\linewidth]{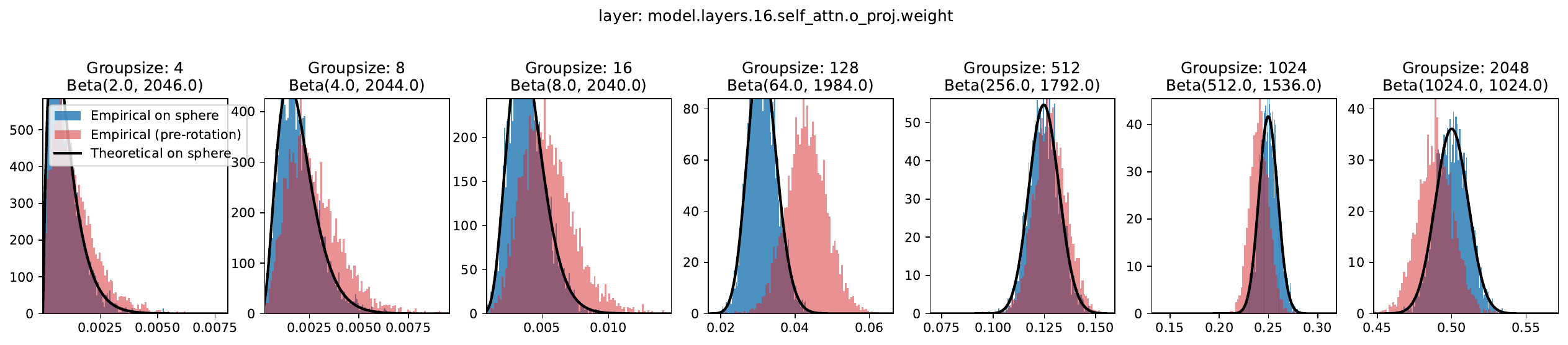} \\
\includegraphics[width=0.22\linewidth]{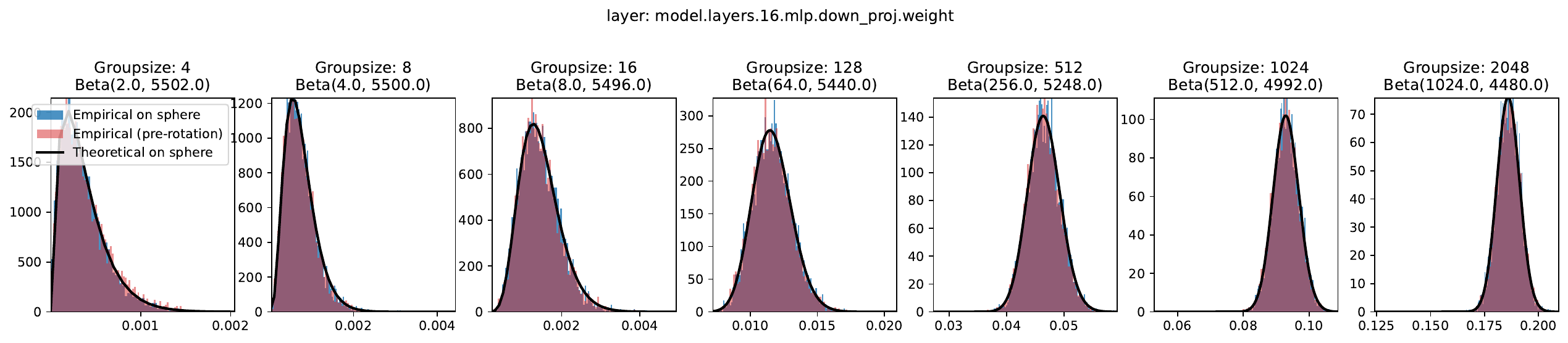} &
\includegraphics[width=0.22\linewidth]{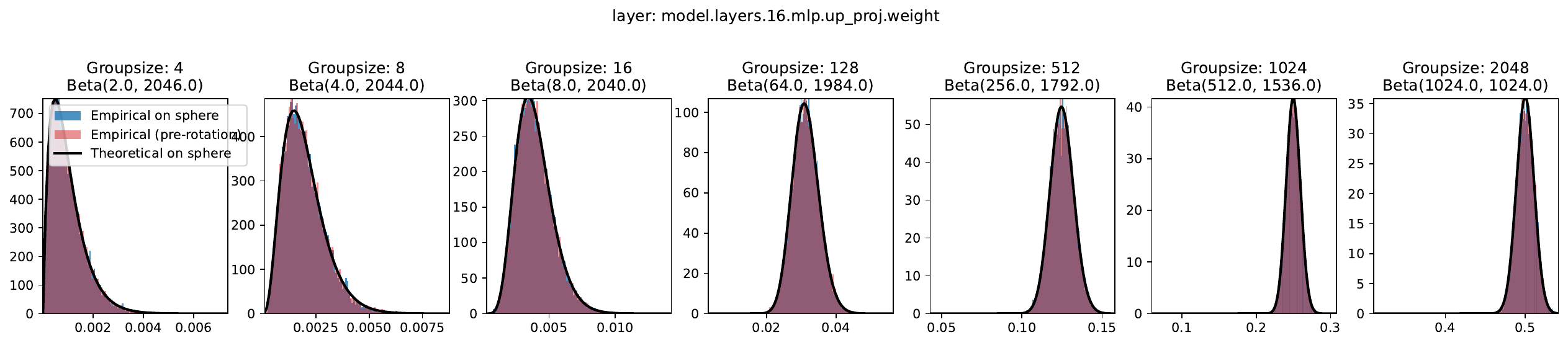} &
\includegraphics[width=0.22\linewidth]{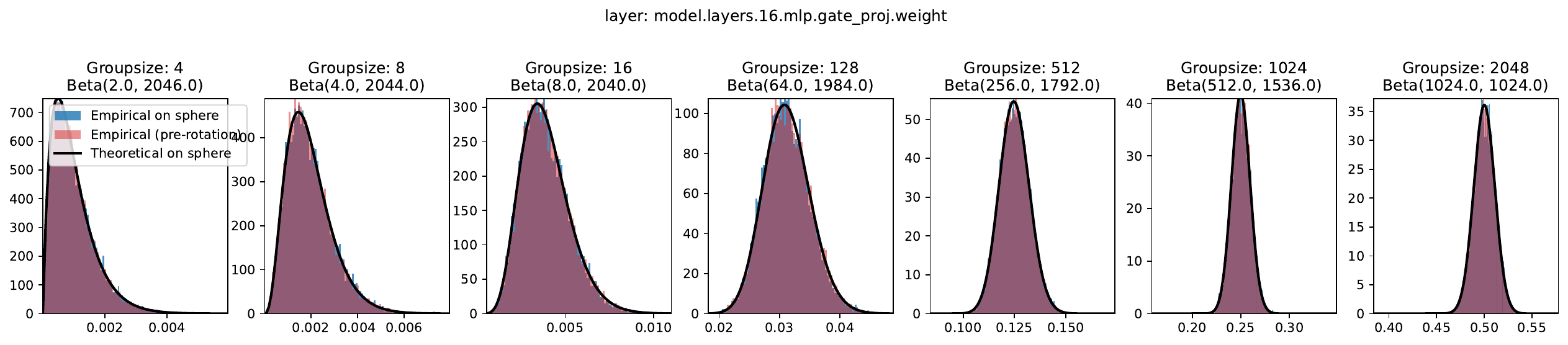} \\
\includegraphics[width=0.22\linewidth]{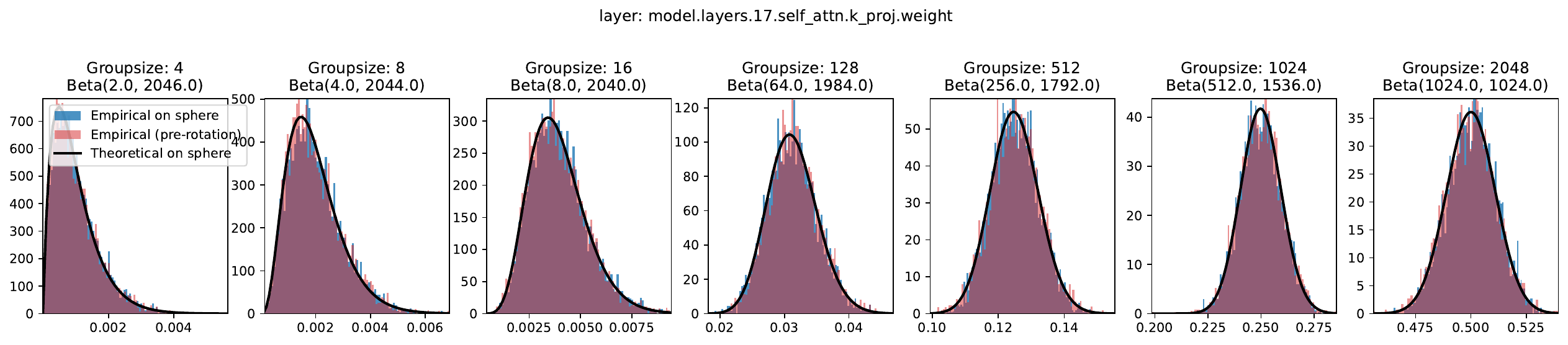} &
\includegraphics[width=0.22\linewidth]{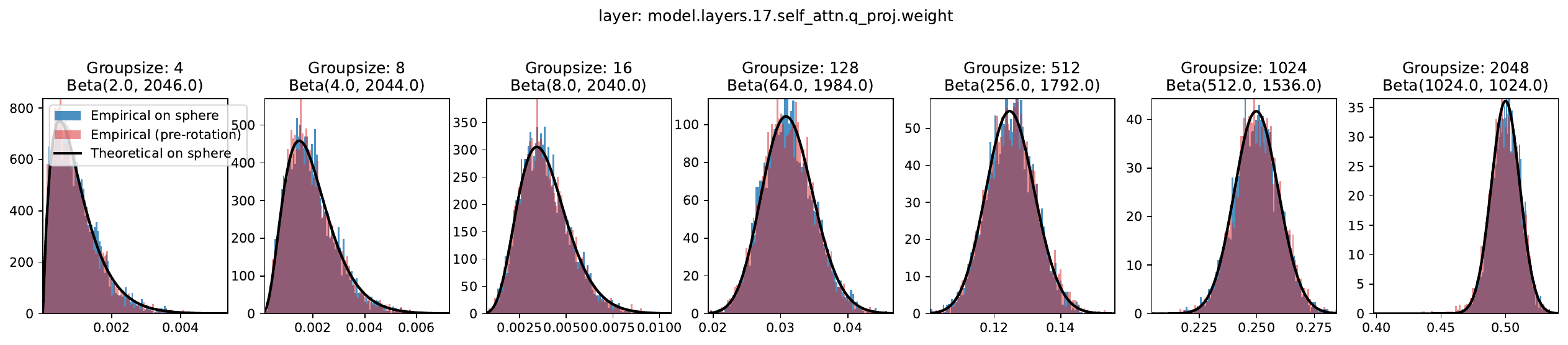} &
\includegraphics[width=0.22\linewidth]{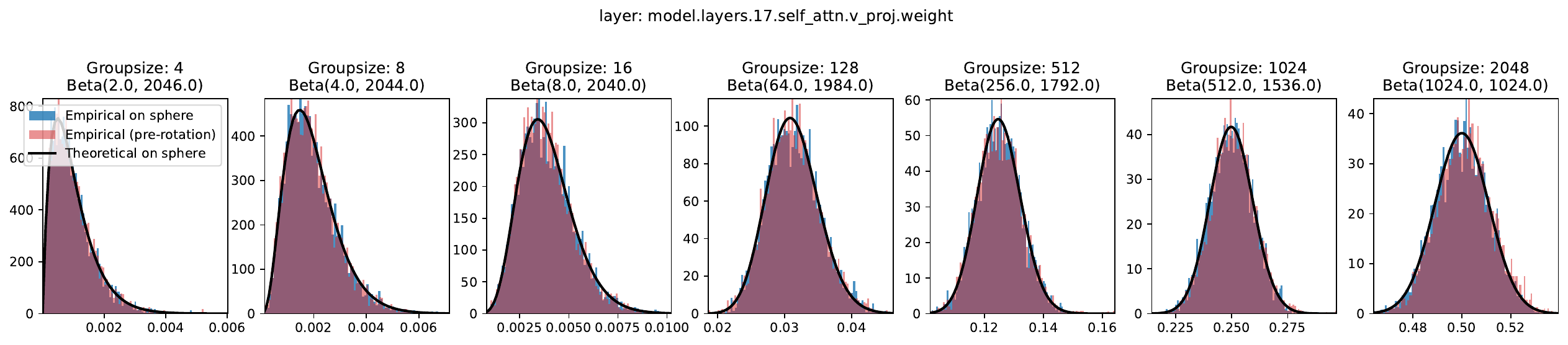} &
\includegraphics[width=0.22\linewidth]{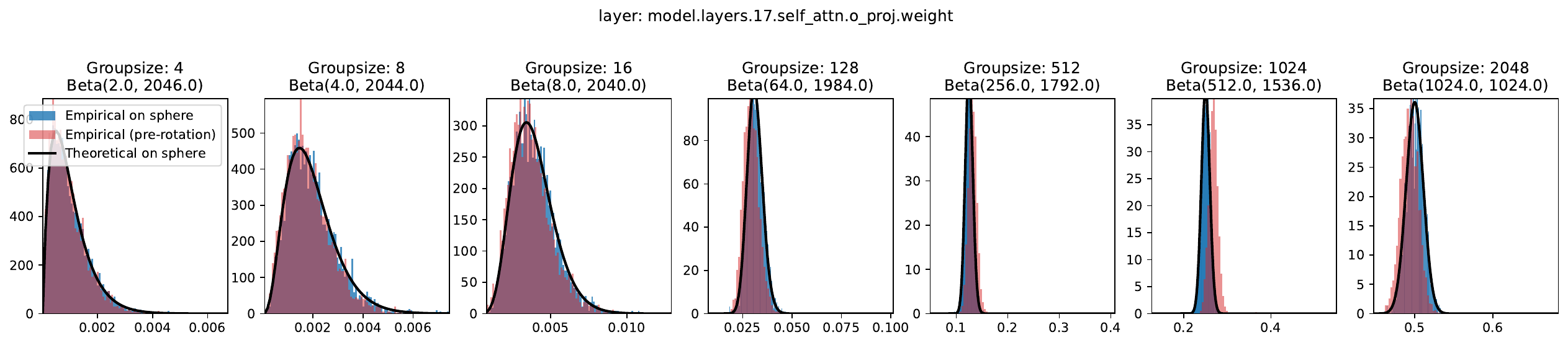} \\
\includegraphics[width=0.22\linewidth]{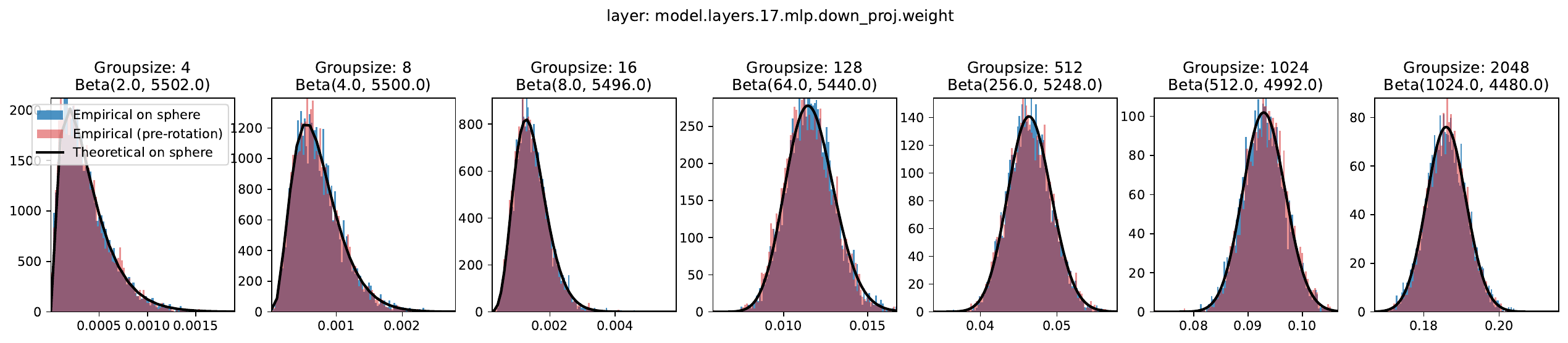} &
\includegraphics[width=0.22\linewidth]{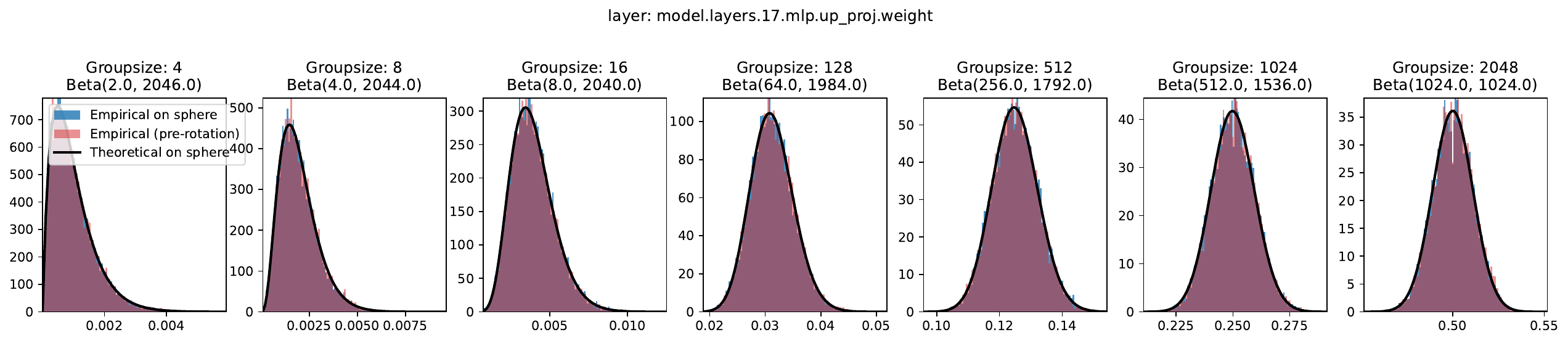} &
\includegraphics[width=0.22\linewidth]{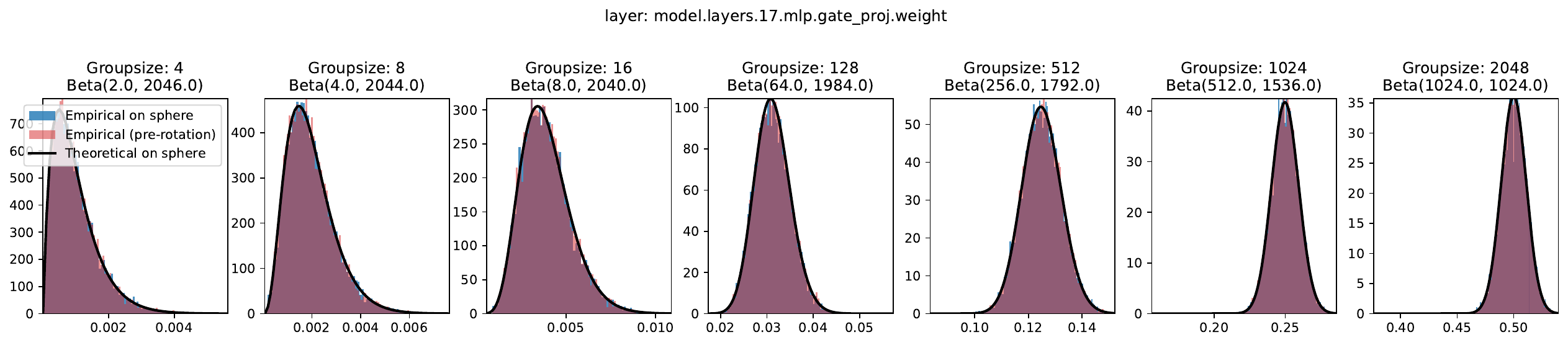} \\
\includegraphics[width=0.22\linewidth]{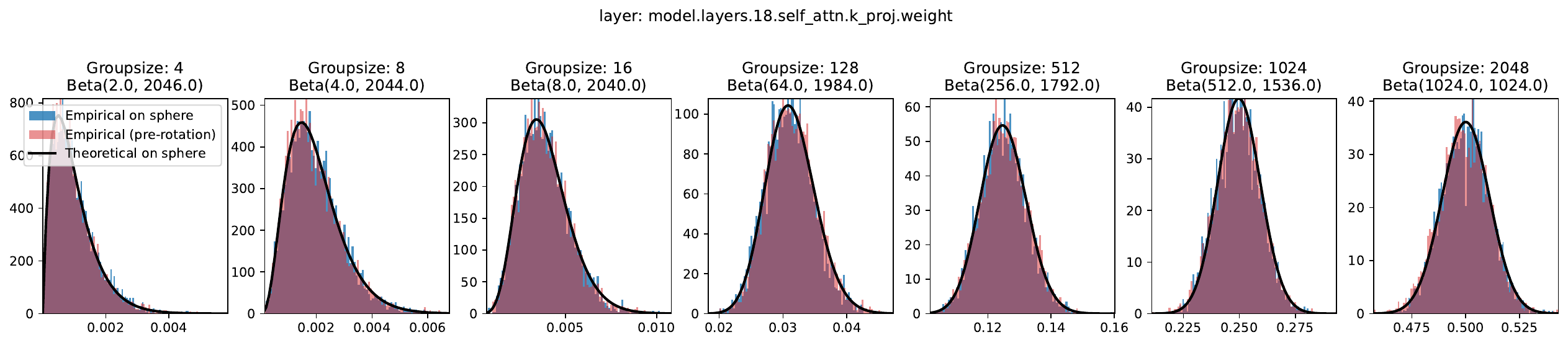} &
\includegraphics[width=0.22\linewidth]{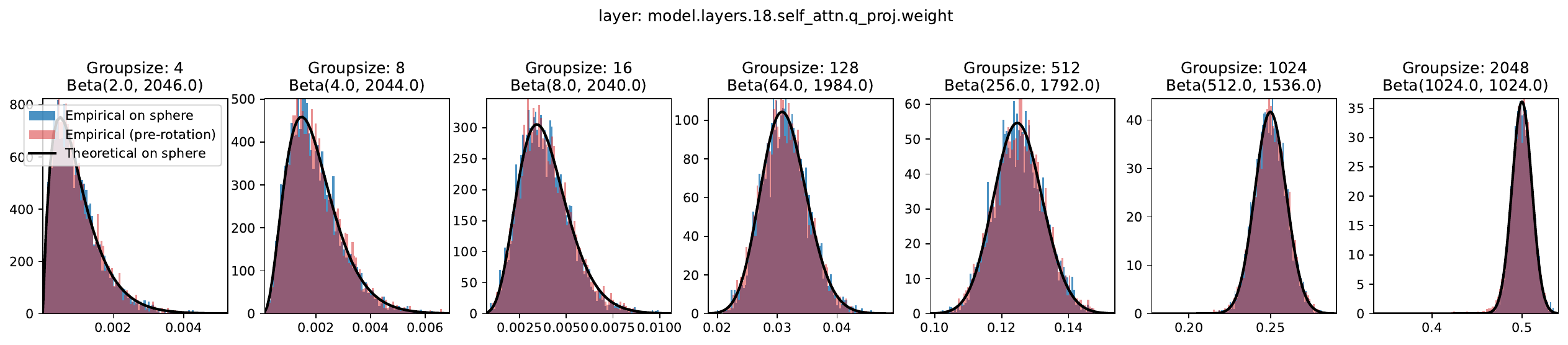} &
\includegraphics[width=0.22\linewidth]{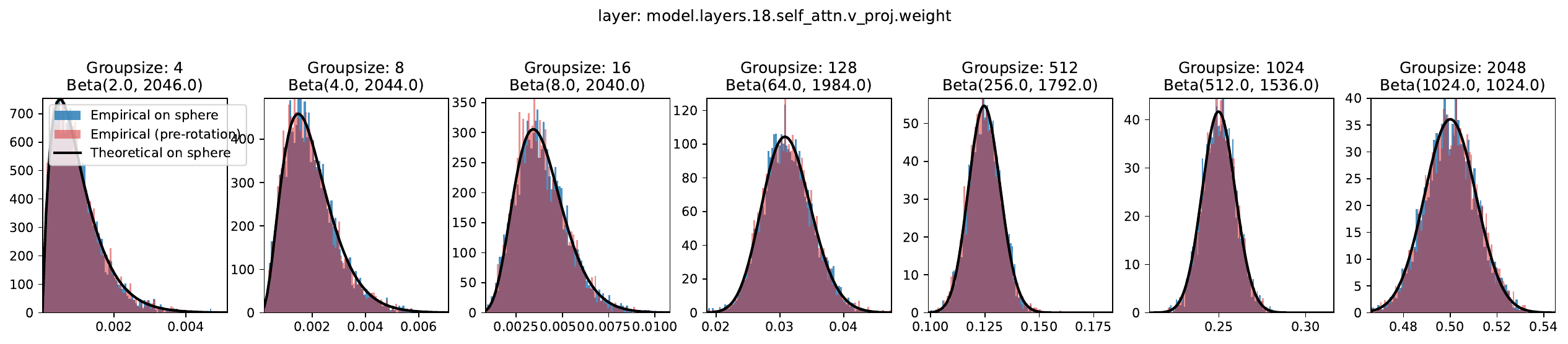} &
\includegraphics[width=0.22\linewidth]{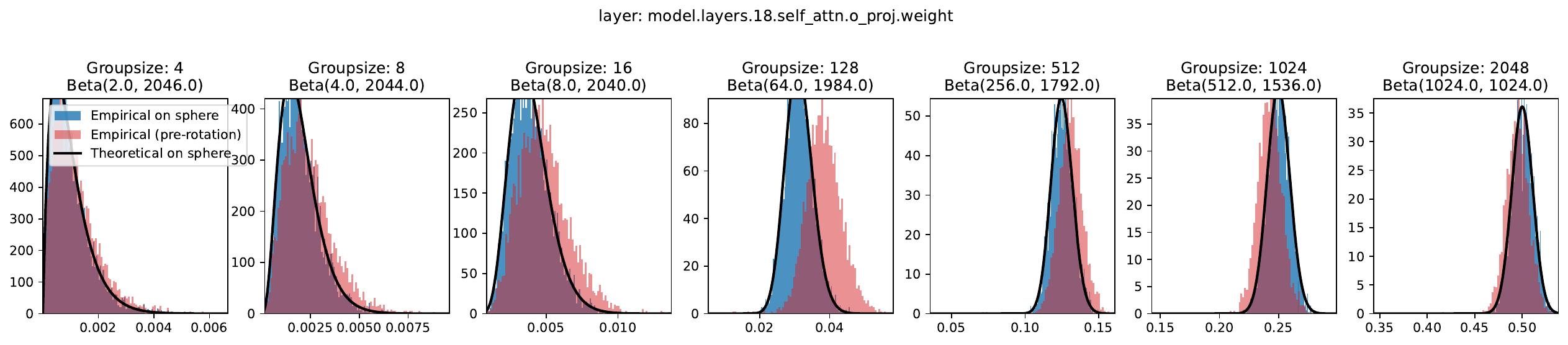} \\
\includegraphics[width=0.22\linewidth]{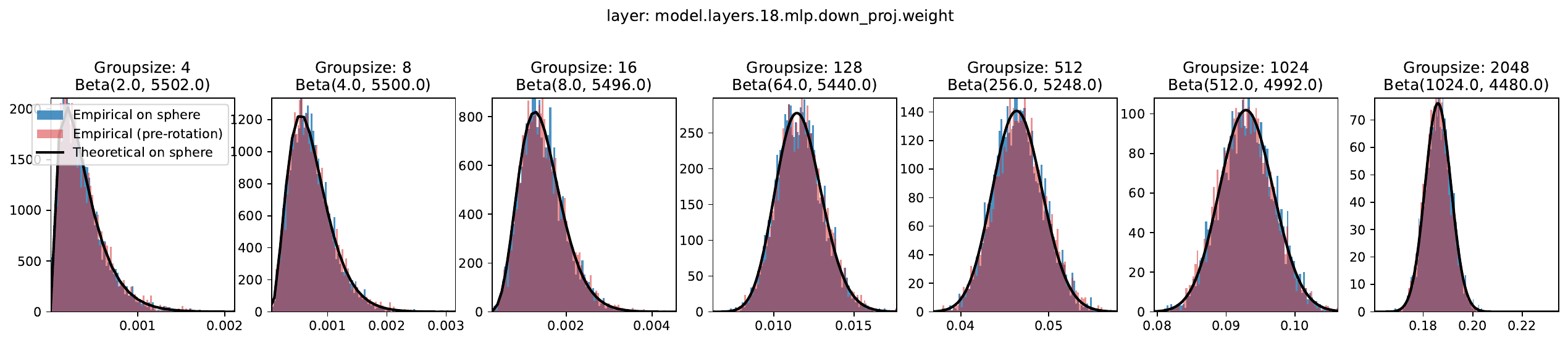} &
\includegraphics[width=0.22\linewidth]{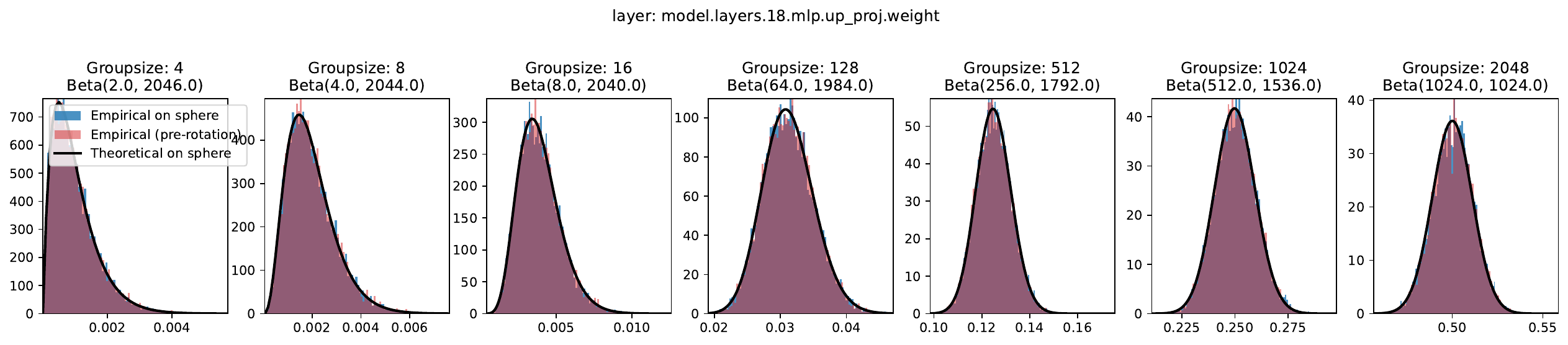} &
\includegraphics[width=0.22\linewidth]{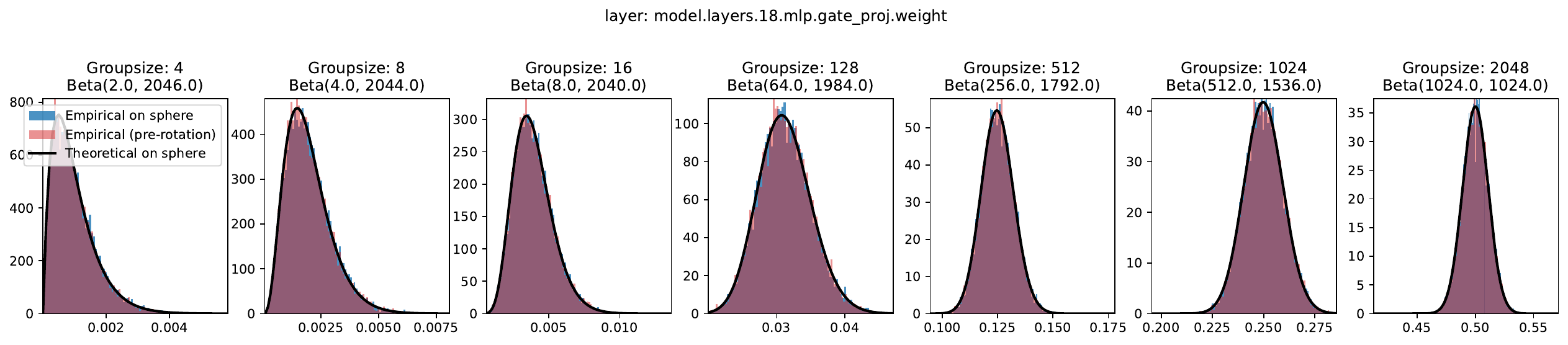} \\
\includegraphics[width=0.22\linewidth]{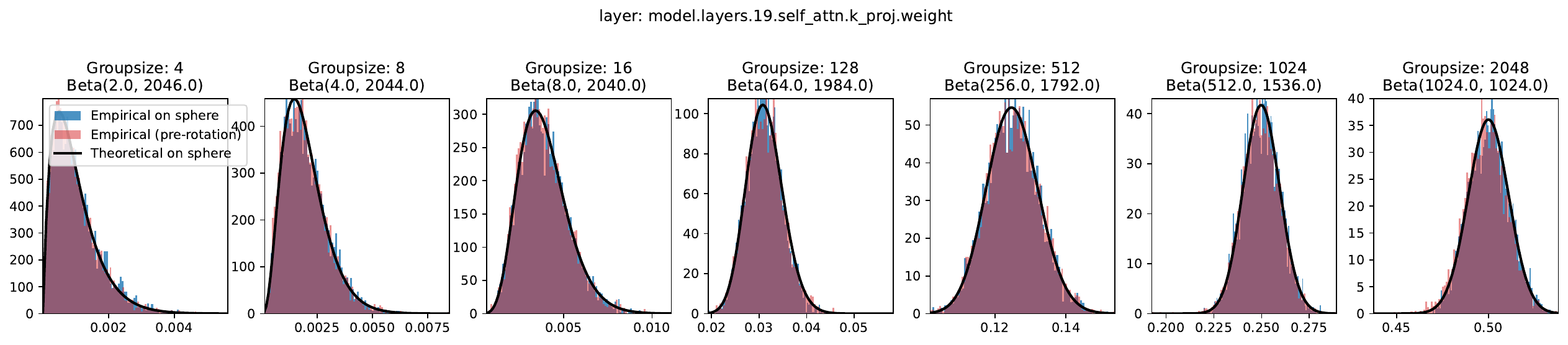} &
\includegraphics[width=0.22\linewidth]{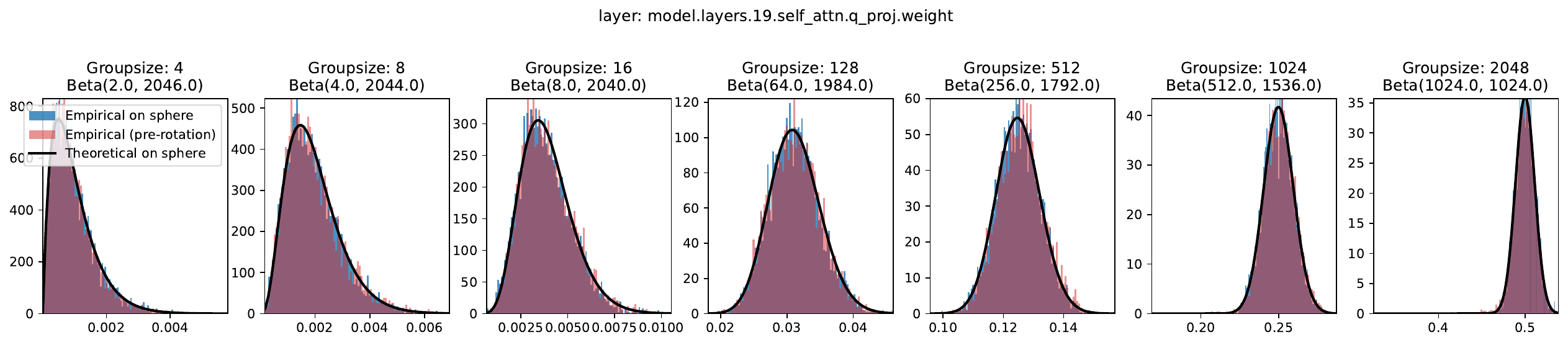} &
\includegraphics[width=0.22\linewidth]{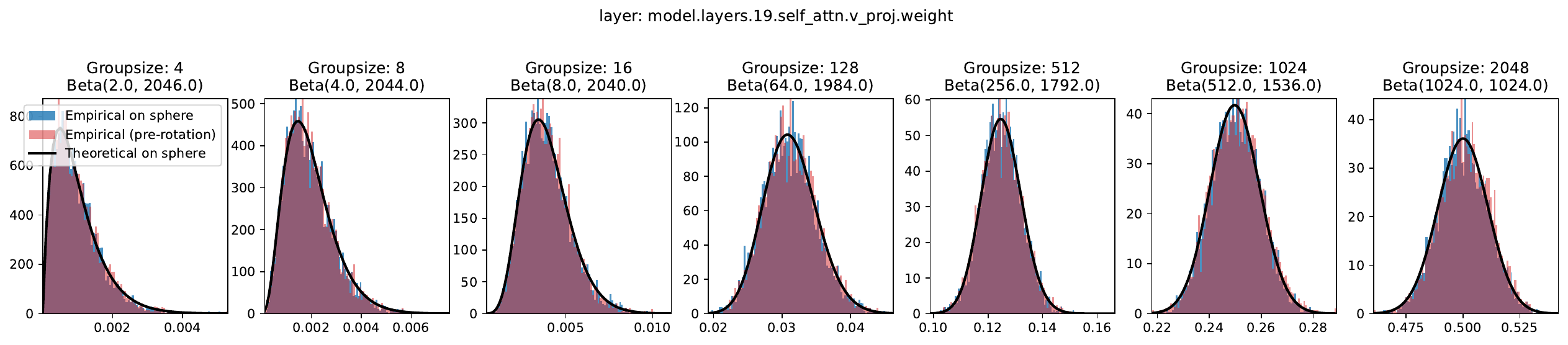} &
\includegraphics[width=0.22\linewidth]{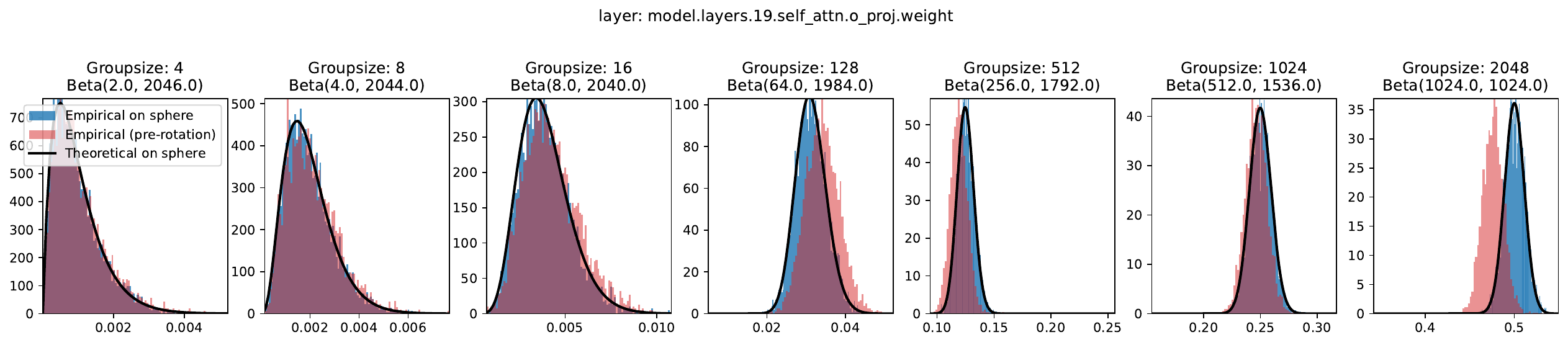} \\
\includegraphics[width=0.22\linewidth]{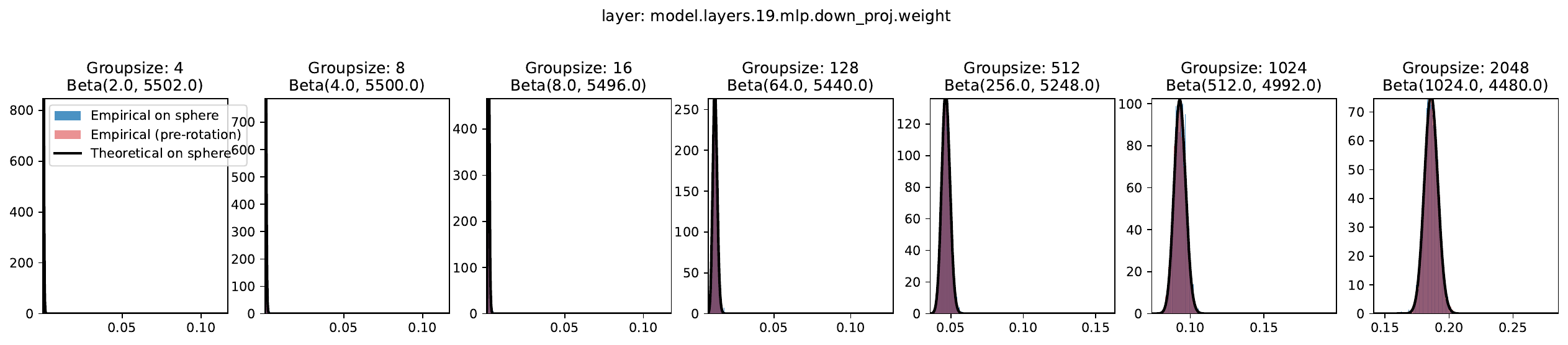} &
\includegraphics[width=0.22\linewidth]{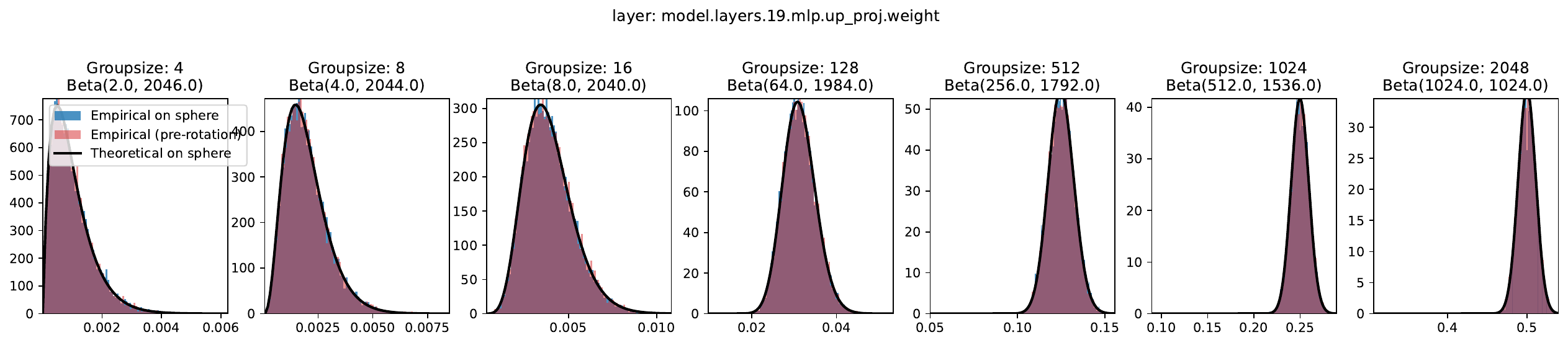} &
\includegraphics[width=0.22\linewidth]{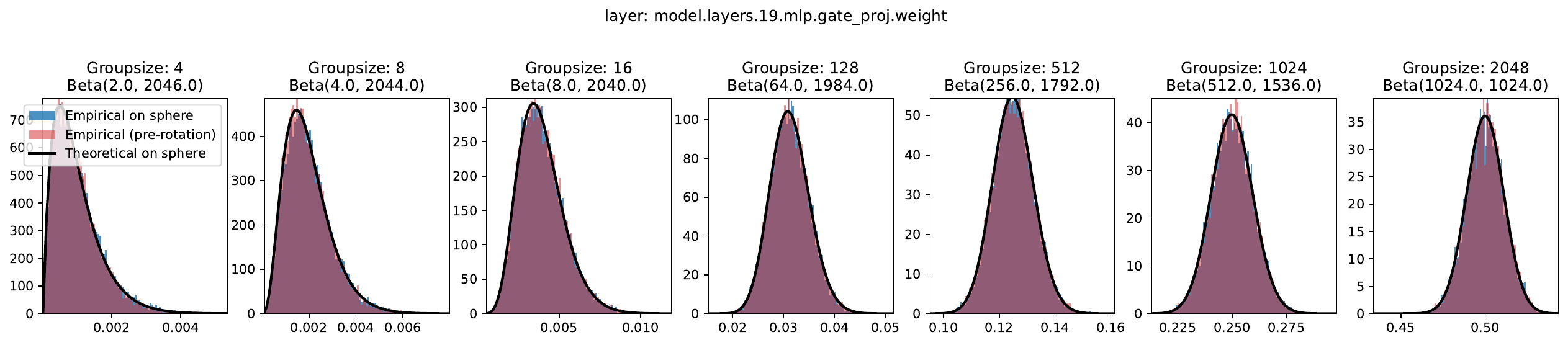} \\
\includegraphics[width=0.22\linewidth]{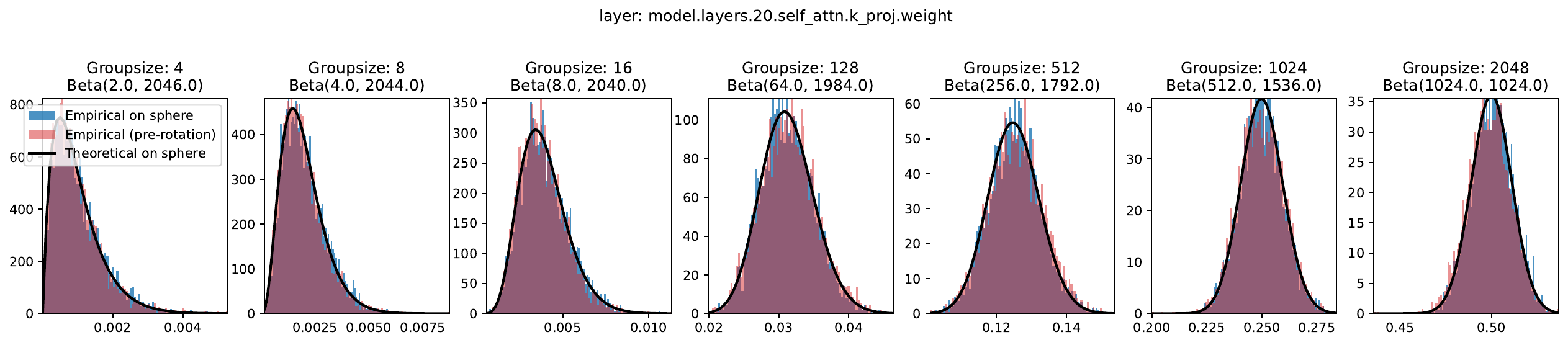} &
\includegraphics[width=0.22\linewidth]{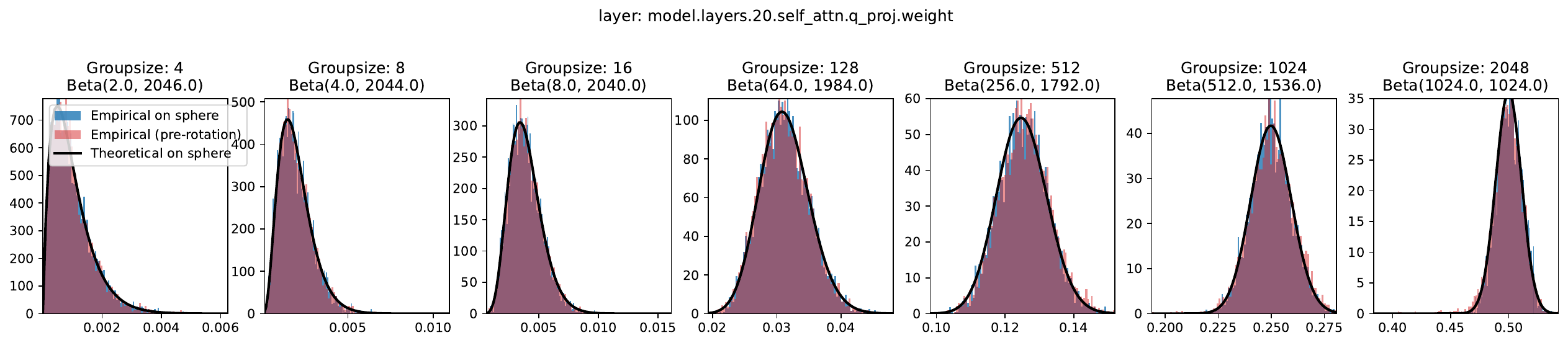} &
\includegraphics[width=0.22\linewidth]{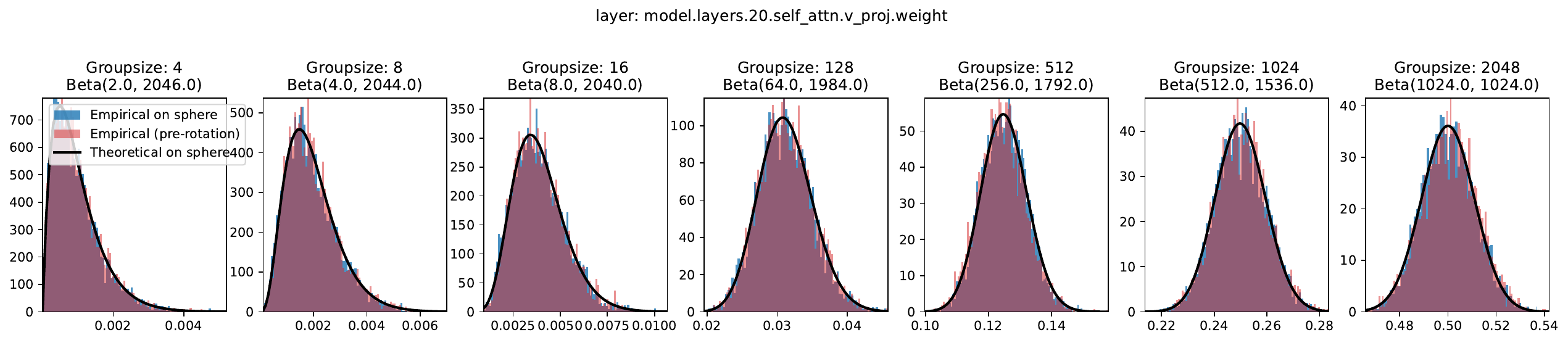} &
\includegraphics[width=0.22\linewidth]{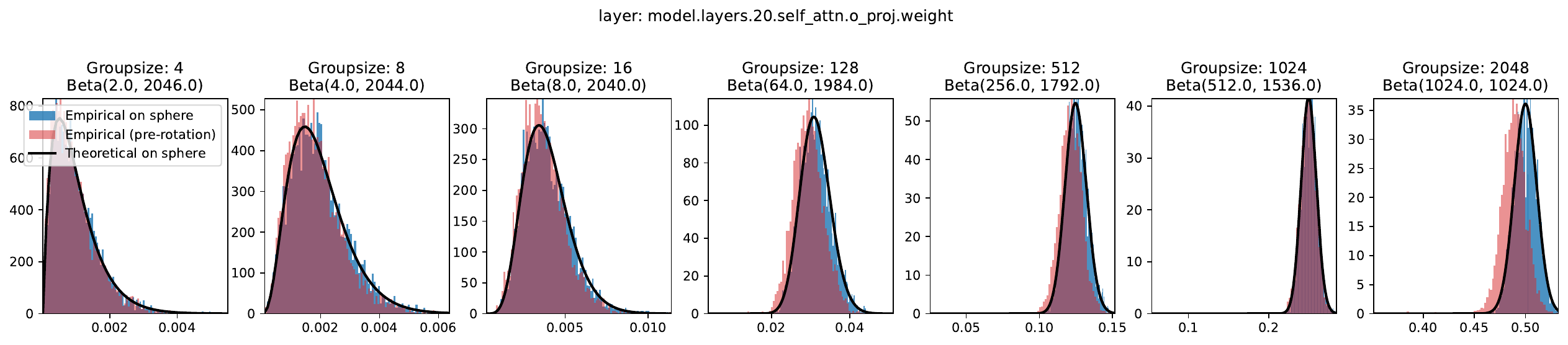} \\
\includegraphics[width=0.22\linewidth]{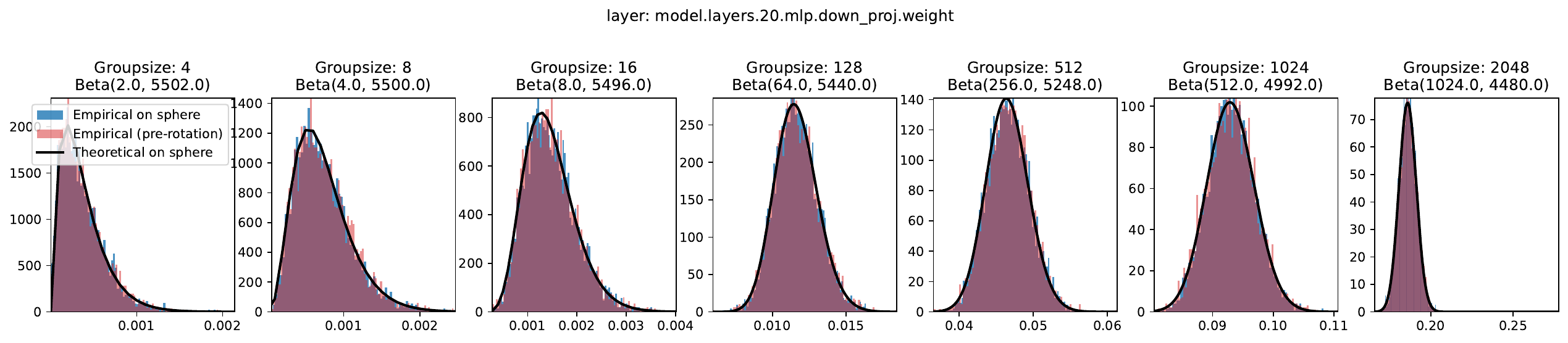} &
\includegraphics[width=0.22\linewidth]{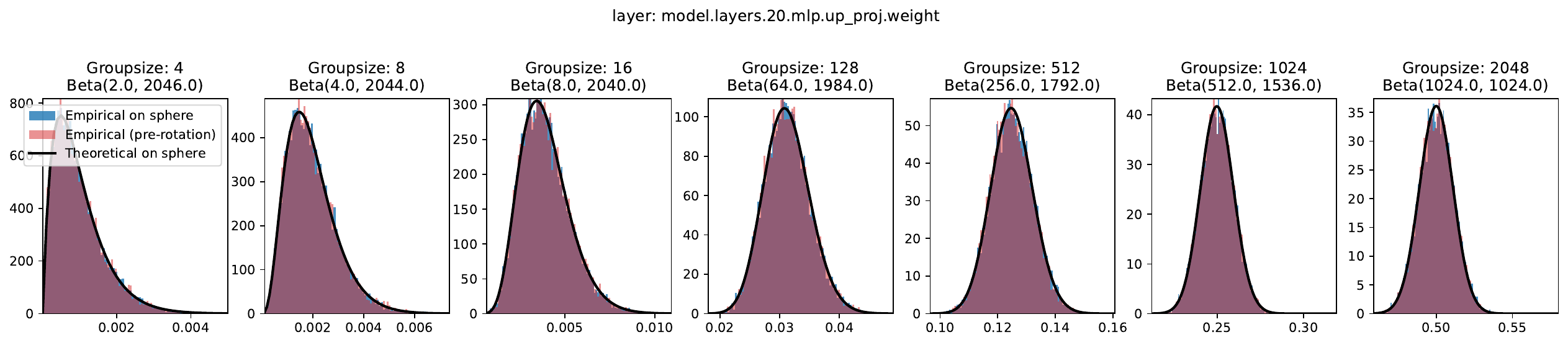} &
\includegraphics[width=0.22\linewidth]{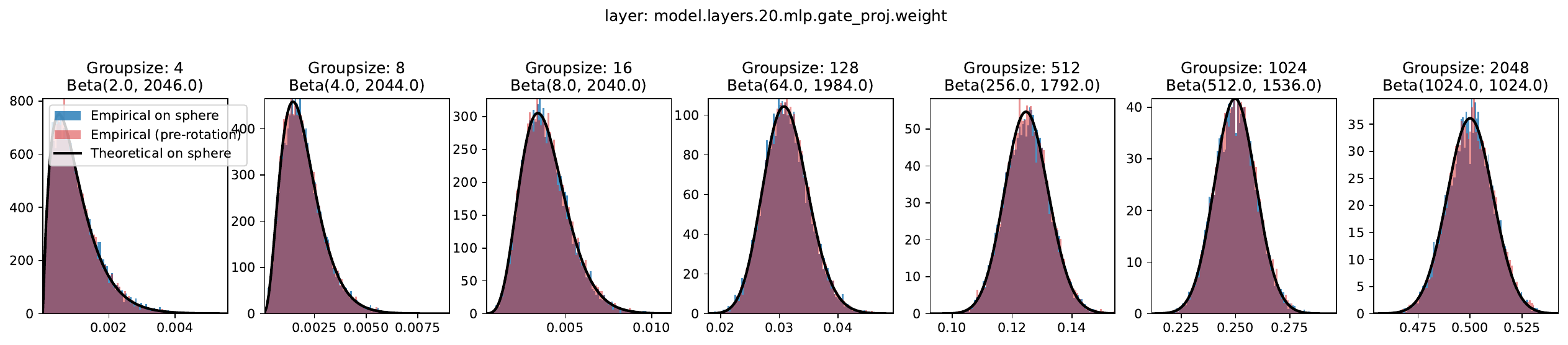} \\
\includegraphics[width=0.22\linewidth]{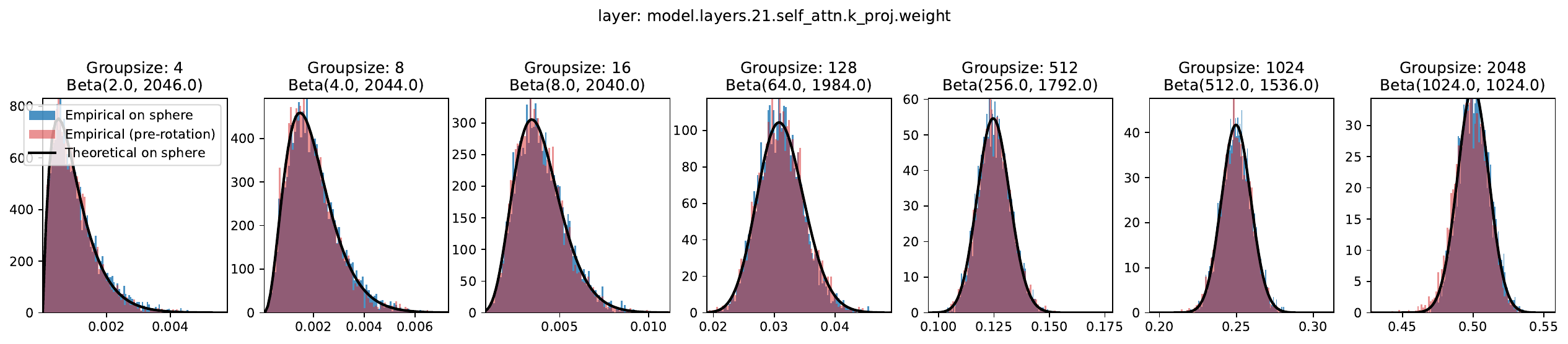} &
\includegraphics[width=0.22\linewidth]{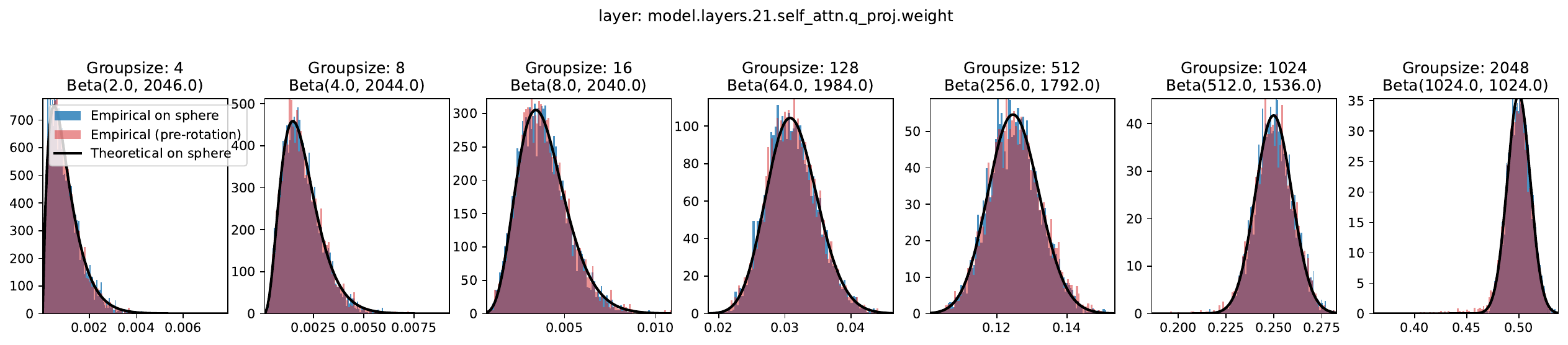} &
\includegraphics[width=0.22\linewidth]{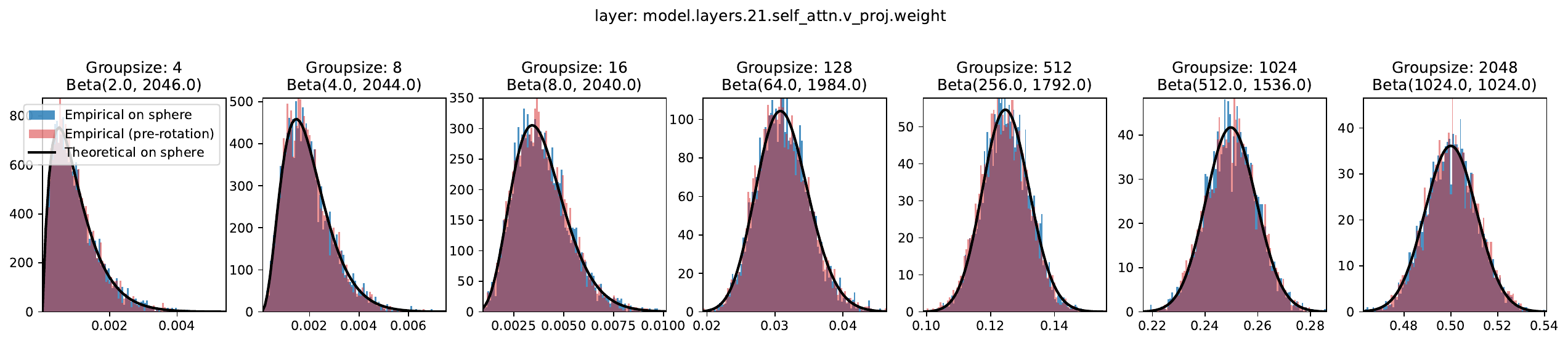} &
\includegraphics[width=0.22\linewidth]{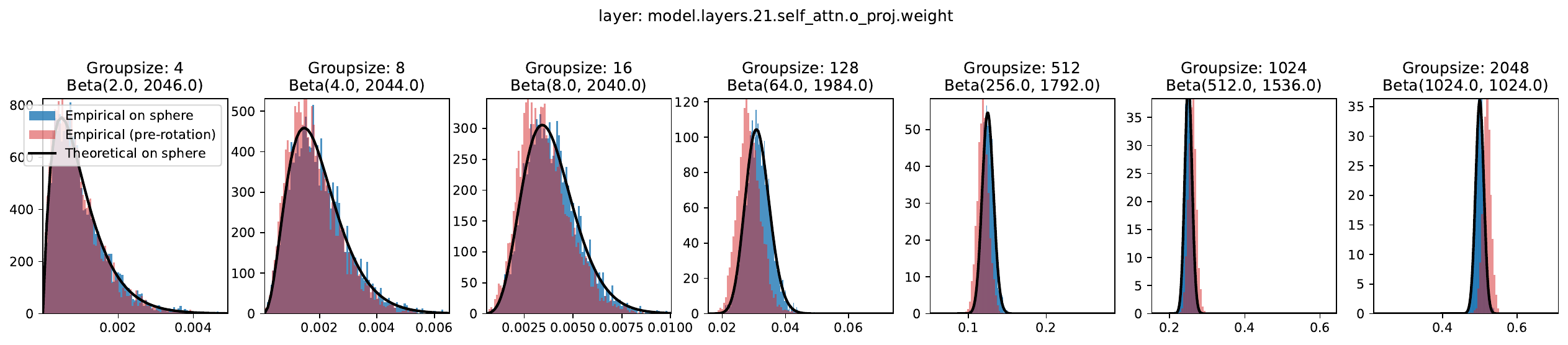} \\
\includegraphics[width=0.22\linewidth]{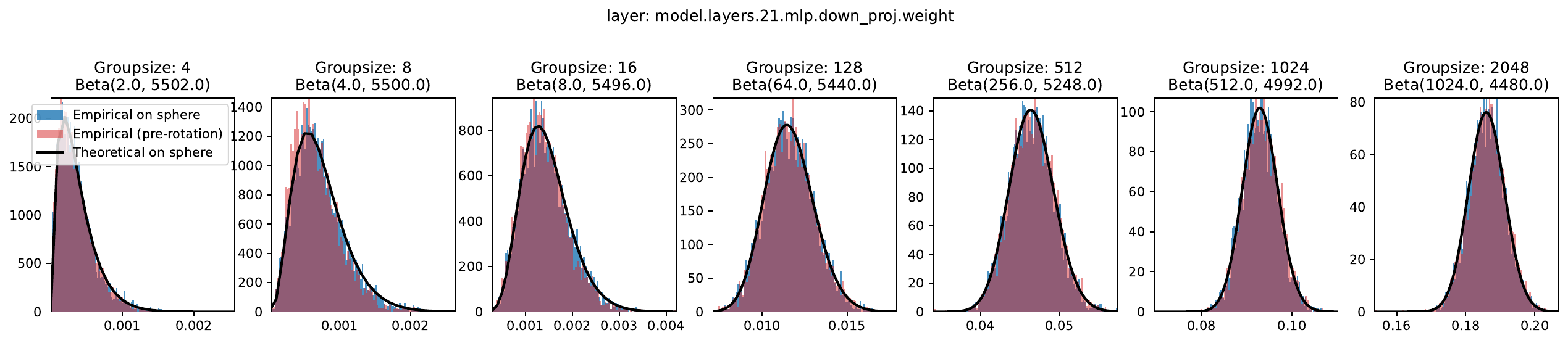} &
\includegraphics[width=0.22\linewidth]{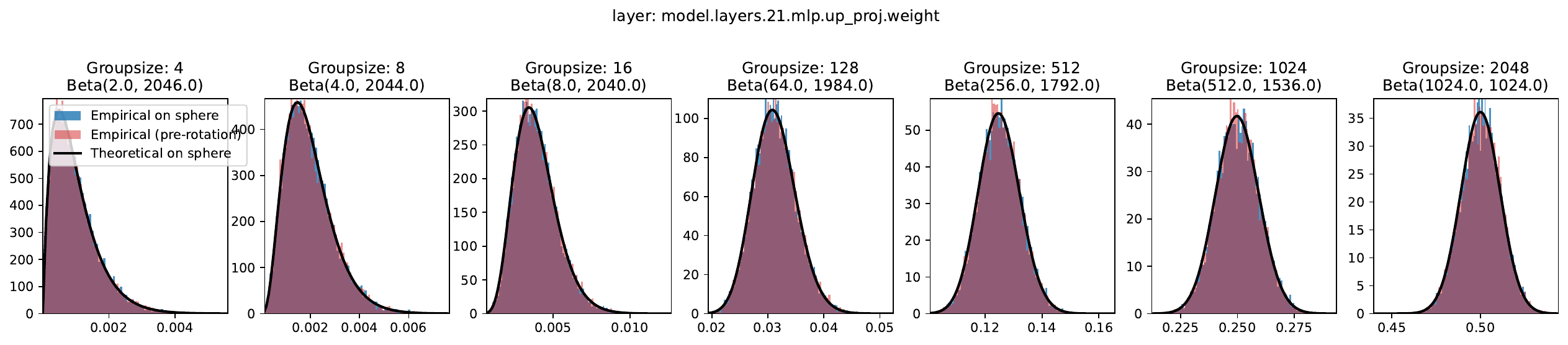} &
\includegraphics[width=0.22\linewidth]{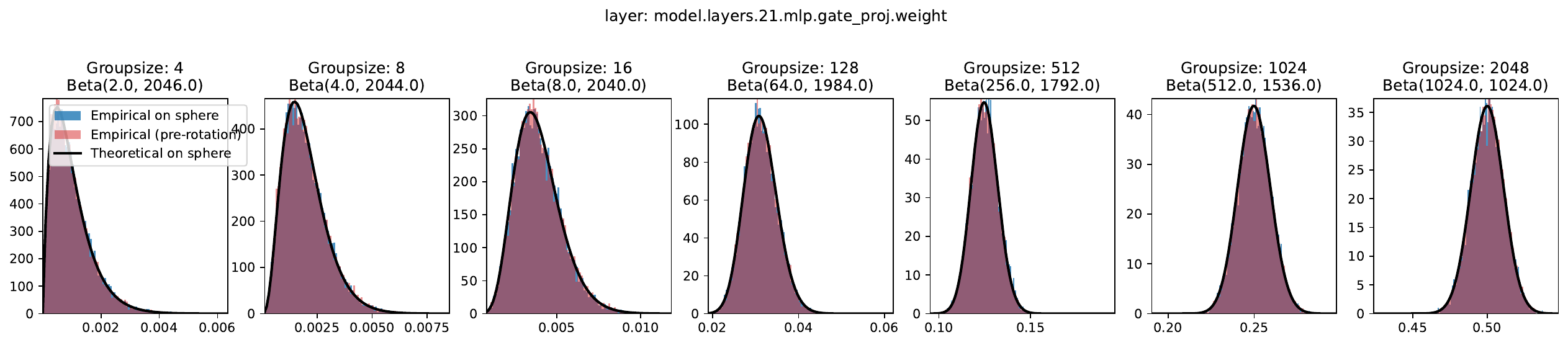} \\
\includegraphics[width=0.22\linewidth]{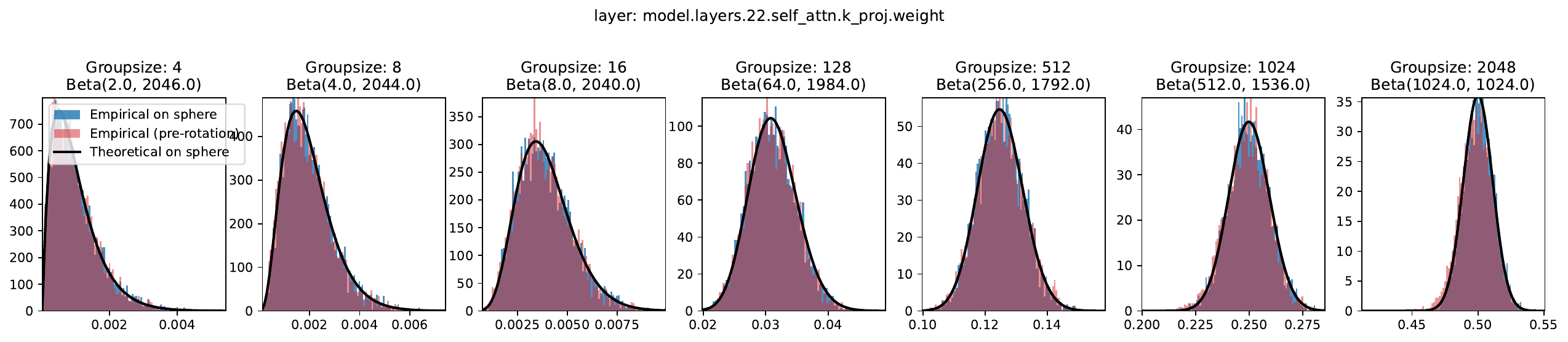} &
\includegraphics[width=0.22\linewidth]{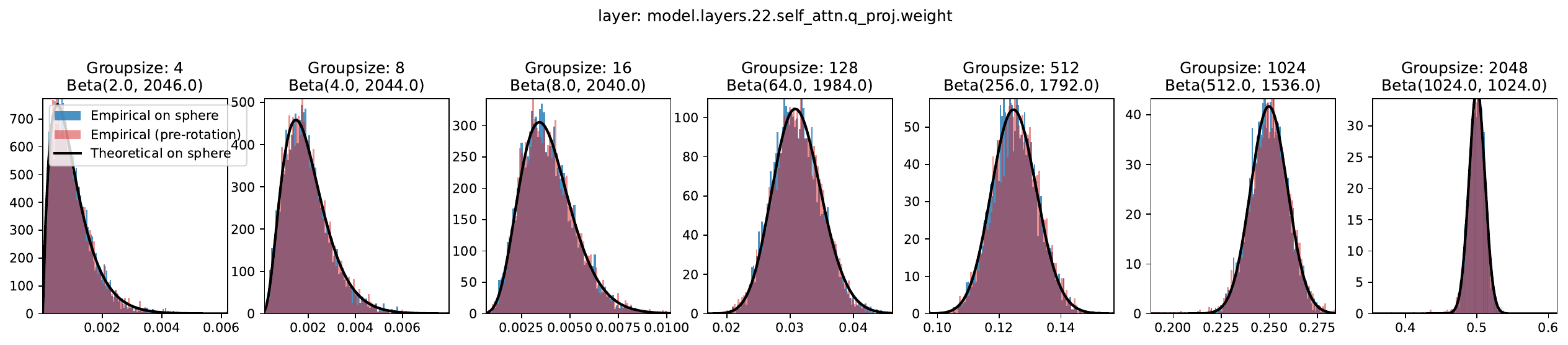} &
\includegraphics[width=0.22\linewidth]{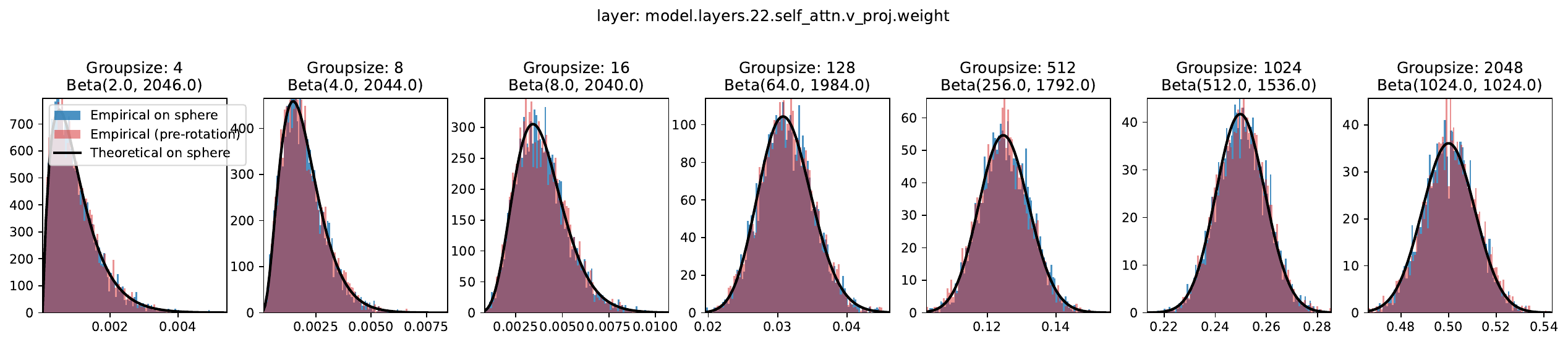} &
\includegraphics[width=0.22\linewidth]{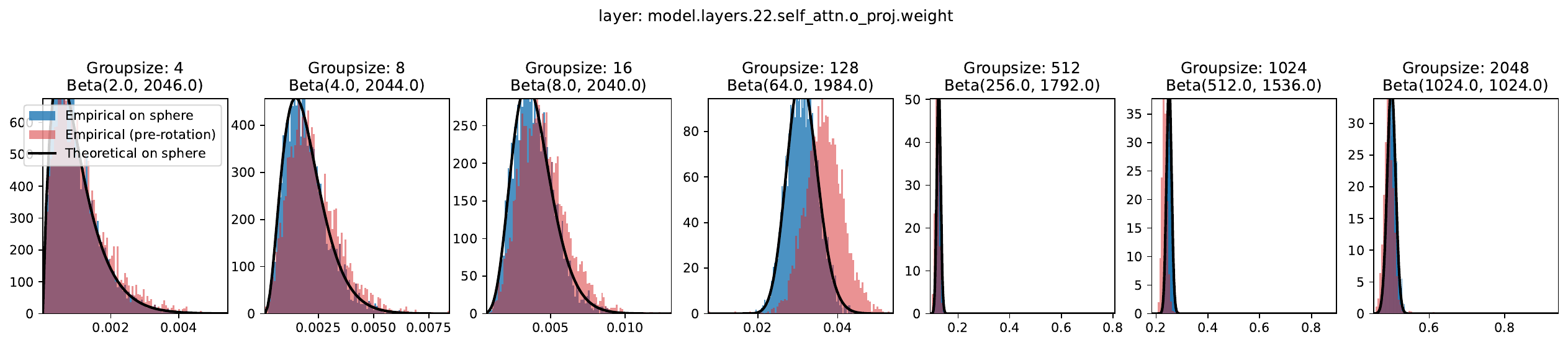} \\
\includegraphics[width=0.22\linewidth]{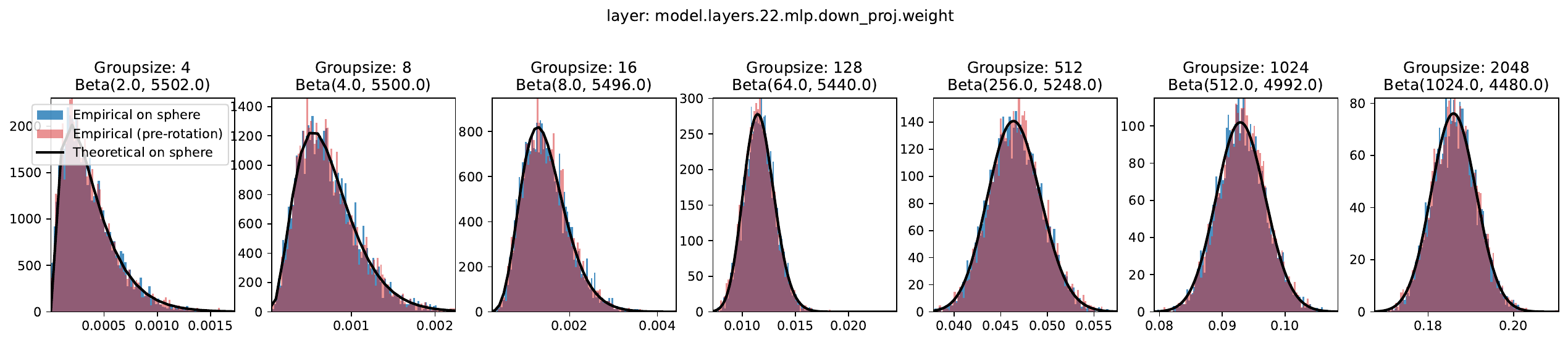} &
\includegraphics[width=0.22\linewidth]{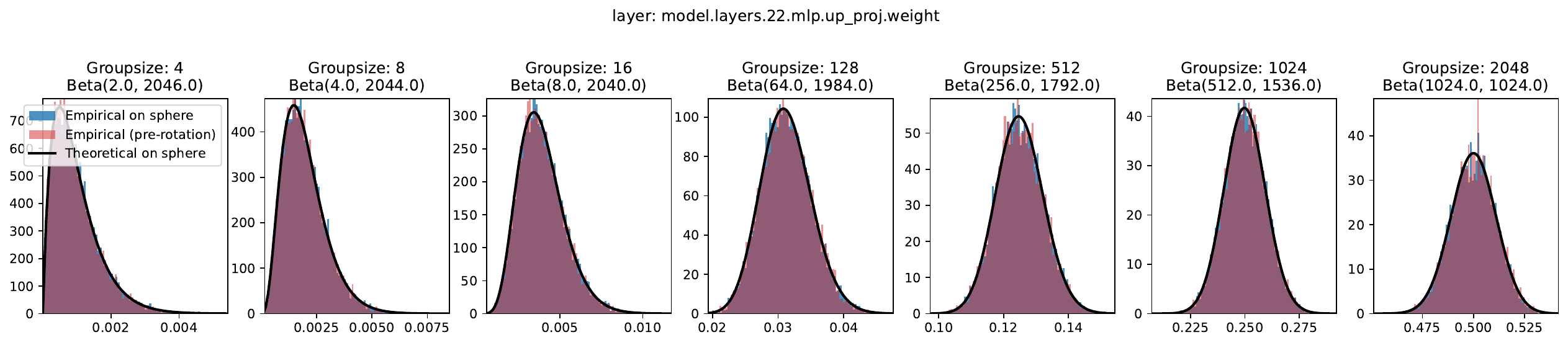} &
\includegraphics[width=0.22\linewidth]{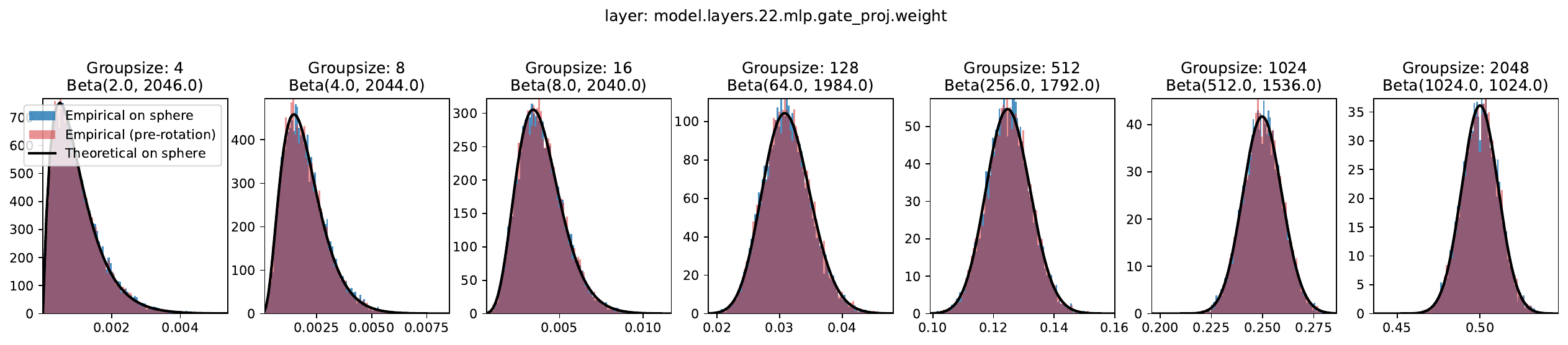} \\
\end{tabular}
\end{table}
\newpage
\begin{table}[H]
%\resizebox{!}{\linewidth}{
\begin{tabular}{llll}
\includegraphics[width=0.22\linewidth]{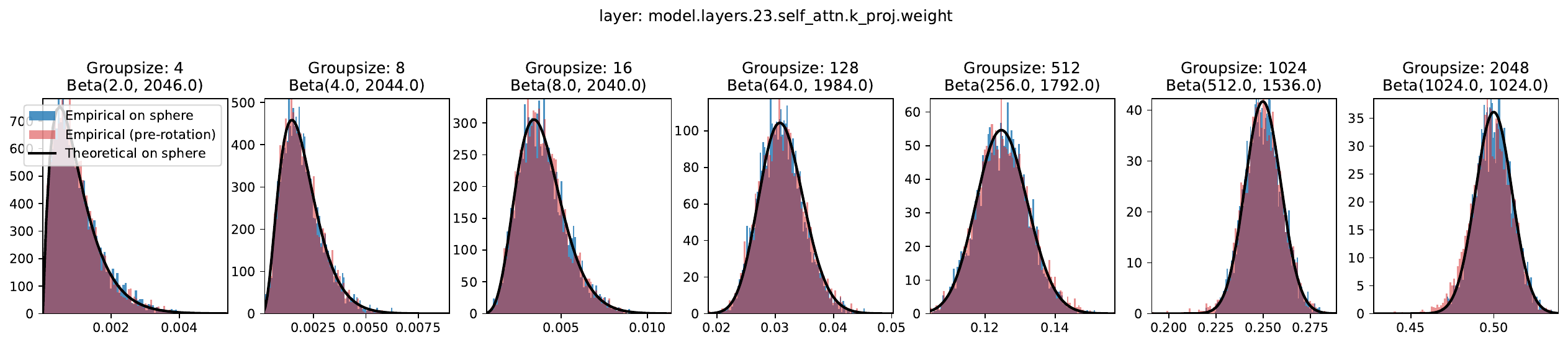} &
\includegraphics[width=0.22\linewidth]{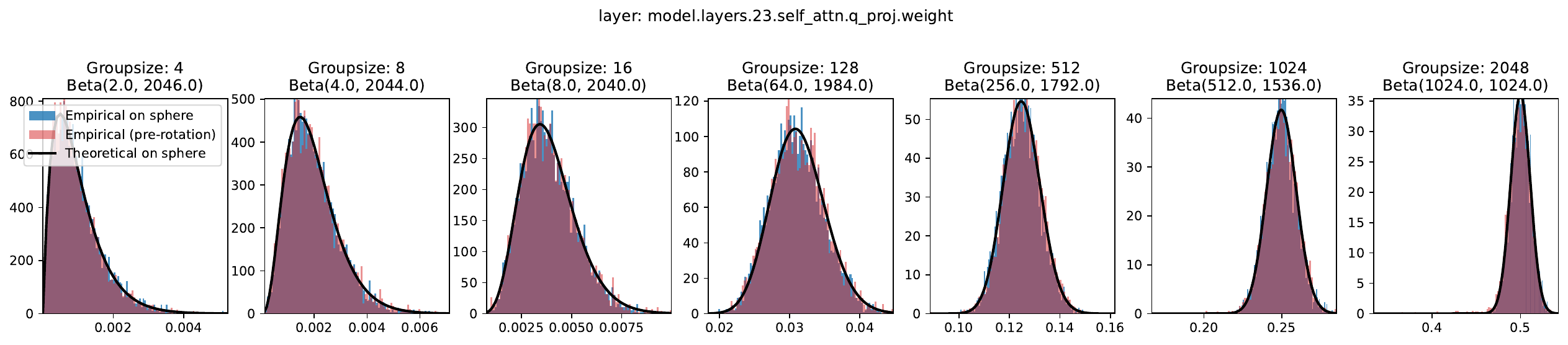} &
\includegraphics[width=0.22\linewidth]{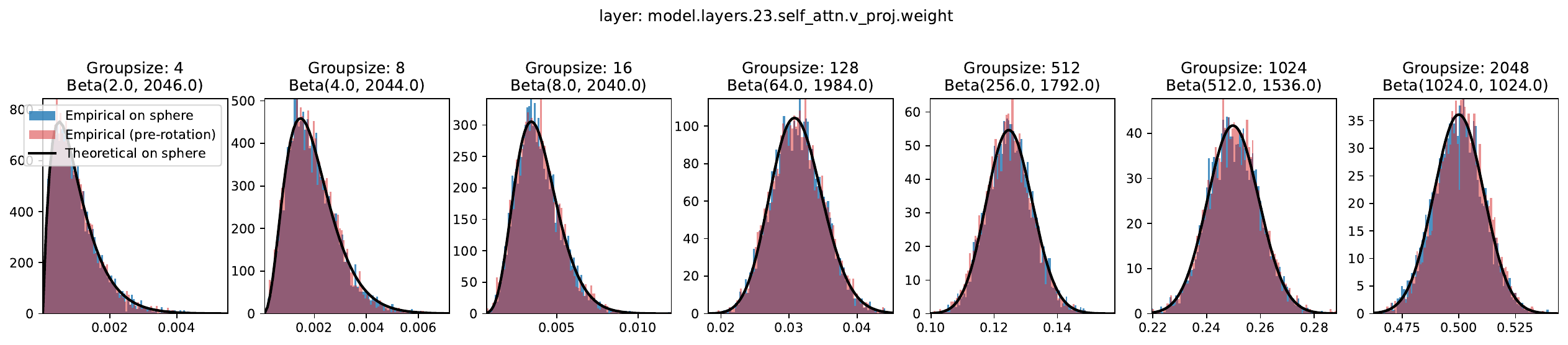} &
\includegraphics[width=0.22\linewidth]{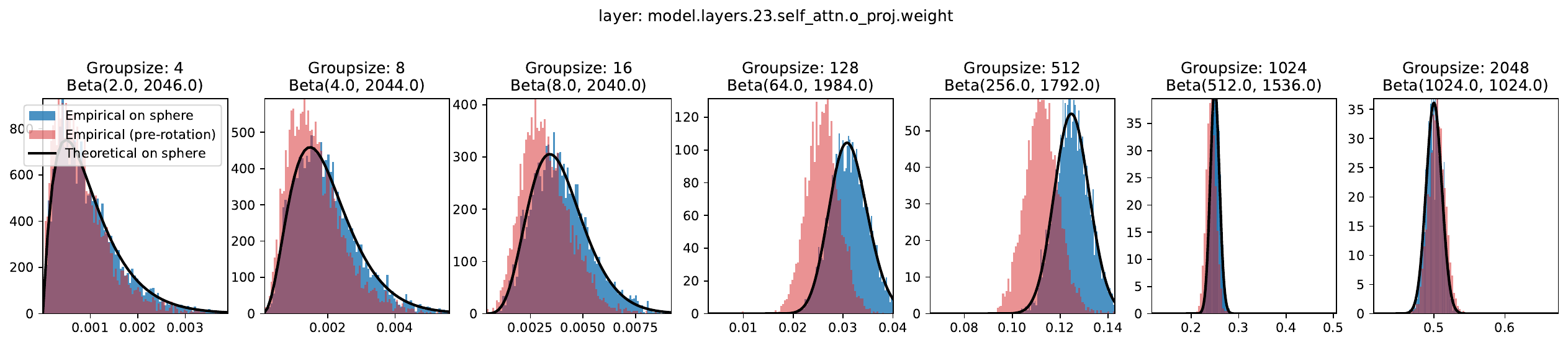} \\
\includegraphics[width=0.22\linewidth]{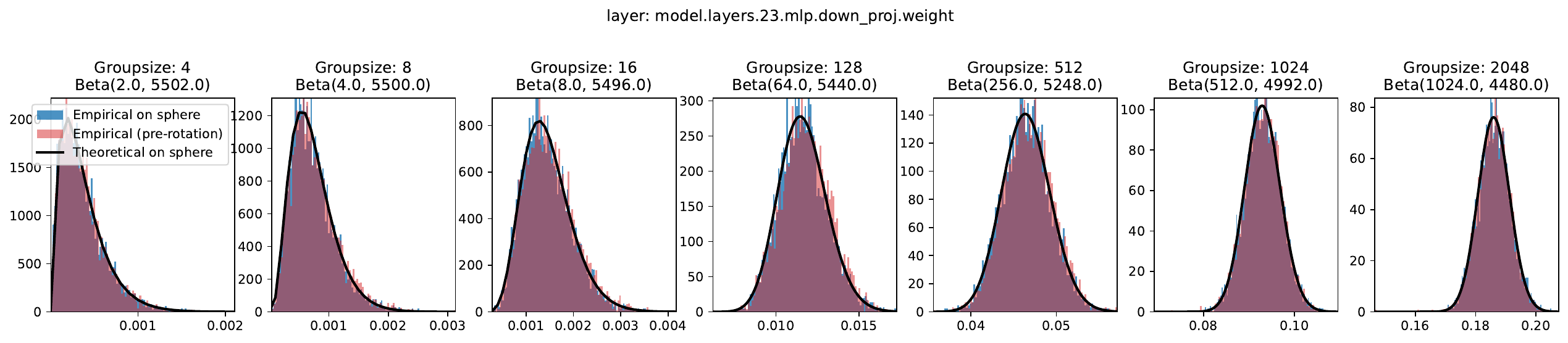} &
\includegraphics[width=0.22\linewidth]{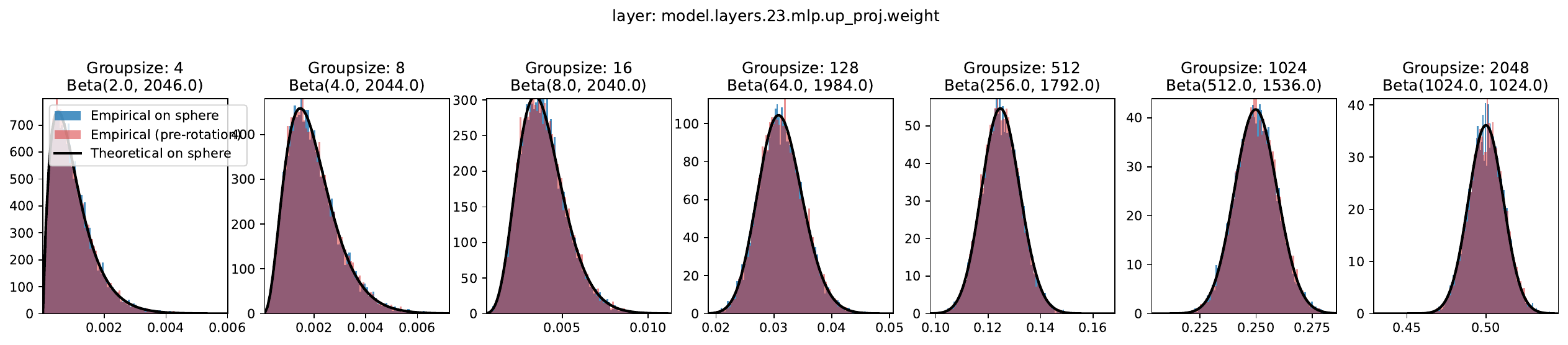} &
\includegraphics[width=0.22\linewidth]{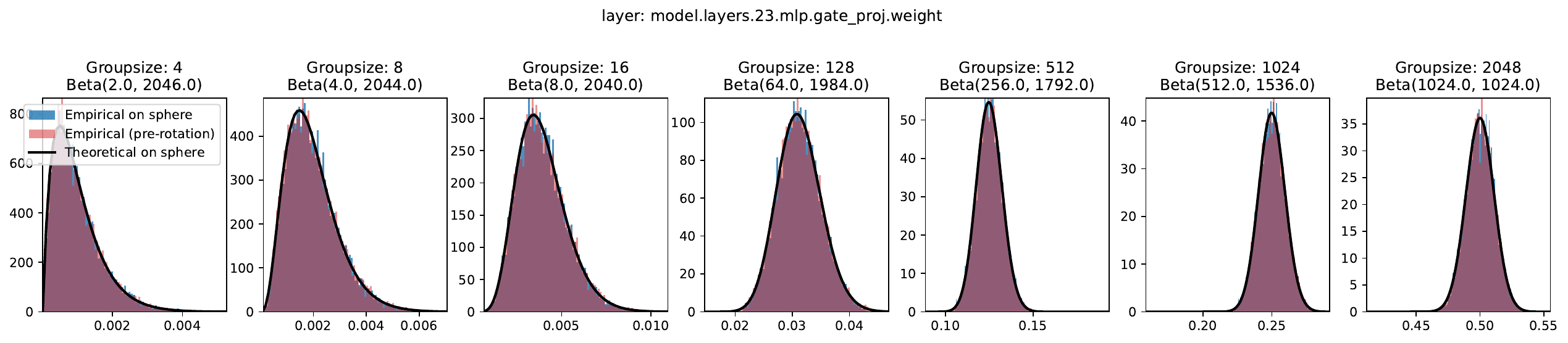} \\
\includegraphics[width=0.22\linewidth]{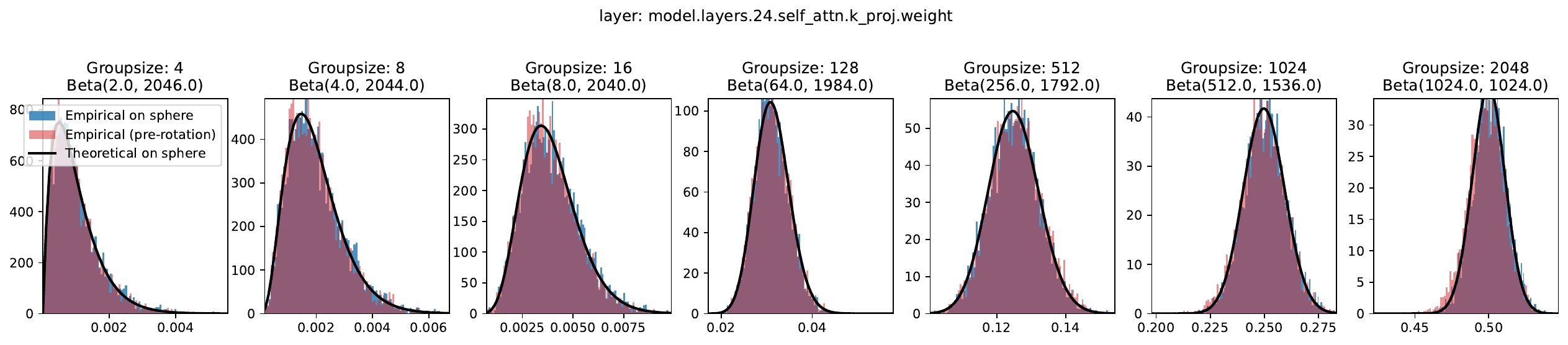} &
\includegraphics[width=0.22\linewidth]{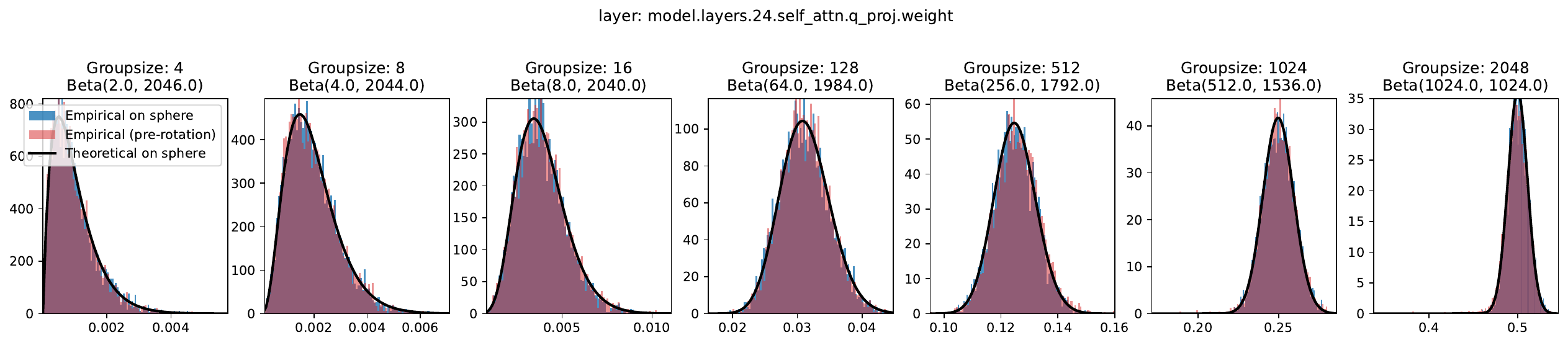} &
\includegraphics[width=0.22\linewidth]{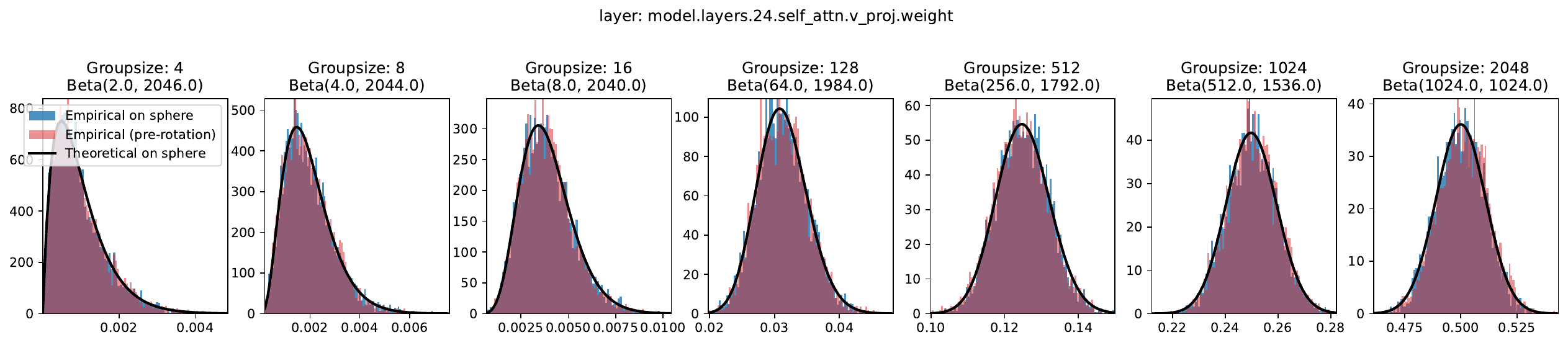} &
\includegraphics[width=0.22\linewidth]{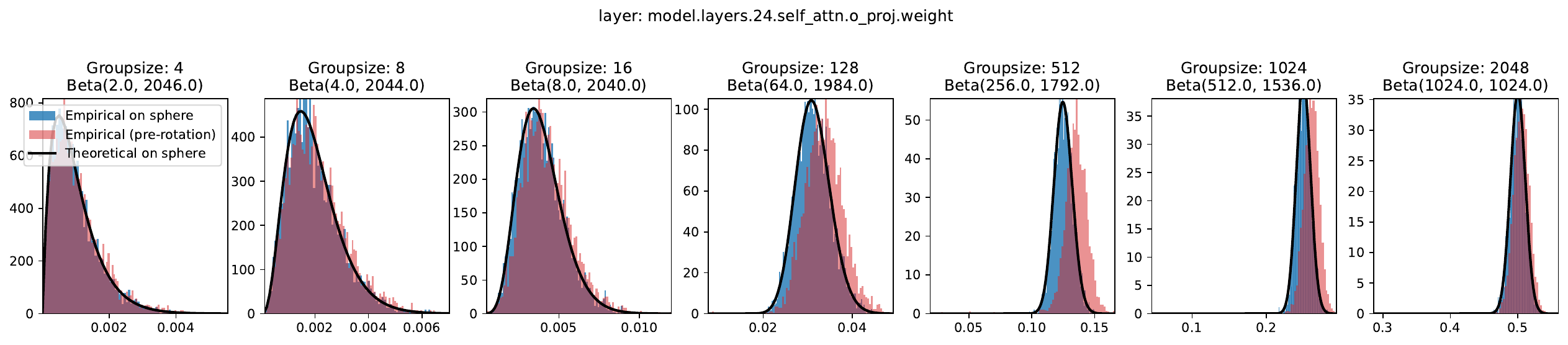} \\
\includegraphics[width=0.22\linewidth]{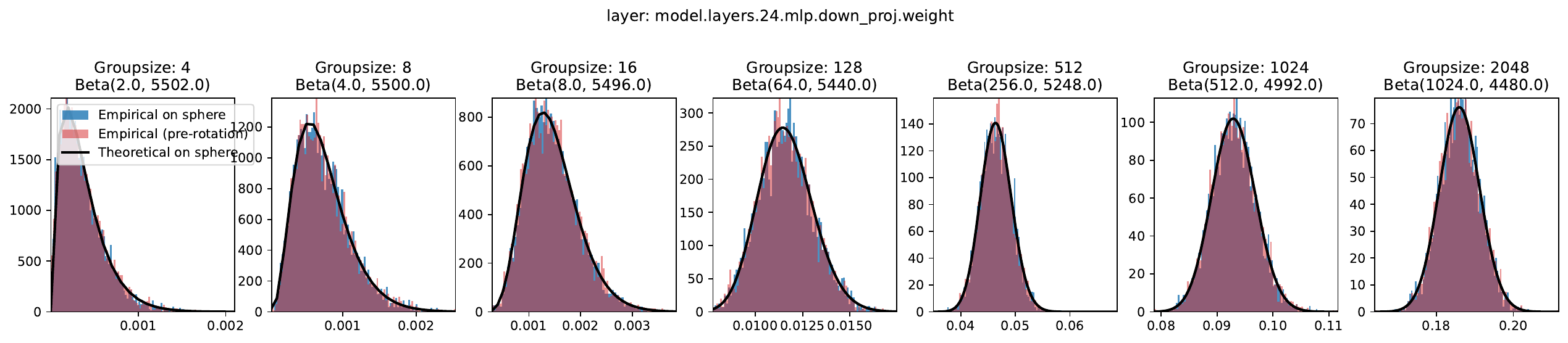} &
\includegraphics[width=0.22\linewidth]{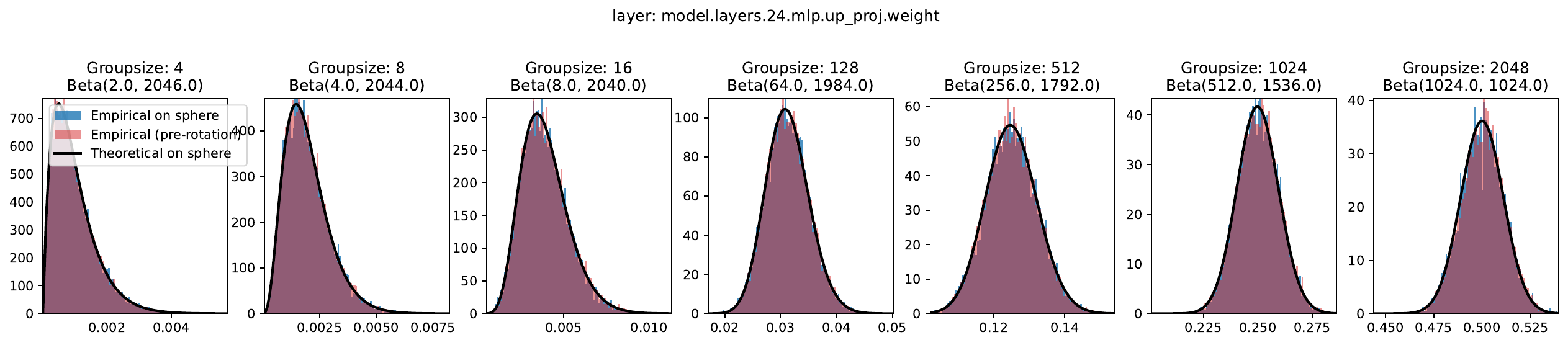} &
\includegraphics[width=0.22\linewidth]{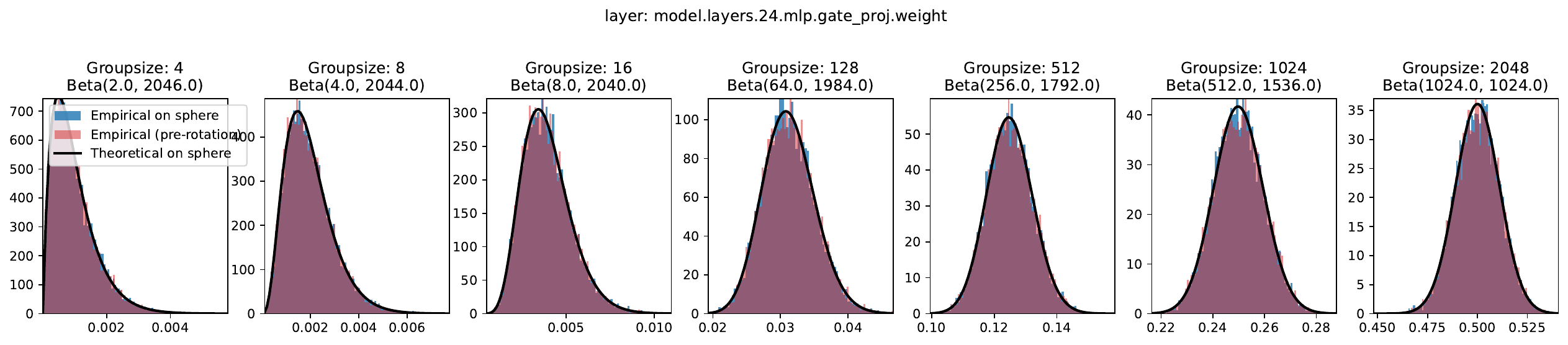} \\
\includegraphics[width=0.22\linewidth]{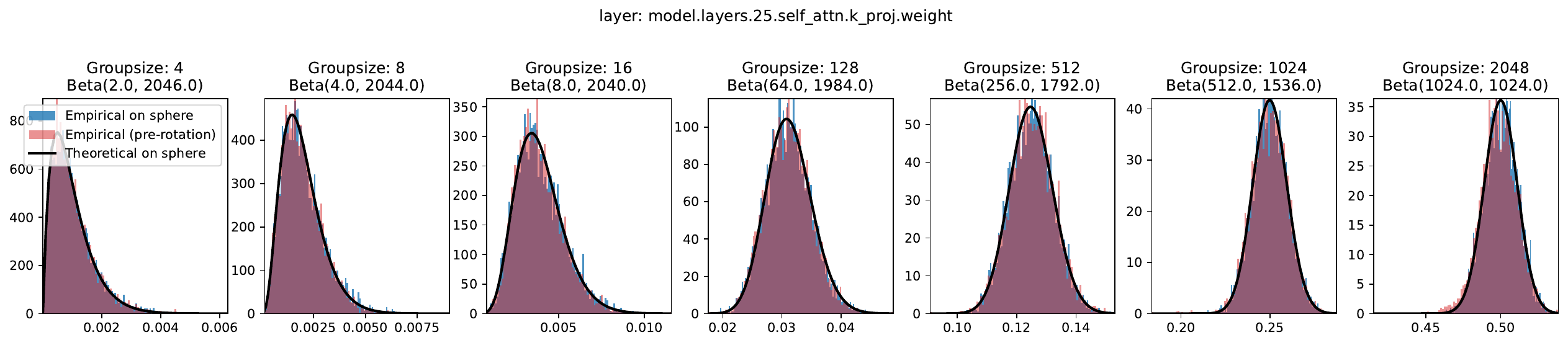} &
\includegraphics[width=0.22\linewidth]{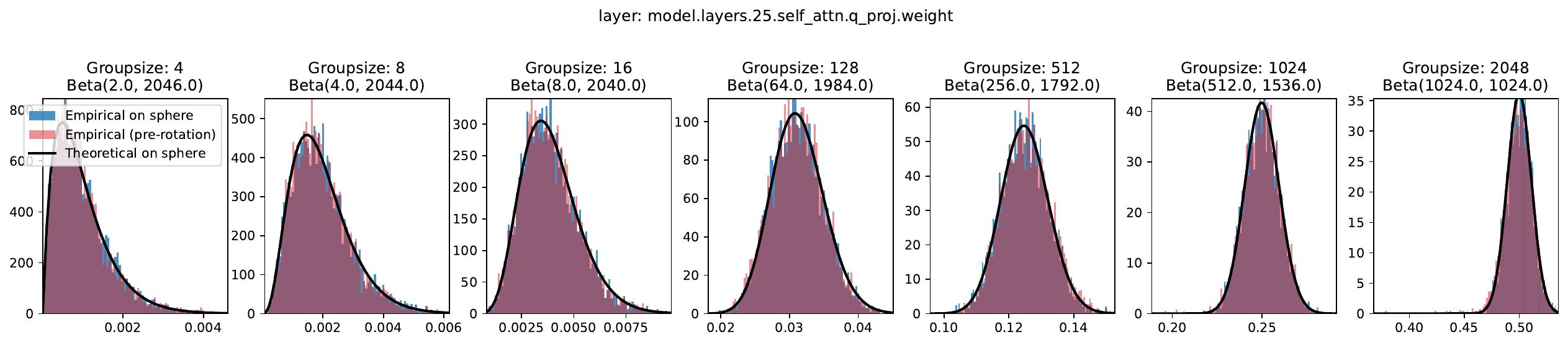} &
\includegraphics[width=0.22\linewidth]{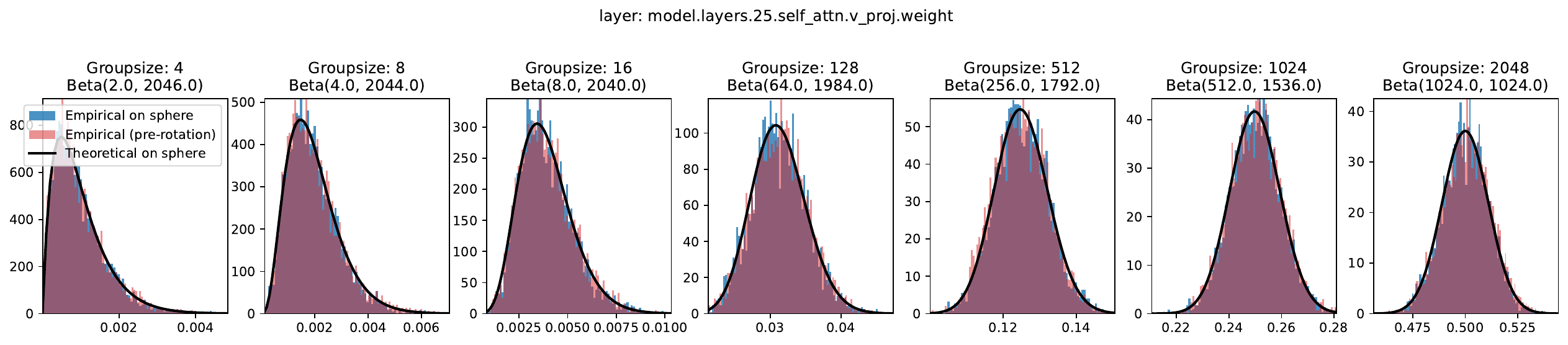} &
\includegraphics[width=0.22\linewidth]{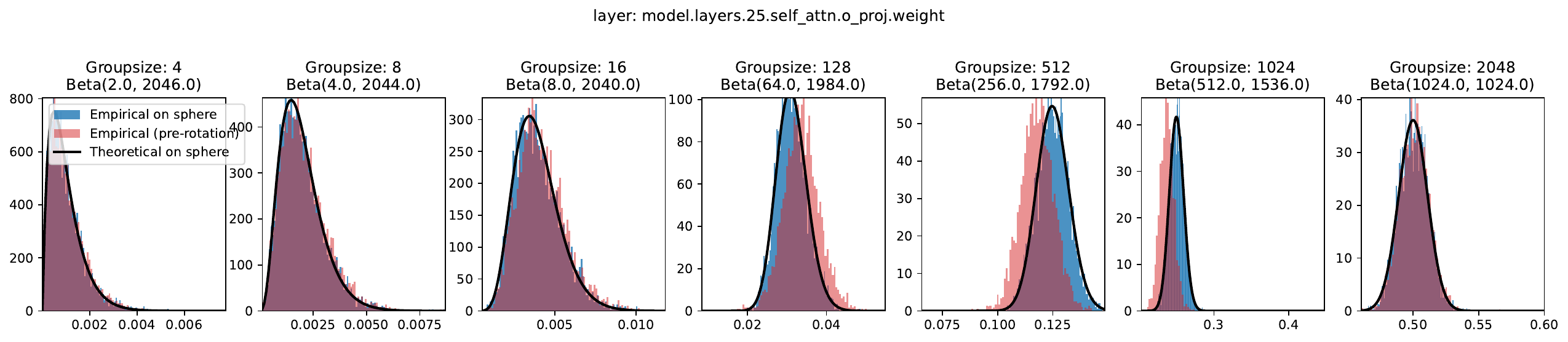} \\
\includegraphics[width=0.22\linewidth]{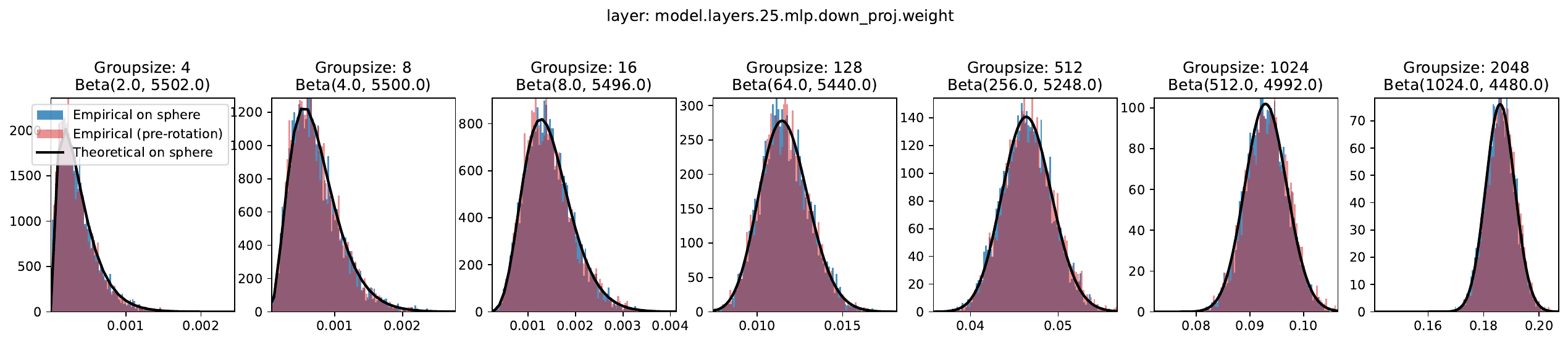} &
\includegraphics[width=0.22\linewidth]{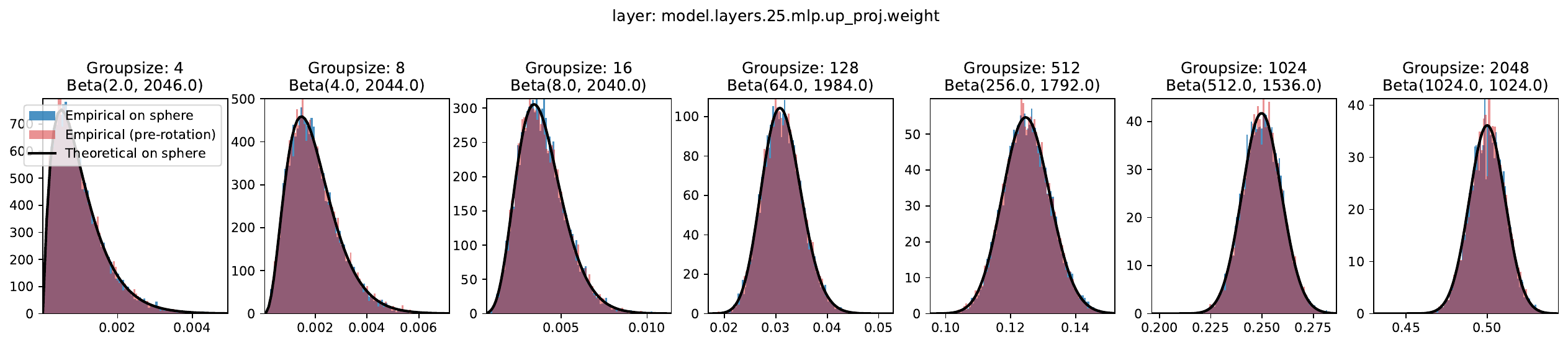} &
\includegraphics[width=0.22\linewidth]{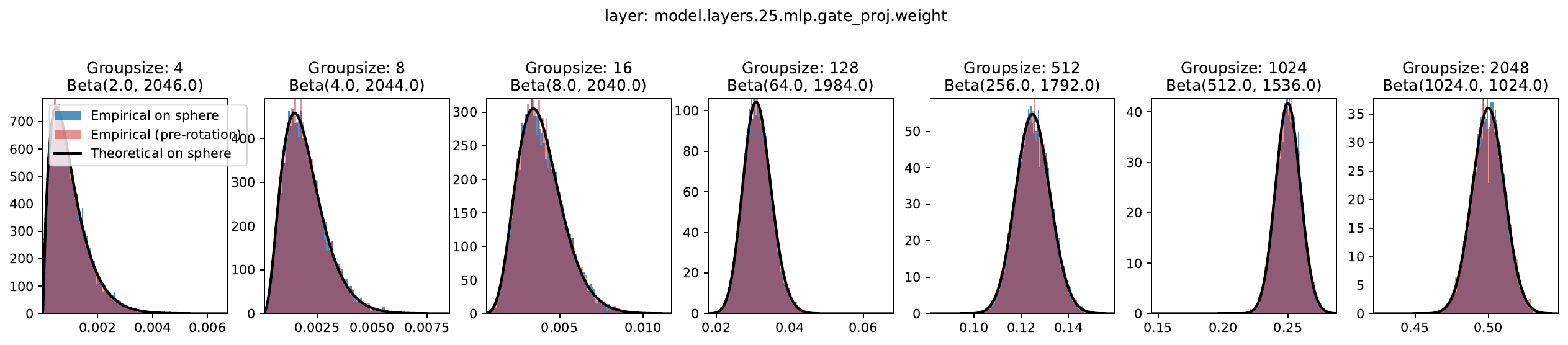} \\
\includegraphics[width=0.22\linewidth]{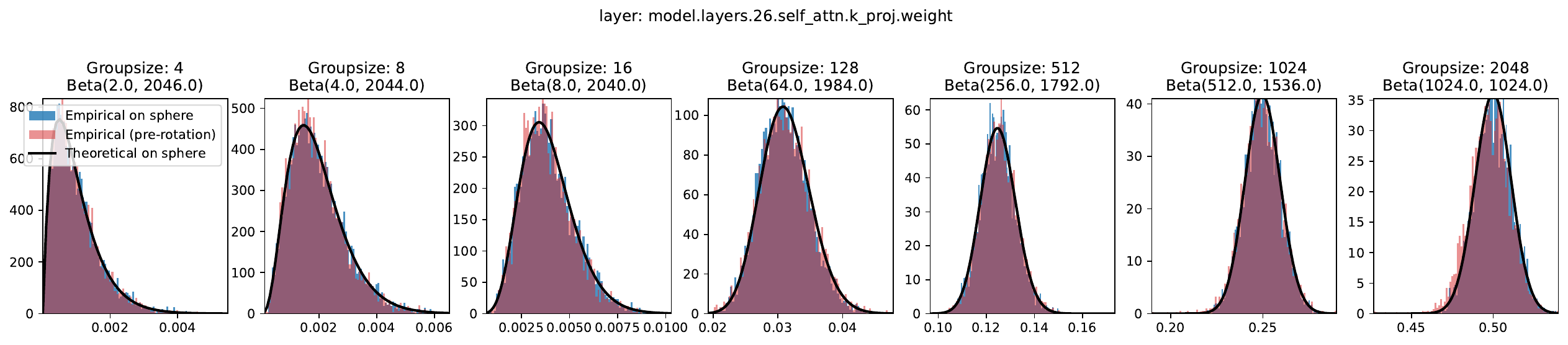} &
\includegraphics[width=0.22\linewidth]{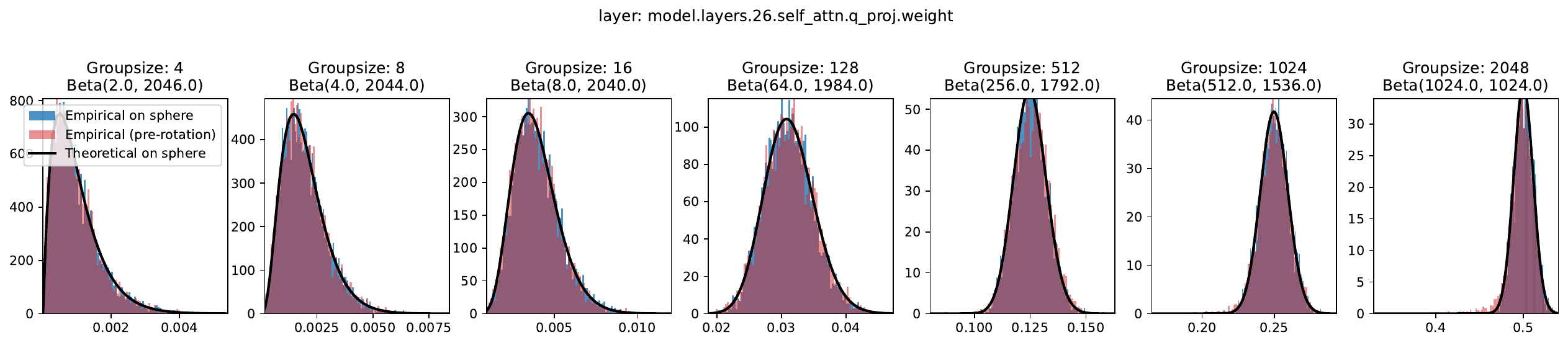} &
\includegraphics[width=0.22\linewidth]{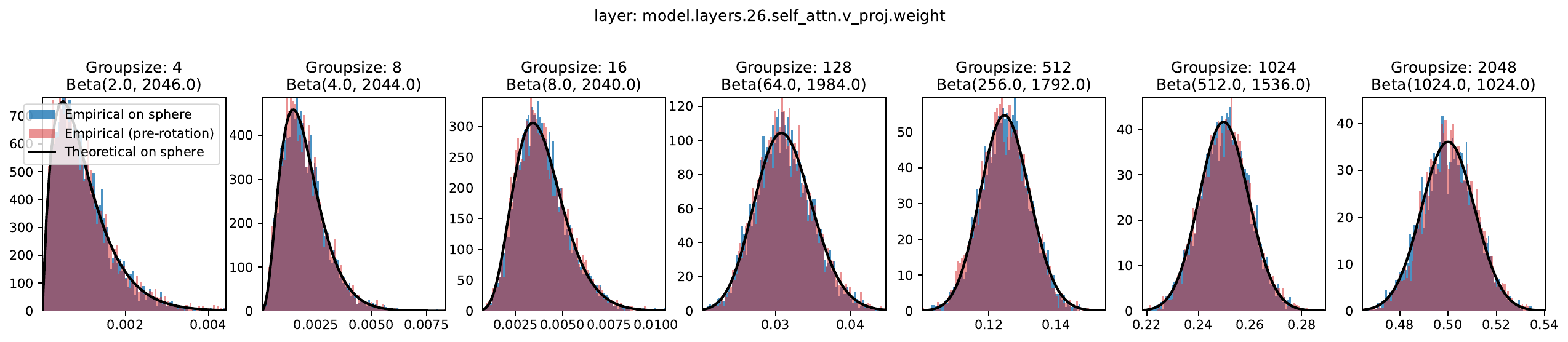} &
\includegraphics[width=0.22\linewidth]{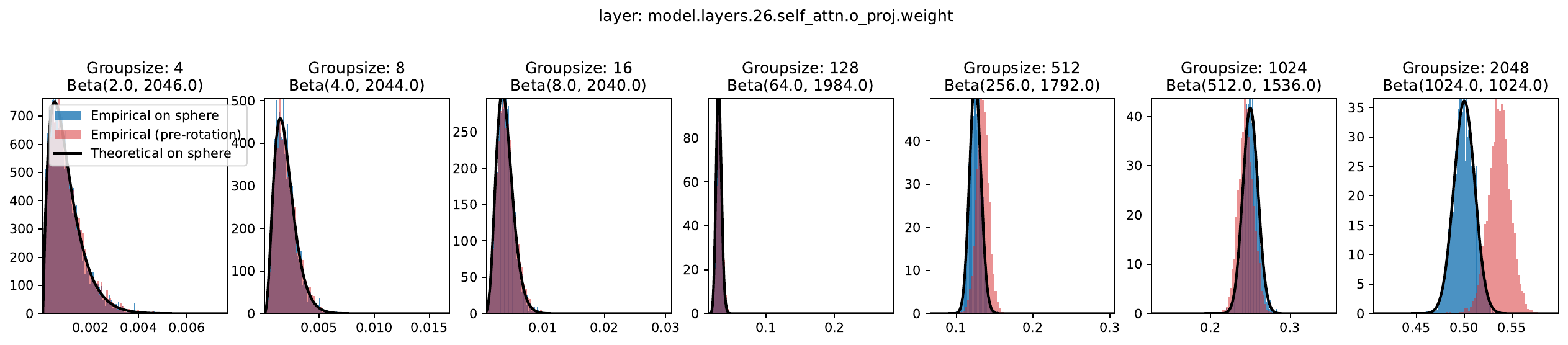} \\
\includegraphics[width=0.22\linewidth]{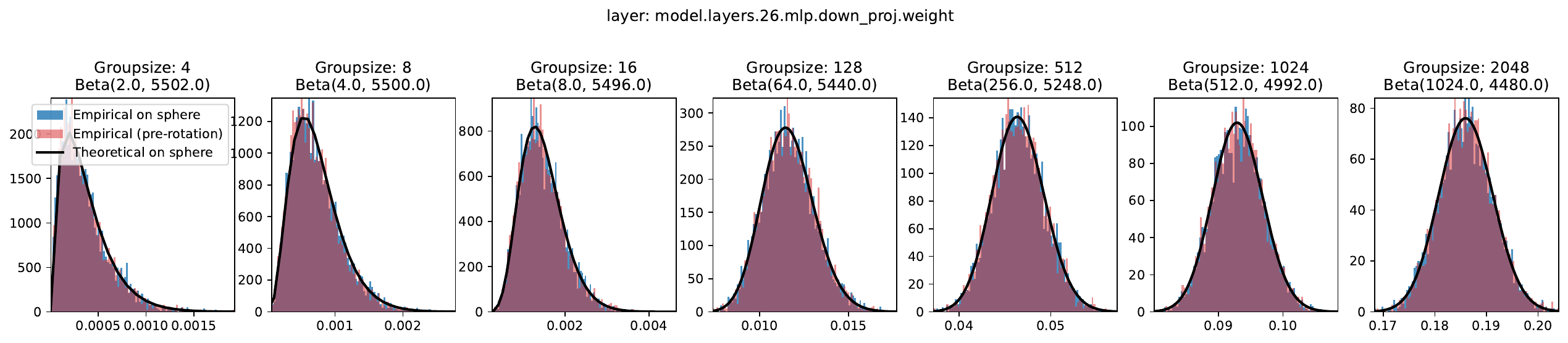} &
\includegraphics[width=0.22\linewidth]{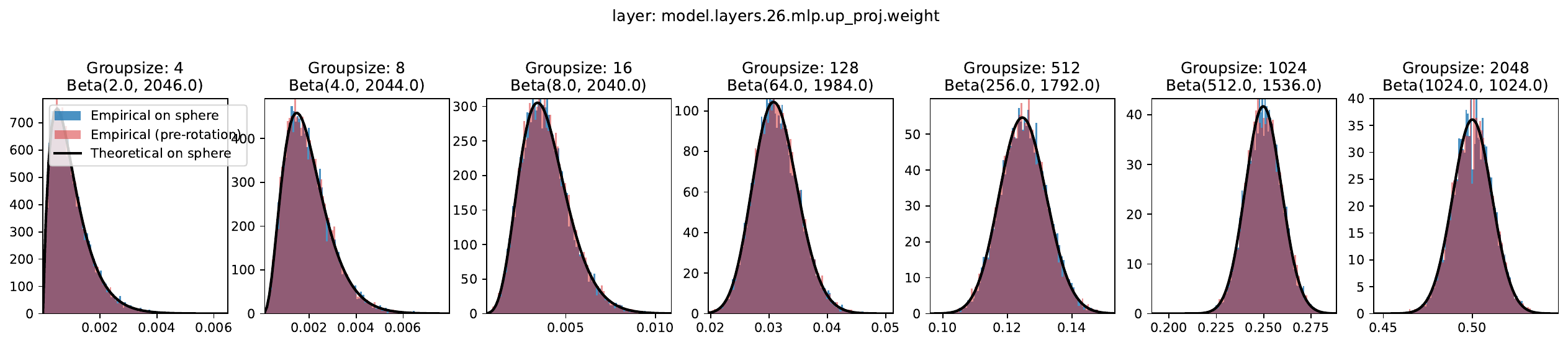} &
\includegraphics[width=0.22\linewidth]{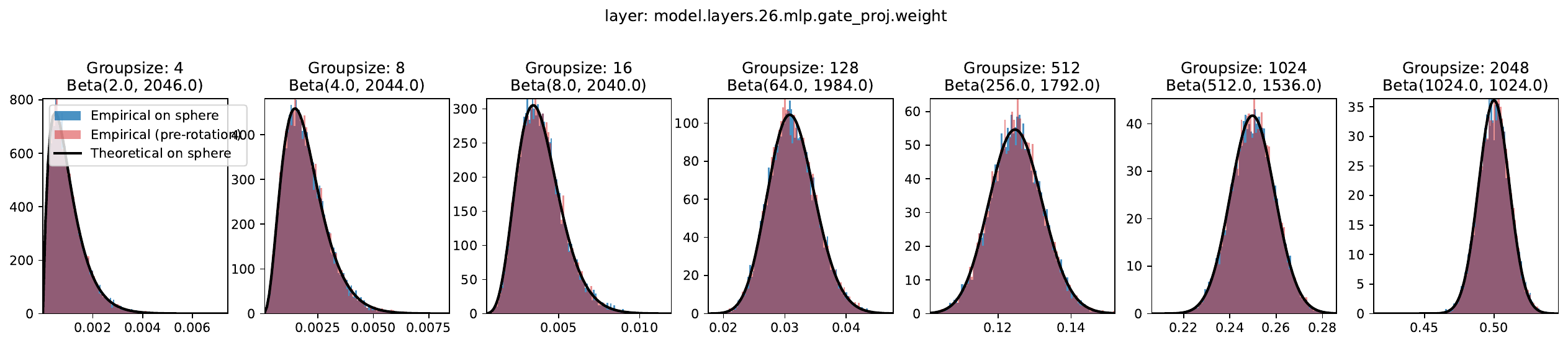} \\
\includegraphics[width=0.22\linewidth]{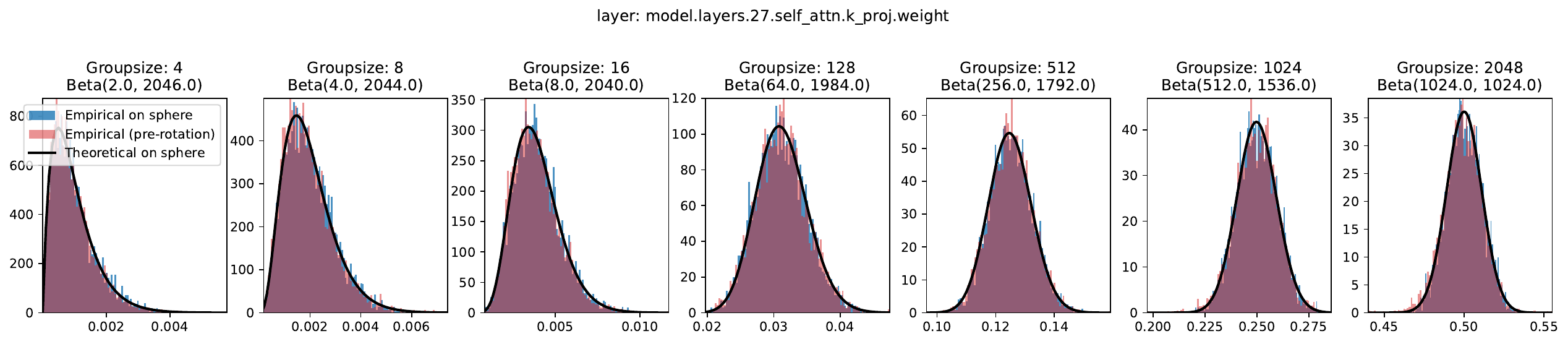} &
\includegraphics[width=0.22\linewidth]{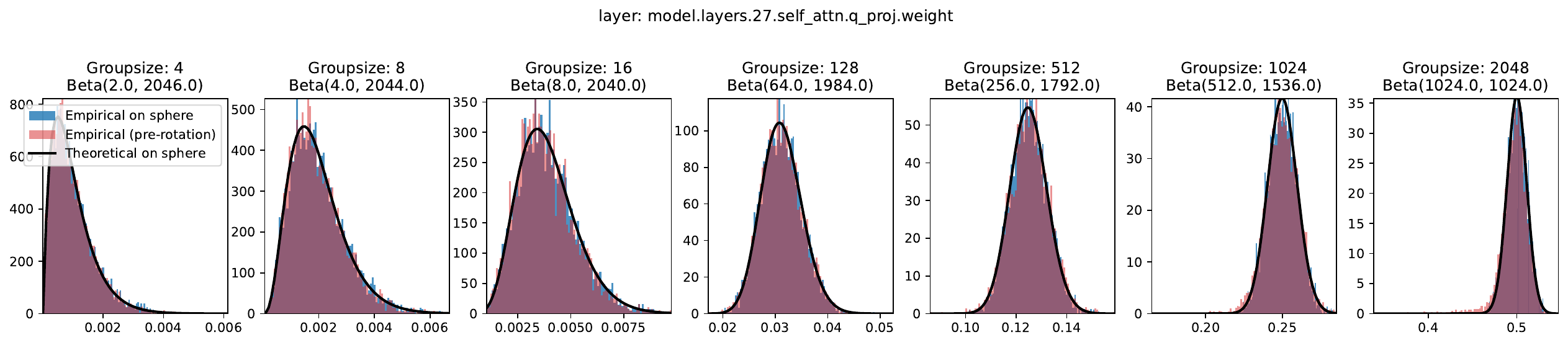} &
\includegraphics[width=0.22\linewidth]{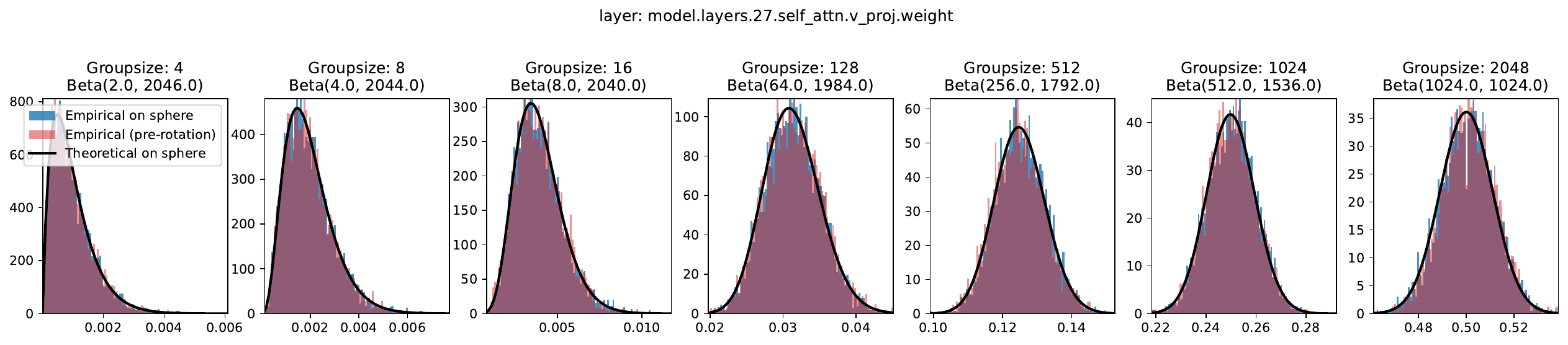} &
\includegraphics[width=0.22\linewidth]{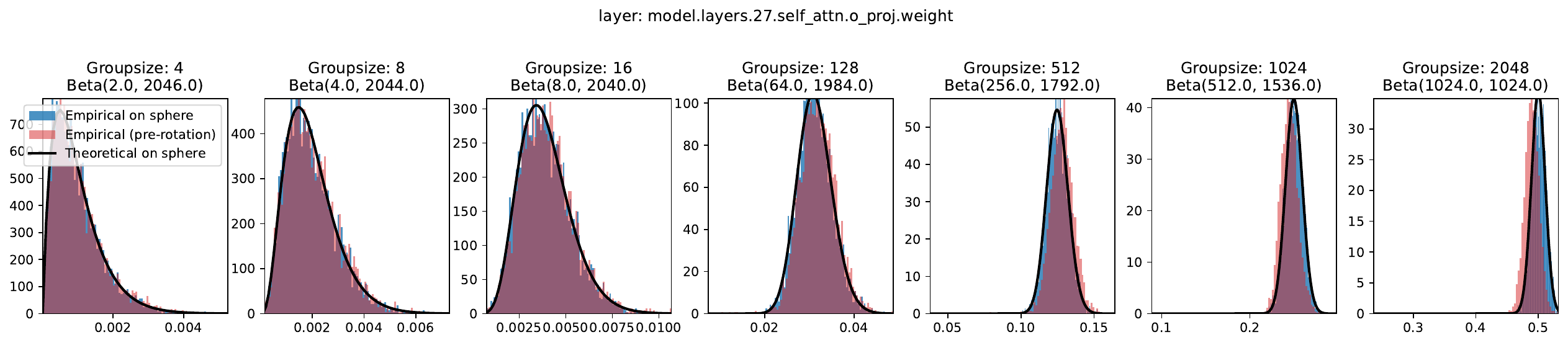} \\
\includegraphics[width=0.22\linewidth]{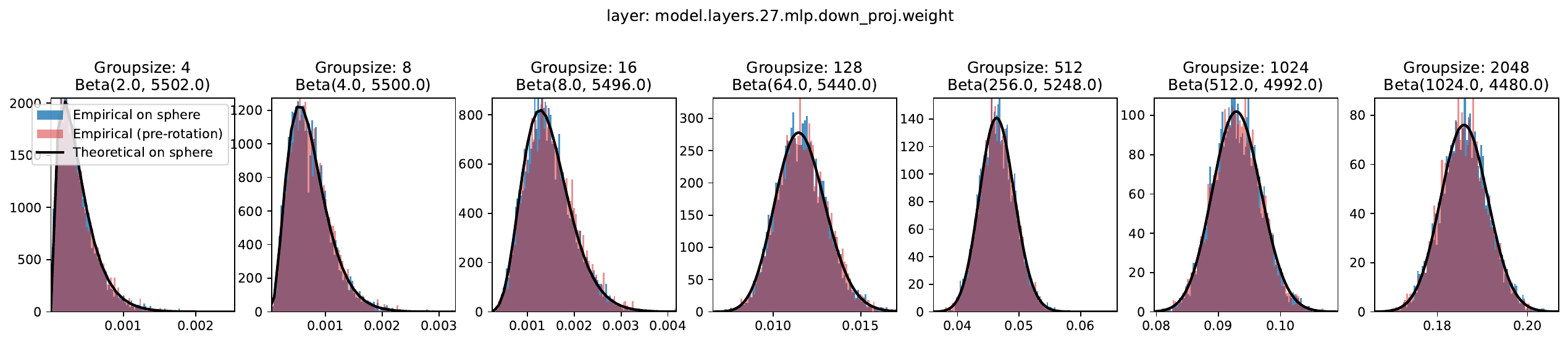} &
\includegraphics[width=0.22\linewidth]{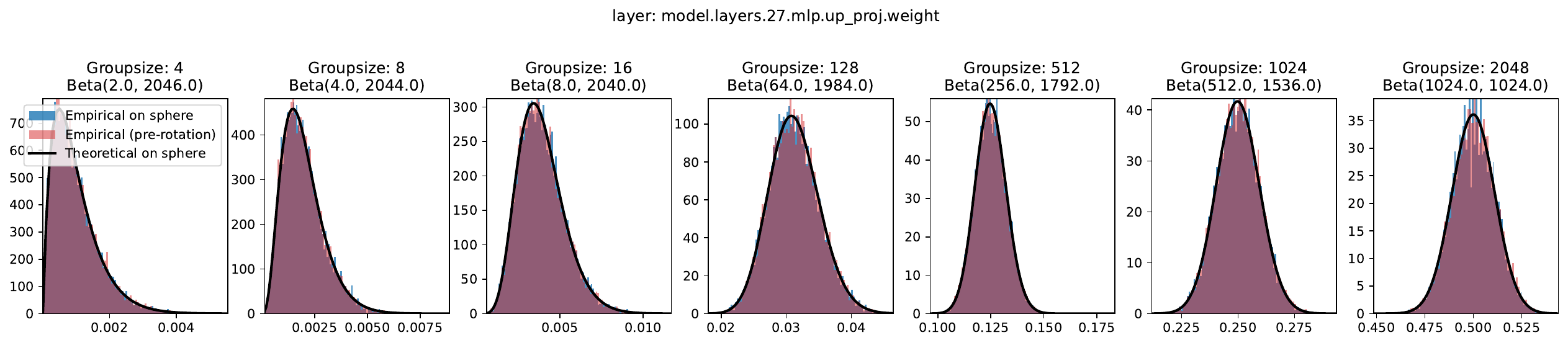} &
\includegraphics[width=0.22\linewidth]{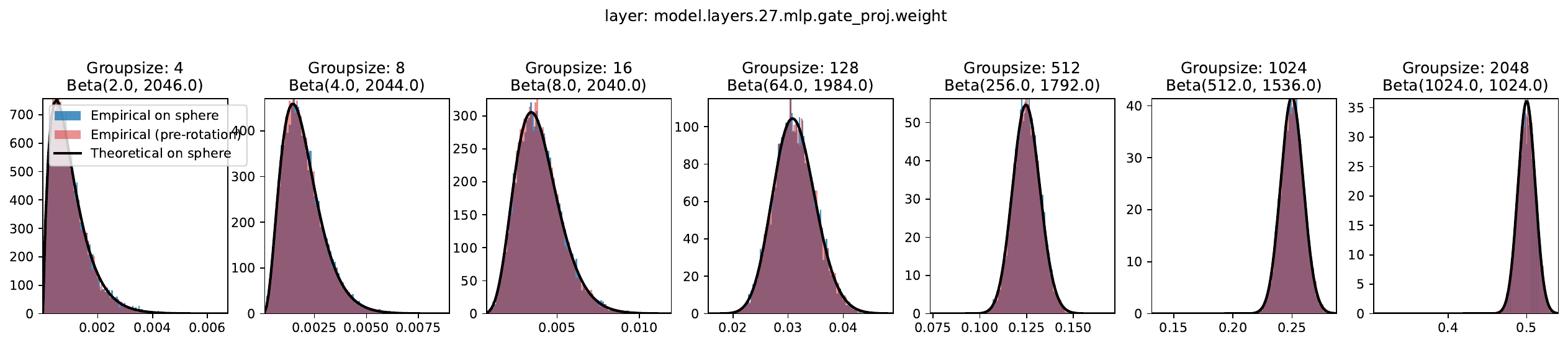} \\
\includegraphics[width=0.22\linewidth]{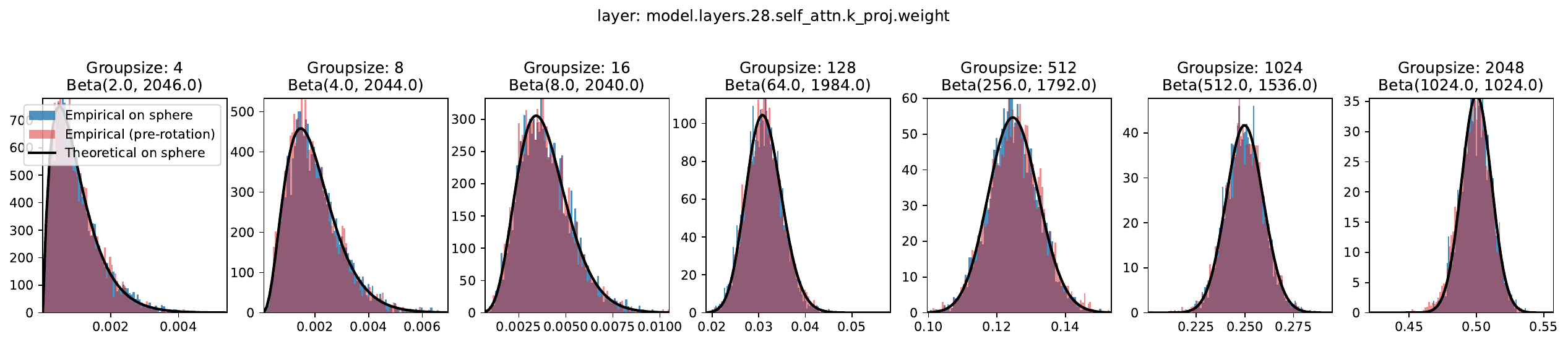} &
\includegraphics[width=0.22\linewidth]{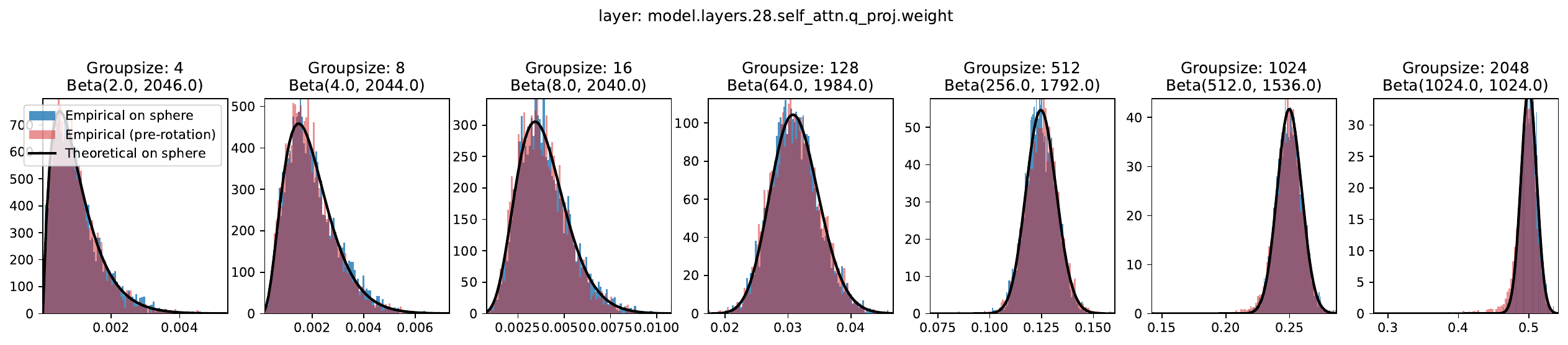} &
\includegraphics[width=0.22\linewidth]{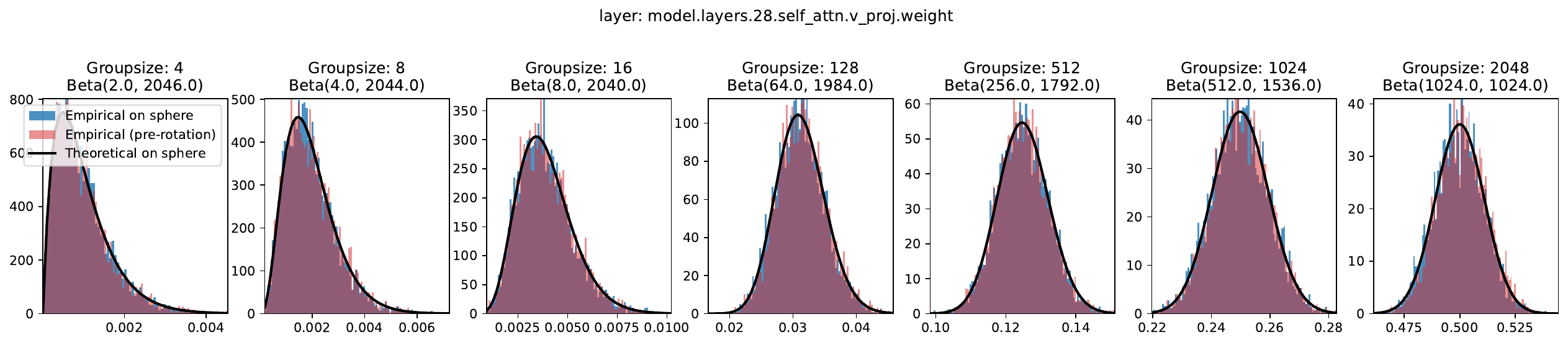} &
\includegraphics[width=0.22\linewidth]{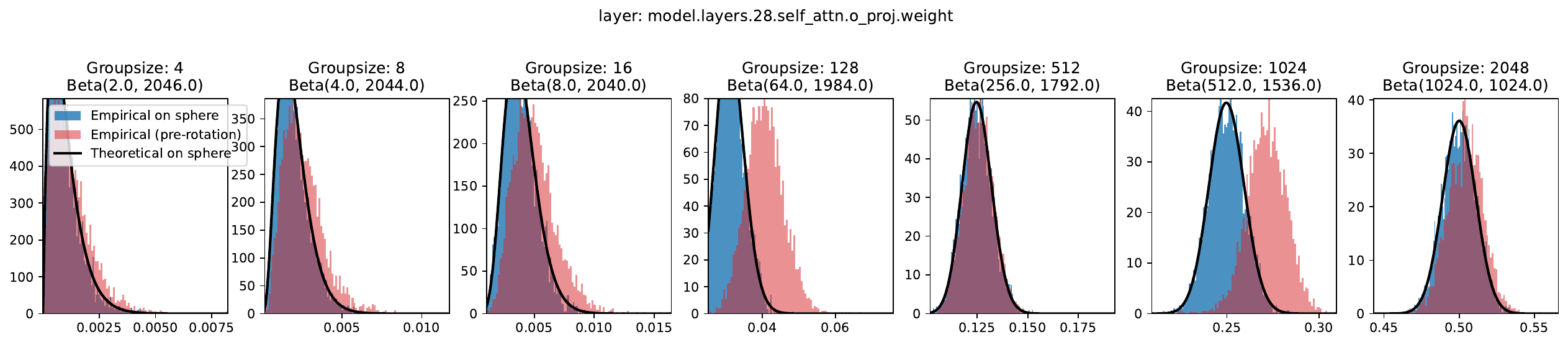} \\
\includegraphics[width=0.22\linewidth]{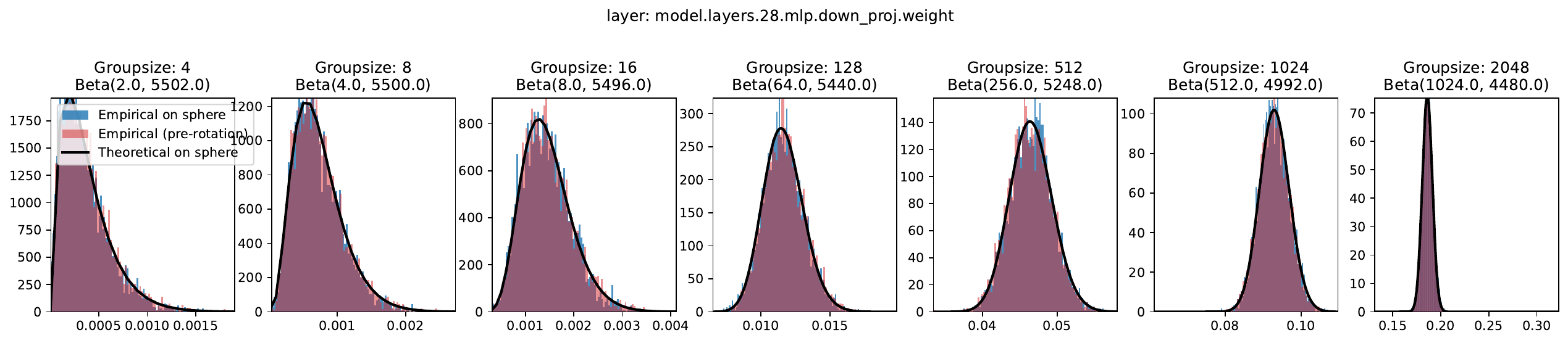} &
\includegraphics[width=0.22\linewidth]{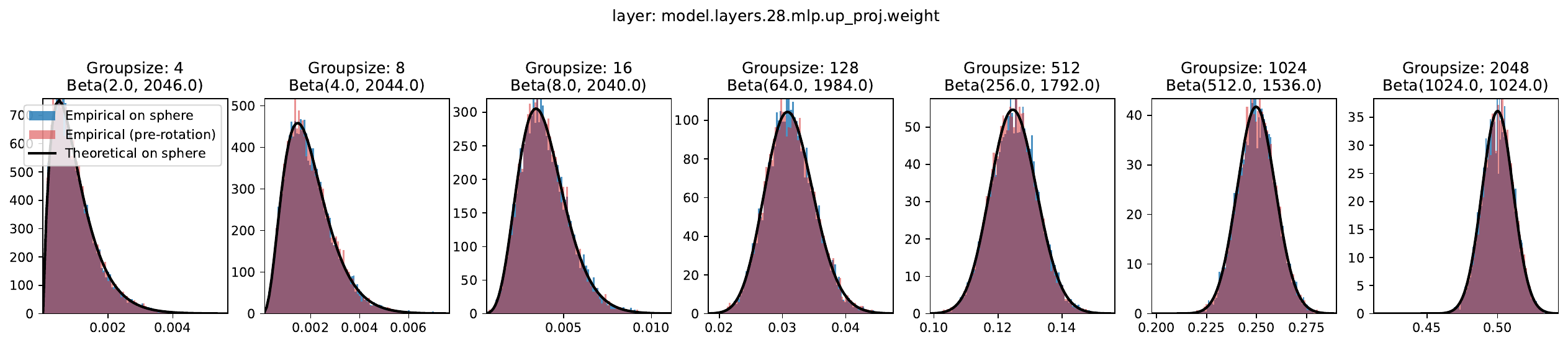} &
\includegraphics[width=0.22\linewidth]{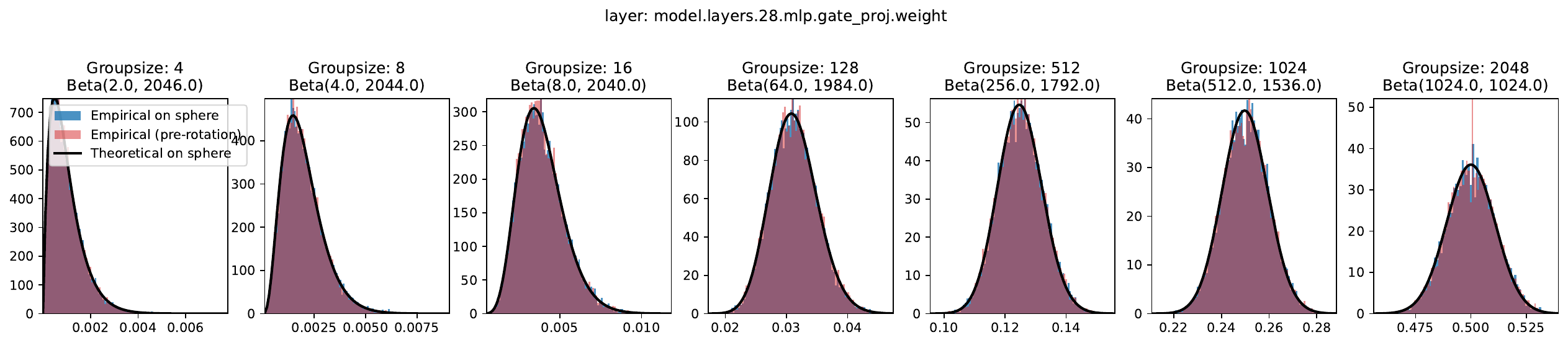} \\
\includegraphics[width=0.22\linewidth]{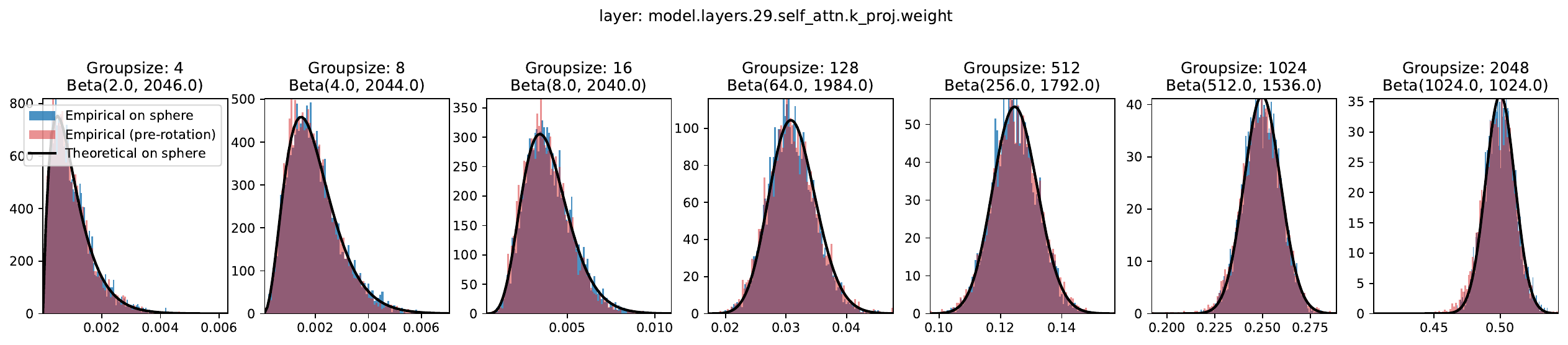} &
\includegraphics[width=0.22\linewidth]{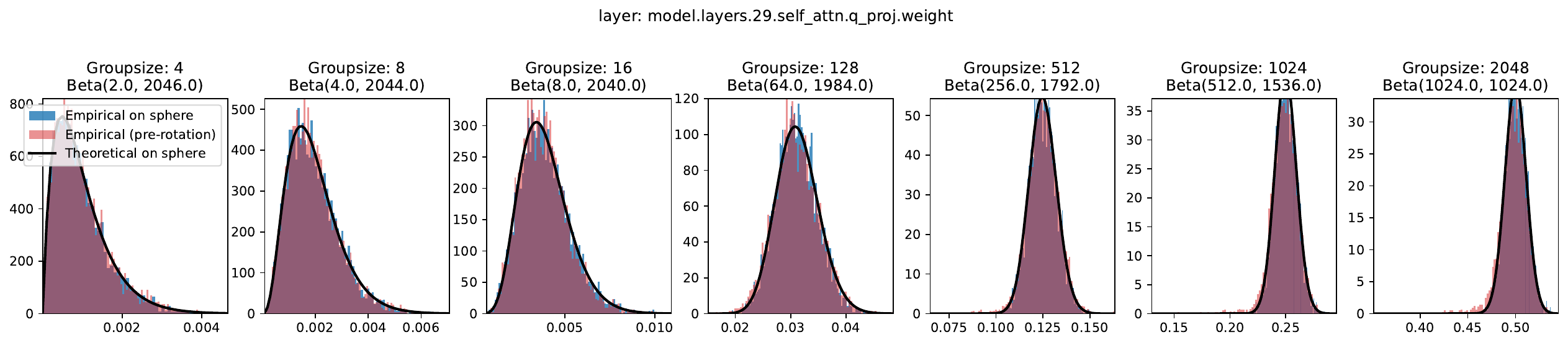} &
\includegraphics[width=0.22\linewidth]{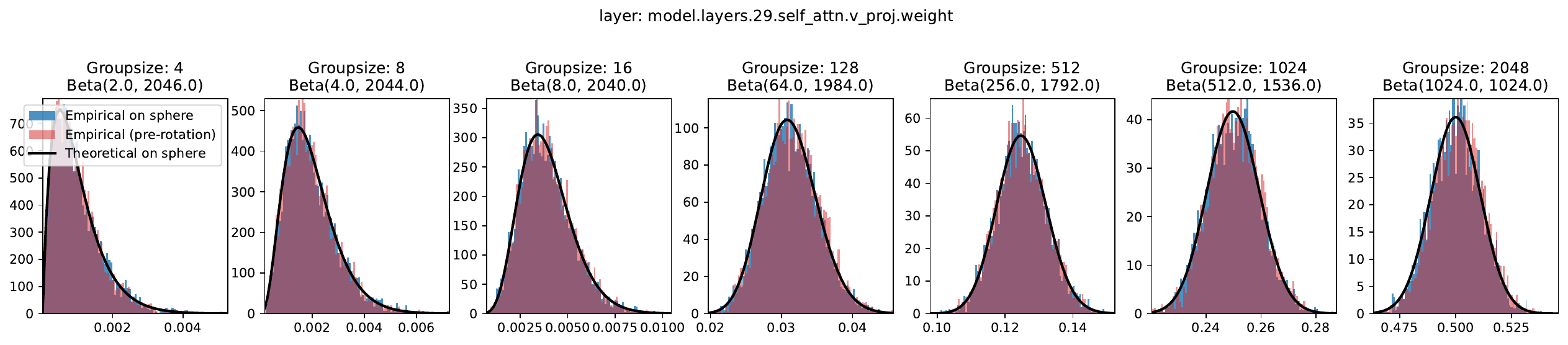} &
\includegraphics[width=0.22\linewidth]{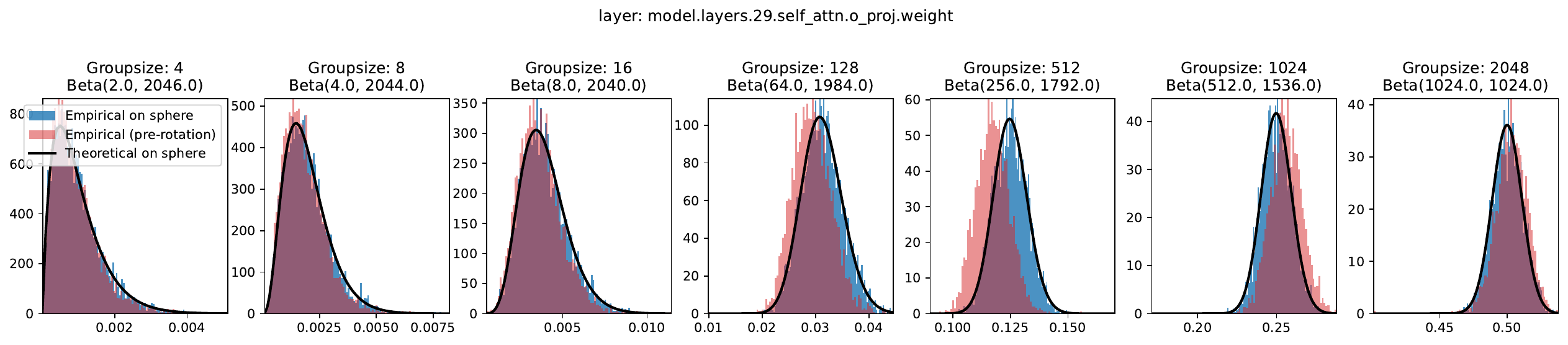} \\
\includegraphics[width=0.22\linewidth]{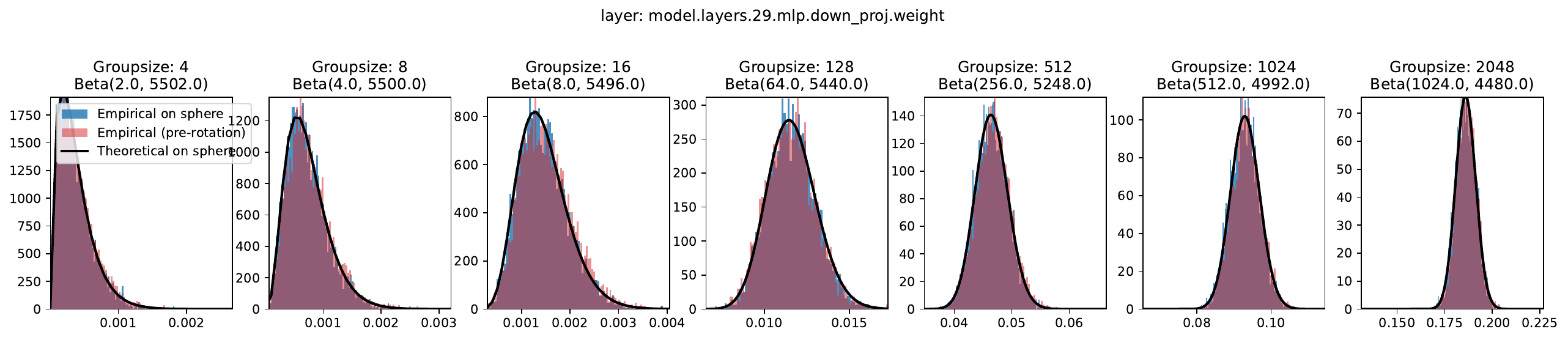} &
\includegraphics[width=0.22\linewidth]{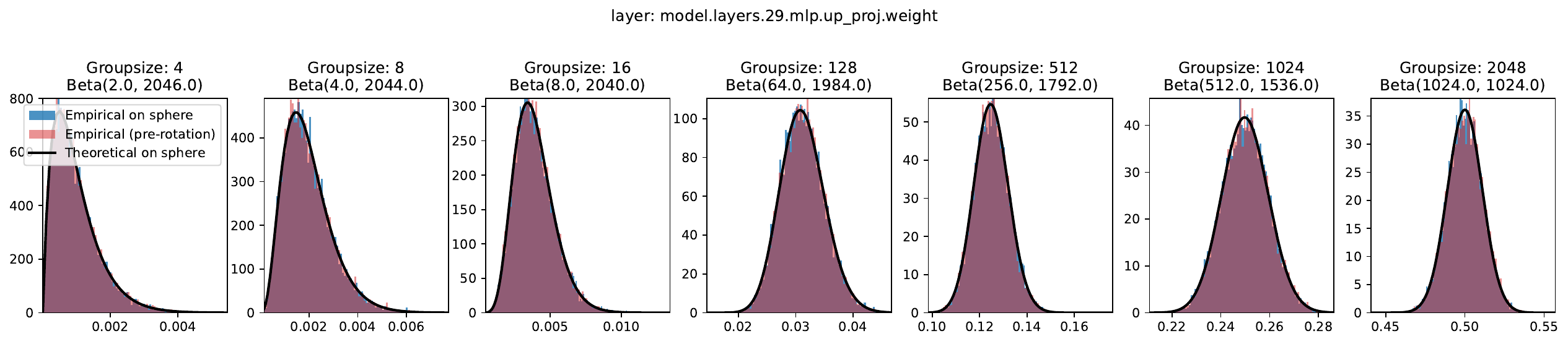} &
\includegraphics[width=0.22\linewidth]{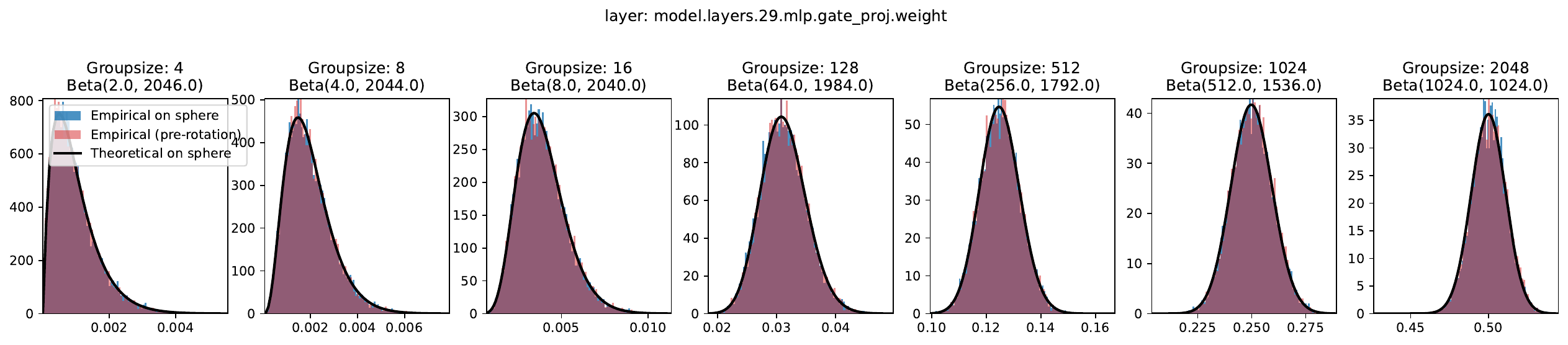} \\
\includegraphics[width=0.22\linewidth]{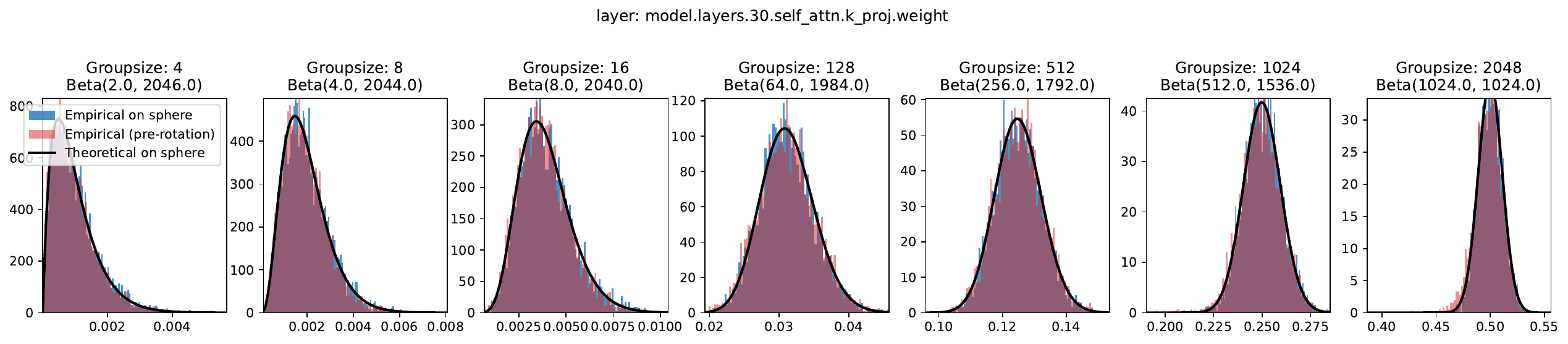} &
\includegraphics[width=0.22\linewidth]{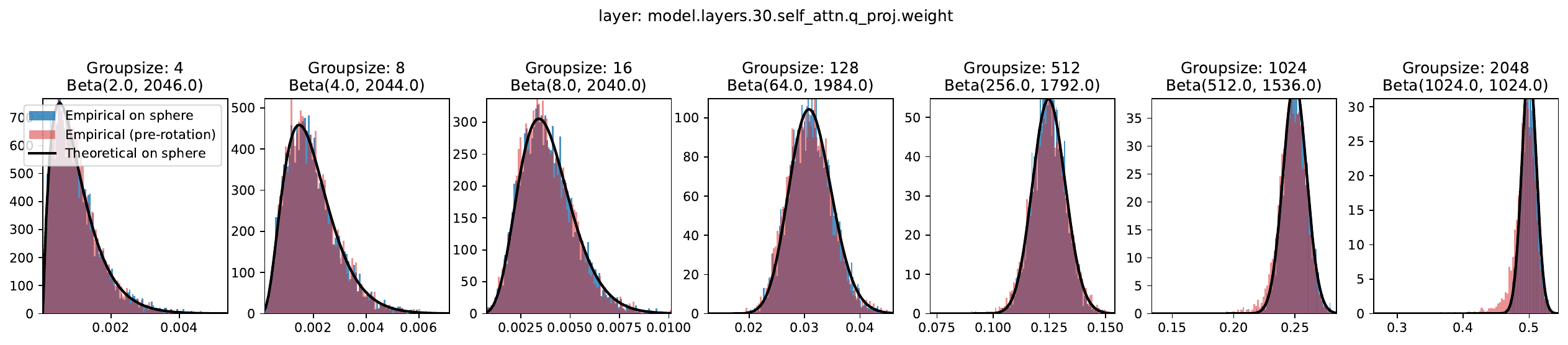} &
\includegraphics[width=0.22\linewidth]{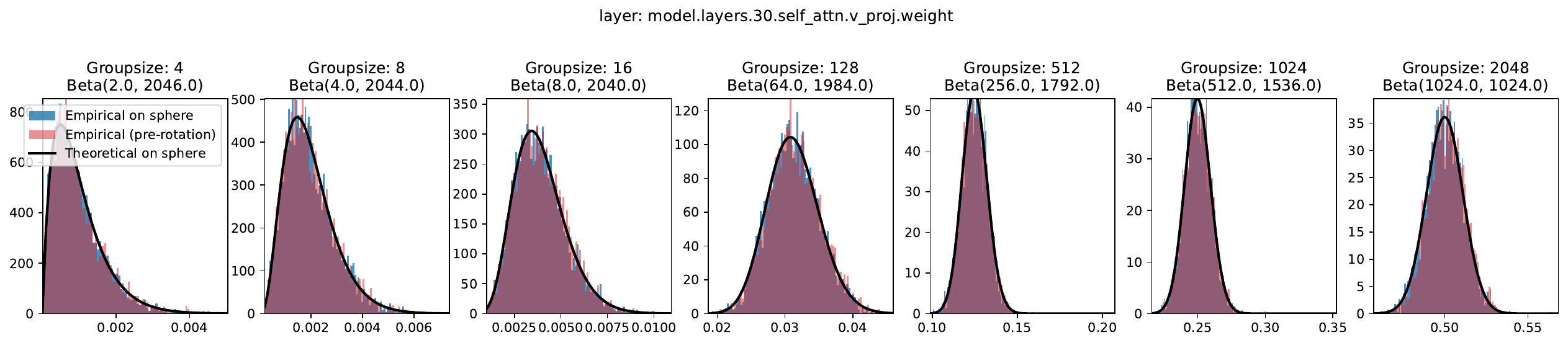} &
\includegraphics[width=0.22\linewidth]{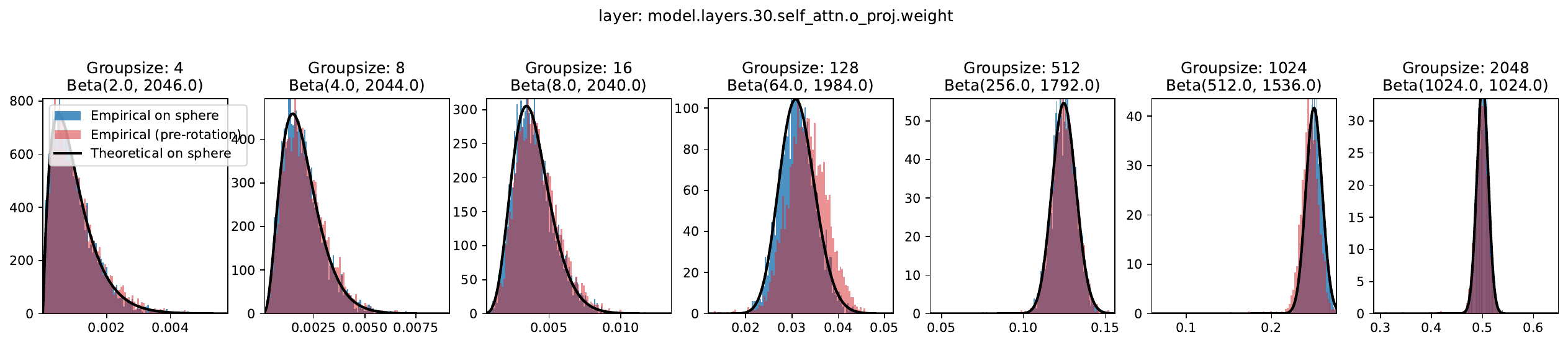} \\
\includegraphics[width=0.22\linewidth]{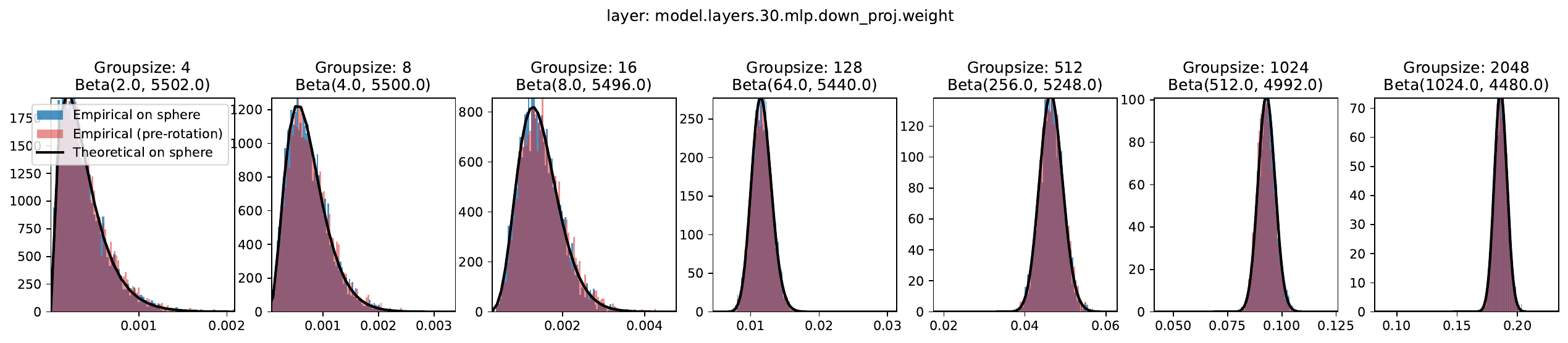} &
\includegraphics[width=0.22\linewidth]{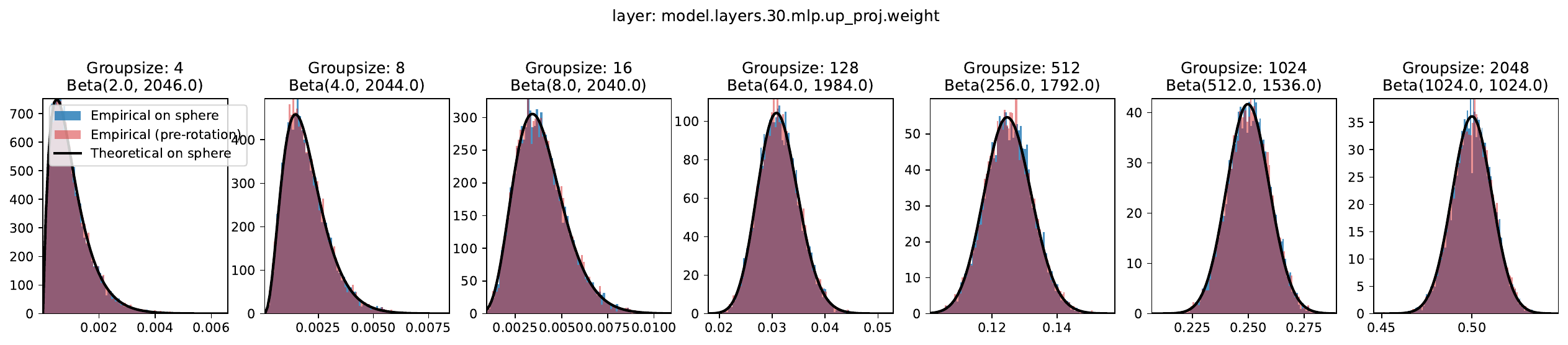} &
\includegraphics[width=0.22\linewidth]{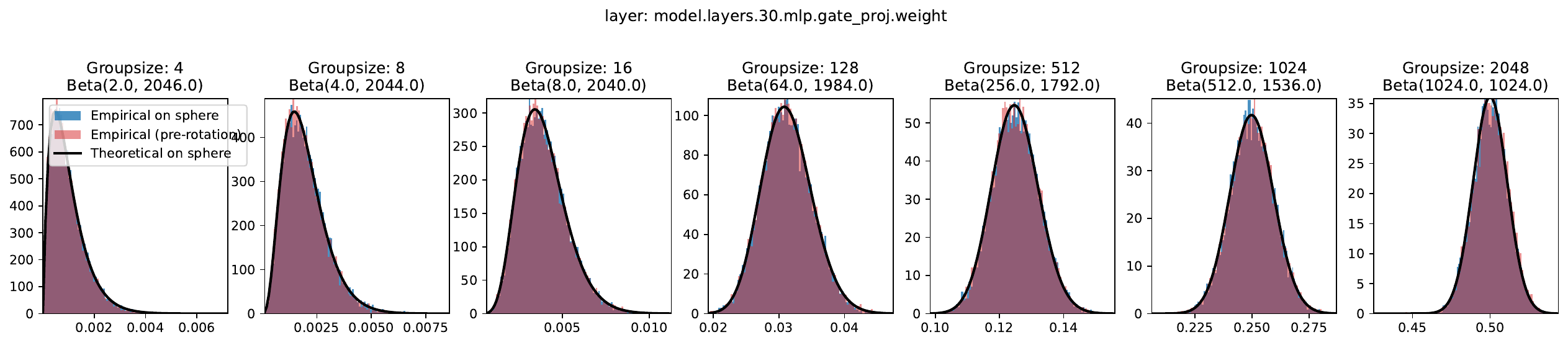} \\
\includegraphics[width=0.22\linewidth]{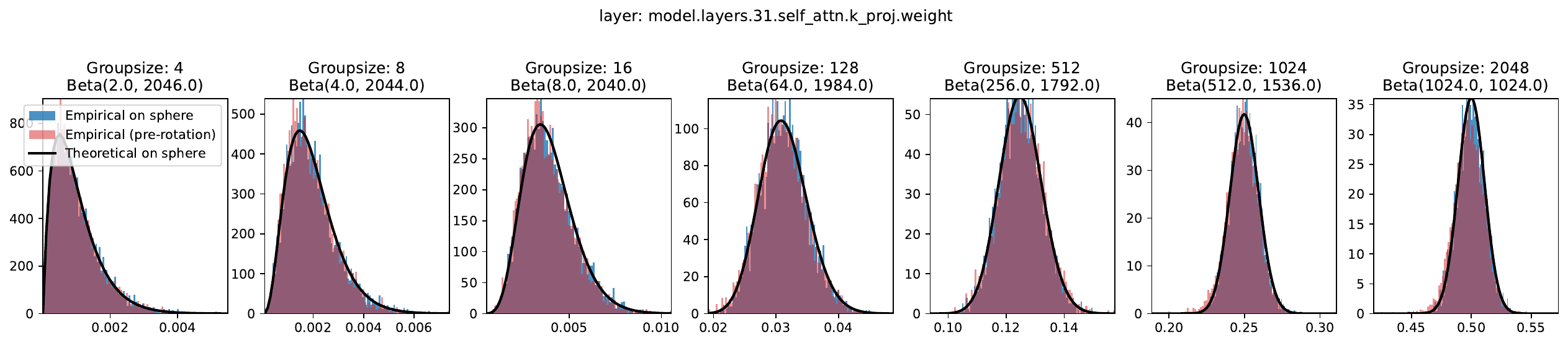} &
\includegraphics[width=0.22\linewidth]{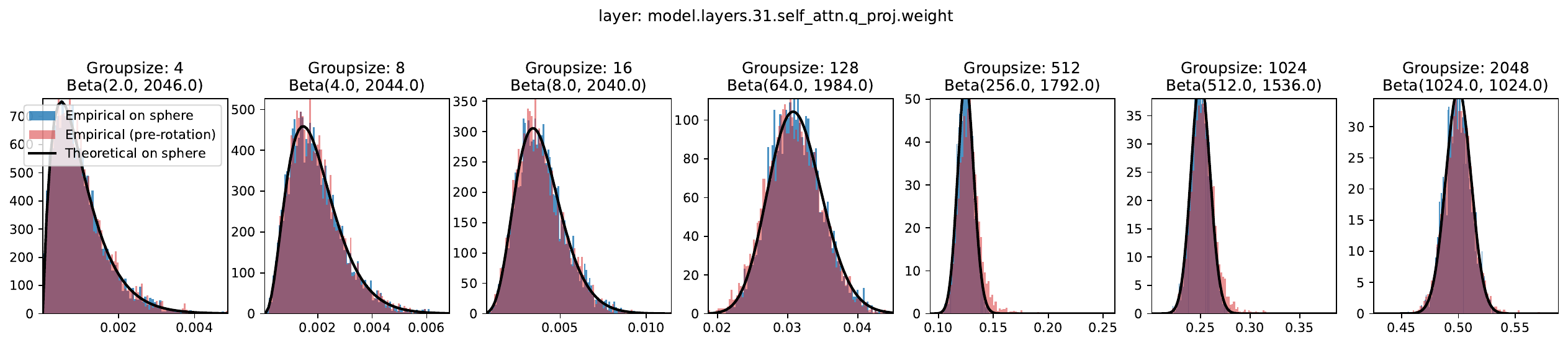} &
\includegraphics[width=0.22\linewidth]{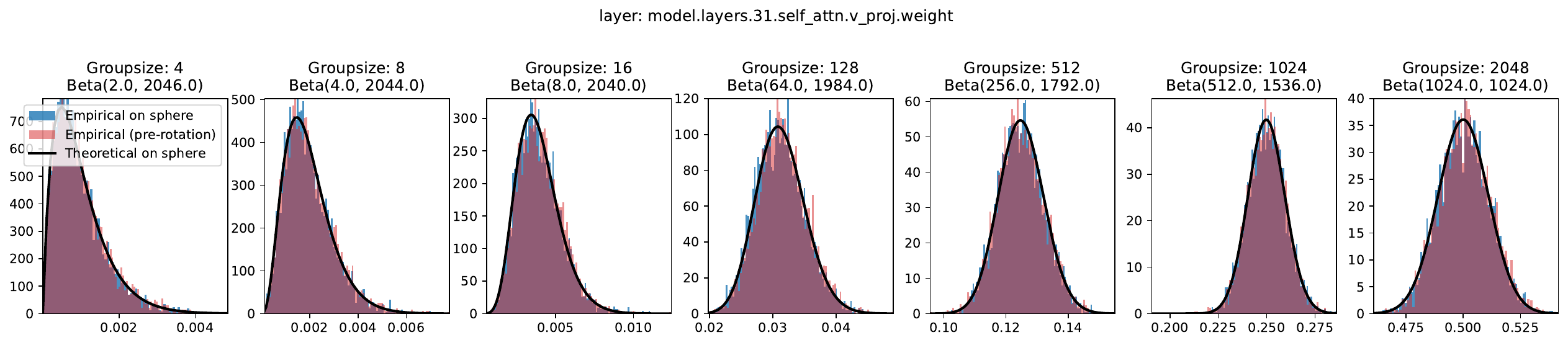} &
\includegraphics[width=0.22\linewidth]{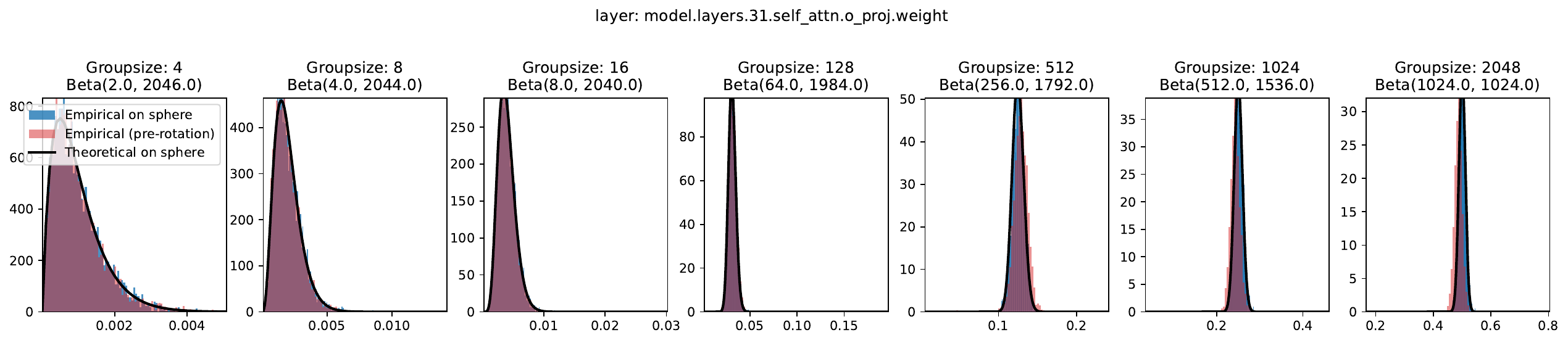} \\
\includegraphics[width=0.22\linewidth]{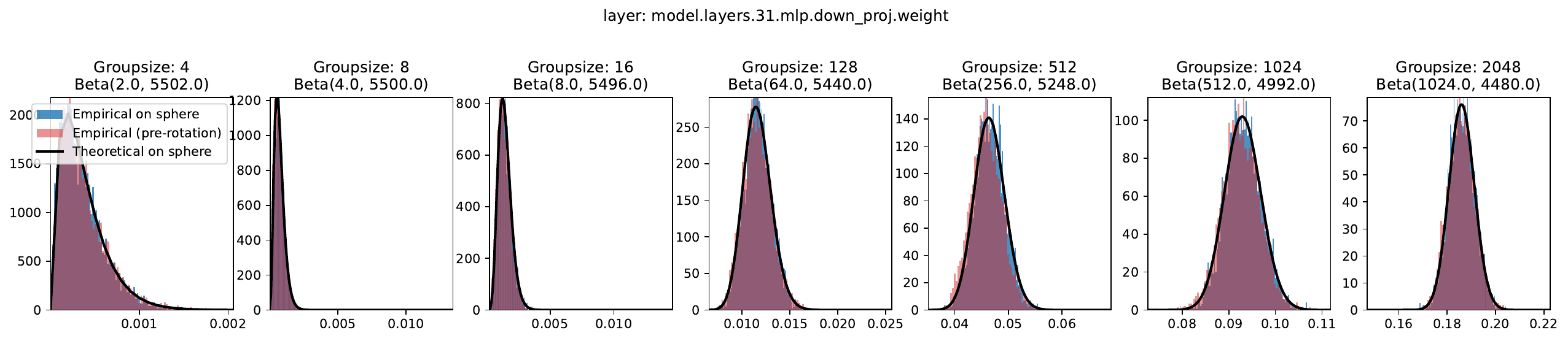} &
\includegraphics[width=0.22\linewidth]{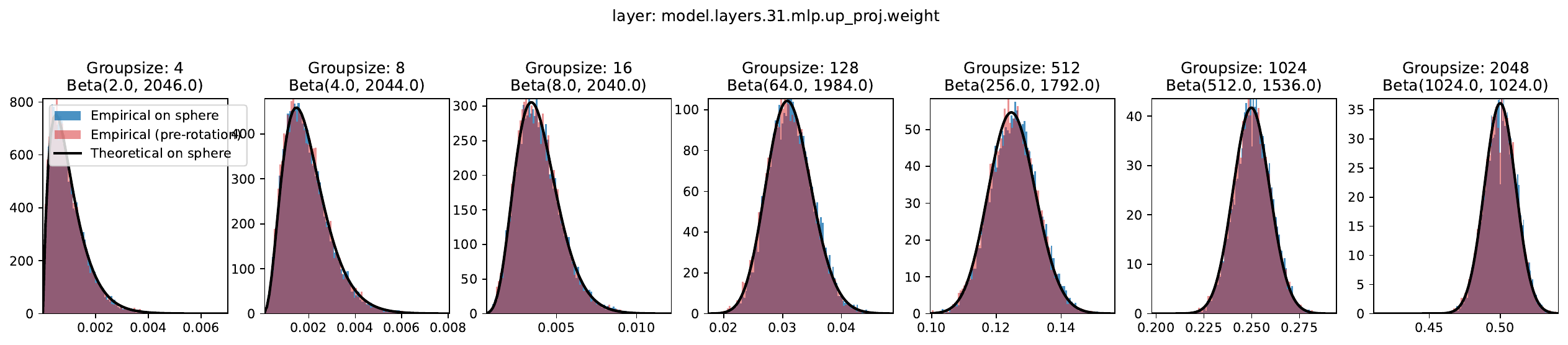} &
\includegraphics[width=0.22\linewidth]{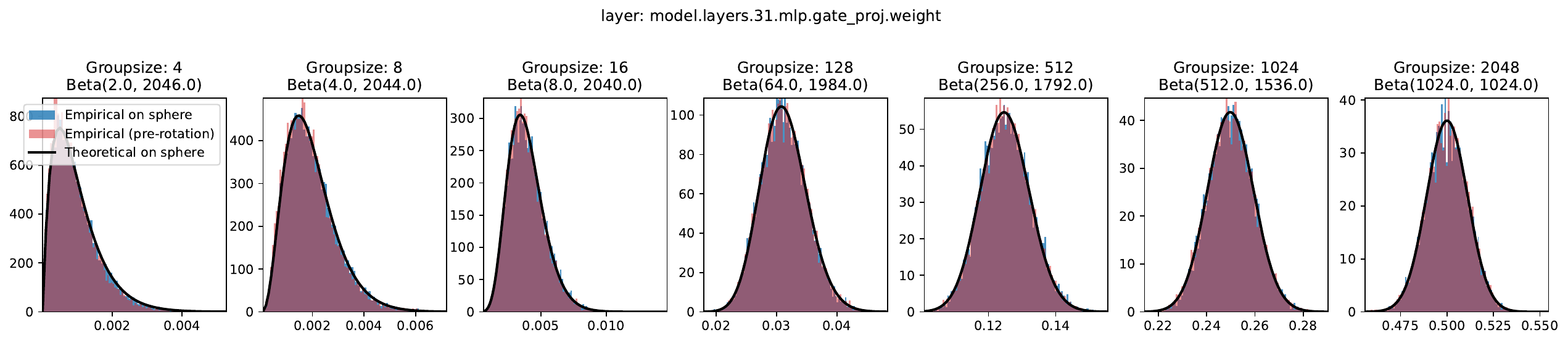} 
\end{tabular}
\end{table}

% \newpage
% \section{Experimental details}
% \label{sec:experimental-details}

\newpage
\newpage
\section{Additional results}

\subsection{Additional weight-only experiments (direction only)}
\label{sec:additional-weight-only}

\begin{table}[h!]
\vspace{-1.0em}
\resizebox{\linewidth}{!}{
\begin{tabular}{l r c c c c | c | cccccc | c}
& & & & &  
& \multicolumn{7}{c}{Phi-3} \\
Method & & Groupsize & Hessian & Spherical & BPW &
PPL $\uparrow$ &
PQ $\uparrow$ &
WG $\uparrow$ &
HS $\uparrow$ &
A-e $\uparrow$ &
A-c $\uparrow$ &
LA $\uparrow$ &
Avg. $\uparrow$ \\
\hline
Original & & & & & 16.000 & 6.01 & 0.81 & 0.73 & 0.78 & 0.79 & 0.57 & 0.65 & 0.72 \\
\hline RTN & & & & & 3.125 & 19.03 & 0.72 & 0.58 & 0.62 & 0.60 & 0.41 & 0.25 & 0.53 \\
GPTQ & & 128 & \checkmark & & 3.125 & 7.36 & 0.77 & 0.66 & 0.69 & 0.74 & 0.51 & 0.53 & 0.65 \\
\quarot & & 128 & \checkmark & \checkmark & 3.125 & 7.17 & 0.77 & 0.69 & 0.71 & 0.74 & 0.50 & 0.62 & 0.67 \\
\hline PVQ [3 bit directions, 16 bit amplitudes] & & 128 & \checkmark & \checkmark & 3.125 & 6.85 & 0.79 & 0.71 & 0.72 & 0.76 & 0.51 & 0.62 & 0.68 \\
\hline
\end{tabular}
}
\vspace{-0.5em}
\caption{Performance on downstream tasks. We compare performance on zero-shot downstream tasks after quantizing weights using different weight quantization methods.}
\end{table}

\begin{table}[h!]
\vspace{-1.0em}
\resizebox{\linewidth}{!}{
\begin{tabular}{l r c c c c | c | cccccc | c}
& & & & &  
& \multicolumn{7}{c}{Mixtral 8x7B} \\
Method & & Groupsize & Hessian & Spherical & BPW &
PPL $\uparrow$ &
PQ $\uparrow$ &
WG $\uparrow$ &
HS $\uparrow$ &
A-e $\uparrow$ &
A-c $\uparrow$ &
LA $\uparrow$ &
Avg. $\uparrow$ \\
\hline
Original & & & & & 16.000 & 3.84 & 0.84 & 0.76 & 0.84 & 0.83 & 0.60 & 0.78 & 0.78 \\
\hline RTN & & & & & 3.125 & 8.95 & 0.78 & 0.66 & 0.69 & 0.71 & 0.45 & 0.64 & 0.66 \\
GPTQ & & 128 & \checkmark & & 3.125 & 8.40 & 0.69 & 0.60 & 0.47 & 0.47 & 0.30 & 0.60 & 0.52 \\
\quarot & & 128 & \checkmark & \checkmark & 3.125 & 4.29 & 0.82 & 0.76 & 0.83 & 0.82 & 0.58 & 0.78 & 0.76 \\
\hline PVQ [3 bit directions, 16 bit amplitudes] & & 128 & \checkmark & \checkmark & 3.125 & 4.20 & 0.83 & 0.76 & 0.82 & 0.81 & 0.58 & 0.79 & 0.77 \\
\hline
\end{tabular}
}
\vspace{-0.5em}
\caption{Performance on downstream tasks. We compare performance on zero-shot downstream tasks after quantizing weights using different weight quantization methods.}
\end{table}

\begin{table}[h!]
\vspace{-1.0em}
\resizebox{\linewidth}{!}{
\begin{tabular}{l r c c c c | c | cccccc | c}
& & & & &  
& \multicolumn{7}{c}{Llama-3-8B} \\
Method & & Groupsize & Hessian & Spherical & BPW &
PPL $\uparrow$ &
PQ $\uparrow$ &
WG $\uparrow$ &
HS $\uparrow$ &
A-e $\uparrow$ &
A-c $\uparrow$ &
LA $\uparrow$ &
Avg. $\uparrow$ \\
\hline
Original & & & & & 16.000 & 6.13 & 0.81 & 0.73 & 0.79 & 0.78 & 0.53 & 0.76 & 0.73 \\
\hline RTN & & & & & 3.125 & 29.41 & 0.64 & 0.55 & 0.42 & 0.41 & 0.25 & 0.22 & 0.41 \\
GPTQ & & 128 & \checkmark & & 3.125 & 17.77 & 0.63 & 0.59 & 0.35 & 0.43 & 0.26 & 0.17 & 0.40 \\
\quarot & & 128 & \checkmark & \checkmark & 3.125 & 7.62 & 0.77 & 0.71 & 0.73 & 0.75 & 0.46 & 0.71 & 0.69 \\
\hline PVQ [3 bit directions, 16 bit amplitudes] & & 128 & \checkmark & \checkmark & 3.125 & 7.01 & 0.80 & 0.73 & 0.76 & 0.78 & 0.50 & 0.75 & 0.72 \\
\hline
\end{tabular}
}
\vspace{-0.5em}
\caption{Performance on downstream tasks. We compare performance on zero-shot downstream tasks after quantizing weights using different weight quantization methods.}
\end{table}

\subsection{Additional quantizing direction and amplitude experiments}
\label{sec:additional-direction-and-amplitude}

\begin{figure}[h]
\vspace{-0.5em}
\resizebox{\linewidth}{!}{
\includegraphics[width=1.0\linewidth]{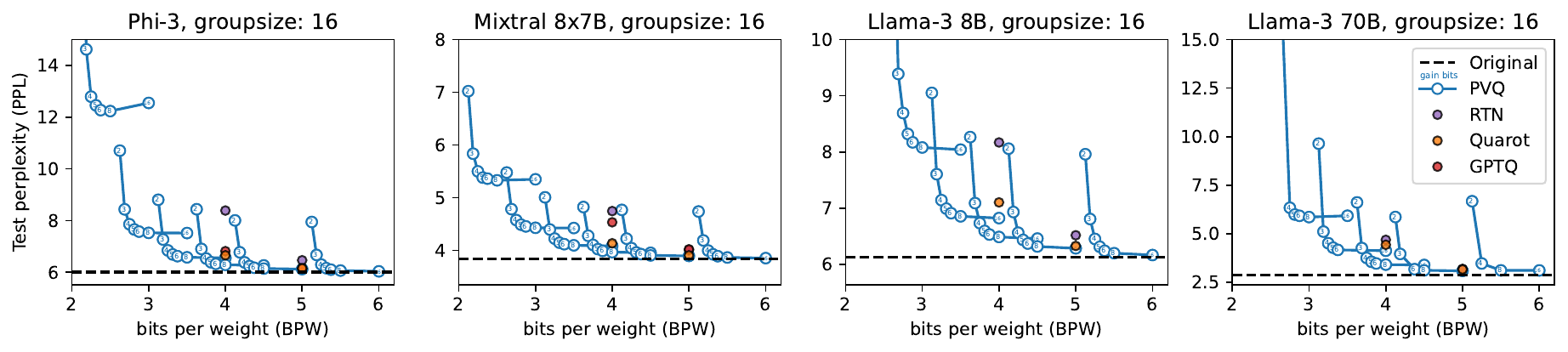}
}
\vspace{-2em}
\caption{direction and amplitude quantization. Test perplexities (PPL) with different quantization methods at various bit levels for amplitudes (within connected sets) and for direction bits (between connected sets).}
\end{figure}
\begin{figure}[h]
\vspace{-0.5em}
\resizebox{\linewidth}{!}{
\includegraphics[width=1.0\linewidth]{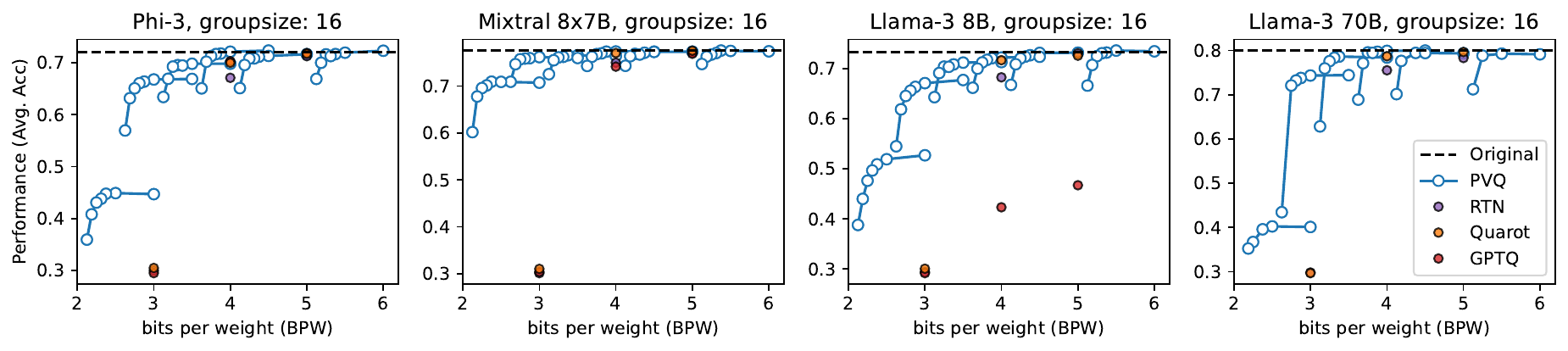}
}
\vspace{-2em}
\caption{direction and amplitude quantization. Average accuracies (Avg. Acc.) after quantizing with different methods at different bits for amplitudes (within connected set) and various direction bits (between connected sets).}
\end{figure}

\newpage
\subsection{Additional weight and activations experiments}
\label{sec:additional-weights-and-activations}

\begin{figure}[H]
\resizebox{\linewidth}{!}{
\includegraphics[width=1.15\linewidth]{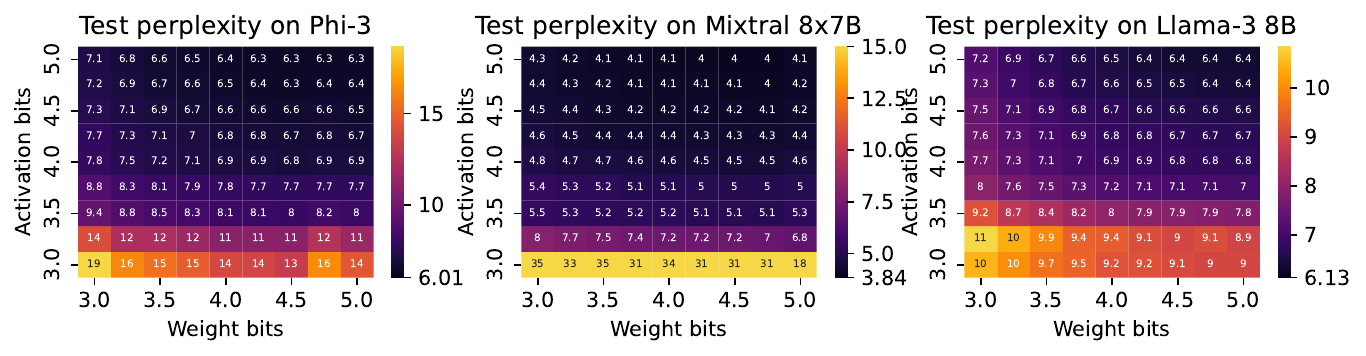}
}
\vspace{-2em}
\caption{Weights and activations. Comparing test perplexity at different bits per weight and bits per activations. From minimal compression (top right) to high levels of compression (bottom left).}
\label{fig:activations-plot}
\end{figure}

\begin{figure}[H]
\resizebox{\linewidth}{!}{
\includegraphics[width=1.15\linewidth]{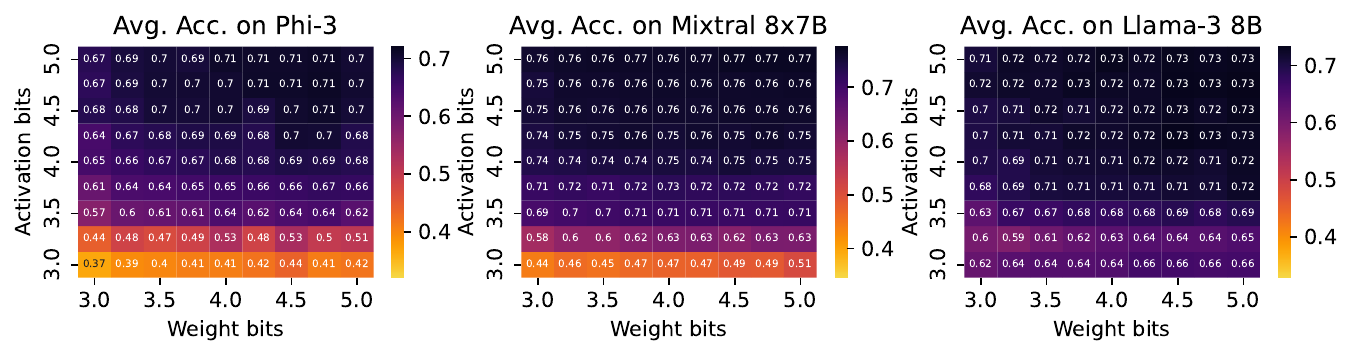}
}
\vspace{-2em}
\caption{Weights and activations. Comparing average accuracy at different bits per weight and bits per activations. From minimal compression (top right) to high levels of compression (bottom left).}
\label{fig:activations-plot}
\end{figure}

\newpage
\subsection{Additional Zero-shot downstream task experiments}
\label{sec:additional-downstream-tasks}

\begin{table}[h!]
\vspace{-1.0em}
\resizebox{\linewidth}{!}{
\begin{tabular}{l r c c c c | c | cccccc | c}
& & & & &  
& \multicolumn{7}{c}{Phi-3} \\
Method & & Groupsize & Hessian & Spherical & BPW &
PPL $\uparrow$ &
PQ $\uparrow$ &
WG $\uparrow$ &
HS $\uparrow$ &
A-e $\uparrow$ &
A-c $\uparrow$ &
LA $\uparrow$ &
Avg. $\uparrow$ \\
\hline
Original & & & & & 16.00 & 6.01 & 0.81 & 0.73 & 0.78 & 0.79 & 0.57 & 0.65 & 0.72 \\
\hline RTN & & & & & 4.00 & 8.39 & 0.77 & 0.68 & 0.74 & 0.76 & 0.53 & 0.55 & 0.67 \\
GPTQ & & 16 & \checkmark & & 4.00 & 6.82 & 0.79 & 0.71 & 0.74 & 0.79 & 0.57 & 0.63 & 0.70 \\
\quarot & & 16 & \checkmark & \checkmark & 4.00 & 6.66 & 0.78 & 0.72 & 0.73 & 0.78 & 0.55 & 0.64 & 0.70 \\
\hline PVQ [3.0 bit directions, 4 bit amplitudes] & & 16 & \checkmark & \checkmark & 3.25 & 6.85 & 0.79 & 0.72 & 0.74 & 0.77 & 0.54 & 0.60 & 0.69 \\
PVQ [3.5 bit directions, 6 bit amplitudes] & & 16 & \checkmark & \checkmark& \textbf{3.88} & \textbf{6.33} & \textbf{0.81} & \textbf{0.72} & \textbf{0.76} & \textbf{0.80} & \textbf{0.57} & \textbf{0.64} & \textbf{0.72} \\
\hline
\end{tabular}
}
\vspace{-0.5em}
\caption{Performance on downstream tasks after quantizing different open source LLM models using various post-training quantization methods. PVQ yields the highest performance after quantization.}
\label{tab:downstream-mixtral}
\end{table}

\begin{table}[h!]
\vspace{-1.0em}
\resizebox{\linewidth}{!}{
\begin{tabular}{l r c c c c | c | cccccc | c}
& & & & &  
& \multicolumn{7}{c}{Mixtral 8x7B} \\
Method & & Groupsize & Hessian & Spherical & BPW &
PPL $\uparrow$ &
PQ $\uparrow$ &
WG $\uparrow$ &
HS $\uparrow$ &
A-e $\uparrow$ &
A-c $\uparrow$ &
LA $\uparrow$ &
Avg. $\uparrow$ \\
\hline
Mixtral 8x7B
Original & & & & & 16.00 & 3.84 & 0.84 & 0.76 & 0.84 & 0.83 & 0.60 & 0.78 & 0.78 \\
\hline RTN & & & & & 4.00 & 4.74 & 0.83 & 0.76 & 0.81 & 0.80 & 0.55 & 0.75 & 0.75 \\
GPTQ & & 16 & \checkmark & & 4.00 & 4.53 & 0.81 & 0.76 & 0.78 & 0.79 & 0.55 & 0.76 & 0.74 \\
\quarot & & 16 & \checkmark & \checkmark & 4.00 & 4.13 & 0.83 & 0.76 & 0.83 & 0.83 & 0.59 & 0.79 & 0.77 \\
\hline PVQ [3.0 bit directions, 4 bit amplitudes] & & 16 & \checkmark & \checkmark & 3.25 & 4.22 & 0.83 & 0.76 & 0.83 & 0.81 & 0.57 & 0.78 & 0.76 \\
PVQ [3.5 bit directions, 6 bit amplitudes] & & 16 & \checkmark & \checkmark& \textbf{3.88} & \textbf{3.99} & \textbf{0.84} & \textbf{0.76} & \textbf{0.84} & \textbf{0.82} & \textbf{0.60} & \textbf{0.78} & \textbf{0.77} \\
\hline
\end{tabular}
}
\vspace{-0.5em}
\caption{Performance on downstream tasks after quantizing different open source LLM models using various post-training quantization methods. PVQ yields the highest performance after quantization.}
\label{tab:downstream-mixtral}
\end{table}

\begin{table}[h!]
\vspace{-1.0em}
\resizebox{\linewidth}{!}{
\begin{tabular}{l r c c c c | c | cccccc | c}
& & & & &  
& \multicolumn{7}{c}{Llama-3-8B} \\
Method & & Groupsize & Hessian & Spherical & BPW &
PPL $\uparrow$ &
PQ $\uparrow$ &
WG $\uparrow$ &
HS $\uparrow$ &
A-e $\uparrow$ &
A-c $\uparrow$ &
LA $\uparrow$ &
Avg. $\uparrow$ \\
\hline
Original & & & & & 16.00 & 6.13 & 0.81 & 0.73 & 0.79 & 0.78 & 0.53 & 0.76 & 0.73 \\
\hline RTN & & & & & 4.00 & 8.17 & 0.78 & 0.71 & 0.74 & 0.70 & 0.45 & 0.70 & 0.68 \\
GPTQ & & 16 & \checkmark & & 4.00 & 1415.44 & 0.66 & 0.58 & 0.36 & 0.57 & 0.34 & 0.03 & 0.42 \\
\quarot & & 16 & \checkmark & \checkmark & 4.00 & 7.10 & 0.79 & 0.73 & 0.76 & 0.76 & 0.51 & 0.75 & 0.72 \\
\hline PVQ [3.0 bit directions, 4 bit amplitudes] & & 16 & \checkmark & \checkmark & 3.25 & 7.14 & 0.78 & 0.72 & 0.76 & 0.74 & 0.49 & 0.74 & 0.70 \\
PVQ [3.5 bit directions, 6 bit amplitudes] & & 16 & \checkmark & \checkmark& \textbf{3.88} & \textbf{6.53} & \textbf{0.80} & \textbf{0.73} & \textbf{0.77} & \textbf{0.76} & \textbf{0.51} & \textbf{0.75} & \textbf{0.72} \\
\hline
\end{tabular}
}
\vspace{-0.5em}
\caption{Performance on downstream tasks. We compare after quantizing different open source LLM models using various post-training quantization methods. PVQ yields the highest performance after quantization.}
\label{tab:downstream-mixtral}
\end{table}

\end{document}